%% file: Main.tex
\documentclass[Afour,sageh,times]{sagej}
\usepackage{moreverb,url}
\usepackage[nice]{nicefrac}
\usepackage{xfrac}
\usepackage{ulem}
\usepackage{subcaption}
\usepackage{sidecap}
\usepackage{microtype}
\usepackage[T1]{fontenc}
\usepackage{algpseudocode}
\usepackage[table,xcdraw]{xcolor}
\usepackage{booktabs} % For modern table lines
\usepackage{array} % For better column alignment
\usepackage[linesnumbered,ruled,vlined]{algorithm2e}

\input{math_commands.tex}

\input{math_commands_ncds.tex}
\usepackage[colorlinks,bookmarksopen,bookmarksnumbered,citecolor=red,urlcolor=red]{hyperref}

\newcommand{\etal}{\MakeLowercase{\textit{et al.}}}
\newcommand\BibTeX{{\rmfamily B\kern-.05em \textsc{i\kern-.025em b}\kern-.08em
T\kern-.1667em\lower.7ex\hbox{E}\kern-.125emX}}

\def\volumeyear{2016}
\begin{document}

\runninghead{Beik-Mohammadi \etal}

\title{Extended Neural Contractive Dynamical Systems: On Multiple Tasks and Riemannian Safety Regions}

\author{Hadi Beik-Mohammadi\affilnum{1}\affilnum{2}, S\o{}ren Hauberg\affilnum{3}, Georgios Arvanitidis\affilnum{3}, Gerhard Neumann\affilnum{2}, and Leonel Rozo\affilnum{1} }

\affiliation{\affilnum{1} Bosch Center for Artificial Intelligence (BCAI), Renningen, Germany\\
\affilnum{2} Autonomous Learning Robots Lab, Karlsruhe Institute of Technology (KIT), Karlsruhe, Germany\\
\affilnum{3} Section for Cognitive Systems, Technical University of Denmark (DTU), Lyngby, Denmark}

\corrauth{Hadi Beik-Mohammadi, Bosch Center for Artificial Intelligence (BCAI)
}

\email{hadi.beik-mohammadi@de.bosch.com}

\begin{abstract}
    % Stability guarantees are crucial when ensuring that a fully autonomous robot does not take undesirable or potentially harmful actions.
    % Unfortunately, global stability guarantees are hard to provide in dynamical systems learned from data, especially when the learned dynamics are governed by neural networks. 
    % We propose a novel methodology to learn \emph{neural contractive dynamical systems}, where our neural architecture ensures contraction, and hence, global stability.
    % To efficiently scale the method to high-dimensional dynamical systems, we develop a variant of the variational autoencoder that learns dynamics in a low-dimensional latent representation space while retaining contractive stability after decoding. %
    % We extend our approach to learning contractive systems on the Lie group of rotations to account for full-pose end-effector dynamic motions.
    % The result is the first highly flexible learning architecture that provides contractive stability guarantees with the capability to perform obstacle avoidance.
    % Empirically, we demonstrate that our approach encodes the desired dynamics more accurately than the current state-of-the-art, which provides less strong stability guarantees.
    Stability guarantees are crucial when ensuring that a fully autonomous robot does not take undesirable or potentially harmful actions.
    We recently proposed the \emph{Neural Contractive Dynamical Systems (NCDS)}, which is a neural network architecture that guarantees contractive stability. With this, learning-from-demonstrations approaches can trivially provide stability guarantees.
    However, our early work left several unanswered questions, which we here address. Beyond providing an in-depth explanation of NCDS, this paper extends the framework with more careful regularization, a conditional variant of the framework for handling multiple tasks, and an uncertainty-driven approach to latent obstacle avoidance. Experiments verify that the developed system has the flexibility of ordinary neural networks while providing the stability guarantees needed for autonomous robotics. 
\end{abstract}


\maketitle

\input{Sections/Introduction}

\input{Sections/Approach}

\definecolor{customblue}{HTML}{94c9df}
\definecolor{customred}{HTML}{e98e83}
\definecolor{customgreen}{HTML}{a4efa4}

\section{Experiments}
\input{Sections/Experiments}

\section{Discussion and future work}
\input{Sections/Discussion}
%\subsection{References}

\begin{acks}
This work was supported by a research grant (42062) from VILLUM FONDEN.
This work was partly funded by the Novo Nordisk Foundation through the Center for Basic Research in Life Science (NNF20OC0062606).
This project received funding from the European Research Council (ERC) under the European Union's Horizon Programme (grant agreement 101125003).
\end{acks}

\bibliographystyle{SageH}
\bibliography{Main.bib}

\section{Appendix}
\input{Sections/Appendix}
\end{document}

%% file: math_commands.tex
\usepackage{amsmath,amsfonts,bm}

\def\eqref#1{equation~\ref{#1}}
\def\1{\bm{1}}

\def\vp{{\bm{p}}}

\def\vr{{\bm{r}}}

\def\vz{{\bm{z}}}

\DeclareMathAlphabet{\mathsfit}{\encodingdefault}{\sfdefault}{m}{sl}
\SetMathAlphabet{\mathsfit}{bold}{\encodingdefault}{\sfdefault}{bx}{n}

\newcommand{\R}{\mathbb{R}}

\DeclareMathOperator{\Tr}{Tr}

%% file: math_commands_ncds.tex
\newtheorem{theorem}{Theorem}
\newtheorem{definition}{Definition}
\newcommand{\x}{\bm{x}} % input space variable/position
\newcommand{\z}{\bm{z}} % latent space variable
\newcommand{\dif}[0]{\mathrm{d}} % differential, e.g. dx/dy
\newcommand{\Metric}{\bm{M}} % riemannian metric
\newcommand{\Jac}{\bm{J}} % function Jacobian
\newcommand{\trsp}{\mathsf{T}} % Transpose

\newcommand{\ambient}{\mathcal{X}} % ambient space
\newcommand{\latent}{\mathcal{Z}} % latent space
\newcommand{\Manifold}{\mathcal{M}} % riemannian manifold
\newcommand{\I}{\mathbb{I}} % identity matrix
\newcommand{\Prob}{p} % probability distribution

\newcommand{\Length}{\mathcal{L}} % curve length
\newcommand{\SO}{\mathcal{SO}(3)}
\newcommand{\so}{\mathfrak{so}(3)} 

%% file: Sections/Introduction.tex
\section{Introduction}

Autonomous robotic systems require stability guarantees to ensure safe and reliable operation, especially in dynamic and unpredictable environments. Manually designed robot movements can achieve such stability guarantees, but even skilled engineers struggle to hand-code highly dynamic motions~\citep{Billard16}. In contrast, learning robot skills from human demonstrations is an efficient and intuitive approach for encoding highly dynamic motions into a robot's repertoire \citep{Schaal03}. Unfortunately, learning-based approaches often struggle to ensure stability as they rely on the machine learning model to extrapolate in a controlled manner. In particular, controlling the extrapolation behavior of neural networks has proven challenging~\citep{xu2020neural}, which hampers stability guarantees.

Many tasks require the robot to dynamically follow desired trajectories, e.g.\@ in flexible manufacturing, human-robot interaction, or entertainment settings. In these cases, asymptotic stability, which guarantees stability only with respect to a fixed point~\citep{EuclidFlow2020:Rana, KhansariZadeh2011:StableEstimatorDS,zhang2022RiemannianStableDS}, is insufficient, demanding a more general stability concept. \emph{Contraction theory}~\citep{Lohmiller1998ContractionAnalysis, Bullo2024:ContractionBook} is particularly well-suited, as it ensures that all integral curves, regardless of their initial state, incrementally converge over time. Unfortunately, the mathematical requirements of a contractive system are difficult to ensure in popular neural network architectures.\looseness=-1

To address this problem, our previous work introduced Neural Contractive Dynamical Systems (NCDS, \citet{NCDS:2023BeikMohammadi}), a neural network architecture designed to ensure contractive behavior across all parameter values. 
This model enables robots to maintain stability even when subjected to external perturbations.
The original NCDS (``vanilla NCDS'') framework provides a solid foundation for learning complex contractive dynamical systems. Still, there remains potential for further refinement, particularly in formulating the Jacobian of the learned system dynamics, which is crucial for stability and generalization capabilities. Vanilla NCDS employs a symmetric Jacobian with a constant regularization term to enforce contraction~\citep{Lohmiller1998ContractionAnalysis, Jouffroy2010ConractiontheoryTutorial}. While effective, this approach left room for exploring how different regularization methods and Jacobian formulations could enhance the model's performance both within and beyond the data support region.

\begin{figure}[t]
\centering
    \includegraphics[width=1.0\linewidth]{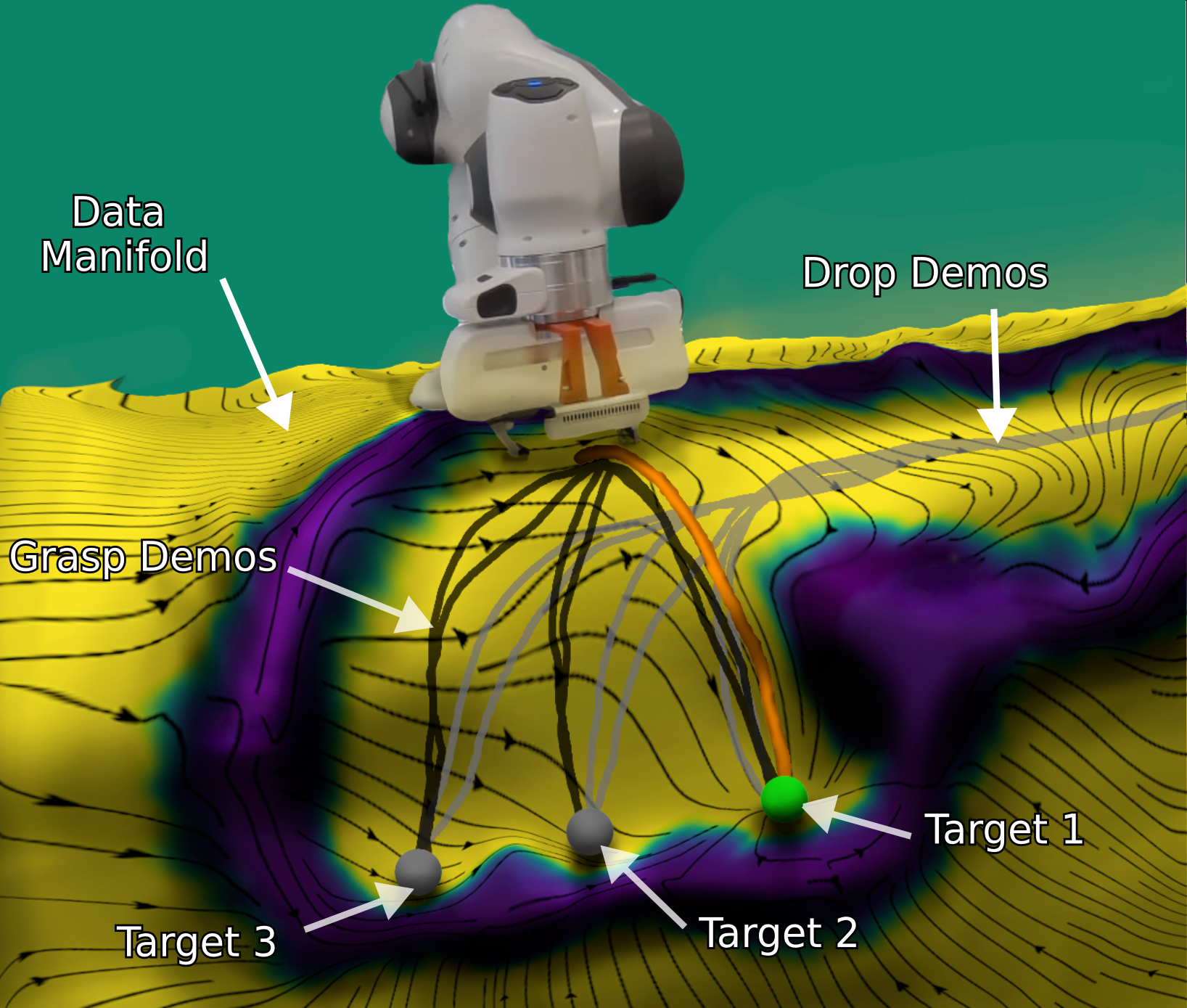}
    \caption{Robot motion generated for a grasping skill using a Conditional Neural Contractive Dynamical System (CNCDS). The robot motion is conditioned on an input image of the object placed on a desk. CNCDS also learns a Riemannian manifold from demonstrations, representing a safety region to navigate through.\looseness=-1}
    \label{fig:teaser}
\end{figure} 

The vanilla NCDS framework learns high-dimensional dynamics via a low-dimensional contractive latent space. However, it performs all obstacle avoidance computations in the ambient (high-dimensional) space, which may be computationally intensive. To the best of our knowledge, no existing method for learning dynamical systems can simultaneously learn contractive stable dynamics while also performing obstacle avoidance in the latent space.
%Meanwhile in~\cite{BeikMohammadi23:ReactiveMotion}, the authors learn non-dynamic motion skills from human demonstrations within a low-dimensional latent space. This approach enables obstacle avoidance directly in the latent space by leveraging a Riemannian pull-back metric, also ensuring trajectories stay within the data support. 
To achieve this, we extend the work of \citet{BeikMohammadi23:ReactiveMotion}, and leverage a Riemannian pullback metric, derived from a VAE injective decoder, to lift the obstacle avoidance problem to the latent space. This ensures the avoidance of unsafe regions outside of data support.
On the multi-task front, the vanilla NCDS framework, like most methods that learn contractive dynamical systems ~\citep{Tsukamoto2021NeuralContractionMetric, Dawson2022LearnedCertificates, jaffe2024:ELCD}, was designed to learn a single skill. Although this may suffice for certain tasks, a more general approach would involve enabling the learning of multiple skills within the same contractive dynamical system. This reduces not only the use of computational resources but also the need to train separate models for each skill.\looseness=-1

\textbf{In this paper}, we introduce several advancements to the vanilla NCDS aimed at improving the model's flexibility and robustness. Our \textbf{first contribution} focuses on the regularization of the Jacobian of the learned system dynamics. We investigate various unexplored strategies, including state-independent and state-dependent regularization vectors, as well as eigenvalue-based approach. These methods aim to fine-tune the contraction properties of the system, leading to more robust generalization and faster convergence within the data region.
Our \textbf{second contribution} explores the potential benefits of introducing asymmetry into the system Jacobian matrix by incorporating both symmetric and skew-symmetric components. This asymmetry aims at increasing the model's flexibility, allowing it to capture more complex dynamical behaviors~\citep{jaffe2024:ELCD}. This investigation builds on the understanding that, while symmetry in the Jacobian provides certain stability benefits, asymmetry could offer more nuanced control over the system's dynamics, particularly relevant in complex dynamic skills.
Our \textbf{third contribution}, ``conditional NCDS'' (CNCDS), extends the model to handle multiple motion skills by conditioning on task-related variables such as target states. This conditional framework allows a single NCDS module to adapt to varying task conditions, broadening its applicability in more complex, multimodal robotic tasks.
Finally, our \textbf{fourth contribution} extends NCDS with latent obstacle avoidance capabilities. Our approach, inspired by modulation matrix techniques, allows the system to navigate around dynamic obstacles while maintaining contractive stability \citep{Huber2022ModulationMatrix, Huber2019:ConvexConcaveObstacles}. 

We frame obstacle avoidance from a Riemannian perspective and define safety regions in the latent space through Riemannian pullback metrics~\citep{BeikMohammadi23:ReactiveMotion}. This designates both obstacles and out-of-data-support regions as unsafe.

%% file: Sections/Approach.tex
\section{Learning contractive vector fields}
\label{sec:contract}
In this section, we revisit the neural contractive dynamical system (NCDS) method introduced by~\cite{NCDS:2023BeikMohammadi}. Later, we explain how NCDS can be improved through the introduction of a new data-driven regularization technique and modifications to the symmetry of its Jacobian formulation. Our main goal is to design a flexible neural architecture that is guaranteed to always output a contractive vector field. First, we introduce contraction theory as it is prerequisite to our design.
\subsection{Background: contractive dynamical systems}
\label{subsec:contraction}
Assume an autonomous dynamical system $\dot{\x}_t = f(\x_t)$, where $\x_t \in \R^D$ is the state variable, $f: \R^D \rightarrow \R^D$ is a $C^1$ function, and $\dot{\x}_t = \nicefrac{\mathrm{d}\x}{\mathrm{d}t}$ denotes temporal differentiation. 
As depicted in Fig.~\ref{fig:example}, contraction stability guarantees that all solution trajectories of a nonlinear system $f$ incrementally converge regardless of initial conditions $\x_0,\dot{\x}_0$, and temporary perturbations \citep{Lohmiller1998ContractionAnalysis}.  
The system stability can, thus, be analyzed differentially, i.e.\@ we can ask if two nearby trajectories converge to one another. 
Specifically, contraction theory defines a measure of distance between neighboring trajectories, known as the \emph{contraction metric}, under which the distance decreases exponentially over time \citep{Jouffroy2010ConractiontheoryTutorial, Tsukamoto2021TutorialContraction}. \looseness=-1

Formally, an autonomous dynamical system yields the differential relation $\delta \dot{\x} = \bm{J}(\x) \delta \x$, where $\bm{J}(\x) = \nicefrac{\partial f}{\partial \x}$ is the system Jacobian and $\delta \x$ is a virtual displacement (i.e., an infinitesimal spatial displacement between the nearby trajectories at a fixed time). 
\begin{figure*}[h!]
  \centering
      \begin{subfigure}{.25\linewidth}
            \centering
            \includegraphics[width=1.\linewidth]{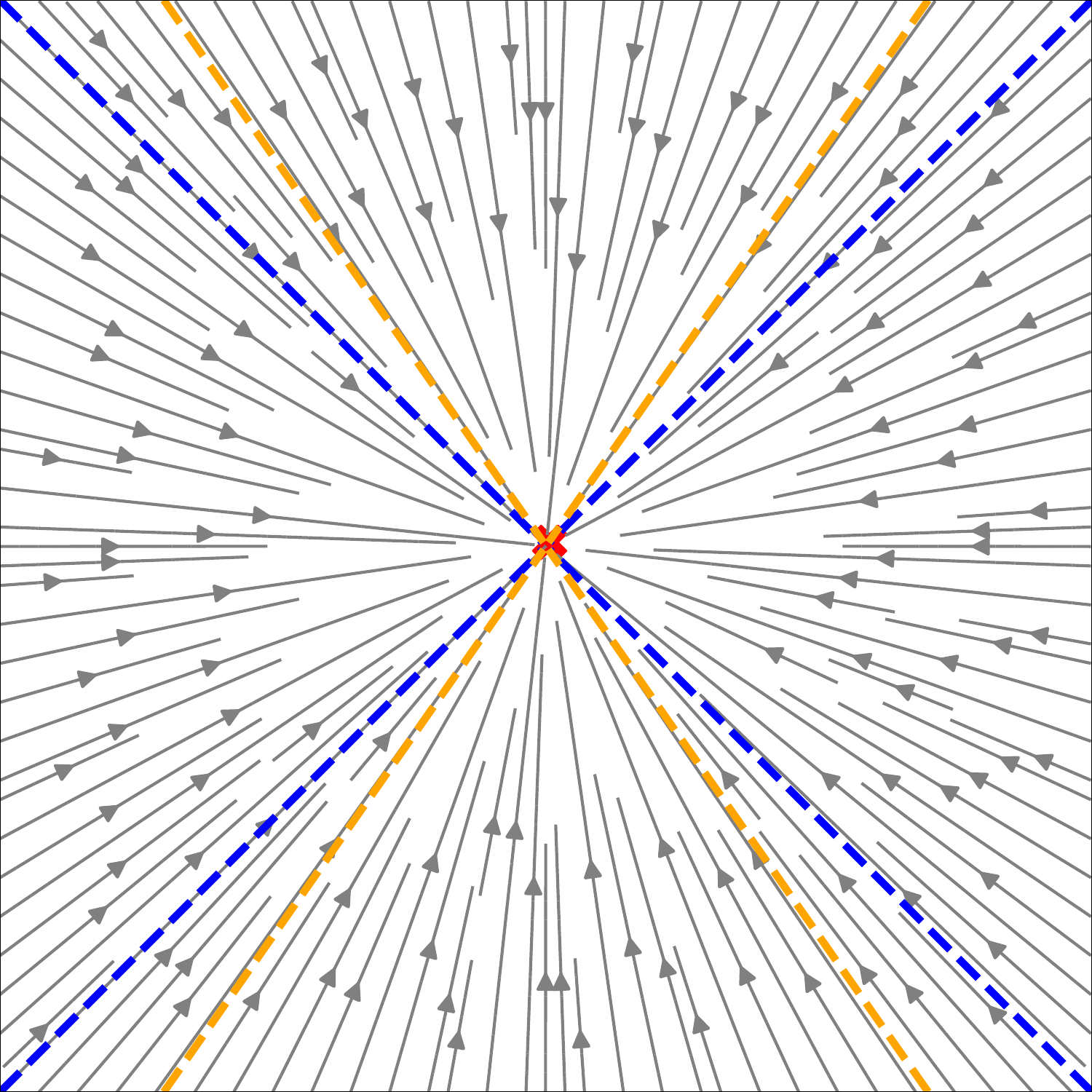}
      \end{subfigure}
        \begin{subfigure}{.25\linewidth}
            \centering
            \includegraphics[width=1.0\linewidth]{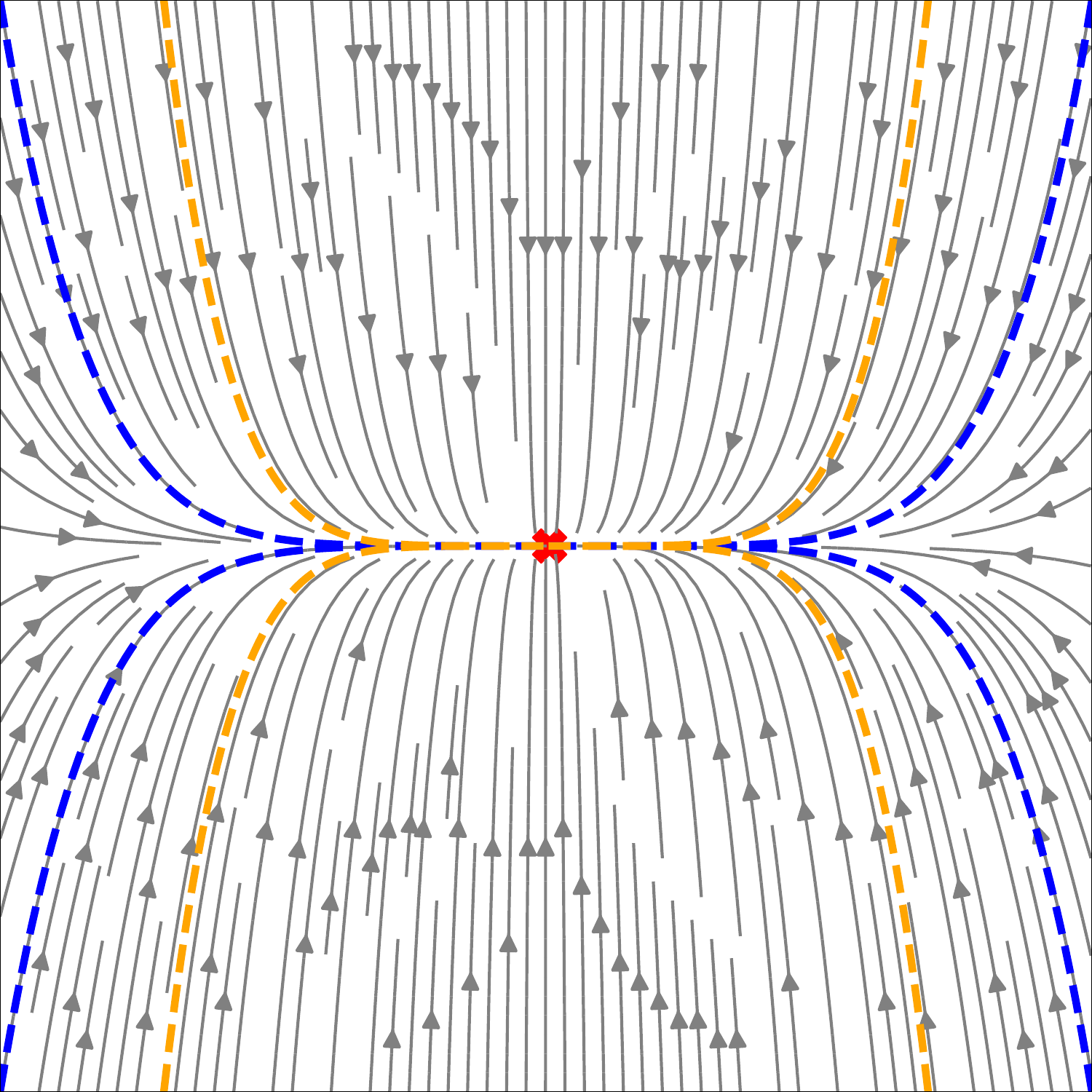}
      \end{subfigure}
          \begin{subfigure}{.25\linewidth}
        \centering
        \includegraphics[width=1.0\linewidth]{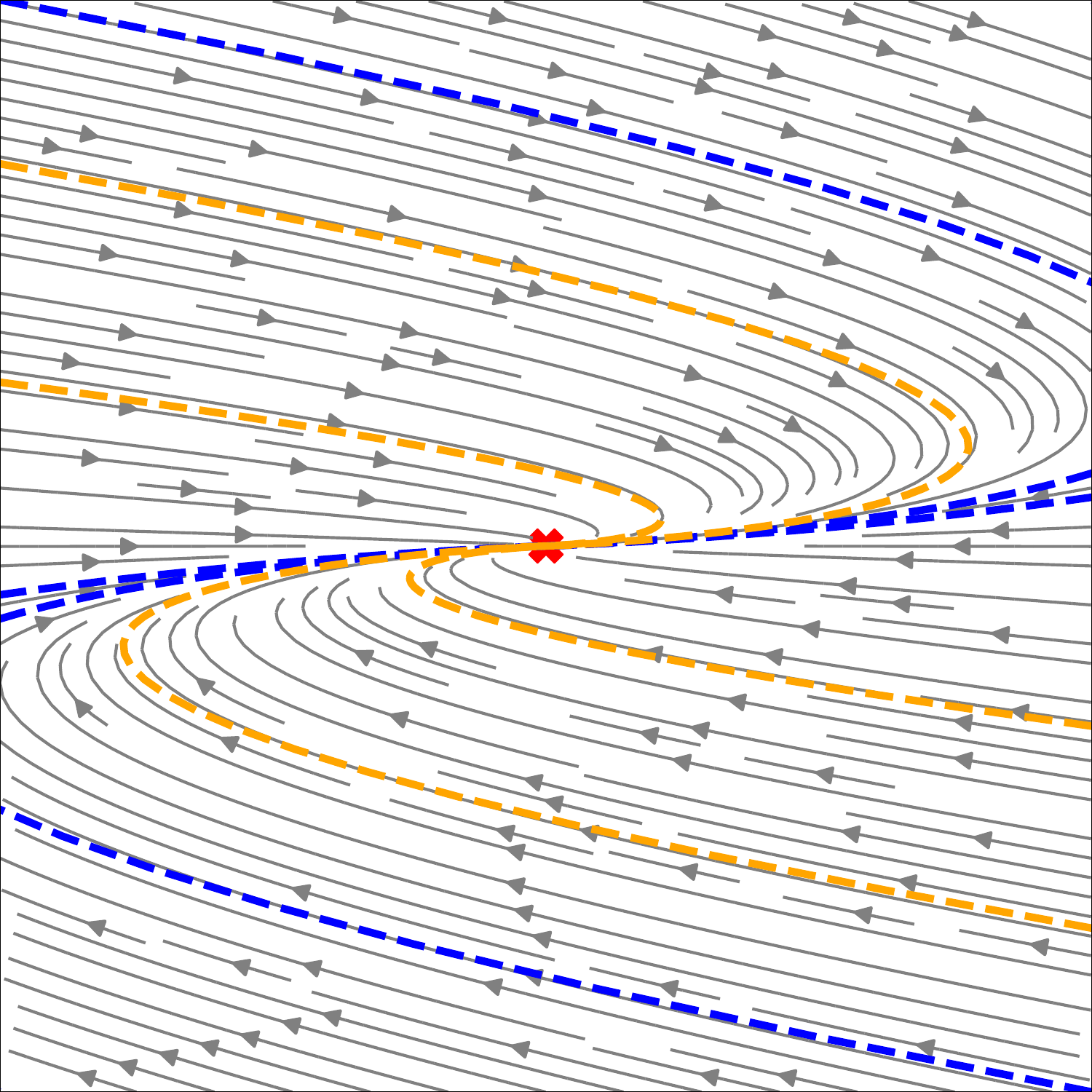}
      \end{subfigure}
      \caption{The effect of eigenvalue differences and asymmetry of the Jacobian on the contraction behavior of the vector field. \emph{Left}: In a contractive dynamical system with a symmetric Jacobian, when the eigenvalues are equal, the system contracts uniformly in all directions. \emph{Middle}: In a contractive dynamical system with a symmetric Jacobian, when the eigenvalues are different, the system contracts more rapidly in the direction associated with the larger eigenvalue. \emph{Right}: An example of a contractive dynamical system with an asymmetric Jacobian.}
        \label{fig:eigenvalues}
\end{figure*}
Note that we have dropped the time index $t$ to limit notational clutter. 
The rate of change of the corresponding infinitesimal squared distance $\delta \x^\trsp \delta \x$ is
\begin{equation}
    \frac{\dif}{\dif t}(\delta \x^\trsp \delta \x) = 2\delta \x^\trsp \delta \dot{\x} = 2 \delta \x^\trsp \bm{J}(\x) \delta \x.
    \label{eq:rate_of_change}    
\end{equation}
It follows that if the symmetric part of the Jacobian $\bm{J}(\bm{x})$ is negative definite, then the infinitesimal squared distance $\delta \x^\trsp \delta \x$ between neighboring trajectories decreases over time. 
This is formalized as follows.
\begin{definition}[Contraction stability \citep{Lohmiller1998ContractionAnalysis}]
    \label{th:contraction_stability}
    An autonomous dynamical system $\dot{\x} = f(\x)$ exhibits a contractive behavior if its Jacobian $\bm{J}(\x) = \nicefrac{\partial f}{\partial \x} $ is uniformly negative definite, or equivalently if its symmetric part is negative definite. This means that there exists a constant $\tau > 0 $ such that $\delta \x^\trsp \delta \x$ converges to zero exponentially at rate $2\tau$, i.e., $\lVert \delta \x \rVert \leq e^{-\tau t} \lVert \delta \x_0 \rVert $. This can be summarized as,
    \begin{equation}
        \exists\,\tau > 0   \enspace \text{s.t.} \enspace \forall \x, \enspace \cfrac{1}{2}\Big(\Jac(\x)+\Jac(\x)^\trsp \Big) \prec -\tau \I \prec 0 .
        \label{eq:negative_def_Jacobian}
    \end{equation} 
\end{definition}
The above analysis can be generalized to account for a more general notion of distance of the form $\delta \x^\trsp \bm{M}(\x) \delta \x$, where $\bm{M}(\x)$ is a positive-definite matrix known as the Riemannian contraction metric \citep{Tsukamoto2021TutorialContraction, Dawson2022LearnedCertificates}.
In our work, we learn a dynamical system $f$ so that it is inherently contractive since its Jacobian $\bm{J}(\x)$ fulfills the condition given in~\eqref{eq:negative_def_Jacobian}. 
Consequently, it is not necessary to separately learn a contraction metric as an identity matrix suffices. 

\subsection{Neural contractive dynamical systems}
\label{subsec:learning_jacobian}
From Definition~\ref{th:contraction_stability}, we seek a flexible neural network architecture, such that the symmetric part of its Jacobian is negative definite. 
We consider the dynamical system $\dot{\x}~=~f(\x)$, where $\x \in \R^D$ denotes the system's state and $f: \R^D \rightarrow \R^D$ is a neural network. 
Note that it is not trivial to impose a negative definiteness constraint on a network's Jacobian without compromising its expressiveness. 
To overcome this challenge, we first design a neural network $\hat{\Jac_f}$ representing directly the Jacobian of our final network. 
This produces matrix-valued negative definite outputs. 
The final neural network, parametrizing $f_{\bm{\theta}}$, will then be formed by integrating the Jacobian network. 
Specifically, we define the Jacobian as,
\begin{equation}
  \hat{\Jac_f}(\x)
    %= \cfrac{\dif f}{\dif \x}
    = - (\Jac_{\bm{\theta}}(\x)^\trsp \Jac_{\bm{\theta}}(\x) + \epsilon\ \I_D) ,
    \label{eq:Jacobian_simple}
\end{equation}
where $\Jac_{\bm{\theta}}: \R^D \rightarrow \R^{D \times D}$ is a neural network parameterized by $\bm{\theta}$, $\epsilon \in \R^+$ is a small positive constant, and $\I_D$ is an identity matrix of size $D$. 
Intuitively, $\Jac_{\bm{\theta}}$ can be interpreted as the (approximate) square root of $\hat{\Jac_f}$. 
Clearly, $\hat{\Jac_f}$ is negative definite as all eigenvalues are bounded from above by $-\epsilon$.  
Next, we take inspiration from \citet{lorraine2019jacnet} and integrate $\hat{\Jac_f}$ to produce a function $f$, which is implicitly parametrized by $\bm{\theta}$, and has Jacobian $\hat{\Jac_f}$. 
The fundamental theorem of calculus for line integrals tells us that we can construct such a function by a line integral,

\begin{SCfigure}[][t]
\centering
  \includegraphics[width=0.6\linewidth]{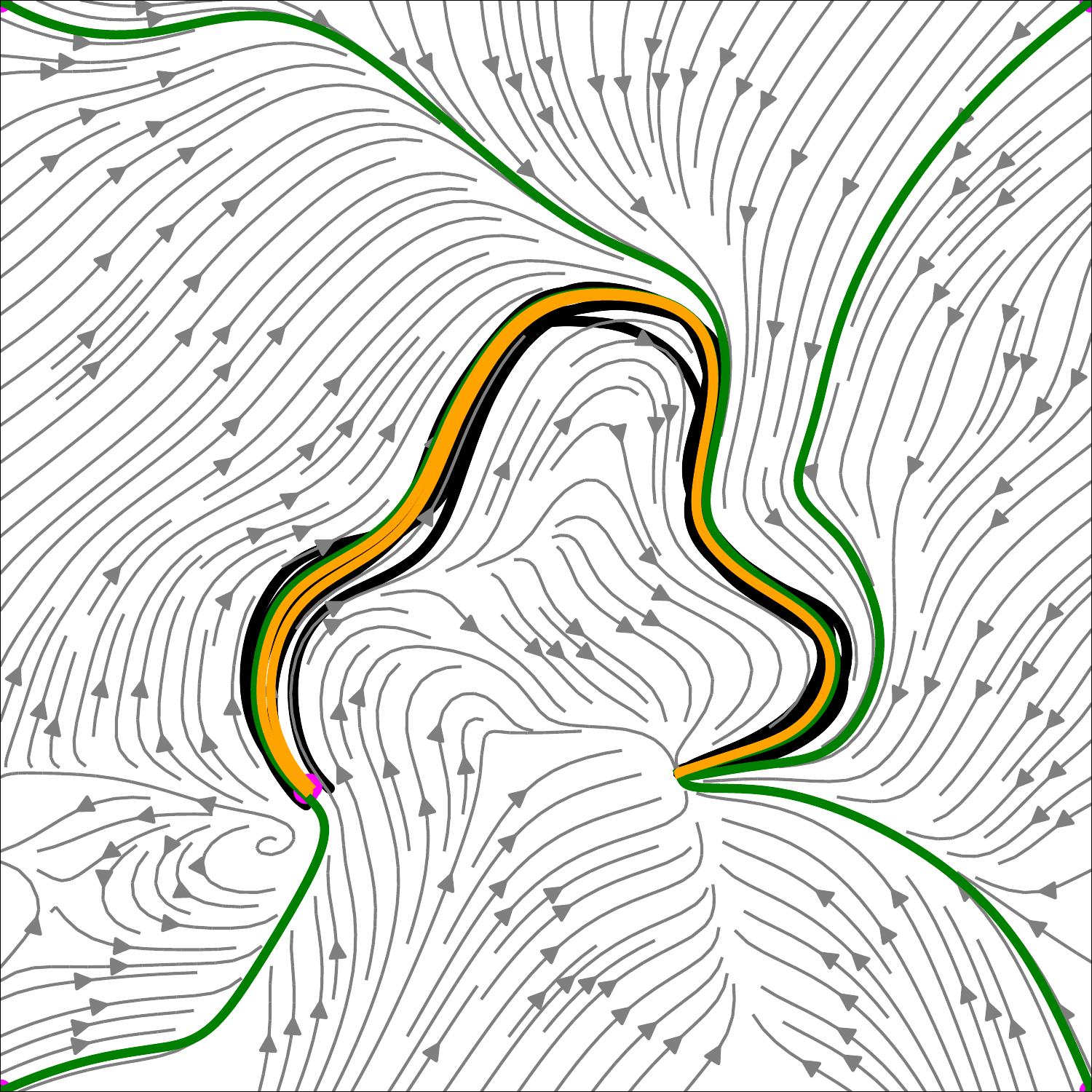}
  \caption{The learned vector field (grey) and demonstrations (black). Yellow and green trajectories show integral curves starting from demonstration starting points and random points, respectively.}
  \label{fig:example}
\end{SCfigure}

\begingroup\abovedisplayskip=0pt
\begin{align}
    \label{eq:integral}
    \dot{\x} = f(\x)
    &= \dot{\x}_0 + \int_{0}^{1} \hat{\Jac_f} \left(c \left(\x, t, \x_0 \right) \right) \dot{c}(\x, t, \x_0)  \dif t ,
\end{align}
    \begin{align*}
    \text{with} \quad c( \x, t, \x_0) &= \left( 1 - t \right) \x_0 + t \x , \\ \dot{c}(\x, t, \x_0) &= \x - \x_0 ,
\end{align*}
\endgroup
where $\x_0$ and $\dot{\x}_0 = f(\x_0)$ represent the initial conditions of the state variable and its first-order time derivative, respectively. 
The input point $\x_0$ can be chosen arbitrarily (e.g.\@ as the mean of the training data or it can be learned), while the corresponding function value $\dot{\x}_0$ has to be estimated along with the parameters $\bm{\theta}$.
This is justified by the fundamental theorem of line integrals, which ensures that the final result is independent of the specific choice of path and initial conditions~\citep{lorraine2019jacnet}. Joint optimization allows the model to correct for numerical integration errors and to select anchor points that help minimize the global overall velocity reconstruction loss.

Therefore, given a set of demonstrations denoted as $\mathcal{D} = \{\x_i, \dot{\x}_i\}$, our objective is to learn a set of parameters $\bm{\theta}$ along with the initial conditions $\x_0$ and $\dot{\x}_0$, such that the integration in~\eqref{eq:integral} enables accurate reconstruction of the velocities $\dot{\x}_i$ given the state $\x_i$.
This is achieved through the velocity reconstruction loss, %$\mathcal{L}_{\text{vel}}$, defined as:
\begin{equation}
\mathcal{L}_{\text{vel}} = \frac{1}{N} \sum_{i=1}^{N} \left\| \dot{\x}_i - \hat{\dot{\x}}_i \right\|^2,
\end{equation}
where $\dot{\x}_i$ denotes the demonstrated velocity, $\hat{\dot{\x}}_i$ is the predicted velocity, and $N$ represents the number of data points.
This process is shown in block $\mathsf{B}$ of the architecture in Fig.~\ref{fig:architecture}.\looseness=-1

\begin{figure*}[htb!]
    \centering
        \begin{subfigure}{.24\linewidth}
            \includegraphics[width=1.0\linewidth]{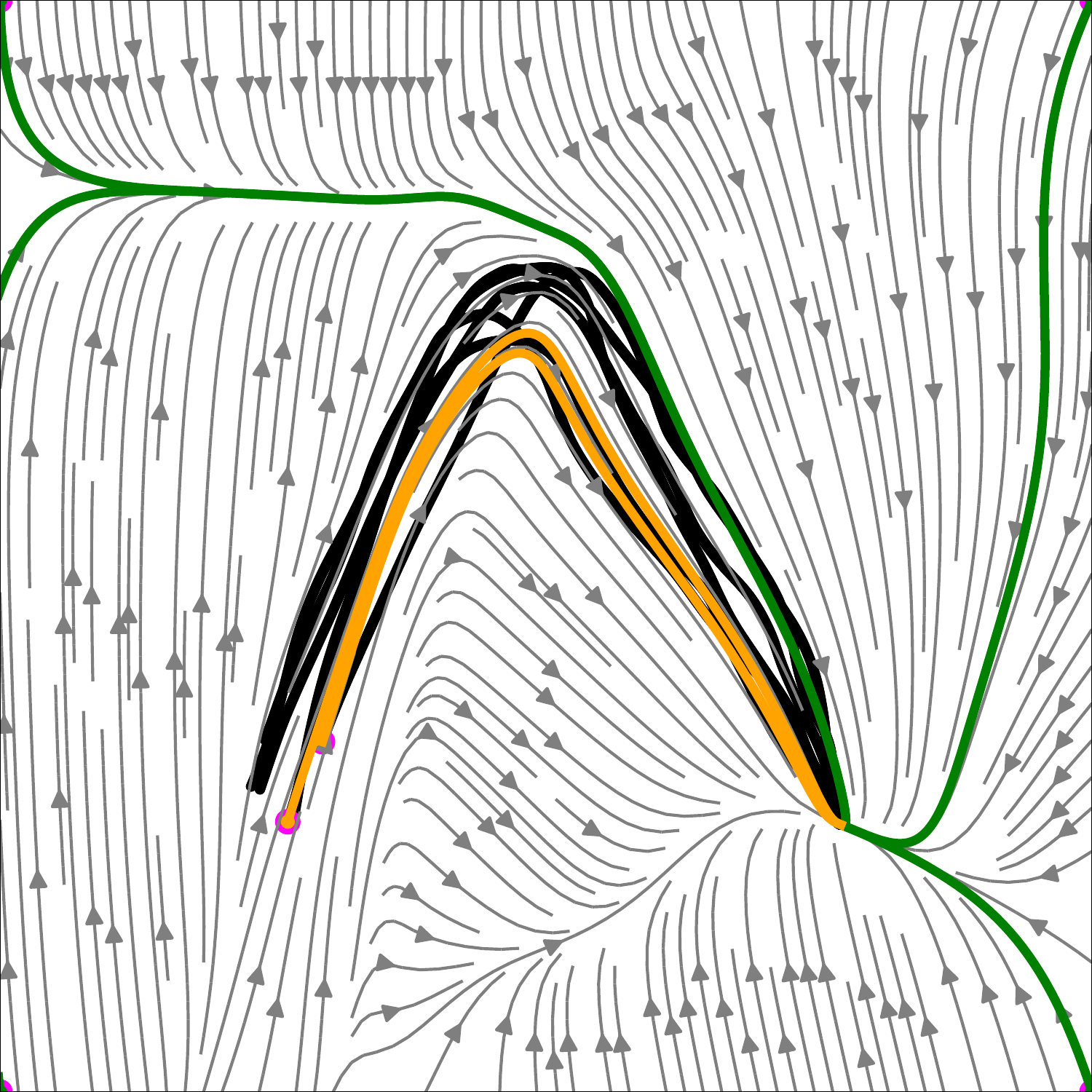}
        \end{subfigure}
        \begin{subfigure}{.24\linewidth}
            \includegraphics[width=1.0\linewidth]{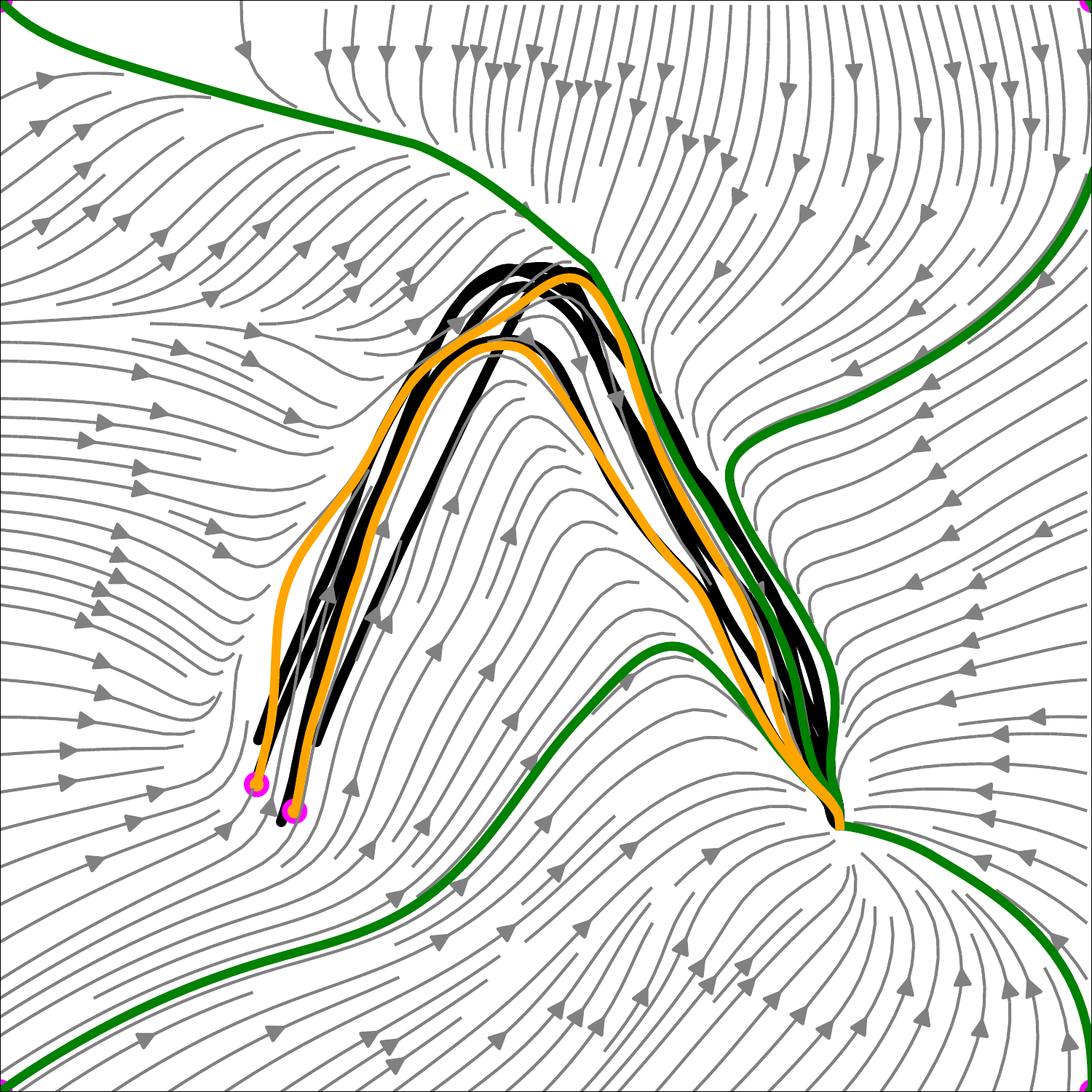}
        \end{subfigure}
        \begin{subfigure}{.24\linewidth}
            \includegraphics[width=1.0\linewidth]{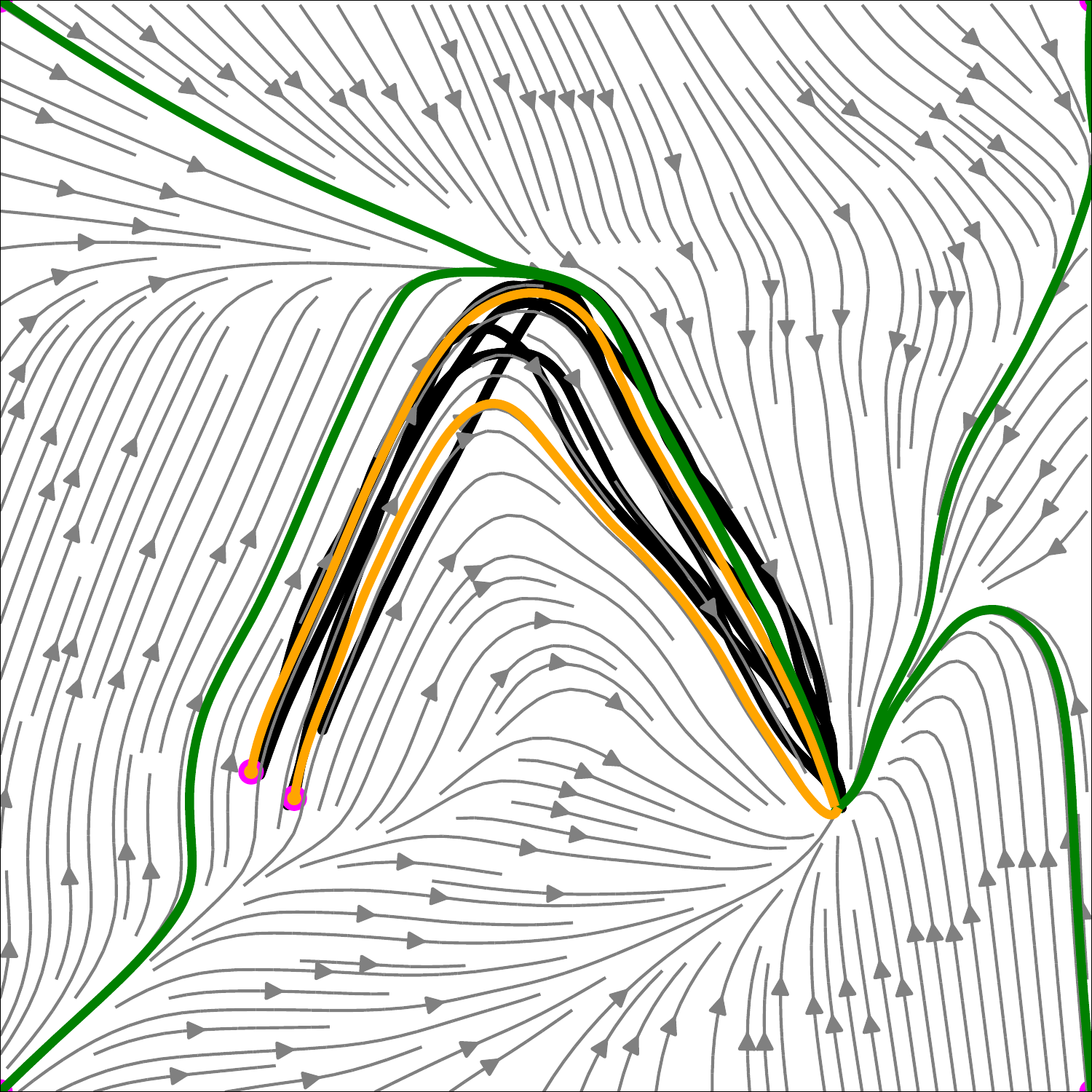}
        \end{subfigure}
        \begin{subfigure}{.24\linewidth}
            \includegraphics[width=1.0\linewidth]{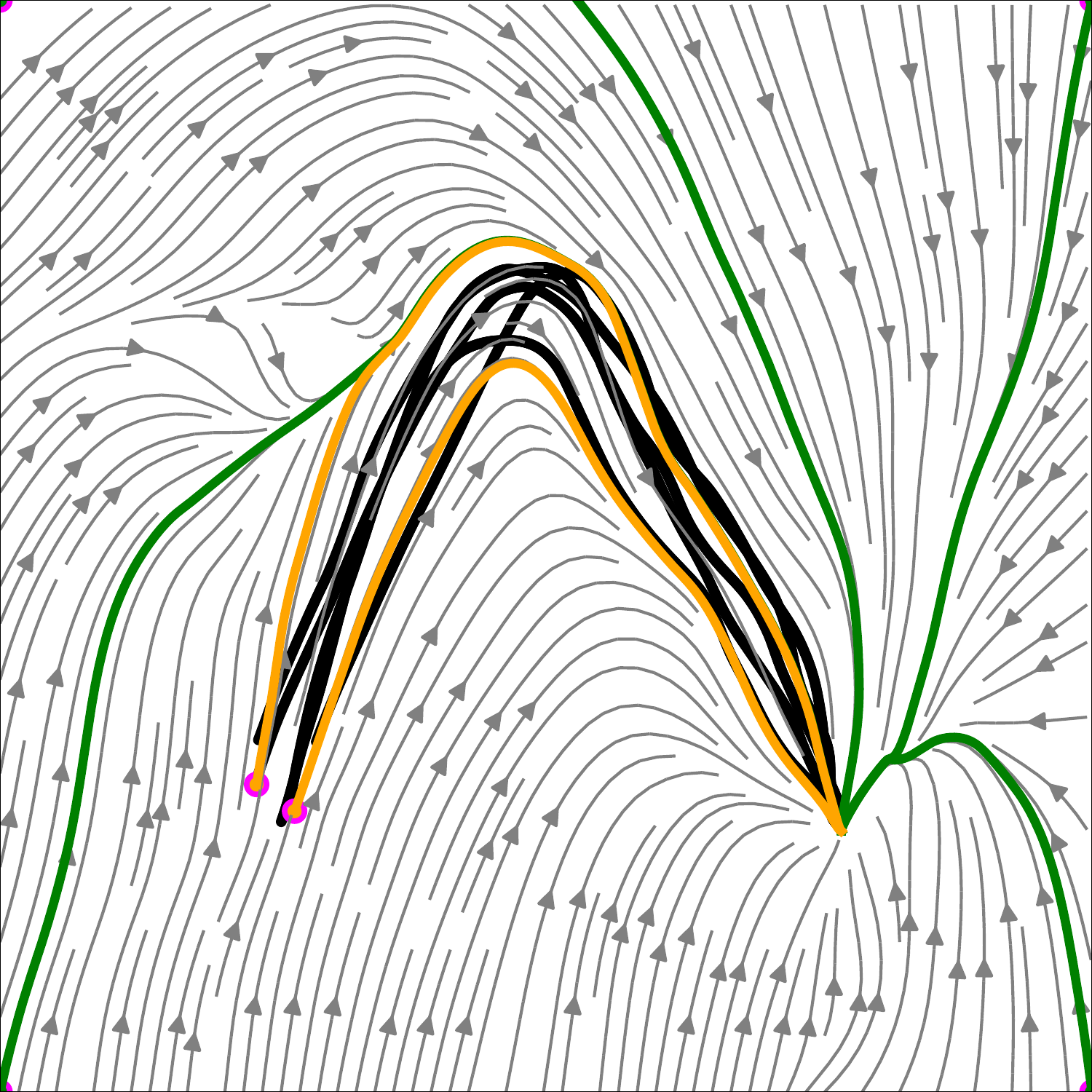}
        \end{subfigure}
        \\
        \begin{subfigure}{.24\linewidth}
            \includegraphics[width=1.0\linewidth]{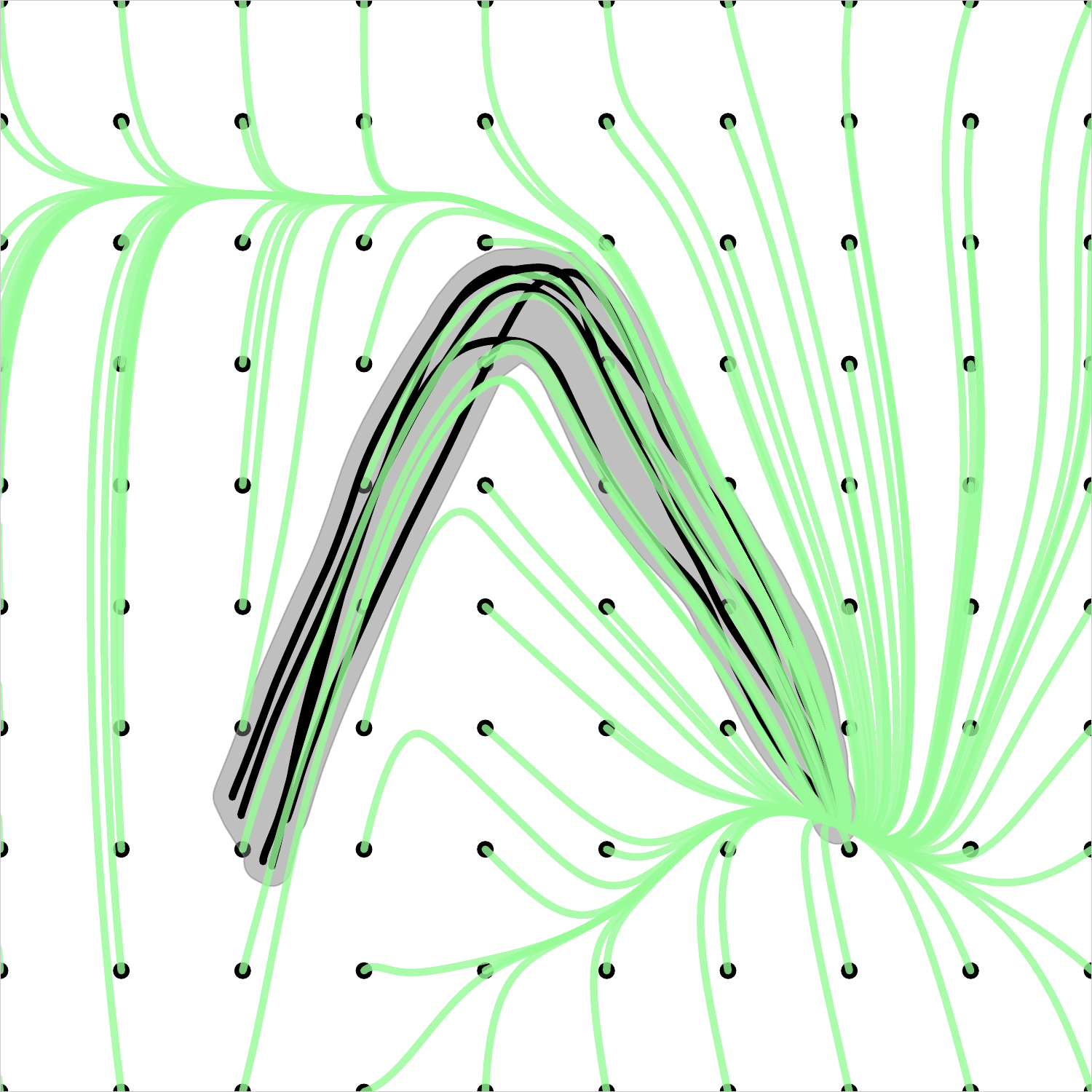}
            \caption{Constant regularization}
        \end{subfigure}
        \begin{subfigure}{.24\linewidth}
            \includegraphics[width=1.0\linewidth]{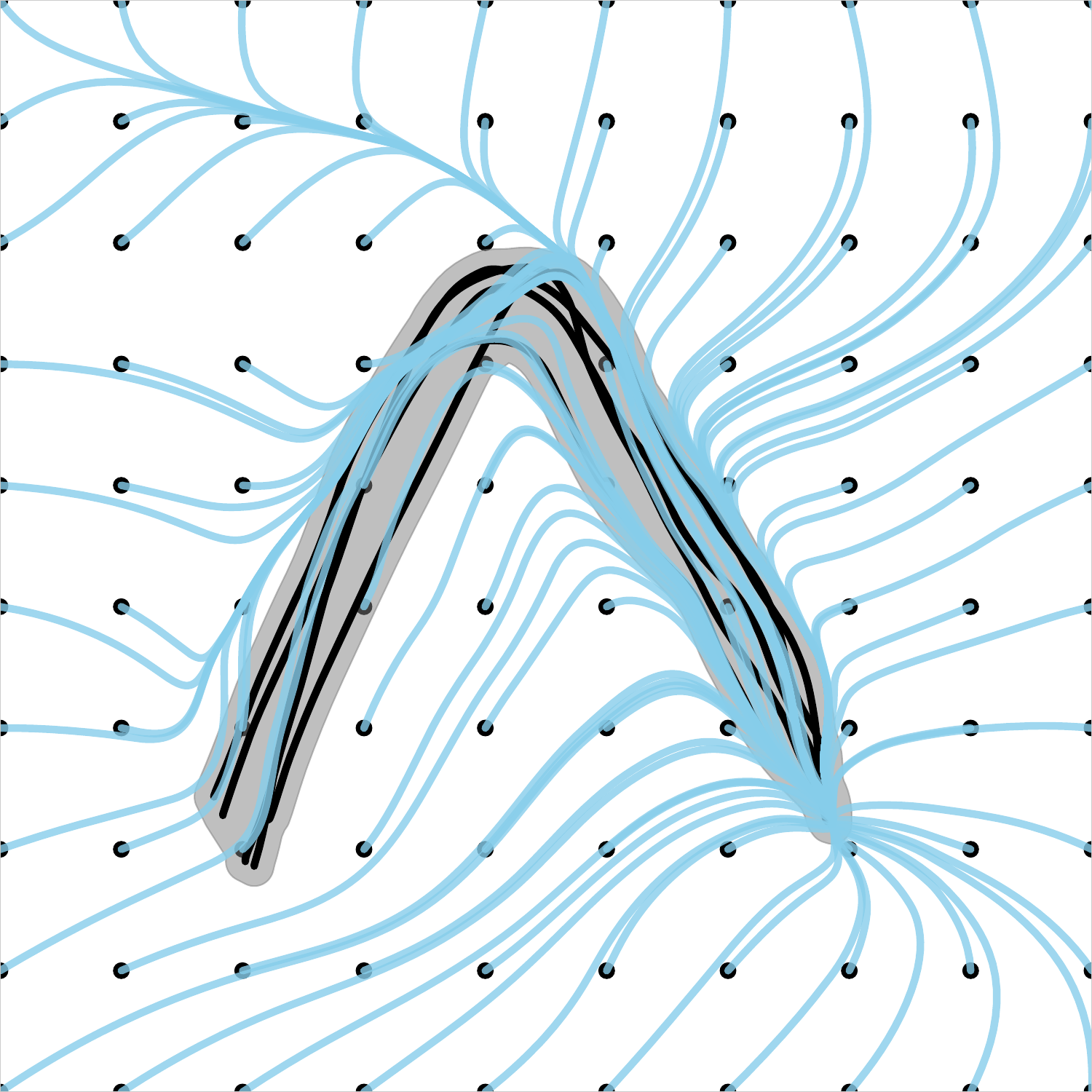}
            \caption{State-independent regularization}
        \end{subfigure}
        \begin{subfigure}{.24\linewidth}
            \includegraphics[width=1.0\linewidth]{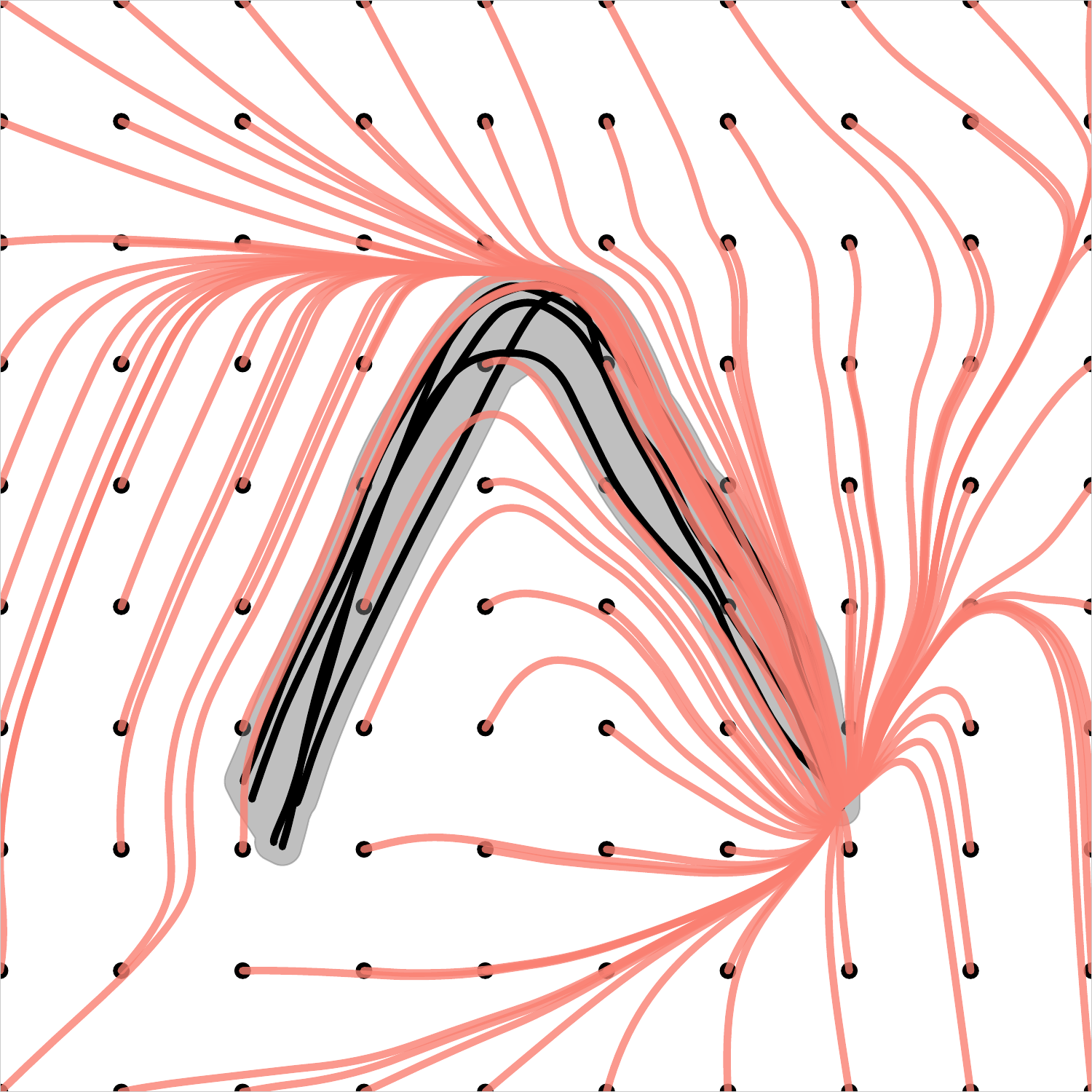}
            \caption{State-dependent regularization}
        \end{subfigure}
        \begin{subfigure}{.24\linewidth}
            \includegraphics[width=1.0\linewidth]{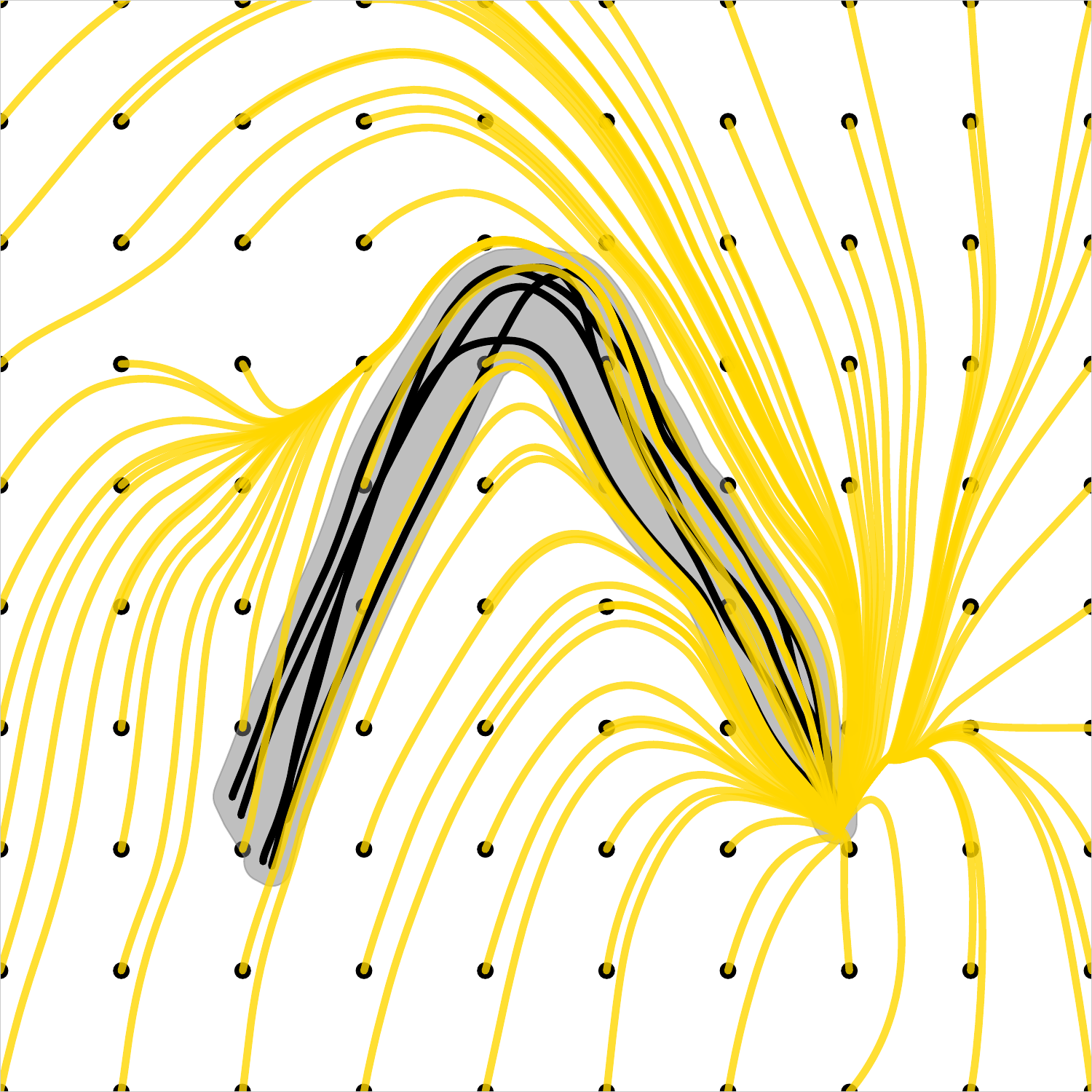}
            \caption{eigenvalue regularization}
        \end{subfigure}
            \caption{Comparison of 2D vector fields learned by using the following regularization approaches: 
            \textbf{(a)} constant regularization,
            \textbf{(b)} state-independent regularization (using basic optimization),
            \textbf{(c)} state-dependent regularization (using a neural network), and
            \textbf{(d)} eigenvalue regularization.
            In the \textit{top row}, the gray background illustrates the learned vector fields, with colored trajectories (integral curves) starting from the initial demonstration point and from the four corners of the figure. In the \textit{second row}, $100$ integral curves are shown, each initiated from an equidistant grid that \textit{encompasses} the demonstration region. This region (shaded gray) is defined by the convex hull around the demonstrations (black curves).
            }
        \label{fig:regularization}
\end{figure*}

Note that the integral in~\eqref{eq:integral} resembles the neural ordinary differential equations \citep{Chen2018:NeuralODE}, with the subtle difference that it is a second-order equation as the outcome pertains to the \emph{velocity} at state $\x$.
We can, thus, view this system as a second-order neural ordinary equation~\citep{norcliffe2021sonode} and solve it using off-the-shelf numerical integrators. 
The resulting function $f_{\bm{\theta}}$ will have a negative definite Jacobian for any choice of $\bm{\theta}$ and is consequently contractive by construction. 
In other words, we can control the extrapolation behavior of the neural network that parameterizes our dynamical system $f$ via the negative-definiteness of $\hat{\Jac_f}(\x)$. 
We call this the \emph{neural contractive dynamical system (NCDS)}, first introduced in our previous work~\citep{NCDS:2023BeikMohammadi}. This approach offers two key benefits: \emph{(1)} it allows the use of any smooth neural network $\Jac_{\bm{\theta}}$ as the base model, and \emph{(2)} training can be performed with ordinary \emph{unconstrained} optimization, unlike previous approaches. \looseness=-1
Figure~\ref{fig:example} shows an example vector field learned by NCDS, which is clearly highly flexible while still providing global contractive stability guarantees. 
With the introduction of NCDS, we propose two ways to improve it by focusing on the formulation of the Jacobian $\hat{\bm{J}}_f(\bm{x})$, as it is the key element that influences the behavior of the system.
The formulation of the Jacobian in~\eqref{eq:Jacobian_simple} is characterized by two main components: the symmetric matrix $\Jac_{\bm{\theta}}(\x)^\trsp \Jac_{\bm{\theta}}(\x)$, and the regularization term $\epsilon\ \I_D$. Each of these components is crucial in determining the behavior of the model on the data support, and more importantly, its ability to generalize outside of it. 
Although NCDS~\citep{NCDS:2023BeikMohammadi} demonstrated significant potential in learning complex contractive dynamical systems, there remains an opportunity to further explore the individual impact of each component. In the following two sections, we will explore the effects of these elements and propose improvements to enhance their performance.
From this point forward, we will refer to the NCDS formulated with equations~\ref{eq:Jacobian_simple} and \ref{eq:integral} as \emph{vanilla NCDS}.

\subsubsection{Regularization:}

The primary objective of the regularization term $\epsilon$ is to ensure that the Jacobian matrix $\hat{\Jac}_f$ associated with NCDS is not semi-definite, thus breaking the contraction guarantees.
Notice that the eigenvalues of the symmetric part of the Jacobian indicate how the system contracts or expands along different eigenvectors. Moreover, the contraction rate of a system is determined by its eigenvalue closest to zero (i.e., the largest eigenvalue as all eigenvalues are negative), representing the minimum rate of convergence that the system exhibits along any direction.
In vanilla NCDS, the regularization method, here referred to as \emph{constant regularization}, adds a small constant to the diagonal terms of the Jacobian $\Jac_{\theta}^\trsp \Jac_{\theta}$. Therefore, the final learned eigenvalues are influenced by both the neural network parameters and the applied regularization. 
Moreover, it indirectly determines the system's contraction rate by providing an upper bound on the eigenvalues.
Note that when these eigenvalues are identical, the system exhibits isotropic contraction, which results in integral curves that are strictly straight as they converge directly to the fixed point, rather than following any curved trajectory (see Fig.~\ref{fig:eigenvalues}--\emph{left}).
Even under the former condition, the system remains contractive. 
Interestingly, by increasing the disparity between the eigenvalues, the system converge more rapidly along certain axes (Fig.~\ref{fig:eigenvalues}--\emph{middle}).

Formally, let $\hat{\Jac_f}$ denote the Jacobian matrix with eigenvalues  $\lambda_1, \lambda_2, \ldots, \lambda_D$. We can obtain different contractive behaviors as a function of $\lambda_i$, as follows:
(1) When $ \lambda_1 = \lambda_2 = \ldots = \lambda_D $, the system contracts uniformly in all directions (see Fig.~\ref{fig:eigenvalues}--\emph{left}); 
(2) When $\lambda_i > \lambda_j$, the system exhibits faster contraction along the direction associated with the larger eigenvalue $\lambda_i$, and slower contraction in the direction corresponding to the smaller eigenvalue $\lambda_j$. As illustrated in Fig.~\ref{fig:eigenvalues}--\emph{middle}, this difference in contraction rates leads to the characteristic curved trajectories of the integral curves, as the system contracts more rapidly along the eigenvector of $\lambda_i$ and more gradually along that of $\lambda_j$. 
% We define the \emph{contraction spread} of the system as the absolute \textcolor{blue}{maximum} difference between \textcolor{blue}{any pair of} eigenvalues \textcolor{blue}{$\max_{n,m} |\lambda_n - \lambda_m|$}, at any point $\x_i$. 

We define the \emph{contraction spread} at a point $\mathbf{x}_i$ as the maximum absolute difference between any pair of eigenvalues of the symmetric part of the Jacobian, that is, $|\lambda_{\max}(\mathbf{x}_i) - \lambda_{\min}(\mathbf{x}_i)|$, where $\lambda_{\max}(\mathbf{x}_i)$ and $\lambda_{\min}(\mathbf{x}_i)$ denote the maximum and minimum eigenvalues at $\mathbf{x}_i$, respectively.

Our goal is not to introduce the contraction spread as a fundamental concept in contraction theory, but rather to use it as a learning bias that encourages stronger contraction in directions (i.e., eigenvectors) not well informed by the data. This promotes stronger contraction, particularly in extrapolative regions, while maintaining a balance between reconstruction accuracy and contraction along data-informed directions.
This contrasts with the conventional \emph{contraction rate}, which only characterizes the slowest contracting eigenvector and fails to provide insight into the overall system's contractive behavior.

\begin{figure*}
    \centering
        \begin{subfigure}{0.54\textwidth}
              \includegraphics[width=1.0\textwidth]{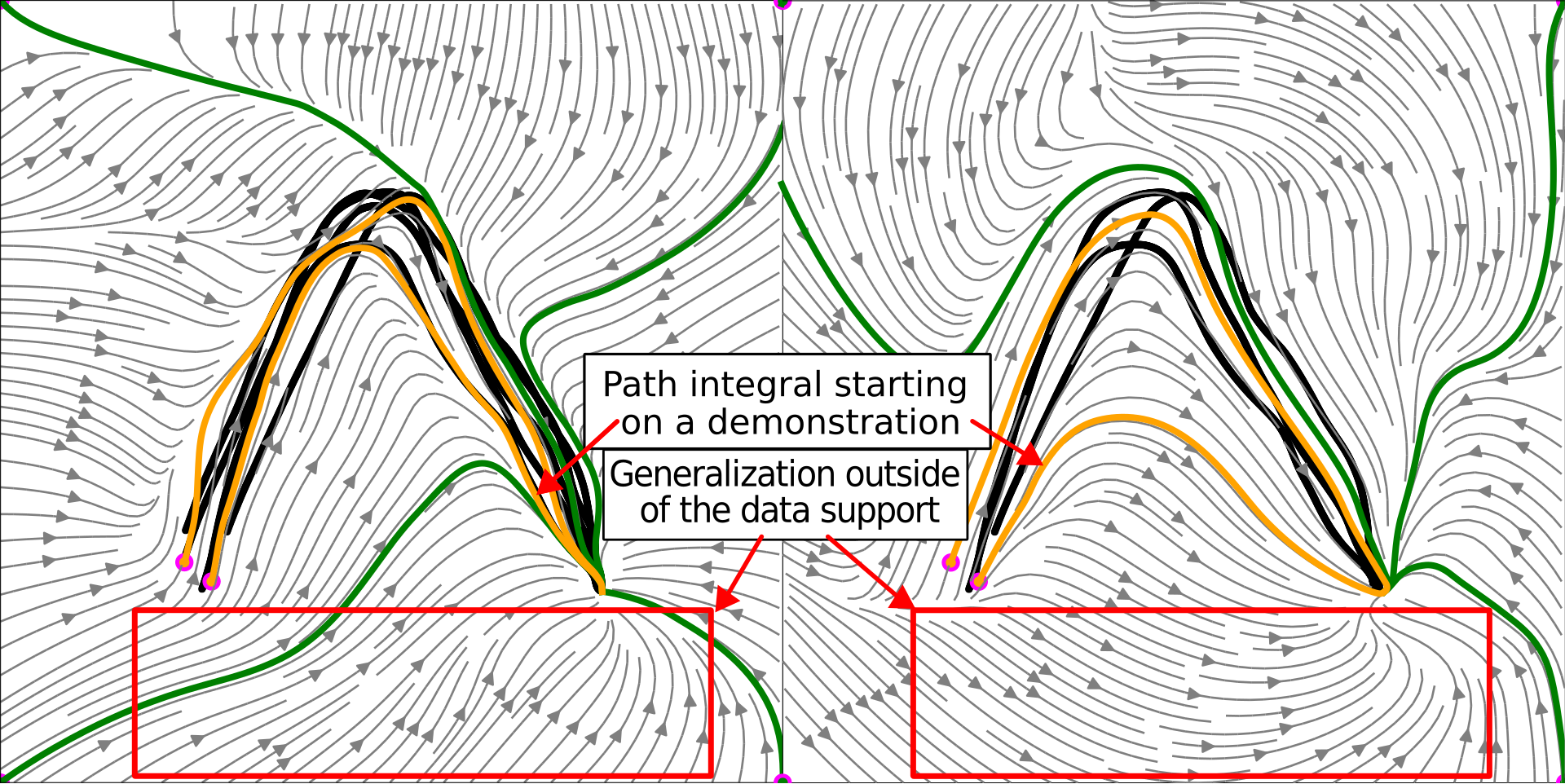}
        \end{subfigure}
        \begin{subfigure}{0.27\textwidth}
            \includegraphics[width=1.0\textwidth]{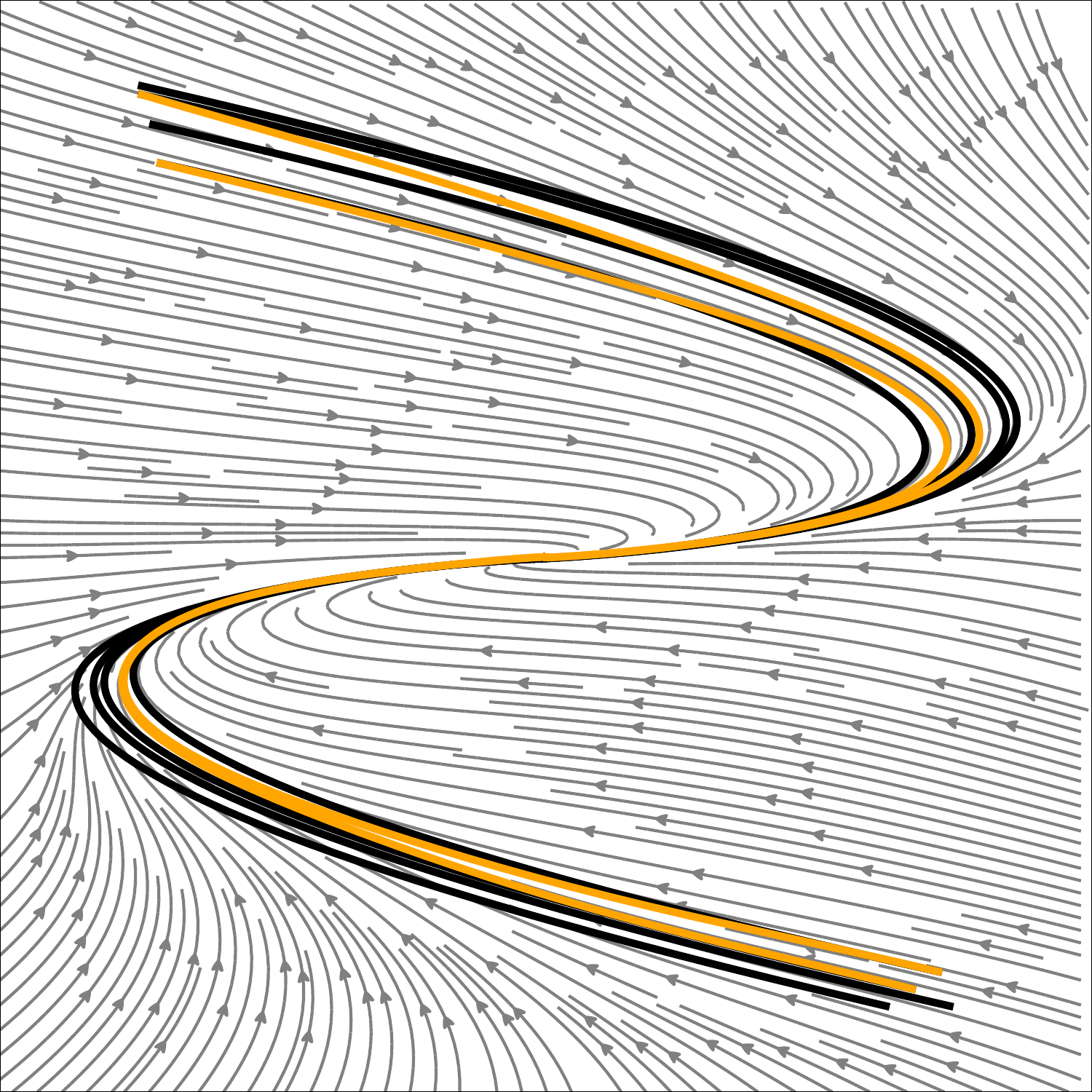}
        \end{subfigure}
        \caption{Comparison of 2D vector fields learned with different Jacobian formulations. \emph{Left:} Vector field obtained using the symmetric Jacobian formulation. \emph{Middle:} Vector field obtained using the asymmetric Jacobian formulation. The learned vector field (grey) and demonstrations (black). Yellow and green trajectories show integral curves starting from demonstration starting points and random points, respectively. \emph{Right:} Vector field with asymmetric Jacobian learned using NCDS with symmetric Jacobian.}
        \label{fig:skewsymmetric}
\end{figure*}

Given this analysis, we propose to better control the system's contraction using the contraction spread and through regularization by considering the following three approaches: \emph{(1) state-independent regularization vector}, \emph{(2) state-dependent regularization vector}, and \emph{(3) eigenvalue regularization}. 
We now proceed to explain these different regularization approaches for NCDS.

\paragraph{Regularization vector:}
Here we treat $\epsilon \in \R^D$ as a vector rather than a constant.
Consequently, the Jacobian is reformulated as follows,
\begin{equation}
  \hat{\Jac_f}(\x) = - (\Jac_{\bm{\theta}}(\x)^\trsp \Jac_{\bm{\theta}}(\x) + \operatorname{diag}(\bm{\epsilon})).
    \label{eq:Jacobian_epsilon_vector}
\end{equation}
We propose two distinct methods to learn the vector $\bm{\epsilon}$.
For a more comprehensive comparison, we employed several metrics, the details of which are provided in Sec.~\ref{sec:metrics}.
\begin{table}[t]
    \centering
    \small % Reduces font size of the entire table
    \renewcommand{\arraystretch}{1.3}
    \setlength{\tabcolsep}{3pt}
    \begin{tabular}{p{3.4cm} cc}
        \rowcolor{gray!15} % Light gray for the header row
        \textbf{Metrics} & \textsf{Contra. rate} & \textsf{Contra. spread (max)} \\
        \rowcolor{gray!5} % Light gray for alternating row
        Constant reg. & $-0.010$ & $1.393$ \\ 
        State-independent reg. & $\mathbf{-0.012}$ & $\mathbf{2.489}$ \\ 
        \rowcolor{gray!5}
        State-dependent reg. & $-10^{-4}$ & $0.796$ \\ 
        eigenvalue reg. & $-10^{-4}$ & $0.801$ \\ 
        \bottomrule
    \end{tabular}
    \caption{The contraction rate and maximum contraction spread computed over the equidistant grids.}
    \label{tab:contraction_reg_metric}
\end{table}
\textbf{State-independent:} 
In this method, $\bm{\epsilon}$ is assumed to be independent of the system's state $\x$. 
The vector $\bm{\epsilon}$ is learned by minimizing the following loss function,
\begin{equation}
\mathcal{L}_{\bm{\epsilon}}  = -\beta \sum_{n=2}^D \left| \epsilon_1 - \epsilon_n \right|^2,
\label{eq:loss_function}
\end{equation}
where $\beta$ is a weight, and $\epsilon_1, \dots, \epsilon_D$ are elements of the regularization vector $\bm{\epsilon}$. Note that, to ensure the strict positive definiteness of $\bm{\epsilon}$, we do not optimize it directly. Instead, we introduce an auxiliary vector $\hat{\bm{\epsilon}}$ with components $ \epsilon_i = \hat{\epsilon}_i^2 + 10^{-10}$, $i=1,\dots,D$. By squaring $\hat{\epsilon}_i$ and adding this very small constant, we guarantee that every $\epsilon_i$ remains strictly positive. 

\begin{figure}
\centering
  \includegraphics[width=1.0\linewidth]{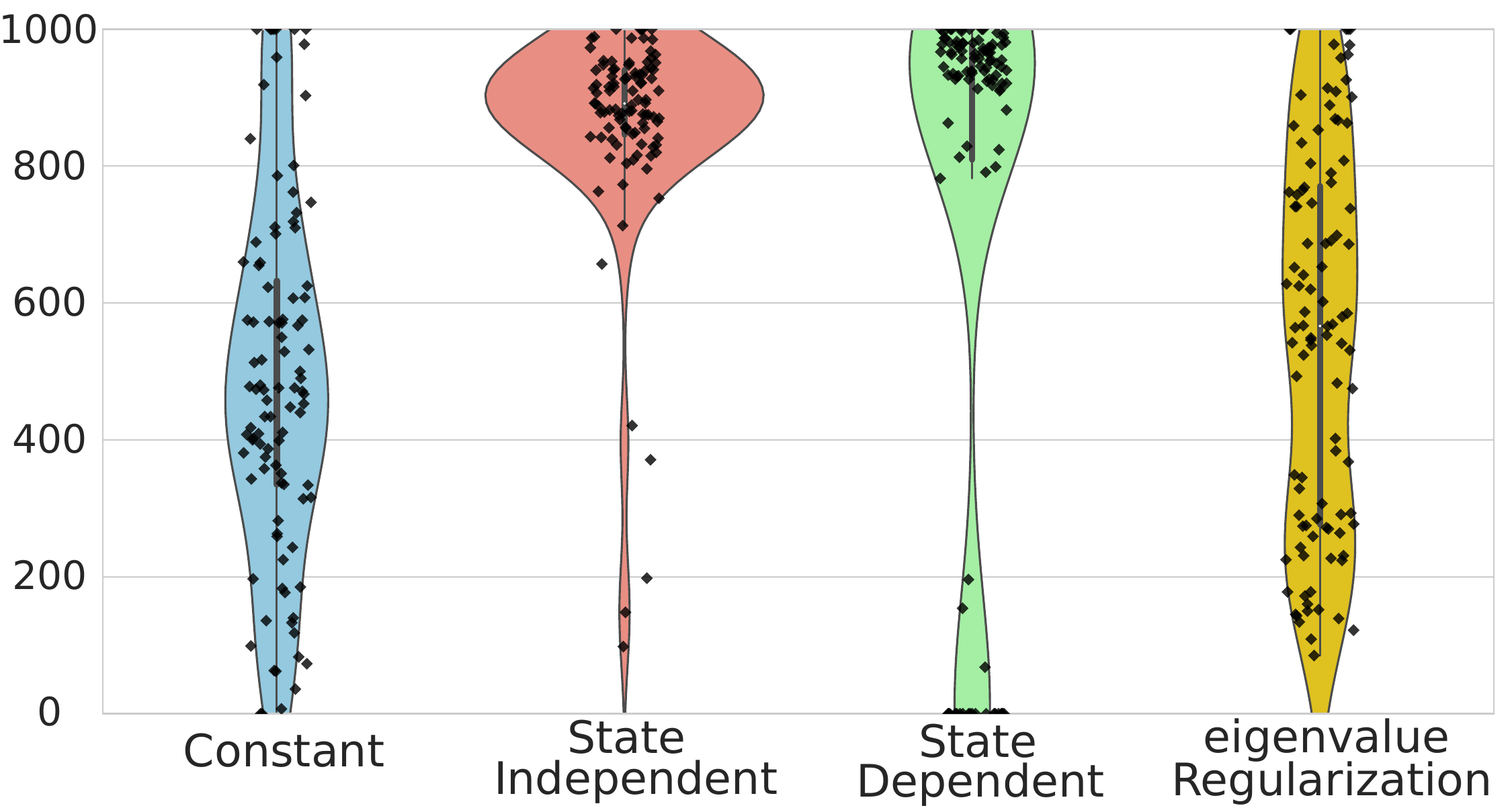}
  \caption{The distribution of the duration, in time steps, that each integral curve remained within the demonstration region across four different regularization methods: \emph{Constant}, \emph{State Independent}, \emph{State Dependent}, and \emph{Eigenvalue Regularization}. A higher count of time steps inside the region indicates that trajectories likely converge faster and exhibit stronger contraction properties. Individual data points are represented by black dots.}
  \label{fig:regularization_comparison_violin}
\end{figure}

Thus, the overall loss is defined as $\mathcal{L}_{\text{total}} = \mathcal{L}_{\text{vel}} + \mathcal{L}_{\bm{\epsilon}}$.
Figure~\ref{fig:regularization}--\emph{b} shows a contractive dynamical system employing this method. 
The integral curves suggest that the system exhibits higher contraction when compared to the naive constant regularization shown in Fig.~\ref{fig:regularization}--\emph{a}.
This is quantitatively supported by the contraction measure statistics reported in Fig.~\ref{fig:regularization_comparison_violin}, showing that the trajectories converge approximately $276\%$ faster towards the data support. 
The reported $276\%$ faster convergence is based on the number of time steps each integral curve spends inside the demonstration region defined by the convex hull $\mathcal{H}$. This measure reflects how quickly trajectories enter and remain within the data support in comparison to vanilla NCDS. 

In conclusion, the loss function enforces the contraction spread, and Table \ref{tab:contraction_reg_metric} shows that the proposed method achieves a higher spread than vanilla NCDS. Although the loss function does not explicitly enforce a higher contraction rate, the method still achieves a higher rate.

\textbf{State-dependent:} In this approach, the regularization vector $\bm{\epsilon} = g(\x)$ is computed as a function of the state $\x$ using a neural network $g_{\bm{\delta}}(\x): \mathbb{R}^D \rightarrow \mathbb{R}^D$ with parameters $\bm{\delta}$. To ensure that $\bm{\epsilon}$ remains strictly positive, we reparameterize it as $\epsilon_i = \hat{\epsilon}_i^2 + 10^{-10}$. Thus, the neural network $g_{\bm{\delta}}$ directly outputs the auxiliary vector $\hat{\epsilon}$, ensuring that the computed $\bm{\epsilon}$ is always positive.
The parameters $\bm{\delta}$ are learned via backpropagation, which aims at minimizing the loss introduced in~\eqref{eq:loss_function}. 
Figure~\ref{fig:regularization}--\emph{c} shows a contractive dynamical system using this state-dependent method, exhibiting faster trajectory convergence compared to the naive constant regularization (Fig.~\ref{fig:regularization}--\emph{a}).
Figure~\ref{fig:regularization_comparison_violin} shows that trajectories converge approximately $100\%$ faster towards the data support. However, as indicated in Table~\ref{tab:contraction_reg_metric}, this method exhibits a lower contraction rate and spread compared to vanilla NCDS.
\begin{figure*}[t!]
  \centering
      \begin{subfigure}{.25\linewidth}
            \centering
            \includegraphics[width=1.\linewidth]{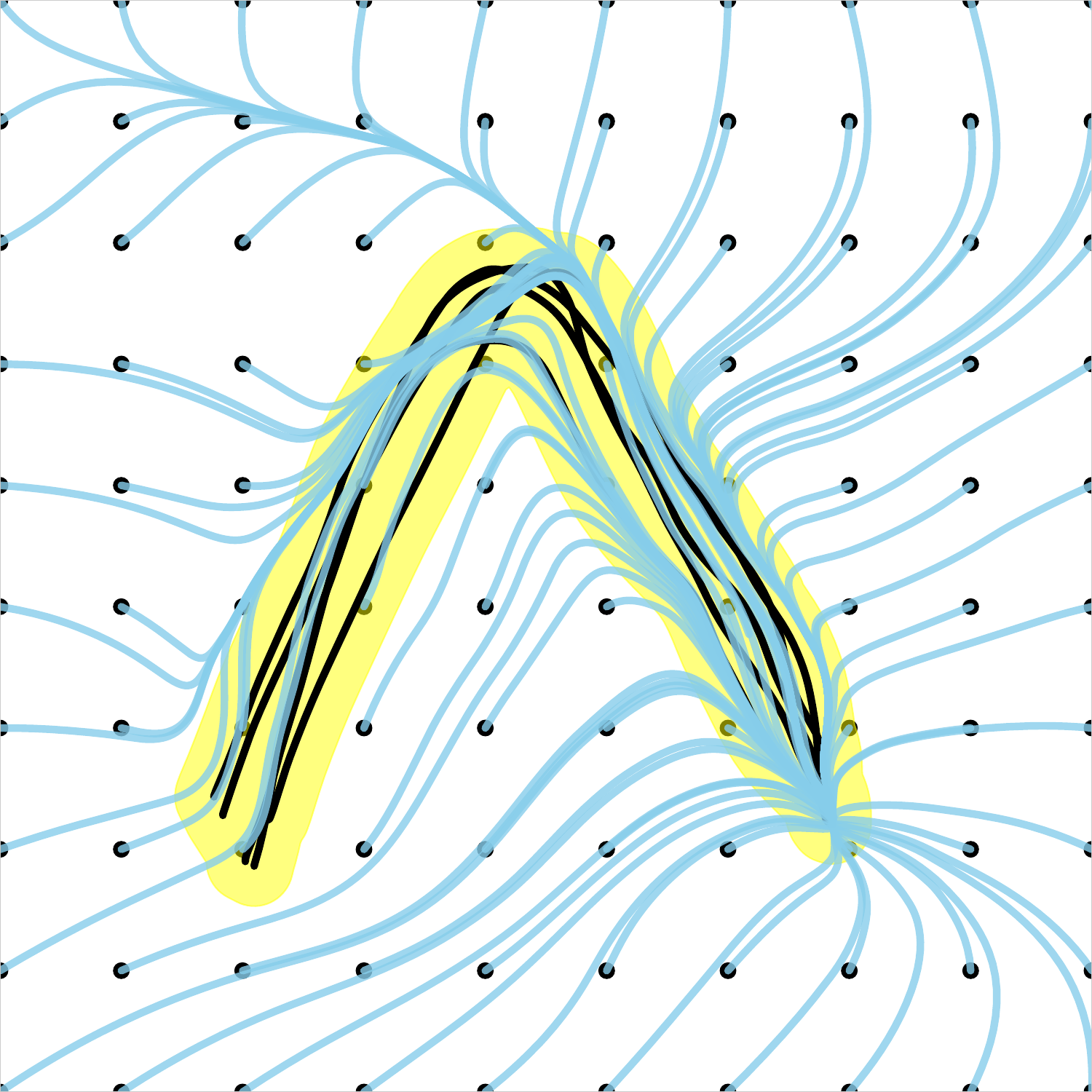}
      \end{subfigure}
        \begin{subfigure}{.25\linewidth}
            \centering
            \includegraphics[width=1.0\linewidth]{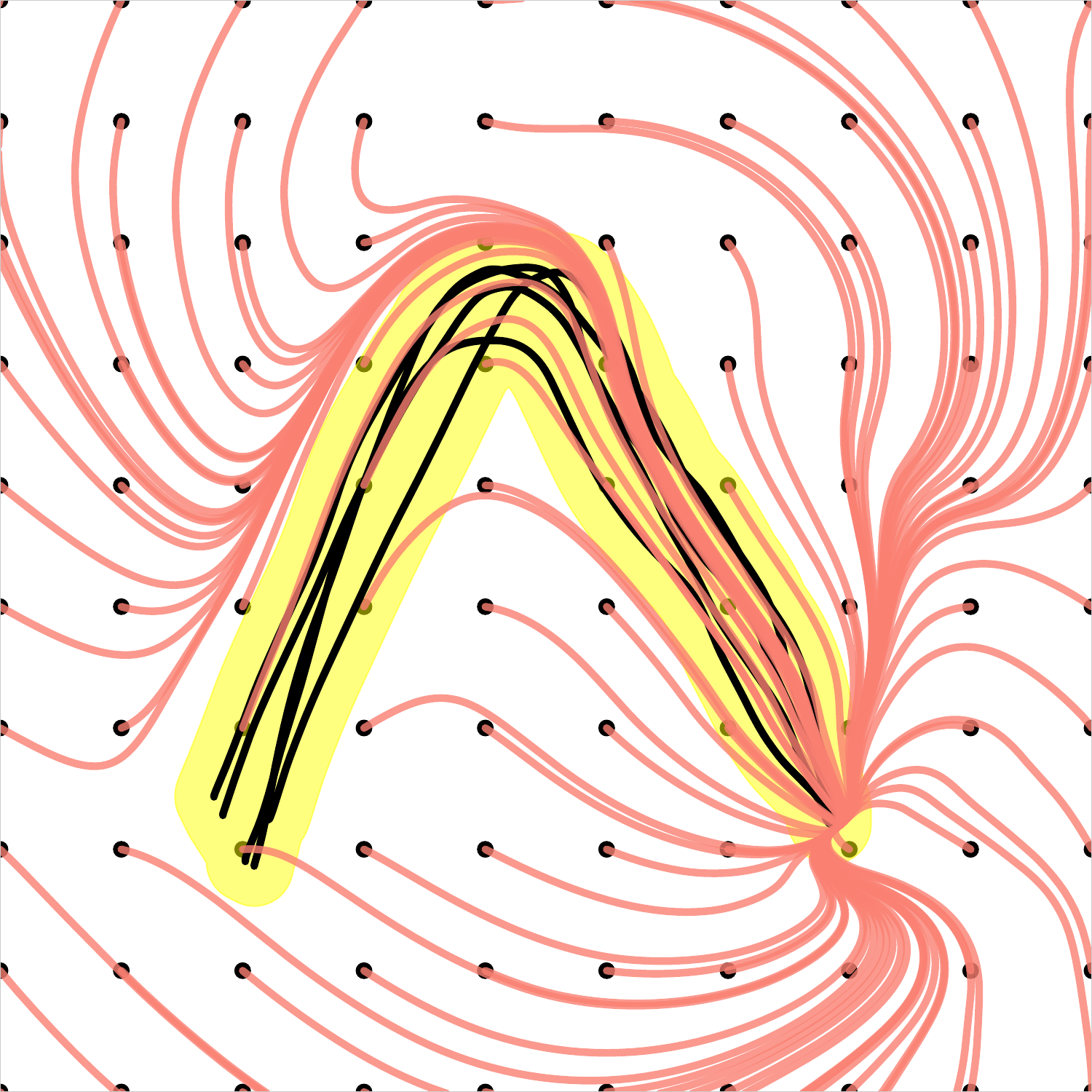}
      \end{subfigure}
          \begin{subfigure}{.257\linewidth}
        \centering
        \includegraphics[width=1.0\linewidth]{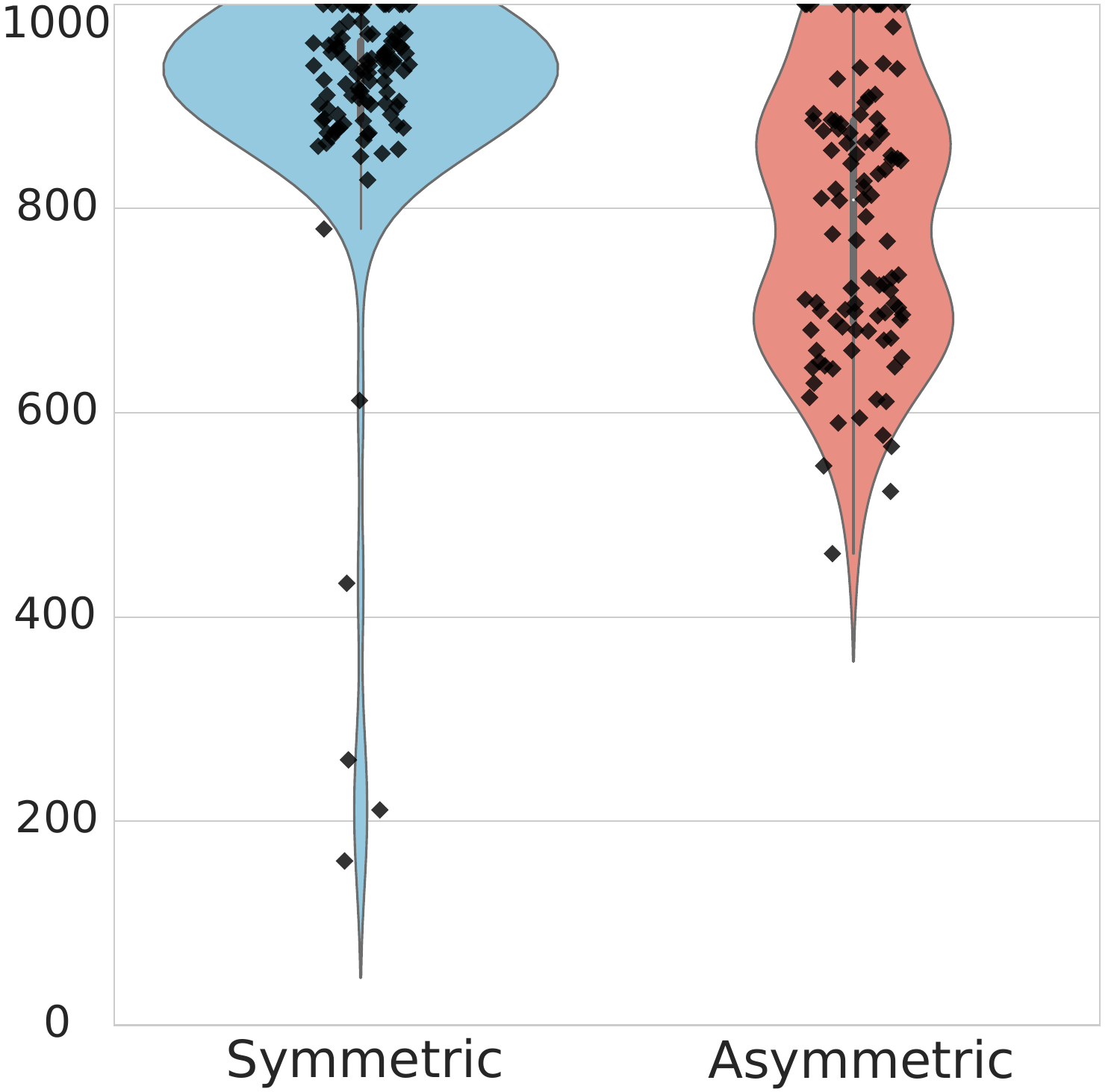}
      \end{subfigure}
      \\
            \begin{subfigure}{.25\linewidth}
            \centering
            \includegraphics[width=1.\linewidth]{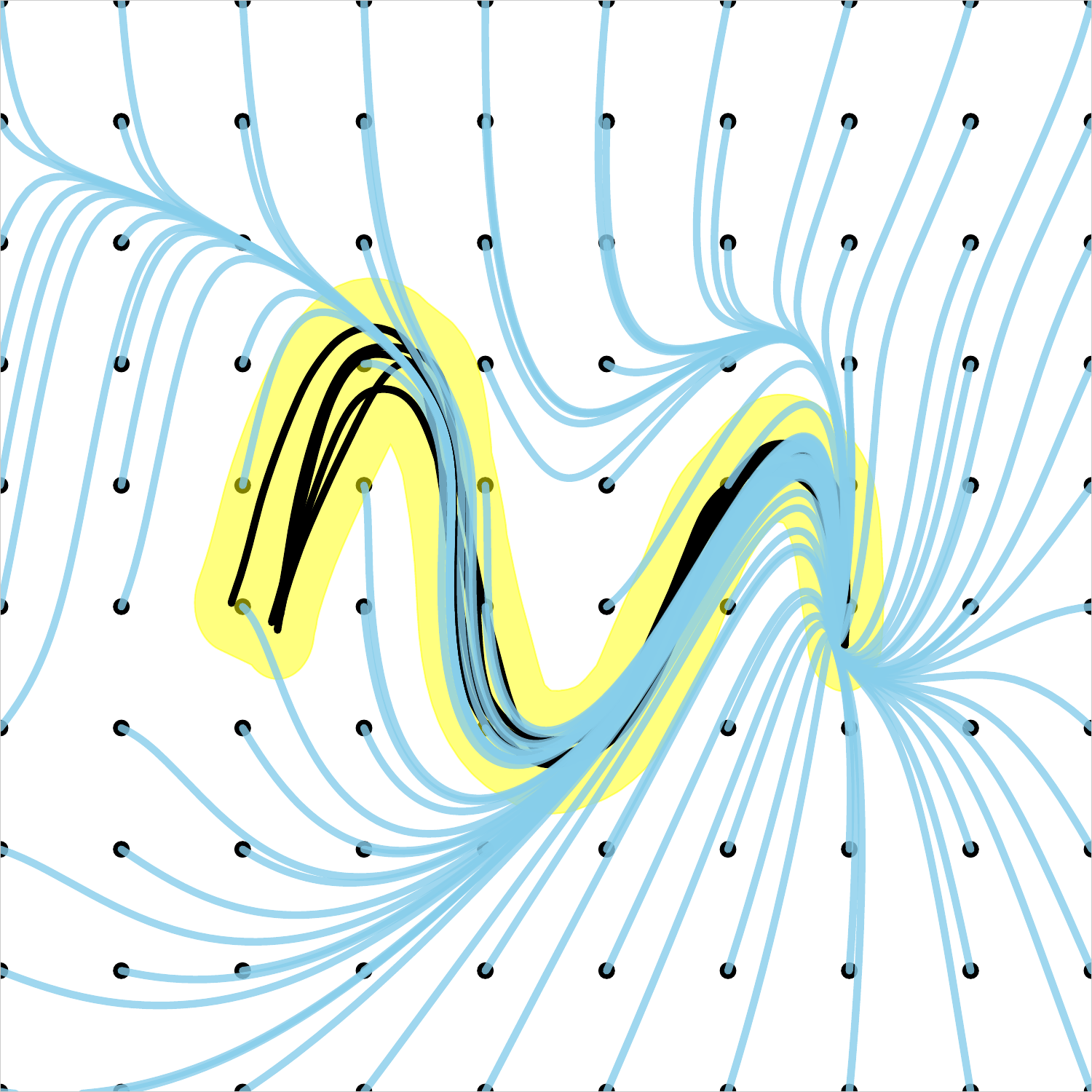}
      \end{subfigure}
        \begin{subfigure}{.25\linewidth}
            \centering
            \includegraphics[width=1.0\linewidth]{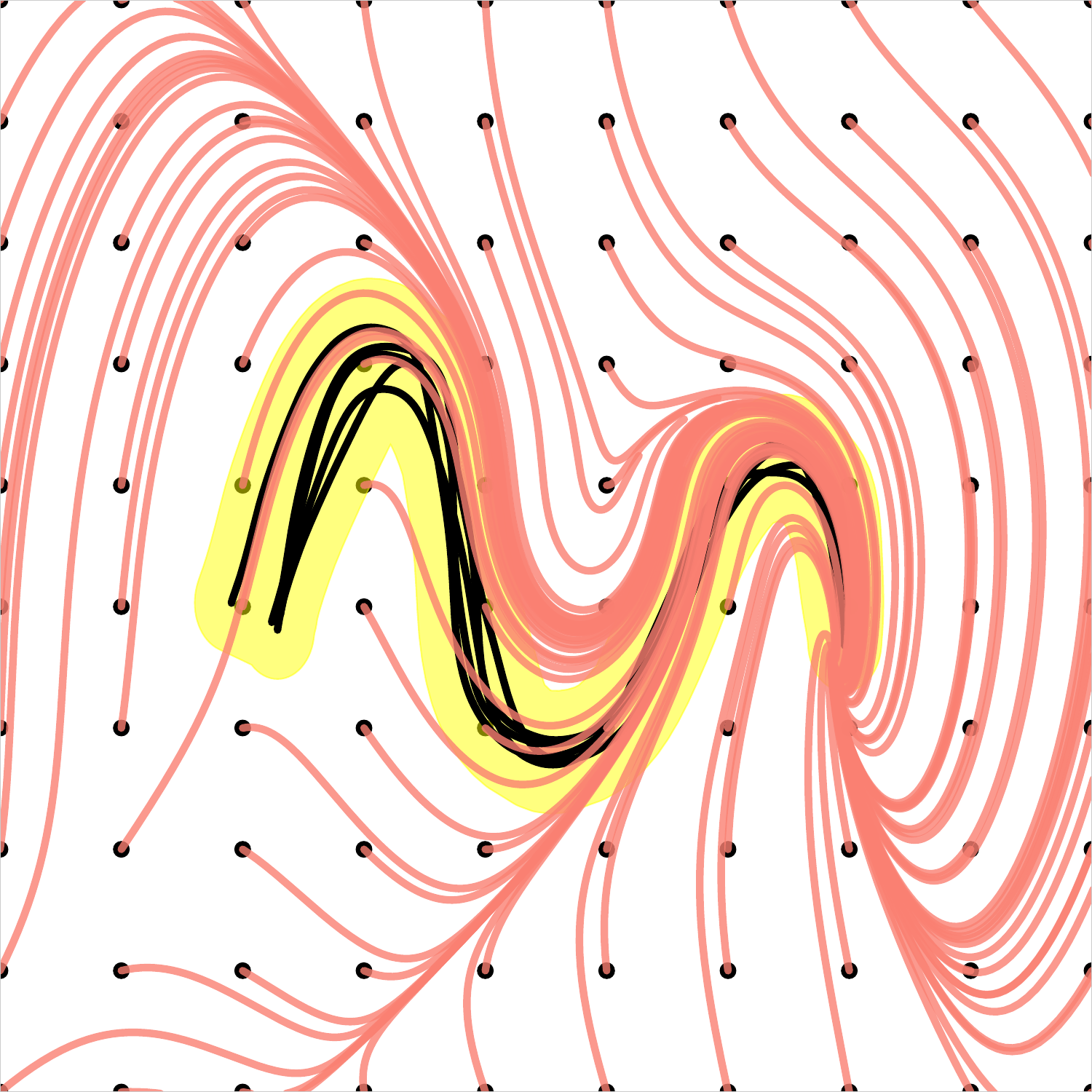}
      \end{subfigure}
          \begin{subfigure}{.257\linewidth}
        \centering
        \includegraphics[width=1.0\linewidth]{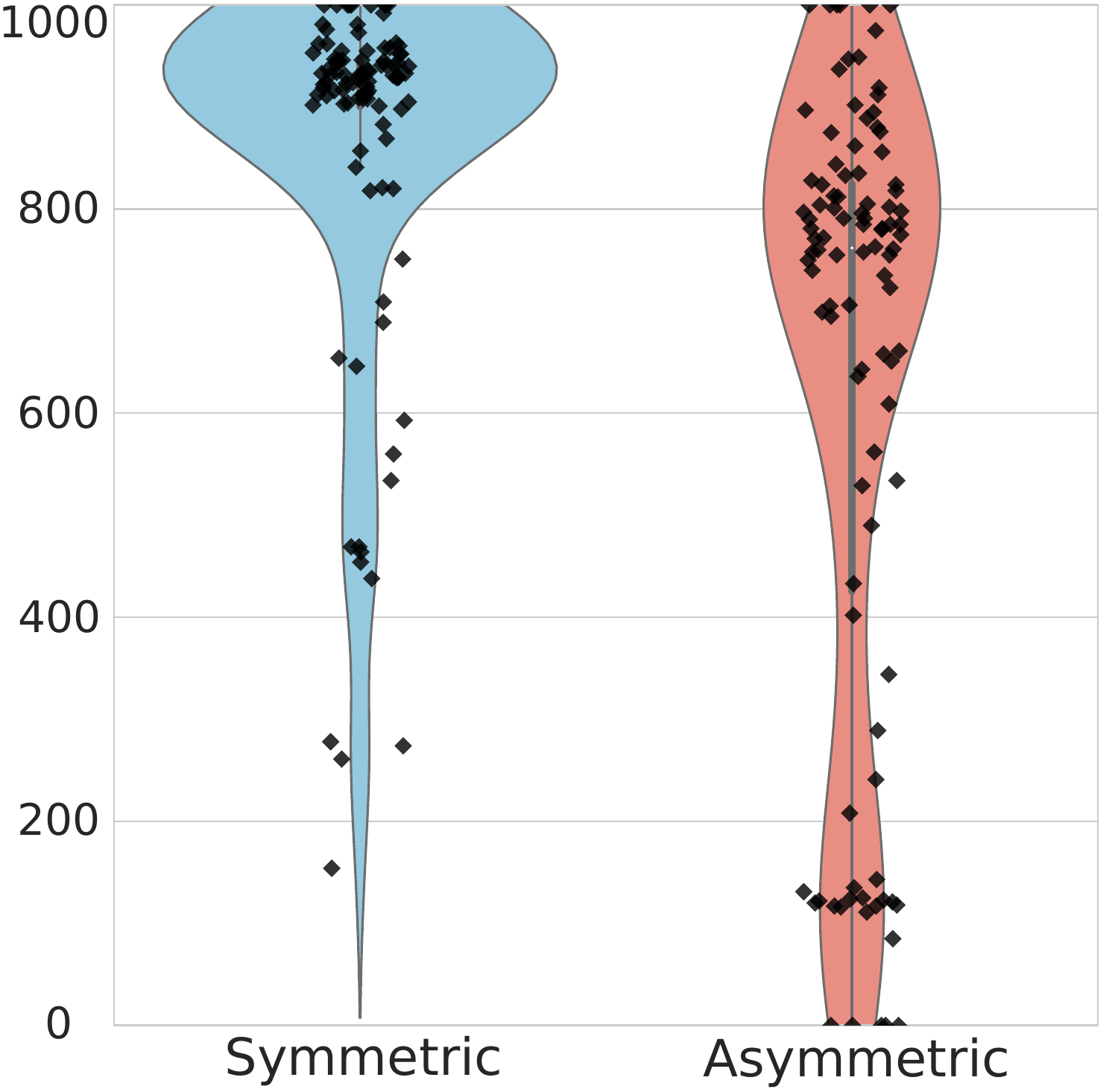}
      \end{subfigure}
      \caption{Comparison between asymmetric and symmetric Jacobians in learning contractive dynamical systems. \emph{Left:} Generalization behavior of a system with a symmetric Jacobian, visualized using integral curves originating from a $10 \times 10$ equidistant grid. The yellow region represents the demonstration area, with black curves denoting the demonstrations. Black dots indicate the initial points of the integral curves, while blue curves illustrate the resulting trajectories. \emph{Middle:} Generalization behavior of a system with an asymmetric Jacobian. Pink curves represent the integral curves. \emph{Right:} Comparison of the number of frames each integral curve spent within the demonstration region. A higher number indicates an integral curve more likely converged to the demonstrations, therefore, better contractive behavior.}
        \label{fig:symmetry_assymetry}
\end{figure*}

\textbf{Eigenvalue regularization:}
Consider the Jacobian matrix $\hat{\Jac_f}$ associated with the dynamical system $f$, whose eigenvalue decomposition is given by $\hat{\Jac_f} = \bm{V} \bm{\Lambda} \bm{V}^{-1}$,
where $\bm{V}$ is the matrix of eigenvectors of $\hat{\Jac_f}$, and $\bm{\Lambda}$ is the diagonal matrix of eigenvalues, $\bm{\Lambda} = \operatorname{diag}(\lambda_1, \lambda_2, \ldots, \lambda_D)$.
To incorporate regularization directly on the eigenvalues, we can use the same loss as in~\eqref{eq:loss_function}, with the difference that we manipulate the eigenvalues directly. 
For example, a simple loss encouraging the deviation of eigenvalues from a chosen eigenvalue $\lambda_{\text{i}}$ is,
\begin{equation}
\mathcal{L}_{\bm{\epsilon}} =  -\beta \sum_{n=1}^D \left| \lambda_i - \lambda_n \right|^2 ,
\label{eq:loss_function_eigen_decomposition}
\end{equation}
where $\beta$ is a weight, and the final loss includes both the velocity reconstruction loss and the regularization loss.
Note that the main limitation of this approach is its computational cost due to the need to backpropagate through the eigenvalue decomposition. 
Figure~\ref{fig:regularization}--{d} illustrates the impact of this method on the system's contraction, showing relatively modest gains on the convergence of trajectories compared to the naive constant regularization approach depicted in Fig.~\ref{fig:regularization}--{a}. This is evident from the contraction measure statistics reported in Fig.~\ref{fig:regularization_comparison_violin}, showing that trajectories converge only $13\%$ faster towards the data support. 

%
% \textbf{In conclusion}, as regularization techniques become more complex, predicting their impact on the system's ability to generalize beyond the training data becomes increasingly challenging. 
% %
% The results in Fig.~\ref{fig:regularization} suggest that the \emph{vector-value state-independent} method provides a better contractive behavior toward the demonstration region and generalization outside of the data support. Also, the computational cost of this approach is lower than the alternatives as it does not rely on a neural network or an eigen-decomposition operation.  

\textbf{In conclusion}, as regularization techniques become more complex, predicting their impact on the system's ability to generalize beyond the training data becomes increasingly challenging. 
Moreover, while the regularization spreads the eigenvalues, its practical benefit lies in promoting stronger contraction along directions that are not well informed by the demonstrations, while also enhancing contraction in directions supported by the data. This leads to improved contractive behavior in extrapolative regions and improved robustness within the data support.
The results in Fig.~\ref{fig:regularization} suggest that the vector-value state-independent method provides a better contractive behavior toward the demonstration region and generalization outside of the data support. Also, the computational cost of this approach is lower than the alternatives as it does not rely on a neural network or an eigen-decomposition operation. Furthermore, Table~\ref{tab:contraction_reg_metric} reports a comparison of the contraction rate and maximum contraction spread, highlighting the superior performance of the state-independent method.
Moreover, while we found that the contraction spread improves the contractive behavior of the system, this does not directly relate to the  contraction rate. Therefore, any observed improvement should not be directly attributed to the regularization loss term $\mathcal{L}_{\bm{\epsilon}}$.
More details on the evaluation of these eigenvalue metrics can be found in App.~\ref{app:reg_maps}. 

Notably, across every dataset and under the considered metrics, the state-independent method consistently outperformed the others. In our experiments, trajectories converged approximately $276\%$ faster to the demonstration region, highlighting the improved performance in convergence and generalization. Based on these findings, we will henceforth adopt this regularization strategy as our default configuration for further analyses.

Next, we turn our attention to the other key component of~\eqref{eq:Jacobian_simple}, namely the symmetric matrix $\Jac_{\bm{\theta}}(\x)^\trsp \Jac_{\bm{\theta}}(\x)$. Our objective is to explore the effects of modifying the Jacobian from its symmetric form to an asymmetric one, with a particular focus on understanding how the introduction of a skew-symmetric component impacts the generalization behavior of the contractive dynamical system.
\begin{figure*}[t!]
  \centering
  \begin{subfigure}{.24\linewidth}
        \includegraphics[width=1.0\linewidth]{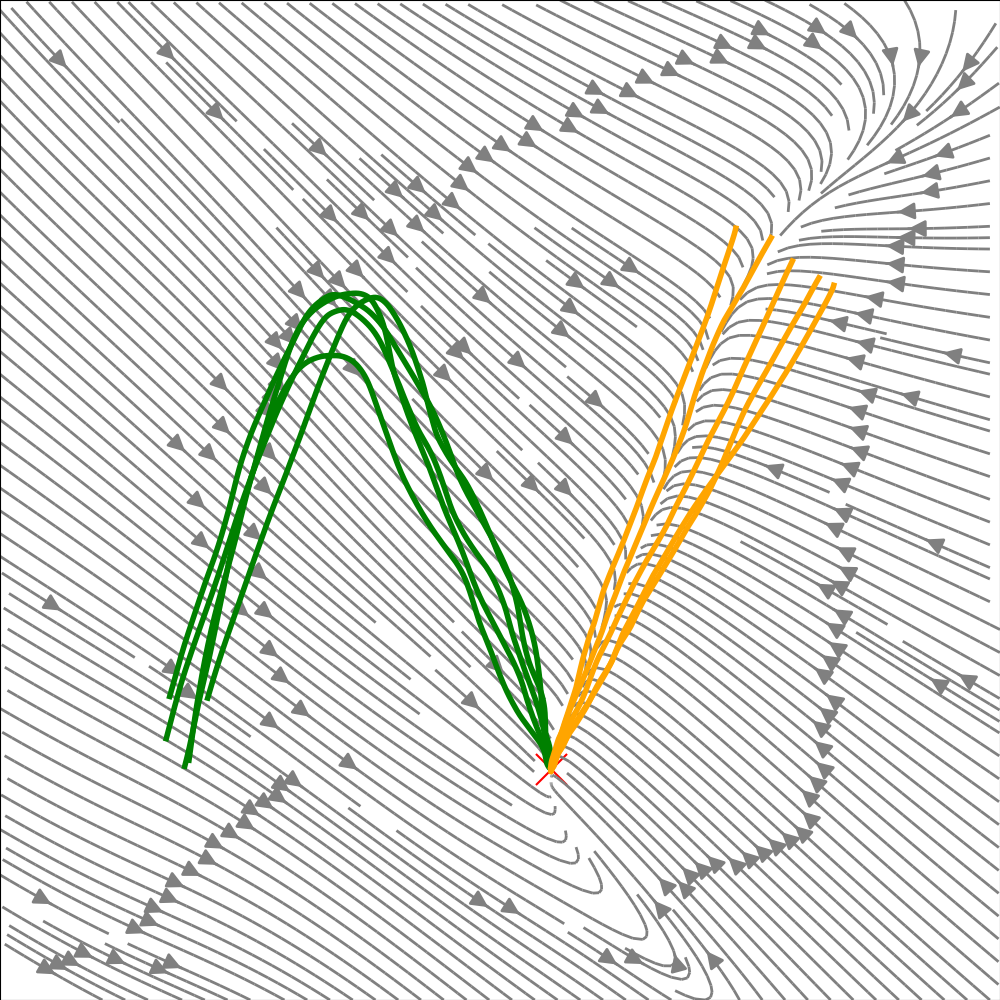}
        \caption{Reproduce \emph{angle} trajectory}
  \end{subfigure}
    \begin{subfigure}{.24\linewidth}
        \includegraphics[width=1.0\linewidth]{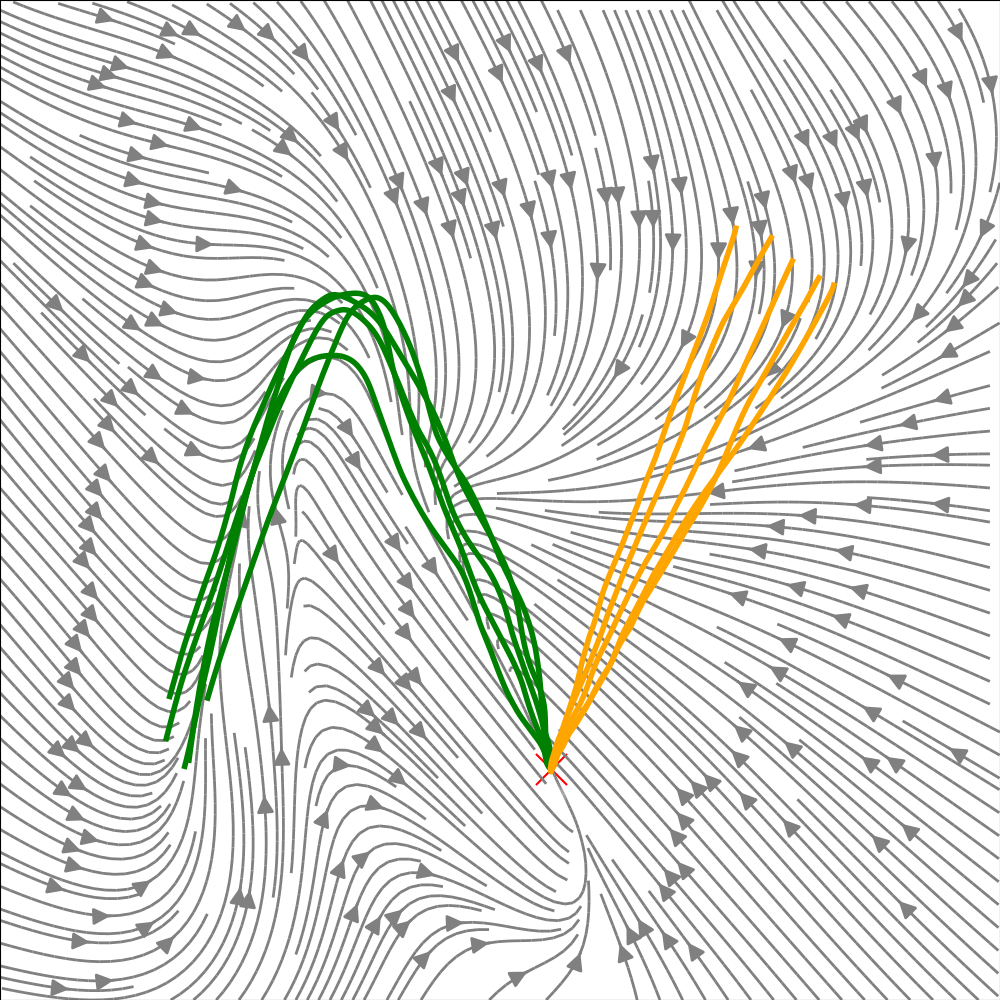}
        \caption{Reproduce \emph{line} trajectory}
  \end{subfigure}
    \begin{subfigure}{.24\linewidth}
        \includegraphics[width=1.0\linewidth]{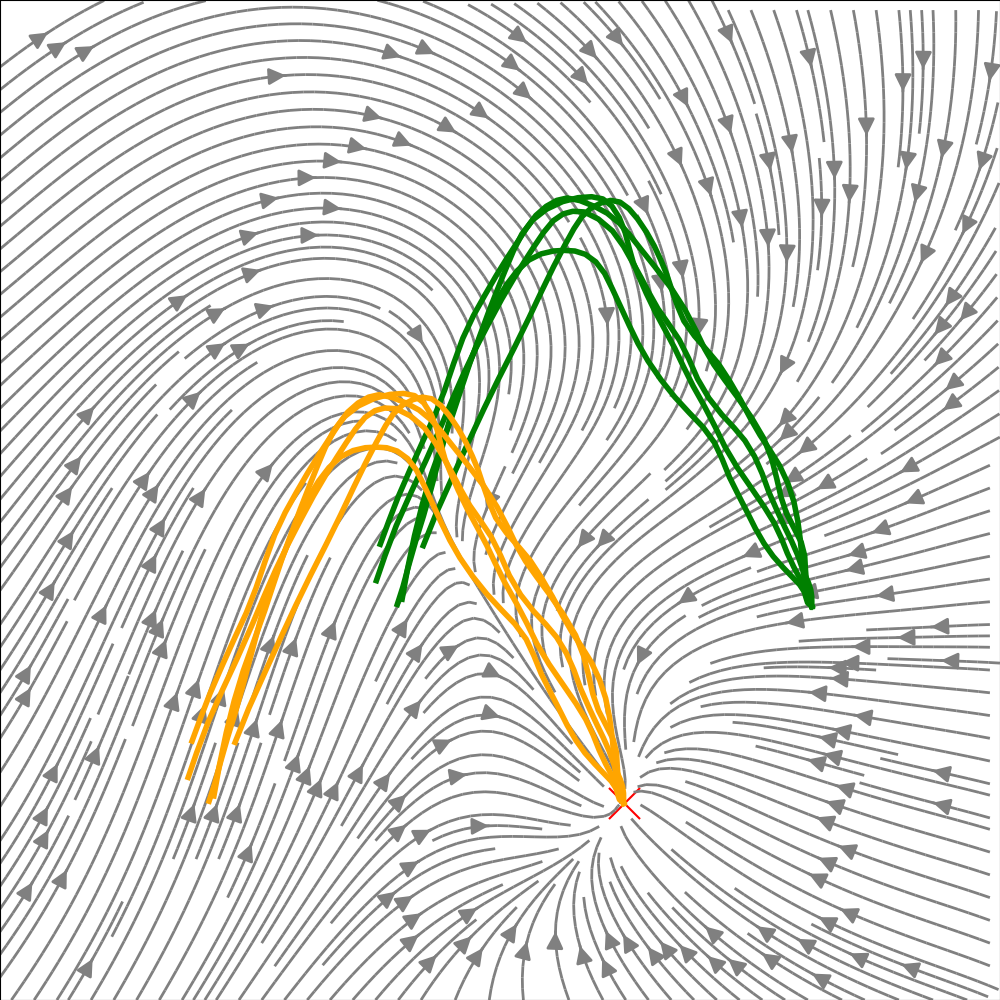}
        \caption{Reach point A}
  \end{subfigure}
    \begin{subfigure}{.24\linewidth}
        \includegraphics[width=1.0\linewidth]{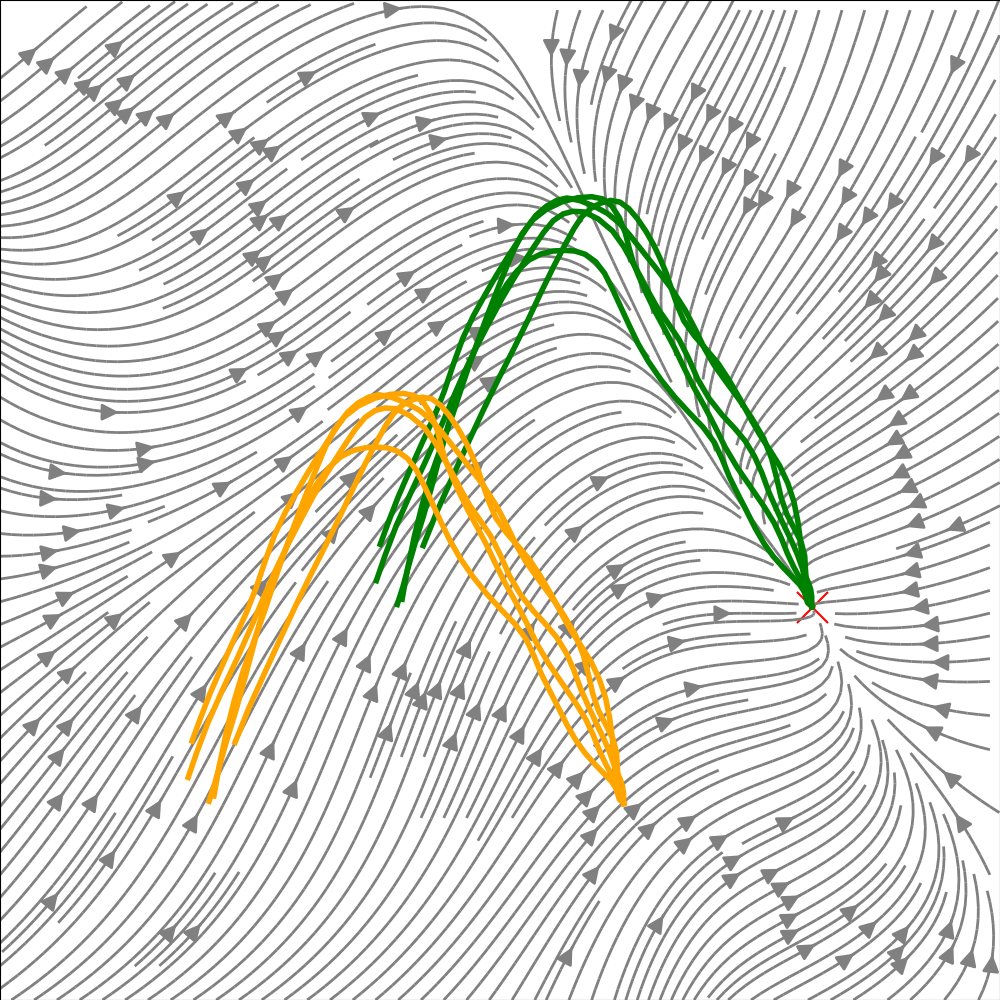}
        \caption{Reach point B}
  \end{subfigure}
  \caption{CNCDS on 2D trajectories from the LASA handwriting dataset. Plots (a) and (b) show CNCDS conditioned on the trajectory shape. Plots (c) and (d) display conditioning on the trajectory target, where the conditional value is chosen to be $0$ and $1$, respectively. }
    \label{fig:conditional_toy_example}
\end{figure*}
\subsubsection{Asymmetry of the Jacobian}:
Vanilla NCDS builds on a symmetric Jacobian, as formulated in~\eqref{eq:Jacobian_simple}. 
However, it is important to note that for a dynamical system to be contractive, its Jacobian does not need to be symmetric~\citep{jaffe2024:ELCD}. 
As \cite{jaffe2024:ELCD} discuss, a simple form of a contractive vector field with an asymmetric Jacobian can be expressed as the following linear system: 
$\dot{\bm{x}} = \bm{A}\bm{x}$ with $\bm{A} = \begin{pmatrix} -1 & 4 \\ 0 & -1 \end{pmatrix}$.
The vector field produced by this system is shown in Fig.~\ref{fig:eigenvalues}--\emph{right}.
\begin{table}[h!]
    \centering
    \small % Reduces font size of the entire table
    \renewcommand{\arraystretch}{1.3}
    \setlength{\tabcolsep}{3pt}
    \begin{tabular}{p{2.8cm} cc}
        \rowcolor{gray!15} % Light gray for the header row
        \textbf{Metrics} & \textsf{Contraction rate} & \textsf{Contraction spread (max)} \\
        \rowcolor{gray!5} 
        Angle (Sym.) & $\mathbf{-0.034}$ & $\mathbf{2.489}$ \\ 
        \rowcolor{gray!5}
        Angle (Asym.) & $-0.006$ & $0.362$ \\ 
        Sine (Sym.) & $\mathbf{-0.042}$ & $\mathbf{1.457}$ \\ 
        Sine (Asym.) & $-0.003$ & $0.439$ \\ 
        \bottomrule
    \end{tabular}
    \caption{The contraction rate and maximum contraction spread computed over the equidistant grids.}
    \label{tab:contraction_asy_metric}
\end{table}
Therefore, we can enhance the NCDS flexibility by reformulating the system's Jacobian to incorporate both a symmetric and a skew-symmetric component. 
This can be achieved by parameterizing the Jacobian using two separate neural networks: one generating the symmetric Jacobian, denoted as $\Jac_{\bm{\bm{\theta}}}$, and the other generating the skew-symmetric Jacobian, denoted as $\Jac_{\bm{\phi}}$.
The combined Jacobian is then % expressed as follows,
\begin{equation}
\hat{\Jac}_{\bm{\theta}, \bm{\phi}}(\x) = \hat{\Jac}_{\bm{\theta}}(\x) + \hat{\Jac}_{\bm{\phi}}(\x),
\label{eq:assym_jac}
\end{equation}
where the skew-symmetric matrix $\hat{\Jac}_{\bm{\phi}}(\x)$ is given by,
\begin{equation}
\hat{\Jac}_{\bm{\phi}}(\x) = 
\begin{bmatrix} 
0 & -\Jac_{\bm{\phi},0}(\x) & \Jac_{\bm{\phi},1}(\x) \\
\Jac_{\bm{\phi},0}(\x) & 0 & -\Jac_{\bm{\phi},2}(\x) \\
-\Jac_{\bm{\phi},1}(\x) & \Jac_{\bm{\phi},2}(\x) & 0 
\end{bmatrix}.
\end{equation}
Here, $\Jac_{\bm{\phi}}(\x)$ is parameterized by a neural network that outputs the three independent components $[\Jac_{\bm{\phi}, 0}(\x), \Jac_{\bm{\phi}, 1}(\x), \Jac_{\bm{\phi}, 2}(\x)]$ of the skew-symmetric matrix. Note that this formulation is tailored for the three-dimensional case and does not readily generalize to higher dimensions. 
While the reformulation in~\eqref{eq:assym_jac} can enhance the expressiveness and flexibility of NCDS, it is essential to assess its impact on the generalization behavior of the learned vector field. 
Specifically, Fig.~\ref{fig:skewsymmetric}--\emph{middle} illustrates the behavior of a learned contractive dynamical system with an asymmetric Jacobian, $\hat{\Jac}_{\bm{\theta}, \bm{\phi}}(\x)$, as described in~\eqref{eq:assym_jac}. This is compared to the learned contractive vector field shown in Fig.~\ref{fig:skewsymmetric}--\emph{left}, which has only a symmetric Jacobian, as described in~\eqref{eq:Jacobian_simple}.
In these plots, the arrows indicate integral curves that originate from one of the initial points in the demonstrations. For the system with an asymmetric Jacobian (Fig.~\ref{fig:skewsymmetric}--\emph{middle}), the integral curve diverges from the data support, effectively taking a shortcut. In contrast, the integral curve for the system with a symmetric Jacobian (Fig.~\ref{fig:skewsymmetric}--\emph{left}) closely follows the data trend and remains within the data region.
The red rectangles emphasize an area far from the data, requiring network generalization. In the asymmetric Jacobian case (Fig.~\ref{fig:skewsymmetric}--\emph{middle}), the system initially diverges the integral curves before redirecting them back to the target. Conversely, the system with the symmetric Jacobian (Fig.~\ref{fig:skewsymmetric}--\emph{left}) guides the integral curves directly towards the data region without diverging first.
To further analyze the contraction and generalization behavior of NCDS under both conditions, we computed multiple integral curves starting from an equidistant grid around the demonstration trajectories. As shown in Fig.~\ref{fig:symmetry_assymetry}--\emph{left} and~\ref{fig:symmetry_assymetry}--\emph{middle}, the yellow region defines the area where the demonstrations, depicted as black curves, reside. 
Figures~\ref{fig:symmetry_assymetry}--\emph{left} and~\ref{fig:symmetry_assymetry}--\emph{middle} display all the integral curves generated by NCDS with symmetric and asymmetric Jacobians, respectively. Figure~\ref{fig:symmetry_assymetry}--\emph{right} shows the number of time steps spent inside the demonstration region. As illustrated, the integral curves with symmetric Jacobians spend more time within this region, i.e. the trajectories reached the data support faster, suggesting a more effective contractive behavior.
%
% \textcolor{blue}{Figure~\ref{fig:assymetric_contraction_metric} shows different eigenvalue metric heatmaps of the learned contractive dynamics. The top row shows that symmetric Jacobian approach results in increased contraction along the ``dominant'' eigenvector. In the middle row, we observe that even the ``non-dominant'' eigenvector has stronger contraction under this method. Finally, the bottom row shows a higher overall contraction spread, indicating a more pronounced pull toward a specific eigenvector. 
Additionally, as confirmed by the results in Table~\ref{tab:contraction_asy_metric}, both the contraction spread and contraction rate for the \emph{Angle} and \emph{Sine} datasets exhibit improved contractive behavior, indicating that the learned dynamics are more contractive under symmetric Jacobian. More details on the evaluation of these eigenvalue metrics can be found in App.~\ref{app:sym_maps}.

Notice that all results in Fig.~\ref{fig:skewsymmetric}, Fig.~\ref{fig:symmetry_assymetry} (and Fig.~\ref{fig:assymetric_contraction_metric_heatmaps} in appendix) are obtained using the \emph{state-independent regularization vector}, with the only distinction being the symmetry properties of the Jacobian matrix.\looseness=-1
Our formulation employs a local approximation paradigm where the network estimates a distinct Jacobian at each point in state space, rather than assuming a ``universal'' Jacobian for the entire space. This local tailoring proves experimentally sufficient for approximating contractive vector fields, even when using a symmetric Jacobian. Although a non-symmetric Jacobian may offer greater theoretical expressivity, our empirical results suggest that the symmetric Jacobian exhibits superior generalization properties, making it the preferred choice in practice.

This is shown in Fig.~\ref{fig:skewsymmetric}--\emph{right}, where NCDS with a symmetric Jacobian successfully learns the dynamics of an asymmetric system, shown in Fig.~\ref{fig:eigenvalues}--\emph{right}.

\textbf{In conclusion}, although a model with an asymmetric Jacobian could theoretically learn a broader range of contractive dynamical systems, our preliminary results presented in Fig.~\ref{fig:skewsymmetric} and Fig.~\ref{fig:symmetry_assymetry} did not show any significant improvement in reconstruction accuracy or generalization performance.
Moreover, empirical investigations presented in the results sections, including both the LASA dataset (Sec.~\ref{sec:res:lasa}) and the robotic and human motion results (Sec.~\ref{sec:human_motion}), have not identified any dynamical systems where the symmetric Jacobian approach failed to effectively capture and learn the underlying dynamics.
Having completed our discussion on improving the Jacobian formulation in~\eqref{eq:Jacobian_simple}, we now shift focus to investigating the ability of a single NCDS module to learn and represent multiple contractive vector fields, each corresponding to a conditional value such as trajectory targets or shapes.\looseness=-1
\subsection{Conditional NCDS}
\label{Sec:CNCDS}
Although NCDS has the ability to generate contractive vector fields for executing complex skills, it lacks the ability to handle multiple motion skills, which may be achieved by conditioning on task-related variables such as target states. 
In this context, other methods that leverage contraction stability often depend heavily on rigid optimization processes, which limits their adaptability when confronted with varying conditional values. These methods typically require either finding a new contraction metric each time the condition changes—likely by considering new constraints for optimization~\citep{Tsukamoto2021CVSTEM}—or, in other cases, generating a contraction metric for every condition when using Neural Contraction Metrics (NCMs)~\citep{Tsukamoto2021NeuralContractionMetric}. 
Although~\citet{jaffe2024:ELCD} introduced a contraction method that dynamically adapts to a varying target, due to the extended linearization used to parametrize the vector field, it does not consider other types of conditioning task variables, e.g., variations in trajectory shape or switching between multiple target points.
Here, we present the concept of conditional NCDS (CNCDS), which extends the vanilla NCDS with the ability of learning multimodal tasks, which depend on context variables such as variable targets, using a single NCDS module. 
The conditional variable can account for changes in the task conditions and further expand the NCDS architecture to integrate perception systems such as a vision perception backbone.
To do so, first we introduce a new condition variable $\bm{\varpi}$ in the formulation. 
This variable is concatenated to the state vector $\x$ so that CNCDS retrieves a velocity vector $\dot{\x}$ as a function of the state and the task condition. 
Therefore, we can reformulate~\eqref{eq:Jacobian_simple} as,
\begin{equation}
  \hat{\Jac_f}([\x, \bm{\varpi}])
    = - (\Jac_{\bm{\theta}}([\x, \bm{\varpi}])^\trsp \Jac_{\bm{\theta}}([\x, \bm{\varpi}]) \!+\! \text{{diag}}(\bm{\epsilon})).
\end{equation}
Figure~\ref{fig:conditional_toy_example} illustrates the vector field generated by a CNCDS trained under two distinct conditional settings: In the first, referred to as \emph{shape conditioning}, the conditional values are $0.0$ for angle motion and $1.0$ for line motion. In the second setting, referred to as \emph{target conditioning}, the condition variable $\bm{\varpi}=[x^*, y^*]$ represents the 2D coordinates of the target.\looseness=-1

\begin{figure*}
\centering
    \includegraphics[width=1.0\linewidth]{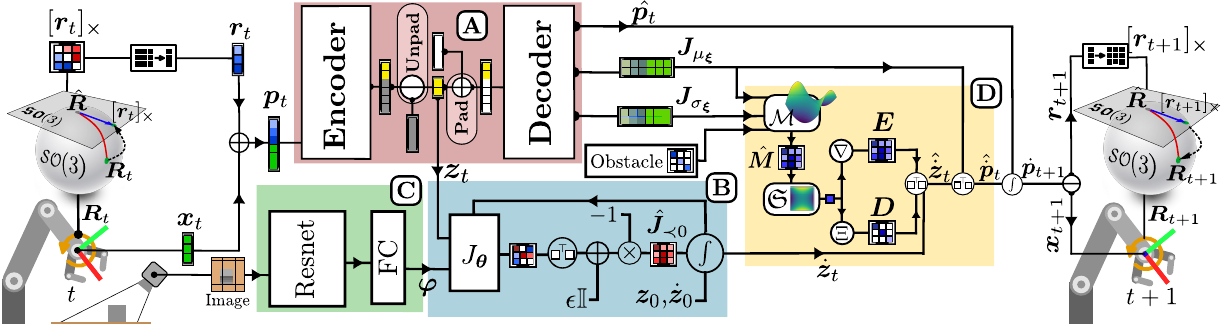}
    \caption{Architecture overview: a single iteration of NCDS simultaneously generating position and orientation dynamics. \textbf{(A) VAE (pink box):} The encoder processes the concatenated position-orientation data $\vp_t$, yielding a resulting vector that is subsequently divided into two components: the latent code $\vz$ (yellow squares) and the surplus (gray squares). The $\operatorname{Unpad}$ function in~\eqref{eq:encoding_process} removes the unused segment. The unpadded latent code $\vz$ is fed to the contraction module and simultaneously padded with zeros (white squares) before being passed to the injective decoder. \textbf{(B) Contraction (blue box):} The Jacobian network output, given the latent codes, is reshaped into a square matrix and transformed into a negative definite matrix using ~\eqref{eq:negative_def_Jacobian}. The numerical integral solver then computes the latent velocity $\dot{\vz}$. Later, using~\eqref{eq:velocity_map}, $\dot{\vz}$ is mapped to the input-space velocity via the decoder's Jacobian $\bm{J}{\mu_{\bm{\xi}}}$. \textbf{(C) Vision backbone (green box):} The vision backbone enhances the model by allowing it to adapt to different task conditions through visual perception. This component processes visual inputs, which are then concatenated with the state vector $\x$, enabling the CNCDS to generate task-specific dynamics. \textbf{(D) Riemannian modulation (yellow box):}  Uses learned Riemannian manifolds in the VAE's latent space to implicitly represent obstacles, dynamically reshaping the vector field via a modulation matrix $\bm{G}_{\mathcal{M}}(\x)$. \looseness=-1
}
    \label{fig:architecture}
\end{figure*}

Figure~\ref{fig:conditional_toy_example}--{a} and Fig.~\ref{fig:conditional_toy_example}--{b} display the vector fields resulting from conditioning on the trajectory shape.
Figure~\ref{fig:conditional_toy_example}--{a} shows the vector field when the conditional vector $\bm{\varpi}$ leads the NCDS model to reconstruct motion characterized by the \emph{angle} motion, whereas panel Fig.~\ref{fig:conditional_toy_example}--{b} displays the vector field for the \emph{line} motion.
Furthermore, Fig.~\ref{fig:conditional_toy_example}--{c} and Fig.~\ref{fig:conditional_toy_example}--{d} show the vector fields conditioned on the trajectory target.
The reported proof-of-concept experiments confirm that conditional variables can be effectively incorporated into our model, allowing a single trained CNCDS to generate motions of different shapes based on varying conditioning inputs. In Sec.~\ref{sec:exp:rob:conditional}, we show how a vision backbone can be used to design CNCDS for image-based applications. This process is visualized in the block $\mathsf{C}$ of the architecture in Fig.~\ref{fig:architecture}.
In the next section, we will focus on how to equip NCDS with obstacle avoidance capabilities using modulation matrices.

\subsection{Obstacle avoidance via matrix modulation}
\label{sec:method:obstacle_avoidance}
In this section, we review the matrix modulation technique employed by vanilla NCDS~\citep{NCDS:2023BeikMohammadi} for obstacle avoidance. Subsequently, in Sec.~\ref{Sec:RiemannianModualtion}, we will explore how this method can be extended using Riemannian manifold learning to navigate obstacles and avoid unsafe regions.
In a dynamic environment with obstacles, a learned contractive dynamical system should effectively adapt to and avoid unseen obstacles, without interfering with the global contracting behavior of the system. 
Vanilla NCDS is equipped with a contraction-preserving obstacle avoidance technique that builds on the dynamic modulation matrix $\bm{G}$ introduced by~\citet{Huber2022ModulationMatrix}. 
%
%This approach locally reshapes the learned vector field in the proximity of obstacles, while preserving contraction properties.
%
This approach locally reshapes the learned vector field in the proximity of obstacles, while preserving contraction properties within the forward invariant safe set. A set $\mathcal{S}\subseteq\mathbb{R}^D$ is forward invariant for the dynamical system $\dot{\bm{x}}=f(\bm{x})$ if, for every initial condition $\bm{x}_0\in\mathcal{S}$, the solution satisfies $\bm{x}_t\in\mathcal{S}$ for all $t\geq0$. In our context, $\mathcal{S}$ is defined as the obstacle-free region, i.e., $\mathcal{S}\cap\mathcal{O}=\emptyset$, where $\mathcal{O}$ denotes the set of points occupied by obstacles.
The discussion above implicitly assumes that the obstacle-free region $\mathcal{S}$ is simply connected, meaning it contains no holes. However, the presence of obstacles $\mathcal{O}$ introduces holes in the ambient space $\mathbb{R}^D$, resulting in a non-simply connected obstacle-free region $\mathcal{S}$ (e.g., when obstacles form ring-shaped structures). This topological complexity may prevent the existence of a continuous vector field that is globally contractive (with a single stable equilibrium). In such cases, our modulation still preserves stability by ensuring that trajectories starting in the same connected component of $\mathcal{S}$ contractively evolve without crossing into obstacle interiors.
Specifically, \citet{Huber2022ModulationMatrix} show how to construct a modulation matrix $\bm{G}$ from the obstacle's location and geometry, such that the vector field $\dot{\x} = \bm{G}(\x) f_{\bm{\theta}}(\x)$ is both contractive and steers around the obstacle. 
Formally, given the modulation matrix $\bm{G}$, we can reshape the vector field
\begin{align}
    \hat{\dot{\x}} &= \bm{G}(\x) f_{\bm{\theta}}(\x) , \\
    \bm{G}(\x) &= \bm{E}(\x) \bm{D}(\x) \bm{E}(\x)^{-1} ,
    \label{eq:modulation}
\end{align}
where $\bm{E}(\x)$ and $\bm{D}(\x)$ are the basis and diagonal eigenvalue matrices computed as,
\begin{align}
    \bm{E}(\x) &= [\bm{n}(\x) \ \mathbf{e}_1(\x) \ldots \mathbf{e}_{d-1}(\x)] ,\\ \bm{D}(\x) &= \text{{diag}}(\lambda_n(\x), \lambda_\tau(\x), \ldots, \lambda_\tau(\x)) ,
    \label{eq:matrixD}
\end{align}
where $\bm{n}(\x) = \frac{\x - \x_r}{\|\x - \x_r\|}$ is a reference direction computed w.r.t.\@ a reference point $\x_r$ on the obstacle, and the tangent vectors $\mathbf{e}_i$ form an orthonormal basis to the gradient of the distance function $\Gamma(\x)$ (see \citep{Huber2022ModulationMatrix} for its full derivation).
Moreover, the components of the matrix $\bm{D}$ are defined as $\lambda_n(\x) = 1 - \left(\frac{1}{{\Gamma(\x)}}\right)^{\frac{1}{\rho}}$, $\lambda_\tau(\x) = 1 + \left(\frac{1}{{\Gamma(\x)}}\right)^{\frac{1}{\rho}}$, where $\rho \in \mathbb{R}^+$ is a reactivity factor.
Note that the matrix $\bm{D}$ modulates the dynamics along the directions of the basis defined by the set of vectors $\bm{n}(\x)$ and $\mathbf{e}(\x)$.
As stated by~\citet{Huber2022ModulationMatrix}, the function $\Gamma(\cdot)$ monotonically increases w.r.t the distance from the obstacle's reference point $\x_r$, and it is, at least, a $C^1$ function. 
\begin{figure}
\centering
  \includegraphics[width=0.7\linewidth]{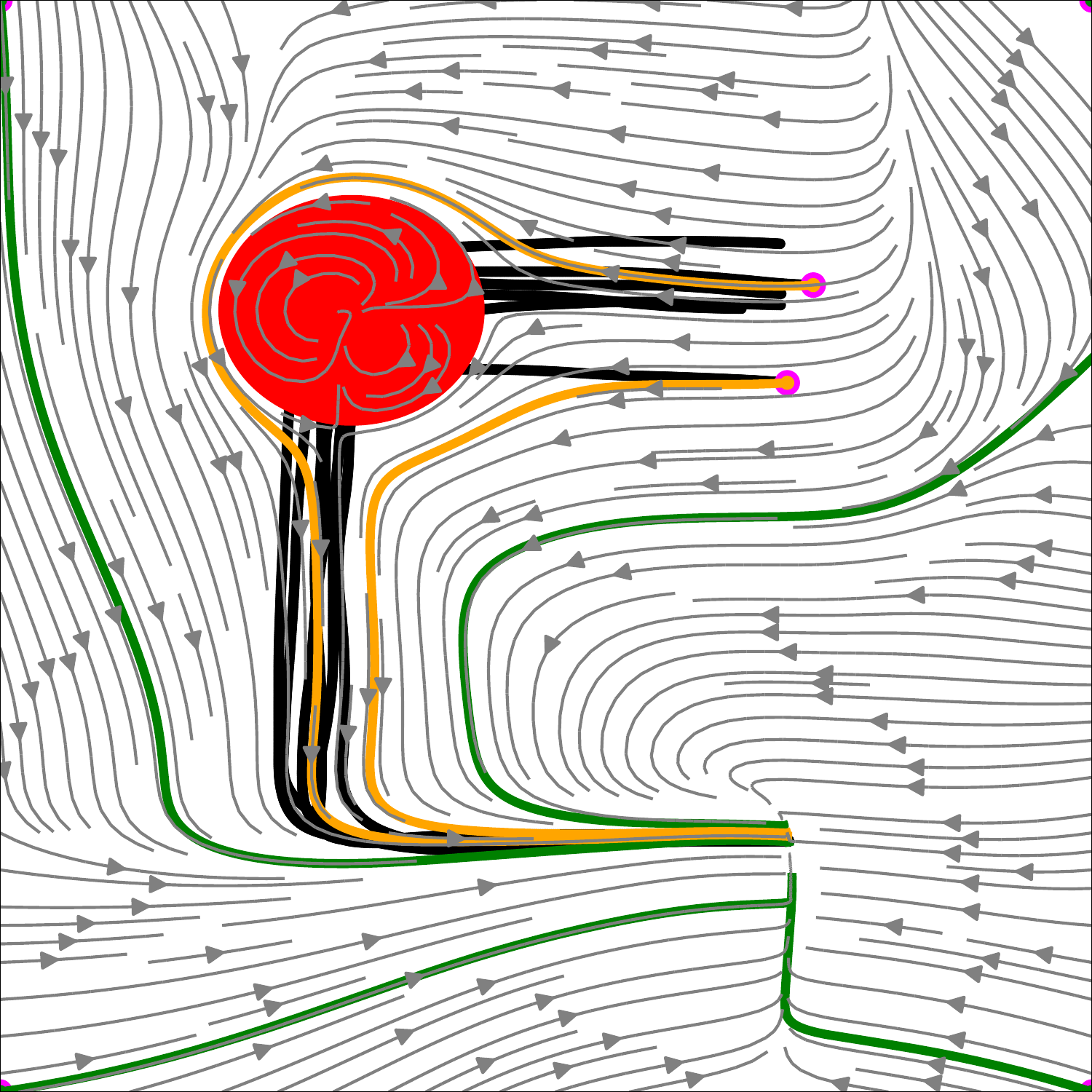}
  \caption{The obstacle locally reshapes the learned vector field using the modulation matrix. The gray contours represent the learned vector field, black trajectories depict the demonstrations, and orange/green trajectories are integral curves starting from both initial points of the demonstrations and plot corners. Magenta and red circles indicate the initial points of the integral curves and the obstacle, correspondingly.}
  \label{fig:obstacleavoidance}
\end{figure}
Importantly, the modulated dynamical system $\hat{\dot{\x}} = \bm{G}(\x) f_{\bm{\theta}}(\x)$ still guarantees contractive stability, which can be proved by following the same proof provided by~\citet{Huber2019:ConvexConcaveObstacles}. 
Figure~\ref{fig:obstacleavoidance} shows the application of a modulation matrix to navigate around an obstacle in a toy example. The obstacle, depicted as a red circle, completely obstructs the demonstrations. Notably, vanilla NCDS successfully generated safe trajectories by effectively avoiding the obstacle and ultimately reaching the target.
%
% Note that this is tailored for obstacle avoidance at the end-effector level. 
% %
% In order to extend this to multiple-limb obstacle avoidance, the modulation matrix must be refined to incorporate the robot body, and NCDS must be trained using joint space trajectories.
%

%
Having introduced all the components of NCDS, we are now prepared to explore how these concepts can be extended by using a latent variable model to learn vector fields within a low-dimensional latent space.
\section{Learning latent contractive dynamics}
\label{Sec:Latent_NCDS}
Learning highly nonlinear contractive dynamical systems in high-dimensional spaces is difficult.
These systems may exhibit complex trajectories with intricate interdependencies among the system variables, making it challenging to capture the underlying dynamics while ensuring a contractive behavior.
In the proof-of-concept examples previously discussed, we showed that NCDS works very well for low-dimensional problems, but when only limited data is available, the approach becomes brittle in higher dimensions. 
A common approach in such cases is to first reduce the data dimensionality, and work as before in the resulting low-dimensional latent space \citep{Chen2018:ActiveLearningVAE,Hung2022:ReachingLatentSpace,BeikMohammadi2021GeodesicMotionSkill}. 
The main challenge is that even if the latent dynamics are contractive, the associated high-dimensional dynamics need not be. 
This we solve next.
To begin with, we review the necessary background concepts.
\subsection{Background: deep generative models and Riemannian manifolds}
\label{subsec:VAE_Riemannian}
To obtain a low-dimensional representation of the demonstration data, the vanilla NCDS  leverages Variational Autoencoders (VAEs) to encode high-dimensional vector fields into a low-dimensional latent space. Therefore, we will review the background on VAEs.
However, while most functionalities of NCDS can be effectively transferred to the latent space, obstacle avoidance remains an exception. 
Therefore, this specific task still needs to be performed in the original high-dimensional space, where the computational complexity can negatively impact the system performance.
To overcome this, Sec.~\ref{Sec:RiemannianModualtion} introduces an approach that equips NCDS with an alternative modulation formulation, enabling the transfer of this computationally-intensive obstacle avoidance process to the latent space, thus improving overall efficiency.
We will later demonstrate that this is achieved by leveraging VAEs and their ability to learn Riemannian manifolds, which represent the underlying geometrical structure of the data in the latent space through pullback metrics.
Therefore, we also introduce the necessary background on Riemannian manifolds.
\paragraph{Variational auto-encoders (VAEs):}
\label{appen:vae}
    VAEs are generative models~\citep{kingma:autoencoding} that learn and reconstruct data by encapsulating their density into a lower-dimensional latent space $\latent$. 
    Specifically, VAEs approximate the training data density $\Prob(\x)$ in an ambient space $\ambient$ through a low-dimensional latent variable $\z$.
    For simplicity, we consider the generative process of a Gaussian VAE defined as,
    \begin{align}
        \Prob(\z) &= \mathcal{N}\left(\z | \bm{0}, \I_d\right), & \z \in \latent ; \\
        \Prob_{\bm{\varrho}}(\x|\z) &= \mathcal{N}\left(\x|\mu_{\bm{\varrho}}(\z), \I_D \sigma_{\bm{\varrho}}^2(\z)\right), & \x \in \ambient.
        \label{eq:vae_gen}
    \end{align}
    where $\mu_{\bm{\varrho}} : \latent \rightarrow \ambient$ and $\sigma_{\bm{\varrho}} : \latent \rightarrow \R^D_+$ are deep neural networks with parameters $\bm{\varrho}$ estimating the mean and the variance of the posterior distribution $\Prob_{\bm{\varrho}}(\x|\z)$, and $\I_D$ and $\I_d$ are identity matrices of size $D$ and $d$, respectively. 
    Since the exact inference of the generative process is in general intractable, a variational approximation of the evidence (marginal likelihood) can be used,
    \begin{align}
    \begin{split}
        \mathcal{L}_{ELBO}(\x)
          &= \mathbb{E}_{q_{\bm{\xi }}(\z|\x)}\left[\log(\Prob_{\bm{\varrho}}(\x|\z))\right]  \\
          &- \mathrm{KL}\left(q_{\bm{\xi }}(\z|\x)||\Prob(\z)\right) ,
        \label{eq:elbo}
    \end{split}
    \end{align}
    where $q_{\bm{\xi}}(\z|\x) = \mathcal{N}(\x | \mu_{\bm{\xi}}(\x), \I_d\sigma^2_{\bm{\xi}}(\x))$ approximates the posterior distribution $p(\z | \x)$ by two deep neural networks $\mu_{\bm{\xi}}(\x) : \ambient \rightarrow \latent$ and $\sigma_{\bm{\xi}}(\x)): \ambient \rightarrow \R^d_+$.
    The posterior distribution $p_{\bm{\xi }}(\z | \x)$ is called \emph{inference} or \emph{encoder} distribution, while the generative distribution $p_{\bm{\varrho}}(\x|\z)$ is known as the \emph{generator} or \emph{decoder}. 
    Later, we employ a VAE to capture a low-dimensional latent representation of the training data, which not only allows us to learn complex high-dimensional skills in a low-dimensional space but also enables the learning of Riemannian manifolds.

\paragraph{Injective flows:}
A limitation of VAEs is that their marginal likelihood is intractable and we have to rely on a bound. When $\mathrm{dim}(\ambient) = \mathrm{dim}(\latent)$, we can apply the change-of-variables theorem to evaluate the marginal likelihood exactly, giving rise to \emph{normalizing flows}~\citep{tabak2013family}. 
This requires the decoder to be diffeomorphic, i.e.\@ a smooth invertible function with a smooth inverse. 
In order to extend this to the case where $\mathrm{dim}(\ambient) > \mathrm{dim}(\latent)$,~\citet{Brehmer2020MFlow} proposed an \emph{injective flow}, which implements a zero-padding operation (see Sec.~\ref{subsec:learning_latent_dyn} and~\eqref{eq:inj}) on the latent variables alongside a diffeomorphic decoder, such that the resulting function is injective. 
Later, using the diffeomorphic properties of these models, we decode the learned contractive vector field from the latent space via the decoder of the injective flows.
\paragraph{Riemannian manifolds:}
\label{sec:back_riemannian_manifolds}
    In differential geometry, Riemannian manifolds are referred to as curved $d$-dimensional continuous and differentiable surfaces characterized by a Riemannian metric~\citep{Lee18Riemann}.
    This metric is characterized by a family of smoothly varying positive-definite inner products acting on the tangent spaces of the manifold, which locally resembles the Euclidean space $\R^d$. 
    In this paper, we use the mapping function $\Omega$ to represent a manifold $\Manifold$ immersed in the ambient space $\ambient$ defined as,
    \begin{align}
      \Manifold = \Omega(\latent) \quad \mathrm{with} \quad \Omega: \latent \rightarrow \ambient,
      \label{eq:mapping}
    \end{align}
    where $\latent$ and $\ambient$ are open subsets of Euclidean spaces with $\dim{\latent} < \dim{\ambient}$.
    An important operation on Riemannian manifolds is the computation of the length of a smooth curve $\zeta: [0, 1] \rightarrow \latent$, % defined as,
    \begin{align}
      \Length_{\zeta}  &= \int_0^1 \| \partial_t \Omega(\zeta(t)) \| \mathrm{d}t .
      \label{eq:length}
    \end{align}
    This length can be reformulated using the chain rule as,
    \begin{align}
      \Length_{\zeta}  &= \int_0^1 \sqrt{\dot{\zeta}(t)^{\trsp} \Metric(\zeta(t)) \dot{\zeta}(t)} \mathrm{d}t,
      \label{eq:length_chain_rule}
    \end{align}
    where $\Metric$ and $\dot{\zeta}_t = \partial_t \zeta$ are the Riemannian metric and curve derivative, respectively. 
    Note that the Riemannian metric corresponds to, 
    \begin{equation}
        \Metric(\z) = \Jac_{\Omega}(\z)^{\trsp} \Jac_{\Omega}(\z) .
        \label{eq:RiemMetric}
    \end{equation}
    Here, $\Jac_{\Omega}(\z)$ is the Jacobian of the mapping function $\Omega$.
    This metric can be used to measure local distances in $\latent$. 
    The shortest path on the manifold, also known as the geodesic, can be computed given the curve length in~\eqref{eq:length_chain_rule}. 
    \paragraph{Learning Riemannian manifolds with VAEs:}
    We here explain the link between VAEs with an injective decoder and Riemannian geometry. 
    It should be mentioned that learning Riemannian manifolds is independent of the injectivity of the decoder. However, the injection is crucial in transferring stability guarantees from the latent space to the ambient space. 
    To begin, we define the VAE generative process of~\eqref{eq:vae_gen} as a stochastic function,
    \begin{equation}
        f_{\bm{\varrho}}(\z) = \mu_{\bm{\varrho}}(\z) + \operatorname{diag}(\varepsilon)\sigma_{\bm{\varrho}}(\z), \quad \varepsilon \sim \mathcal{N}(\bm{0}, \I_D) ,
        \label{eq:StochasticF}
    \end{equation}
    where $\mu_{\bm{\varrho}}(\z)$ and $\sigma_{\bm{\varrho}}(\z)$ are decoder mean and variance neural networks, respectively. 
    Also, $\operatorname{diag}(\cdot)$ is a diagonal matrix, and $\I_D$ is a $D \times D$ identity matrix. 
    The above formulation is referred to as the reparameterization trick~\citep{kingma:autoencoding}, which can be interpreted as samples generated out of a random projection of a manifold jointly spanned by $\mu_{\bm{\varrho}}$ and $\sigma_{\bm{\varrho}}$ \citep{eklund:arxiv:2019}.
    Riemannian manifolds may arise from mapping functions between two spaces as in~\eqref{eq:mapping}. 
    As a result,~\eqref{eq:StochasticF} may be seen as a stochastic version of the mapping function of~\eqref{eq:mapping}, which in turn defines a Riemannian manifold~\citep{Hauberg:OnlyBS}. 
    We can now write the stochastic form of the Riemannian metric of~\eqref{eq:RiemMetric}. 
    To do so, we first recast the stochastic function~\eqref{eq:StochasticF} as follows~\citep{eklund:arxiv:2019},
    \begin{align}
      f_{\bm{\phi}}(\z) &= \begin{pmatrix} \I_D, & \operatorname{diag}(\varepsilon) \end{pmatrix} \begin{pmatrix} \mu_{\bm{\varrho}}(\z) \\ \sigma_{\bm{\varrho}}(\z) \end{pmatrix}
         = \bm{P}\;s(\z) ,
    \end{align} 
    where $\bm{P}$ is a random matrix, and $s(\z)$ is the concatenation of $\mu_{\bm{\varrho}}(\z)$ and $\sigma_{\bm{\varrho}}(\z)$. 
    Therefore, the VAE can be seen as a random projection of a deterministic manifold spanned by $s$. 
    Given that this stochastic mapping function is defined by a combination of mean $\mu_{\bm{\varrho}}(\z)$ and variance $\sigma_{\bm{\varrho}}(\z)$, the expected Riemannian metric is likewise based on a mixture of both as follows,
    \begin{equation}
        \Metric(\z) = \Jac_{\mu_{\bm{\varrho}}}(\z)^{\trsp} \Jac_{\mu_{\bm{\varrho}}}(\z) + \Jac_{\sigma_{\bm{\varrho}}}(\z)^{\trsp} \Jac_{\sigma_{\bm{\varrho}}}(\z),
        \label{eq:VAE_metric}
    \end{equation}
    where $\Jac_{\mu_{\bm{\varrho}}}(\z)$, $\Jac_{\sigma_{\bm{\varrho}}}(\z)$,  $\Jac_{\mu_{\bm{\varrho}}}(\z)^{\trsp}$, and $\Jac_{\sigma_{\bm{\varrho}}}(\z)^{\trsp}$ are respectively the Jacobian of $\mu_{\bm{\varrho}}(\z)$ and $\sigma_{\bm{\varrho}}(\z)$ and their corresponding transpose evaluated at $\z \in \latent$, with $\latent$ being the VAE low-dimensional latent space. Note that we assume the noise $\varepsilon$ is sampled from a normal distribution with zero mean and unit covariance, thus the contribution from the variance term vanishes in expectation.
    With access to this metric, we can locally reshape the manifold by modifying the metric in targeted regions. Specifically, we can increase the local metric volume in areas where obstacles are present -- effectively increasing the local curvature (i.e., the degree of metric distortion that governs how geodesics bend). This adjustment makes paths through these regions less favorable.

    We later discuss how this adjusted curvature can be transformed into a metric that aids in obstacle avoidance within the latent space.
    In this study, we employ a reshaped ambient metric to integrate obstacle information~\citep{BeikMohammadi23:ReactiveMotion}. Specifically, assuming that an obstacle is modeled using a Gaussian-like function, we define the ambient metric as:
    \begin{equation}
    \label{eq:obstacle_metric}
        \Metric_\mathcal{P} (\x) \!=\! \left( 1 \!+\! \mathfrak{w} \exp\left(\! \frac{-\| \bm{x} \!-\! \bm{o} \|^2}{2r^2} \right)\!\right)\I_3, \enspace \bm{x} \in \mathbb{R}^3,
    \end{equation}
    where $\mathfrak{w} > 0$ scales the obstacle cost, and $\bm{o} \in \R^3$, $r > 0$ represent the obstacle's position and radius, respectively. The orientation metric is assumed to be the identity, denoted by $\bm{M}_\mathcal{R}$ (or $\bm{M}_\mathcal{Q}$). For example, if the orientation is represented by quaternions $\bm{q} \in \mathcal{S}^3 \subset \mathbb{R}^4$, then the corresponding metric on the tangent space $\mathcal{T}_{\bm{q}} \mathcal{S}^3$ (which is isomorphic to $\mathbb{R}^3$) is given by the $3\times 3$ identity matrix $\mathbb{I}_3$. Therefore, the full metric is defined as follows,
    
    \begin{equation}
        \Metric_\ambient = 
        \begin{pmatrix}
            \Metric_\mathcal{P} & \mathbf{0} \\[1mm]
            \mathbf{0} & \bm{M}_\mathcal{R} \text{ or } \bm{M}_\mathcal{Q}
        \end{pmatrix}
        \label{eq:metric_ambient_full}
    \end{equation}

    Note that we use a Gaussian-like function to represent the obstacle here; however, more complex representations, such as point cloud-based meshes, can also be utilized.
    Then, the corresponding Riemannian metric in the latent space is then given by,
        \begin{equation}
            \Metric(\z) = \Jac_{\mu_{\bm{\varrho}}}(\z)^{\trsp} \Metric_\ambient \Jac_{\mu_{\bm{\varrho}}}(\z) + \Jac_{\sigma_{\bm{\varrho}}}(\z)^{\trsp}  \Metric_\ambient  \Jac_{\sigma_{\bm{\varrho}}}(\z).
            \label{eq:RiemMetricWithObstacle}
        \end{equation}
    Note that changing the obstacle positions does not require re-training the VAE, as this only changes the ambient metric $\Metric_\ambient (\x)$.

\subsection{Latent NCDS}
\label{subsec:learning_latent_dyn}
As briefly discussed above, we want to reduce the data dimensionality with a VAE, but we further require that any latent contractive dynamical system remains contractive after it has been decoded into the data space. 
To do so, we leverage the fact that contraction is invariant under coordinate changes~\citep{Manchester17:ControlContractionMetric, kozachkov2023:RiemannianContraction}.
This means that the transformation between the latent and data spaces may be generally achieved through a diffeomorphic mapping.
\begin{theorem}[Contraction invariance under diffeomorphisms  \citep{Manchester17:ControlContractionMetric}]
    \label{th:contraction_invariance}
    Given a contractive dynamical system $\dot{\x} = f(\x)$ and a diffeomorphism $\psi$ applied on the state $\x \in \mathbb{R}^D$, the transformed system preserves contraction under the change of coordinates $\bm{y}~=~\psi(\x)$. Equivalently, contraction is also guaranteed under a differential coordinate change $\delta_{\bm{y}} = \frac{\partial \psi}{\partial \x} \delta_{\x}$.
\end{theorem}
Following Theorem~\ref{th:contraction_invariance}, we learn a VAE with an injective decoder $\mu: \latent \rightarrow \ambient$. 
Letting $\mathcal{M} = \mu(\latent)$ denote the image of $\mu$, then $\mu$ is a diffeomorphism between $\latent$ and $\mathcal{M}$, such that Theorem~\ref{th:contraction_invariance} applies. 
Geometrically, $\mu$ spans a $d$-dimensional submanifold of $\ambient$ on which the dynamical system operates.

Here we leverage the zero-padding architecture from \citet{Brehmer2020MFlow} for the decoder.
Formally, an injective flow $\mu: \latent \rightarrow \ambient$ learns an injective mapping between a low-dimensional latent space $\latent$ and a higher-dimensional data space $\ambient$.
Injectivity of the flow ensures that there are no singular points or self-intersections in the flow, which may compromise the stability of the system dynamics in the data space.
The injective decoder $\mu$ is composed of a zero-padding operation on the latent variables followed by a series of $K$ invertible transformations $\iota_k$. 
This means that,
\begin{align}
  \mu = \iota_K \circ \cdots \circ \iota_1 \circ \operatorname{Pad},
  \label{eq:inj}
\end{align}
where $\operatorname{Pad}(\z) = \left[z_1 \cdots z_d \,\, 0 \cdots 0\right]^\trsp$ represents a $D$-dimensional vector $\z$ with additional $D-d$ zeros. 
We emphasize that this decoder is an injective mapping between $\latent$ and $\mu(\latent) \subset \ambient$, such that a decoded contractive dynamical system remains contractive.
Specifically, we propose to learn a latent data representation using a VAE, where the decoder mean $\mu_{\bm{\xi}}$ follows the architecture in~\eqref{eq:inj}. 
Empirically, we have found that training stabilizes when the variational encoder takes the form $q_{\bm{\xi}}(\z|\x) = \mathcal{N}(\z ~|~ \mu^{\sim 1}_{\bm{\xi}}(\x), \, \I_d\sigma^2_{\bm{\xi}}(\x))$, where $\mu^{\sim 1}_{\bm{\xi}}$ is the approximate inverse of $\mu_{\bm{\xi}}$ given by,
\begin{align}
\label{eq:encoding_process}
  \mu^{\sim 1}_{\bm{\xi}} = \operatorname{Unpad} \circ \, \iota_1^{-1} \circ \cdots \circ \iota_K^{-1} , 
\end{align}
where $\operatorname{Unpad}: \R^D \rightarrow \R^d$ removes the last $D\!-\!d$ dimensions of its input as an approximation to the inverse of the zero-padding operation. 
We emphasize that an exact inverse is not required to evaluate a lower bound of the model evidence.
 This process is visualized in block $\mathsf{A}$ of the architecture in Fig.~\ref{fig:architecture}.\looseness=-1
It is important to note that the state $\x$ solely encodes the positional information of the system, disregarding the velocity $\dot{\x}$. 
In order to decode the latent velocity $\dot{\z}$ into the data space velocity $\dot{\x}$, we exploit the Jacobian matrix associated with the decoder mean function $\mu_{\bm{\xi}}$, computed as $\Jac_{\mu_{\bm{\xi}}}(\z) = \nicefrac{\partial\mu_{\bm{\xi}}}{\partial \z}$. 
This enables the decoding process formulated as below,
\begin{equation}
    \dot{\x} = \Jac_{\mu_{\bm{\xi}}}(\z)\dot{\z} .
    \label{eq:velocity_map}
\end{equation}
The above tools let us learn a contractive dynamical system on the latent space $\latent$, where the contraction is guaranteed by employing the NCDS architecture (Sec.~\ref{subsec:learning_jacobian}).
Then, the latent velocities\footnote{For training, the latent velocities are simply estimated by a numerical differentiation w.r.t\@ the latent state $\z$.} $\dot{\z}$ given by such a contractive dynamical system can be mapped to the data space $\ambient$ using~\eqref{eq:velocity_map}. 

Assuming the initial robot configuration $\x_0$ is in $\mathcal{M}$ (i.e., $\x_0 = f(\z_0)$), the subsequent motion follows a contractive system along the manifold. 
If the initial configuration $\x_0$ is not in $\mathcal{M}$, the encoder is used to approximate a projection onto $\mathcal{M}$, producing $\z_0 = \mu^{\sim 1}(\x_0)$. 

Although the transition from $\x_0$ to $\mu(\z_0)$ may not exhibit contractive behavior, this phase is of finite duration; once the state is on $\mathcal{M}$, the contractive dynamics ensure global stability on the decoder's manifold, leading to convergence of the overall system.
This behavior can be observed in high-dimensional data, where initial deviations may occur. However, the contractive dynamics on the manifold guarantee stability, as explained in the example given in Sec.~\ref{sec:Outside_Manifold}.

\subsection{Learning position and orientation dynamics} 
\label{sec:learning_R3So3}
So far, we have focused on Euclidean robot states, but in practice, the end-effector motion also involves rotations, which do not have an Euclidean structure. 
We first review the group structure of rotation matrices and then extend NCDS to handle non-Euclidean data using Theorem~\ref{th:contraction_invariance}.
\subsubsection{Orientation parameterization.}
\label{sec:so3}
\begin{figure}
\centering
\begin{subfigure}{1.0\linewidth}
\centering
    \includegraphics[width=1.0\linewidth]{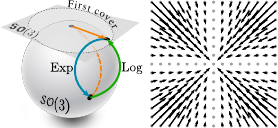}
\end{subfigure}
\caption{\emph{Left}: Aspects of the Lie group $\SO$. \emph{Right}: An illustration of the function $b(\x)$ in two dimensions.}
\label{fig:exp_log_maps}
\end{figure}

Three-dimensional spatial orientations can be represented in several ways, including Euler angles, unit quaternions, and rotation matrices~\citep{Shuster1993:OrientationRepresentation}. 
We focus on the latter approaches.
\paragraph{Rotation matrices $\SO$:}
The set of rotation matrices forms a Lie group, known as the \emph{special orthogonal group} $\SO = \left\{ \bm{R} \in \mathbb{R}^{3 \times 3} \, \middle| \, \bm{R}^\trsp \bm{R} = \mathbb{I}, \det(\bm{R}) = 1 \right\}$.
Every Lie group is associated with its Lie algebra, which represents the tangent space at its origin (see Fig.~\ref{fig:exp_log_maps}--\emph{left}). 
This Euclidean tangent space allows us to operate with elements of the group via their projections on the Lie algebra~\citep{Sola2018:MicroLieTheory}.
In the context of $\SO$, its Lie algebra $\so$ is the set of all $3 \times 3$ skew-symmetric matrices $[\bm{r}]_\times$.
This skew-symmetric matrix exhibits three degrees of freedom, which can be reparameterized as a $3$-dimensional vector $\bm{r}=[r_x, r_y, r_z] \in \mathbb{R}^3$. 
%
% \begin{figure}
%     \centering
%         \includegraphics[width=0.8\linewidth]{Sections/Plots/box2sphere2.pdf}
%         \caption{An illustration of the function $b(\x)$ in two dimensions.}
%         \label{fig:box2sphere}
% \end{figure} 
%

We can map back and forth between the Lie group $\SO$ and its associated Lie algebra $\so$ using the \emph{logarithmic} and \emph{exponential maps}, denoted $\operatorname{Log}: \SO \rightarrow \so$ and $\operatorname{Exp}: \so \rightarrow \SO$, which are defined as follows, 
\begin{align*}
    \operatorname{Exp}([\bm{r}]_\times) &= \mathbb{I} + \frac{\sin(\zeta)}{\zeta}[\bm{r}]_\times + \frac{1 - \cos(\zeta)}{\zeta^2}[\bm{r}]_\times^2 ,\\
    \operatorname{Log}(\bm{R}) &= \zeta\frac{\bm{R} - \bm{R}^\trsp}{2\sin(\zeta)} ,
\end{align*}
where $\zeta = \arccos\left(\frac{\Tr(\bm{R}) - 1}{2}\right)$. 
Moreover, $\bm{R}$ represents the rotation matrix, and $[\bm{r}]_\times$ denotes the skew-symmetric matrix associated with the coefficient vector $\bm{r}$.
Due to wrapping (i.e.,\@ $360^{\circ}$ rotation corresponds to $0^{\circ}$), the exponential map is surjective. 
This implies that the inverse, i.e.\@ $\operatorname{Log}$, is multivalued, which complicates matters. However, for vectors $\vr \in \so$, both $\operatorname{Exp}$ and $\operatorname{Log}$ are diffeomorphic if $\|\vr\|~<~\pi$~\citep{Falorsi19:DistributionsLieGroups,Urain2022:StableLieGroup}. 
This $\pi$-ball $\mathcal{B}_{\pi}$ is known as the \emph{first cover} of the Lie algebra and corresponds to the part where no wrapping occurs.
\begin{figure*}[t!]
  \centering
  \begin{subfigure}{.23\linewidth}
        \centering
        \includegraphics[width=1.\linewidth]{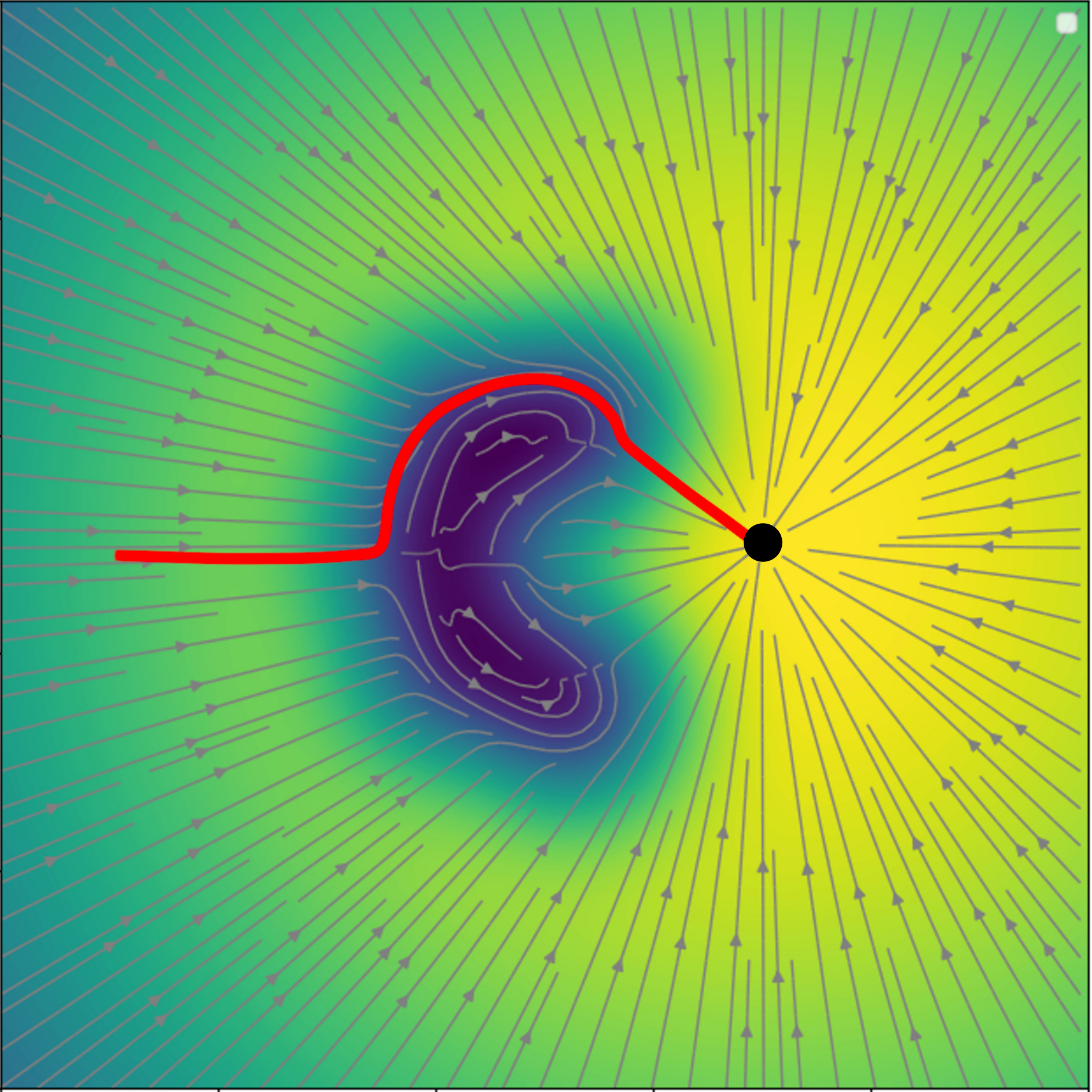}
        \caption{Convex obstacle without tangential vector field}
  \end{subfigure}
    \begin{subfigure}{.23\linewidth}
        \centering
        \includegraphics[width=1.0\linewidth]{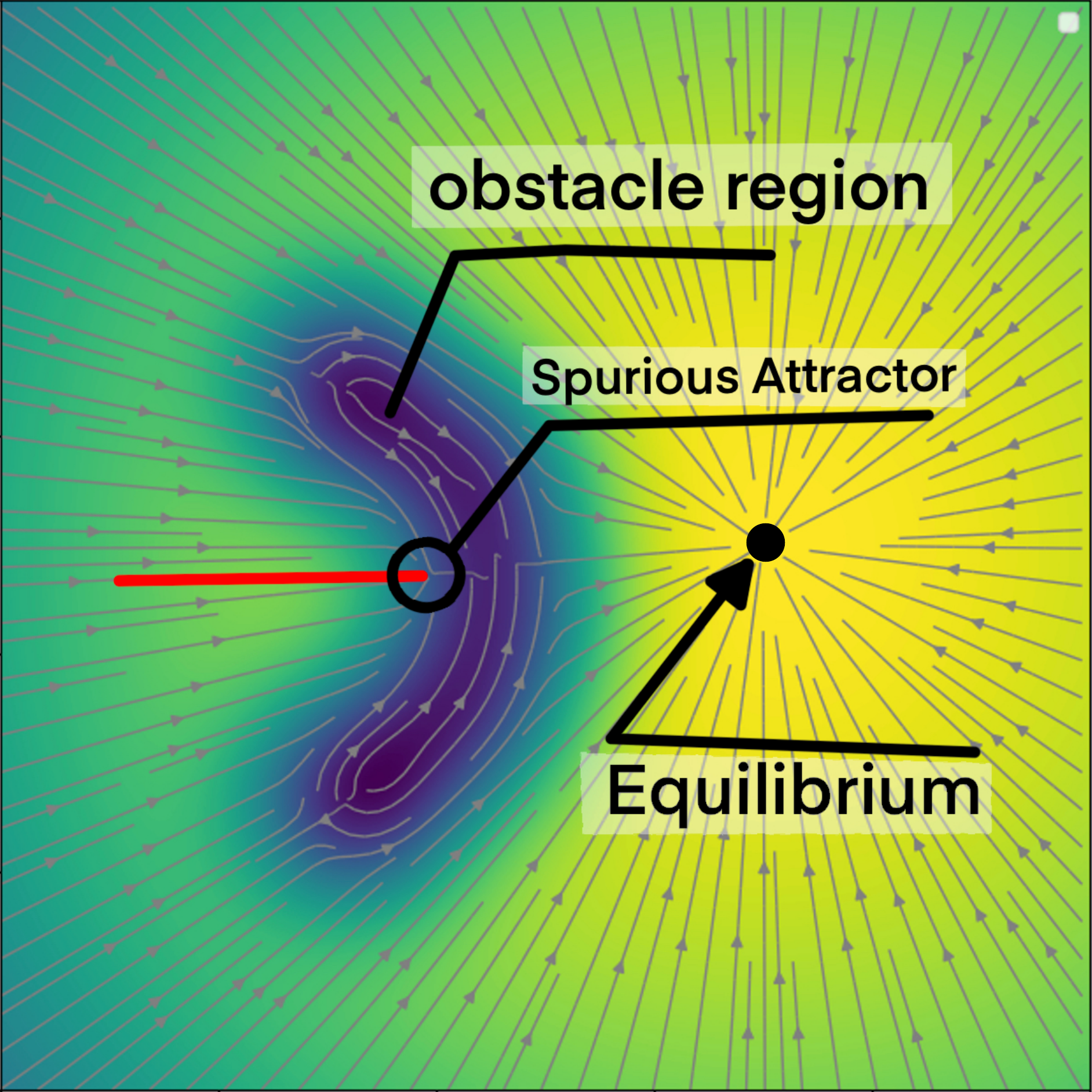}
        \caption{Concave obstacle without tangential vector field}
  \end{subfigure}
    \begin{subfigure}{.23\linewidth}
        \centering
        \includegraphics[width=1.0\linewidth]{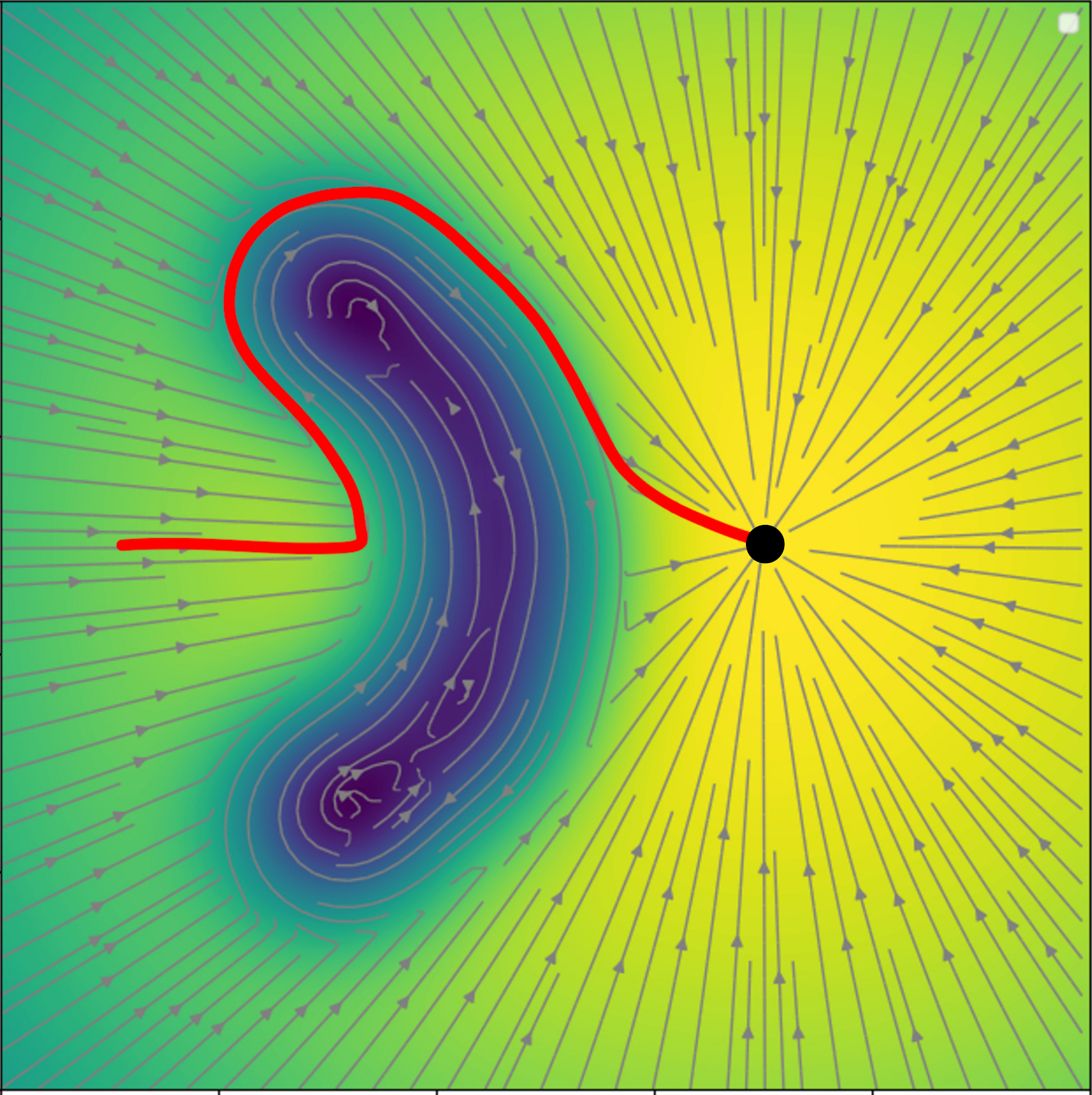}
        \caption{Concave obstacle with left tangential vector field}
  \end{subfigure}
    \begin{subfigure}{.23\linewidth}
        \centering
        \includegraphics[width=1.0\linewidth]{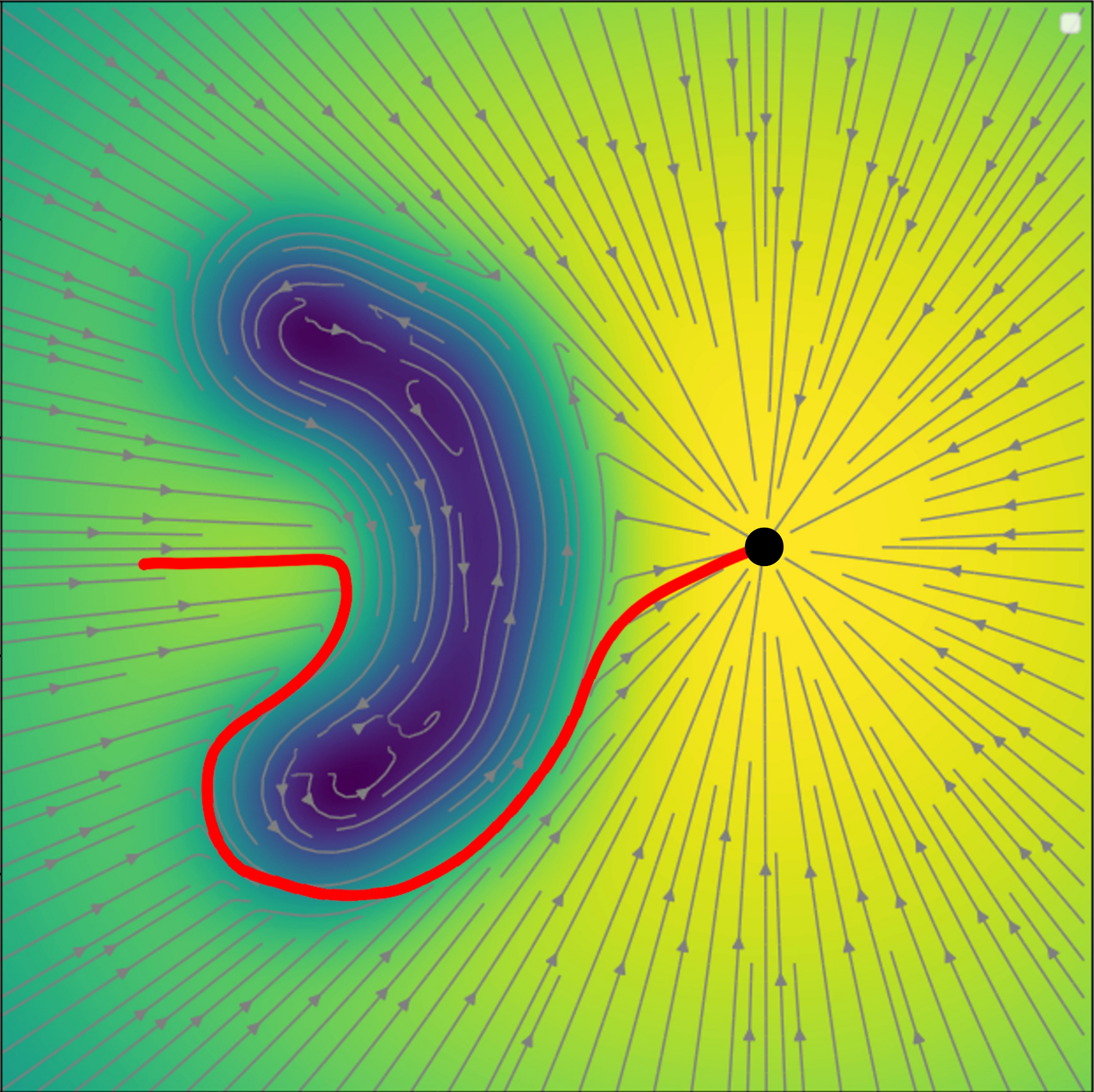}
        \caption{Concave obstacle with right tangential vector field}
  \end{subfigure}
  \caption{Modulated contractive dynamical systems using a pullback metric derived from the decoder's Jacobian.}
    \label{fig:toy_example}
\end{figure*}
\paragraph{Quaternions $\mathcal{S}^3$:}
Quaternions offer an alternative representation of rotations in 3D space and can also be endowed with a Lie group structure, leading to the \emph{unit quaternion group} $\mathcal{S}^3 = \left\{ \bm{q} \in \mathcal{S}^3 \subset \mathbb{R}^4 \, \middle| \, \|\bm{q}\| = 1 \right\}$. Like rotation matrices, the unit quaternion group $\mathcal{S}^3$ is associated with a Lie algebra $\mathfrak{su}(2)$, which represents the tangent space at its origin and corresponds to the set of pure imaginary quaternions, which can be represented as 3D vectors.

A quaternion $\bm{q} \in \mathcal{S}^3$ is typically expressed as $\bm{q} = q_0 + q_x\bm{i} + q_y\bm{j} + q_z\bm{k}$, where $q_0$ is the scalar part and $(q_x, q_y, q_z)$ represents the vector part. The exponential and logarithmic maps, denoted as $\operatorname{Exp}: \mathfrak{su}(2) \rightarrow \mathcal{S}^3$ and $\operatorname{Log}: \mathcal{S}^3 \rightarrow \mathfrak{su}(2)$, provide a way to transition between the Lie group and its algebra. These maps are defined as follows,
\begin{align*}
    \operatorname{Exp}(\bm{v}) &= \cos\left(\frac{\|\bm{v}\|}{2}\right) + \sin\left(\frac{\|\bm{v}\|}{2}\right) \frac{\bm{v}}{\|\bm{v}\|} ,\\
    \operatorname{Log}(\bm{q}) &= 2 \arccos(q_0) \frac{(q_x, q_y, q_z)}{\sqrt{q_x^2 + q_y^2 + q_z^2}} ,
\end{align*}
where $\bm{v} \in \mathbb{R}^3$ is a vector representing an element in the Lie algebra $\mathfrak{su}(2)$, and $\bm{q}$ is a unit quaternion in $\mathcal{S}^3$.

Similar to rotation matrices, the exponential map for quaternions is surjective. Quaternions, which reside on the hypersphere $\mathcal{S}^3$, have special properties when restricted to a half-sphere. In this domain, the exponential and logarithmic maps are diffeomorphic, ensuring a one-to-one correspondence between the quaternion group and its Lie algebra.\looseness=-1

\subsubsection{NCDS on Lie groups.}
Consider the situation where the system state represents its orientation, i.e.\@ $\x \in \SO$ or alternatively $\x \in \mathcal{S}^3$, whose first-order dynamics we seek to model with a latent NCDS. 
From a generative point of view, we first construct a decoder $\mu: \latent \rightarrow \so$ (or $\mu: \latent \rightarrow \mathfrak{su}(2)$ for quaternions) with outputs in the corresponding Lie algebra. 
We can then apply the exponential map to generate either a rotation matrix $\bm{R} \in \SO$ or a quaternion $\bm{q} \in \mathcal{S}^3$, such that the complete decoder becomes $\operatorname{Exp} \circ \mu$ in both cases.
Unfortunately, even if $\mu$ is injective, we cannot ensure that $\operatorname{Exp} \circ \mu$ is also injective (since $\operatorname{Exp}$ is surjective in both cases), which then breaks the stability guarantees of NCDS. 

Here we leverage the result that $\operatorname{Exp}$ is a diffeomorphism as long as we restrict ourselves to the first cover of $\so$ or $\mathfrak{su}(2)$, depending on whether we are using rotation matrices or quaternions. 
Specifically, if we choose a decoder architecture such that $\mu: \latent \rightarrow \mathcal{B}_{\pi}$ is injective and has outputs on the first cover, then $\operatorname{Exp} \circ \mu$ is injective, and stability is ensured for both $\SO$ and $\mathcal{S}^3$.
To ensure that a decoder neural network has outputs over the $\pi$-balls, we introduce a simple layer. 
Let $\mu: \R^d \rightarrow \R^D$ be an injective neural network, where injectivity is only considered over the image of $f$. 
If we add a $\texttt{TanH}$-layer, then the output of the resulting network is the $[-1, 1]^D$ box, i.e. $\operatorname{tanh}(h(\z)): \R^d \rightarrow [-1, 1]^D$. 
We can further introduce the function,
\begin{align*}
    b(\x) &= \begin{cases}
        \frac{\|\x\|_{\infty}}{\|\x\|_2}\x & \x \neq \bm{0} \\
        \x & \x = \bm{0}
    \end{cases},
\end{align*}
which smoothly (and invertibly) deforms the $[-1, 1]^D$ box into the unit ball (see Fig.~\ref{fig:exp_log_maps}--\emph{right}).
Then, the function,
\begin{align*}
h(\z) = \pi b(\operatorname{tanh}(\mu(\z))),
\end{align*}
is injective and has the signature $h: \R^d \rightarrow \mathcal{B}_{\pi}(D)$.

During training, the observed rotation matrices can be mapped directly into the first cover of $\so$ using the logarithmic map, while quaternions can be mapped into the first cover of $\mathfrak{su}(2)$. 
When the system state consists of both orientations and positions, we simply decode to higher-dimensional variables and apply exponential and logarithmic maps on the appropriate dimensions for both rotation matrices $\SO$ and quaternions $\mathcal{S}^3$. 
Figure~\ref{fig:architecture} illustrates the architecture using rotation matrices for convenience, but this is simply a design choice, as the orientation data in the preprocessing and postprocessing stages on the far left and far right of the architecture plot can easily be adapted to represent quaternions.\looseness=-1
The steps for training the VAE and Jacobian network are outlined in Algorithm~\ref{alg:Training}. 
Furthermore, the steps for employing NCDS to control a robot are detailed in Algorithm~\ref{alg:robot_control}. 
As the algorithms show, the training of the latent NCDS is not fully end-to-end. 
In the latter case, we first train the VAE (end-to-end), and then train the latent NCDS using the encoded data.\looseness=-1

\begin{algorithm}[t]
    \caption{Task-Space Training Using $\SO$}
    \label{alg:Training}
    \SetAlgoLined % Adds line numbers to each line
    \KwIn{\textbf{Data}: $\tau_n = \left \{\x_{n, t}, \bm{R}_{n, t} \right \}_{n=1}^N, \quad t \in [1, T_n]$}
    \KwOut{Learned contractive dynamics parameters $\bm{\theta}$.}
    \ForEach{trajectory $n$}{
        \ForEach{time step $t$}{
            $\bm{r}_{n,t} = \operatorname{Log}(\bm{R}_{n,t})$ \hfill \Comment{Extract skew-sym coeffs}
            
            $\bm{p}_{n,t} = [\x_{n,t}, \bm{r}_{n,t}]$ \hfill \Comment{Form new state vector}
        }
    }
    $\bm{\xi}^* = \arg \min_{\bm{\xi}} \mathcal{L}_{\text{ELBO}}(\bm{p}_{n,t})$ \hfill \Comment{Train the VAE}
    
    \ForEach{trajectory $n$}{
        \ForEach{time step $t$}{
            $\z_{n,t} = \mu_{\bm{\xi}^*}(\bm{p}_{n,t})$ \hfill \Comment{Encode poses}
            
            $\dot{\z}_{n,t} = \frac{\z_{n,t+1} - \z_{n,t}}{\Delta t}$ \hfill \Comment{Compute latent velocities}
        }
    }
    $\bm{\theta}^* = \arg \min_{\bm{\theta}} \mathcal{L}_{\text{Jac}}(\bm{p}_{n, t})$ \hfill \Comment{Train Jacobian network}
    
\end{algorithm}

\begin{algorithm}[t]
    \caption{Robot Control Scheme}
    \label{alg:robot_control}
    \SetAlgoLined % Adds line numbers to each line
    
    \KwIn{\textbf{Data}: $[\x_t, \bm{R}_t]$ \hfill \Comment{Current state of the robot}}
    \KwOut{$\dot{\x}_t$ \hfill \Comment{Velocity of the end-effector}}
    \ForEach{time step $t$}{
    $\bm{r}_{t} = \operatorname{Log}(\bm{R}_{t})$ \hfill \Comment{Extract skew-sym coeffs}
    
    $\bm{p}_{t} = [\x_{t}, \bm{r}_{t}]$ \hfill \Comment{Form new state vector}
    
    $\z_t = \mu_{\bm{\xi}}(\bm{p}_t)$ \hfill \Comment{Compute the latent state}
    
    $\hat{\dot{\z}}_t = f(\z_t)$ \hfill \Comment{Compute the latent velocity}
    
    $\bm{J}_{\mu_{\bm{\xi}}}(\z_t) = \frac{\partial \mu_{\bm{\xi}}}{\partial \z_t}$ \hfill \Comment{Compute the Jacobian of the decoder}
    
    $\dot{\x}_t = \bm{J}_{\mu_{\bm{\xi}}}(\z_t)\hat{\dot{\z}}_t$ \hfill \Comment{Compute input space velocity}
    }
\end{algorithm}

\begin{figure*}
  \centering
    \begin{subfigure}{.21\linewidth}
        \centering
        \includegraphics[width=1.0\linewidth]{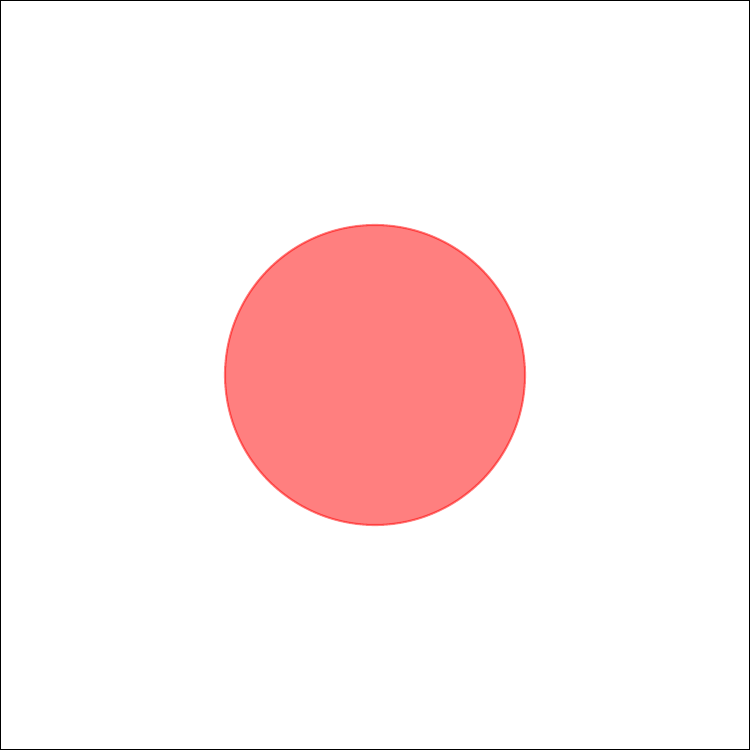}
  \end{subfigure}%
  \quad
  \begin{subfigure}{.25\linewidth}
    \centering
    \includegraphics[width=1.0\textwidth]{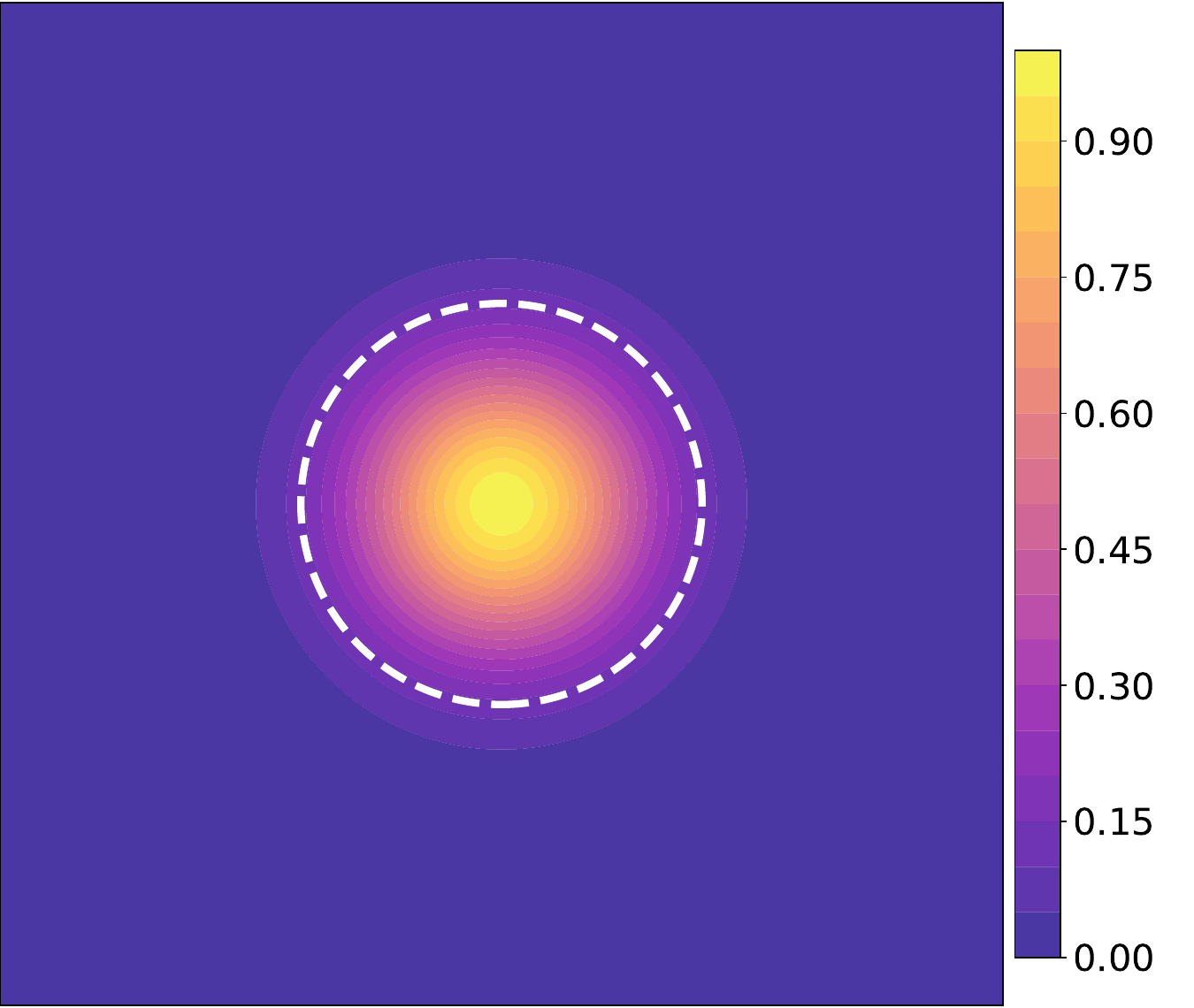}
  \end{subfigure}%
    \begin{subfigure}{.25\linewidth}
    \centering
    \includegraphics[width=1.0\textwidth]{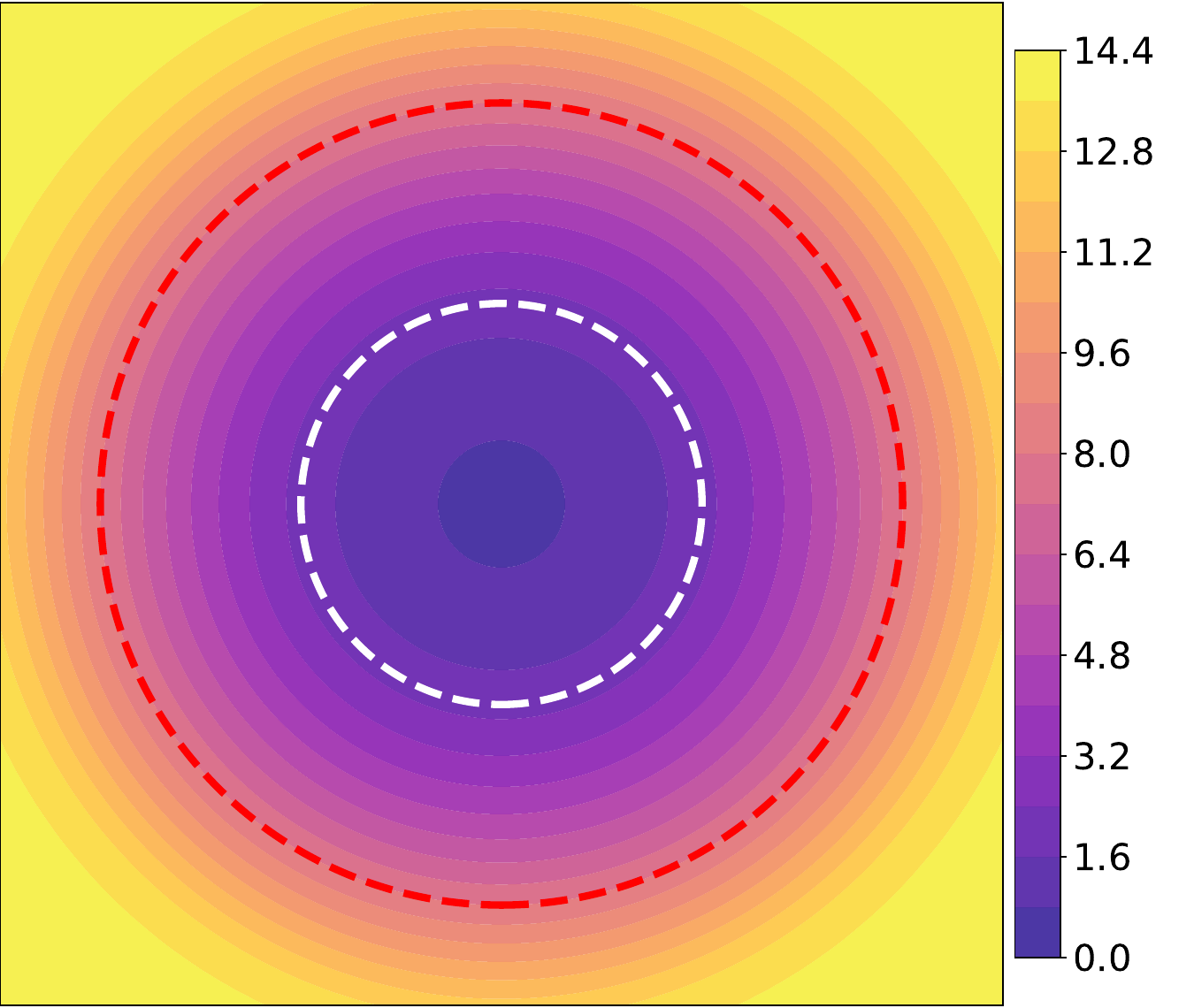}
  \end{subfigure}%
  \begin{subfigure}{.25\linewidth}
    \centering
    \includegraphics[width=1.0\textwidth]{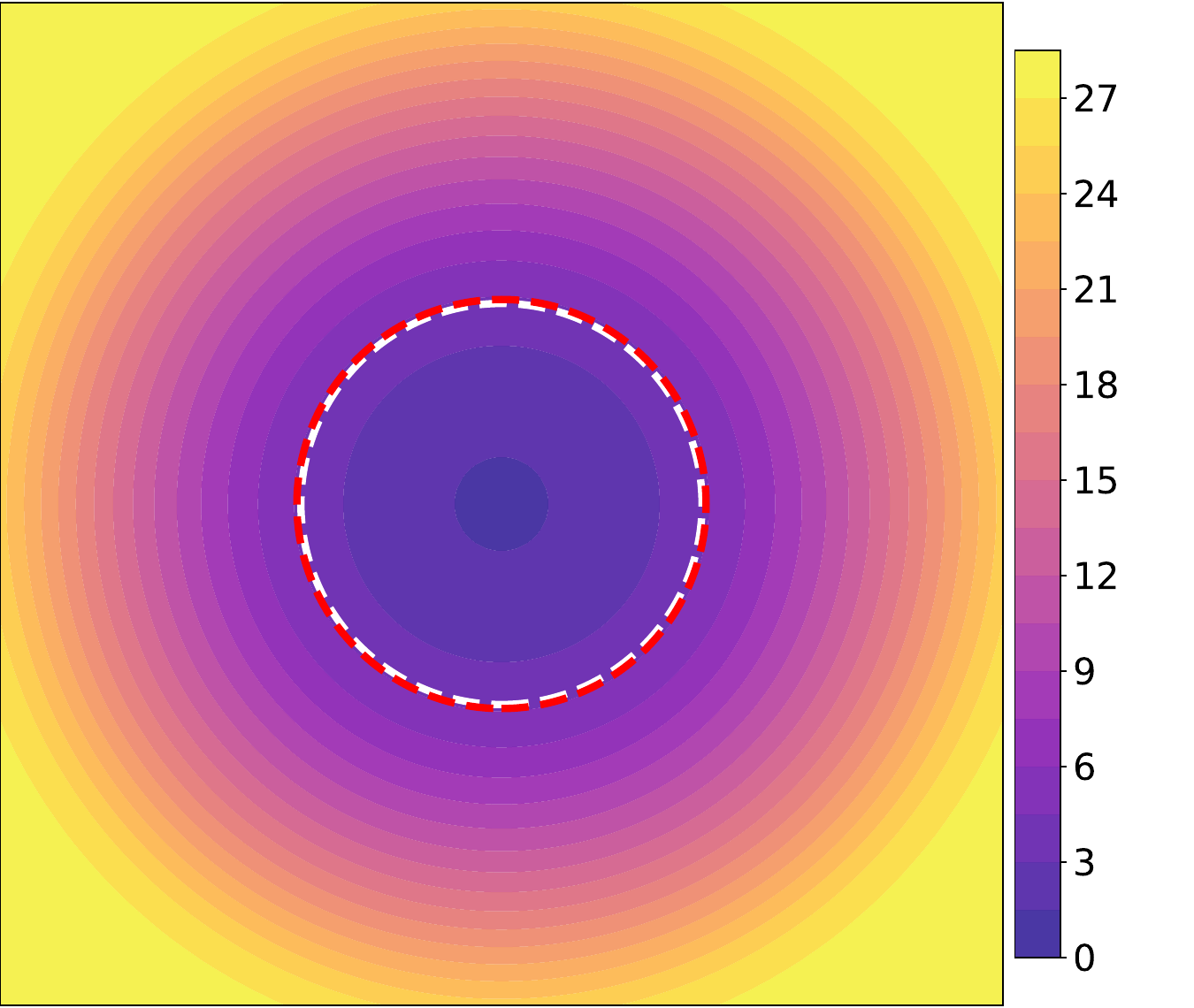}
  \end{subfigure}
  \caption{Illustration of the computation of the obstacle avoidance distance field for designing Riemannian safety regions. From left: (1) Obstacle representation (circle), (2) Metric volume around obstacle, (3) Metric inversion to compute distance map and problem with initial boundary misalignment (white vs. red circle for real vs. approximated boundary), (4) Rescaling with $\alpha$ to correct the boundary.}
  \label{fig:obstacle_definition}
\end{figure*}

\subsection{Latent obstacle avoidance and Riemannian safety regions}
\label{Sec:RiemannianModualtion}
Section~\ref{subsec:VAE_Riemannian} described a Riemannian metric (\eqref{eq:RiemMetricWithObstacle}) in the VAE latent space. Under this metric, shortest paths stay within the data support while avoiding obstacles. Building on the approach for learning Riemannian manifolds from data, we now introduce a \emph{modulation matrix} that ensures our latent NCDS vector field also stays within the data support while avoiding obstacles. Similar to classical robotics, where the configuration space provides a representation distinct from the task space for obstacle avoidance, our approach constructs a low-dimensional representation of the task space on which we perform obstacle avoidance by pulling back information from the ambient (robot's end-effector) task space.
We notice that the volume of the pullback Riemannian metric in \eqref{eq:RiemMetricWithObstacle} increases when moving away from the data manifold and when approaching an obstacle. If we change the NCDS vector field to avoid ``high volume'' areas, our goal will be achieved. We accomplish this via a Riemannian modulation matrix $\bm{G}_{\mathcal{M}}(\z)$ that reshapes the vector field $f$, resulting in an obstacle-free vector field $\hat{f}$,
\begin{equation}
    \hat{f}(\z) = \bm{G}_{\mathcal{M}}(\z) f(\z).
\end{equation}
To design this modulation matrix, we follow the recipe of \citet{Huber2022ModulationMatrix}, described in Sec.~\ref{sec:method:obstacle_avoidance}, and let $\bm{G}_{\mathcal{M}}(\z) = \bm{E}(\z) \bm{D}(\z) \bm{E}(\z)^{-1}$.

First, \underline{the matrix $\bm{D}$} is designed to guarantee \emph{impenetrability} and to ensure the \emph{local effect} of the modulation matrix. We use the formulation of \citet{Huber2022ModulationMatrix}, which is given in \eqref{eq:matrixD}. Specifically, we ensure that $\lambda_n(\z):\R^d \rightarrow [0.0, 1.0]$ decreases towards $0.0$ and $\lambda_\tau(\z):\R^d \rightarrow [1.0, 2.0]$ increases towards $2.0$ when the distance to the obstacle decreases. We give explicit expressions in Appendix~\ref{App:matrixD}.

Second, \underline{the basis matrix $\bm{E}$}, which determines the modulation directions, is defined by stacking the obstacle normal $\bm{n}$ (i.e.\@ a vector orthogonal to the tangent plane of the obstacle surface, pointing outward) and an orthogonal basis vectors $\bm{e}$ that defines a hyperplane tangential to the surface of the obstacle. To allow for obstacles with complex shapes, we compute the normal through a distance field $\mathfrak{S}$ that determines the distance from any point to the obstacle \citep{Koptev2023:ImplicitDistanceFunctions}. The normal vector can then be chosen as,
\begin{equation}
\bm{n}(\z) = \nabla \mathfrak{S}(\z) \quad \text{s.t.} \quad  \bm{e}(\z) \perp \bm{n}(\z),
\label{eq:normal_vector}
\end{equation}
where $\nabla \mathfrak{S}(\z)$ is the gradient of the distance field at $\z$.

To incorporate the Riemannian metric, we rescale the distance field by the inverse Riemannian volume,
\begin{equation*}
\mathfrak{S}_{\text{scaled}}(\z) = \cfrac{\alpha}{\mathcal{V}(\z)} \mathfrak{S}(\z), \quad \text{with} \quad
\mathcal{V}(\z) = \sqrt{|\det({\bm{M}(\z)})|},
\end{equation*}
where $\alpha$ is a scaling factor. Figure~\ref{fig:obstacle_definition} illustrates the computation of the distance field and Appendix~\ref{App:alpha} describes how we select the $\alpha$ factor.

The complete process from obstacle to distance field involves several key steps, as illustrated in Fig.~\ref{fig:obstacle_definition}. The obstacle is presented here as a circle for simplicity (leftmost panel), but it can be represented using more complex shapes, such as meshes, depending on the application. A Gaussian-like function is then applied to reshape the manifold, serving as the ambient metric centered on the obstacle (second panel). This metric models the obstacle's influence on the surrounding space. The metric is inverted to construct a distance map, as shown in the third panel. However, the computed obstacle boundary (red circle) does not align with the real boundary (white circle) due to the nature of the Gaussian function, which skews distance measurements. Finally, in the fourth panel, the distance field is rescaled using the parameter $\alpha$. This parameter adjusts the Gaussian’s influence, ensuring the computed boundary aligns with the real obstacle boundary, regardless of the obstacle’s shape.

As \citet{Koptev2023:ImplicitDistanceFunctions} discuss, the modulation in~\eqref{eq:modulation} can generate spurious attractors around concave regions of the obstacle's surface (Fig.\ref{fig:toy_example}--\emph{b}). \citet{Koptev2023:ImplicitDistanceFunctions} suggests using a secondary tangential vector field that activates only when the velocity is zero to avoid this issue. Specifically,
\begin{equation}
\label{eq:modulation_riemannian}
\hat{f} = \bm{G}_{\mathcal{M}}(\z) \cdot f(\z) + \beta(\x) \bm{G}_{\mathcal{M}}(\z) \cdot g(\z) ,
\enspace \beta \!\in\! [0,1]
\end{equation}
where $\beta$ increases as the velocity generated by the main modulation approaches zero. This tangential vector field $g(\z)$ is computed using the tangent vector on the surface of the obstacle $\bm{n}(\z)$. When multiple tangent vector fields are available, it is crucial to select the one that produces the most optimal modulated vector field. For example, in a two-dimensional space (as illustrated in Fig. \ref{fig:toy_example}--\emph{c} and \emph{d}), two distinct tangent vector fields can be identified. The criteria for choosing the appropriate tangent vector field are highly problem-dependent. For instance, when the criterion is to follow the shortest path to the target, one can compute the geodesic path and then select the tangential direction that most closely aligns with it.

%% file: Sections/Experiments.tex
\begin{figure*}
  \centering
  \begin{subfigure}{.2\textwidth}
        \centering
        \includegraphics[width=1.0\textwidth]{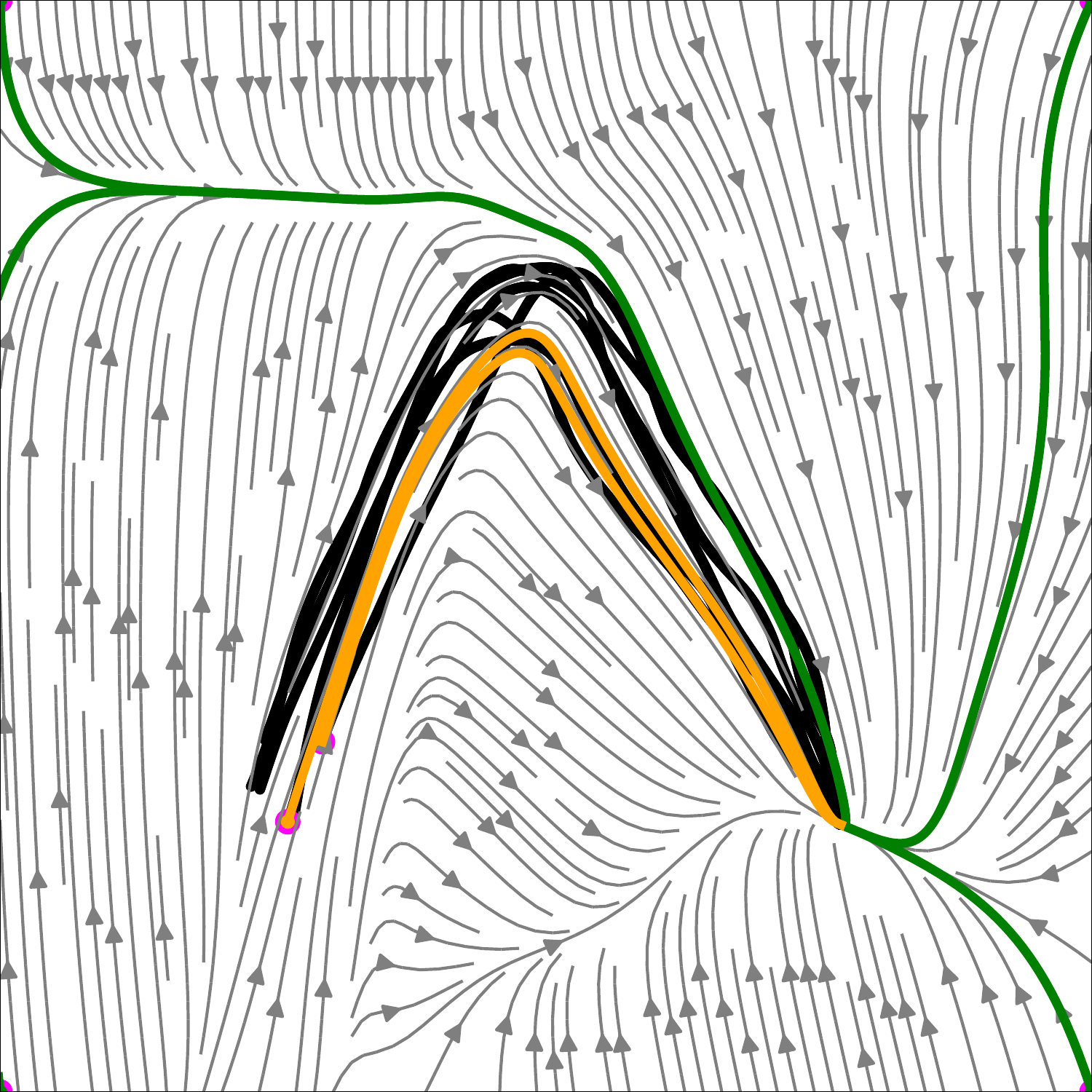}
  \end{subfigure}%
  \begin{subfigure}{.2\textwidth}
    \centering
    \includegraphics[width=1.0\textwidth]{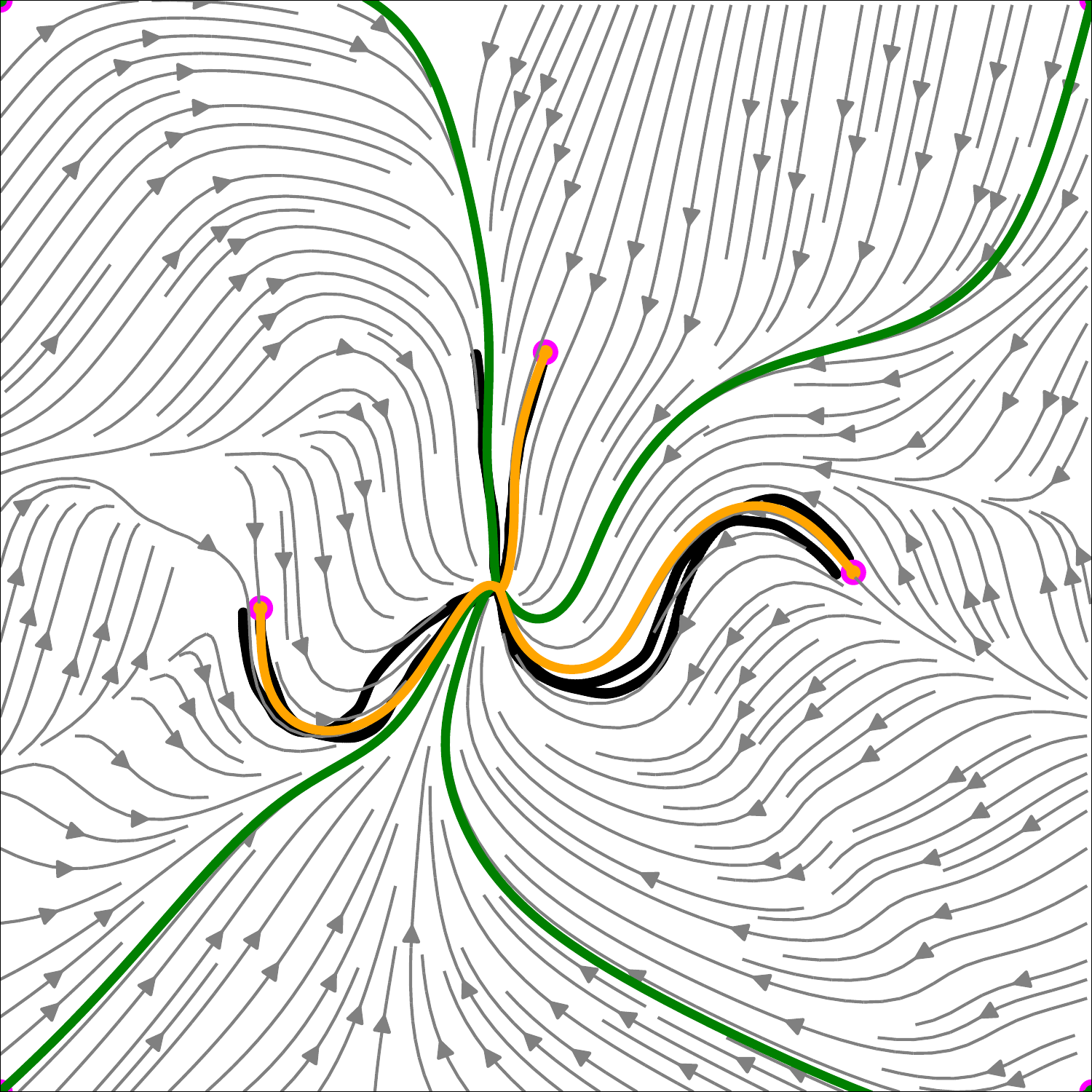}
  \end{subfigure}%probably
    \begin{subfigure}{.2\textwidth}
        \centering
        \includegraphics[width=1.0\textwidth]{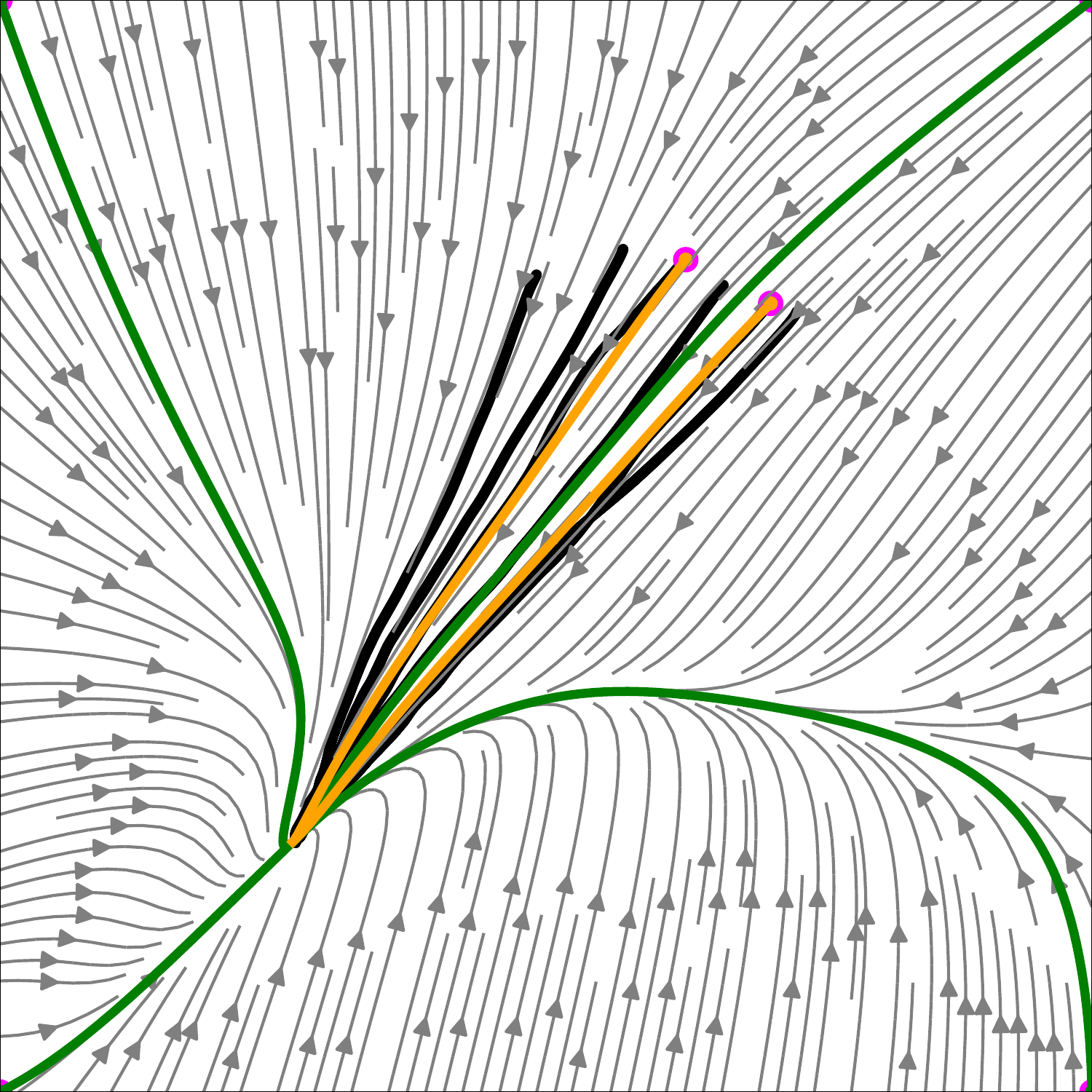}
  \end{subfigure}%
  \begin{subfigure}{.2\textwidth}
    \centering
    \includegraphics[width=1.0\textwidth]{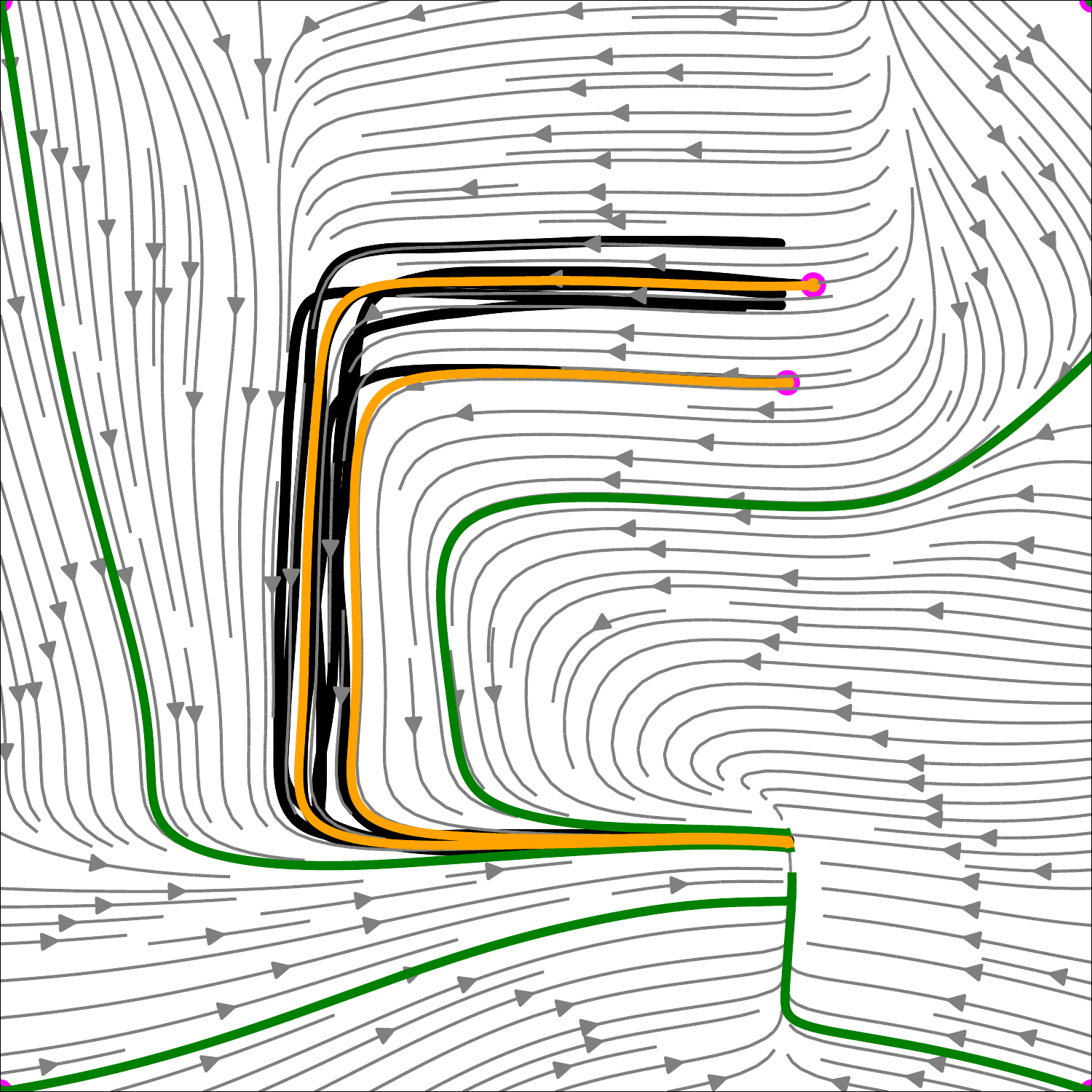}
  \end{subfigure}%
    \begin{subfigure}{.2\textwidth}
    \centering
    \includegraphics[width=1.0\textwidth]{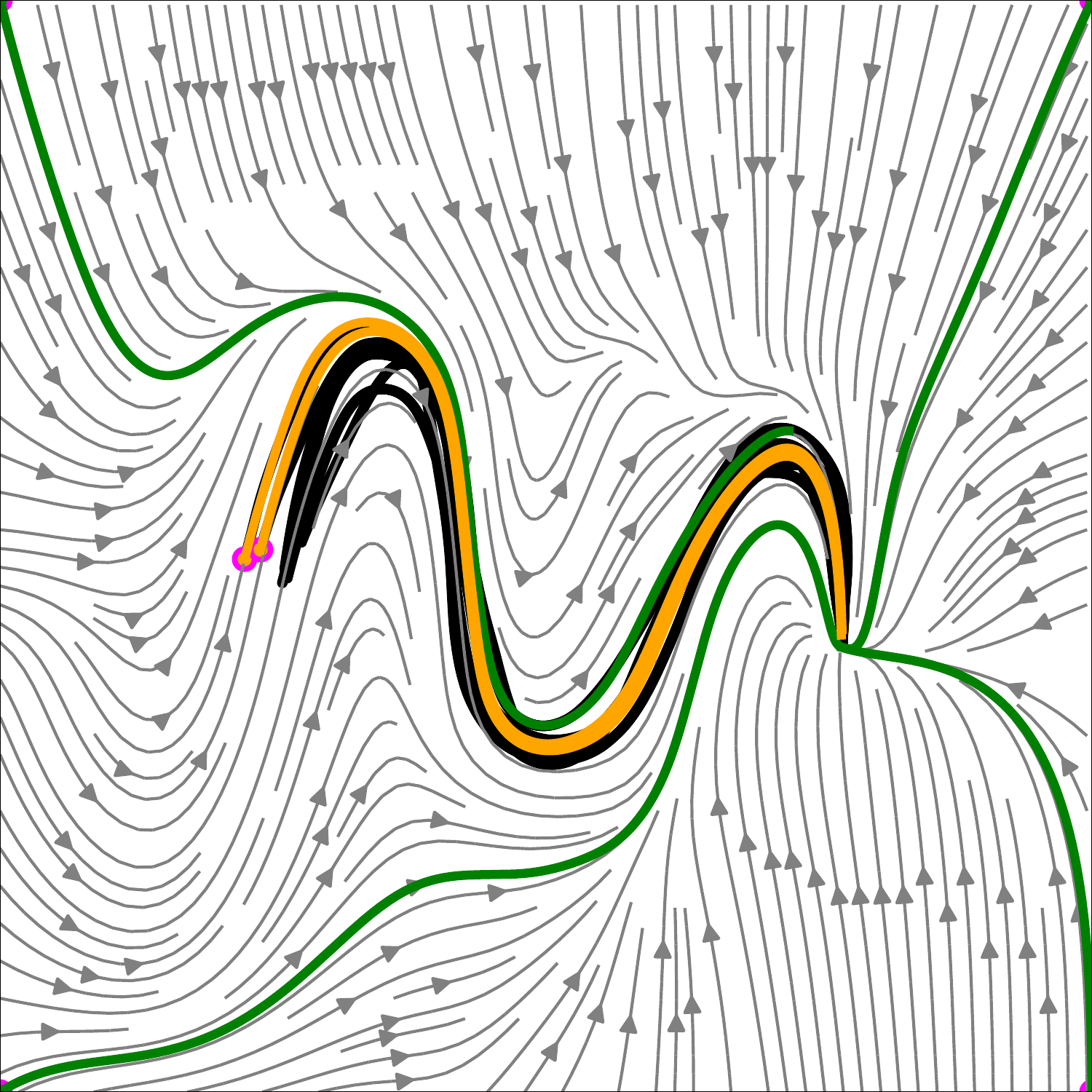}
  \end{subfigure}%
  \\
    \begin{subfigure}{.2\textwidth}
        \centering
        \includegraphics[width=1.0\textwidth]{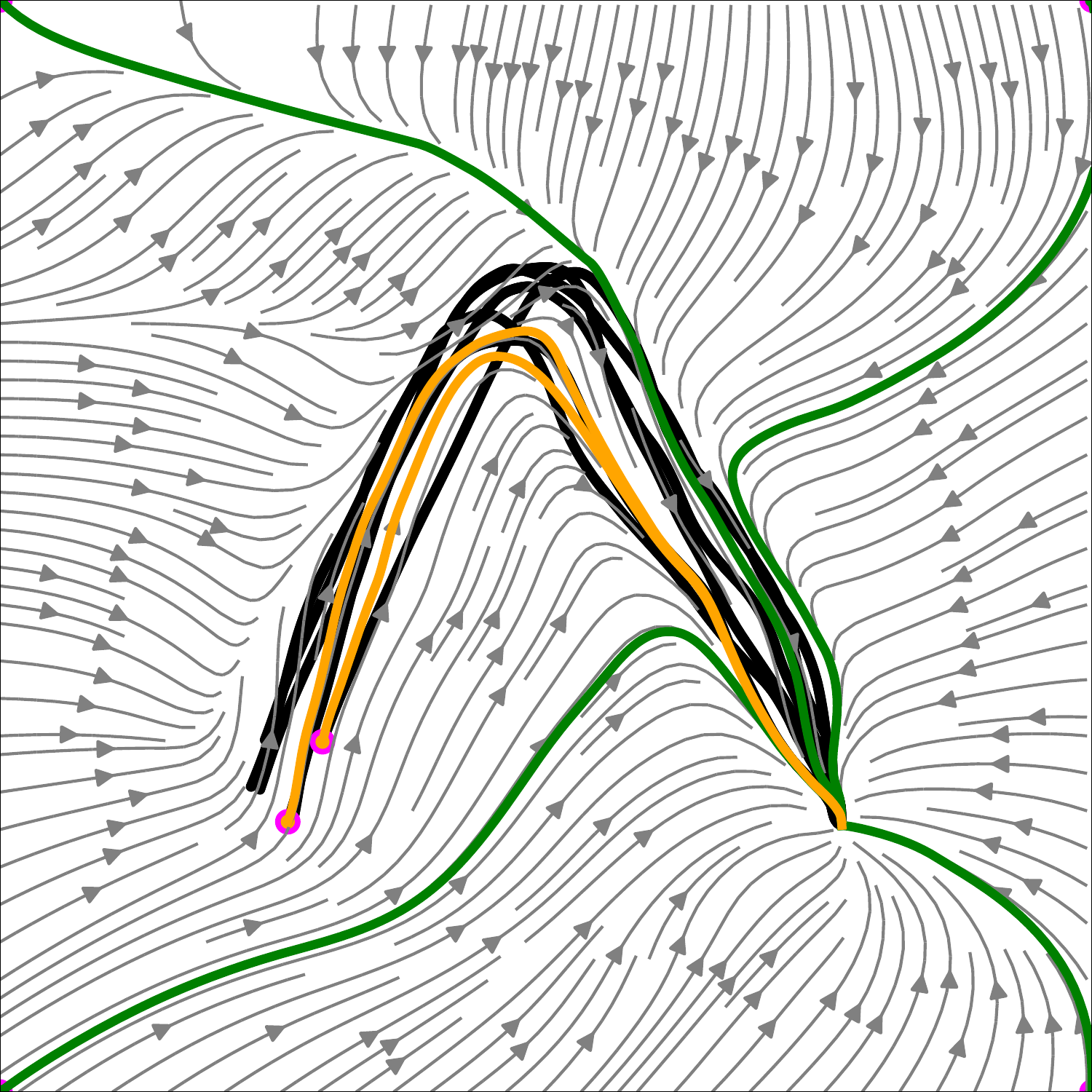}
  \end{subfigure}%
  \begin{subfigure}{.2\textwidth}
    \centering
    \includegraphics[width=1.0\textwidth]{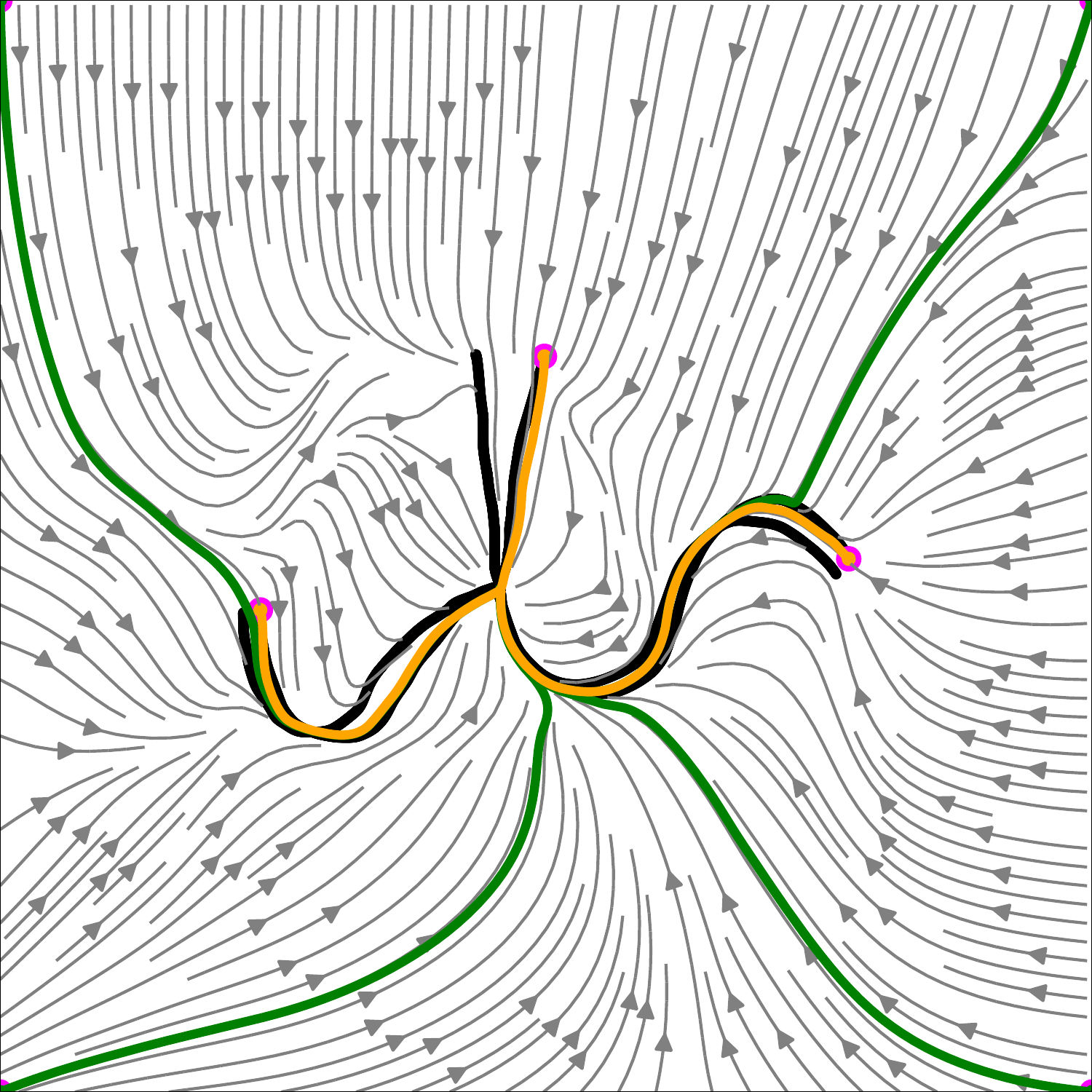}
  \end{subfigure}%
    \begin{subfigure}{.2\textwidth}
        \centering
        \includegraphics[width=1.0\textwidth]{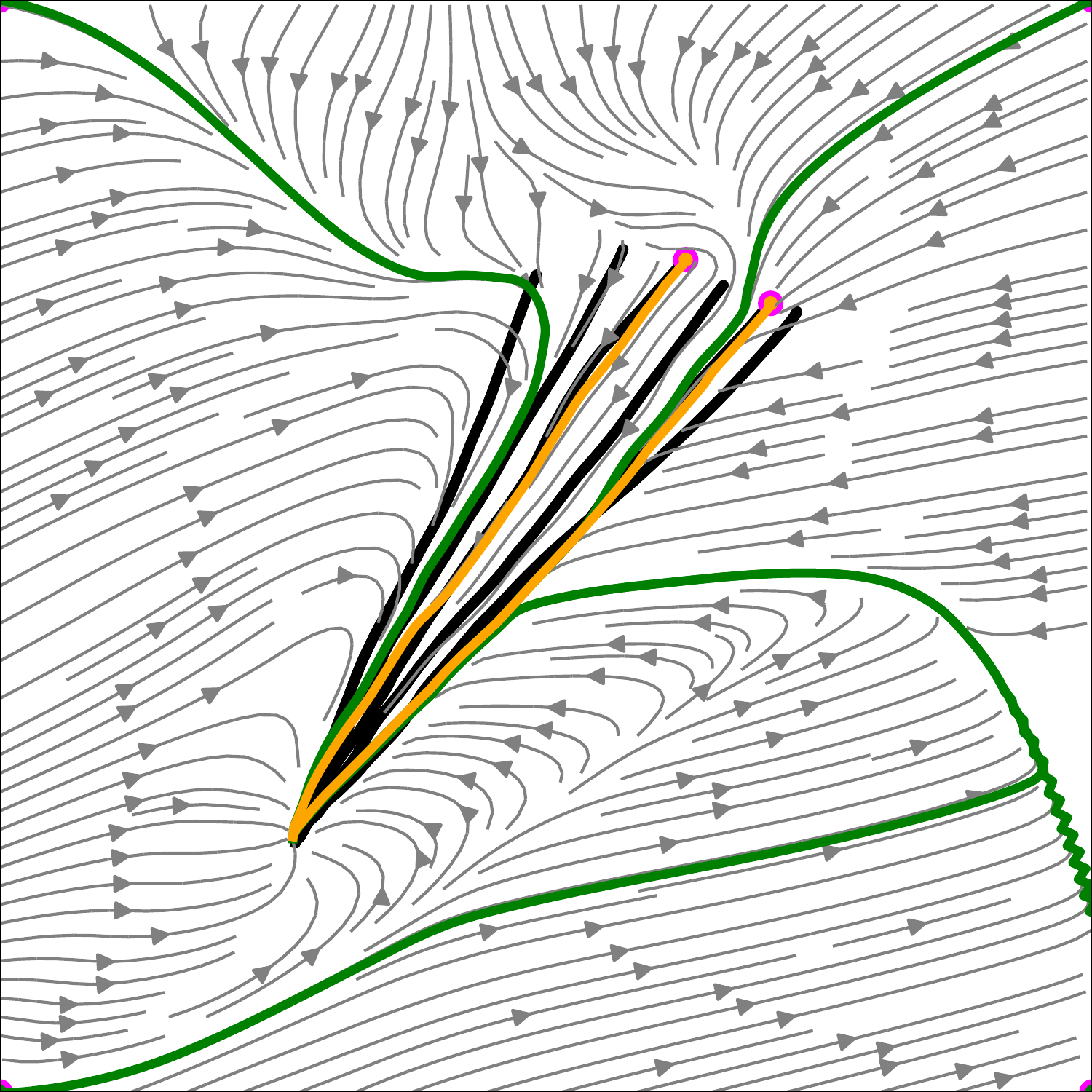}
  \end{subfigure}%
  \begin{subfigure}{.2\textwidth}
    \centering
    \includegraphics[width=1.0\textwidth]{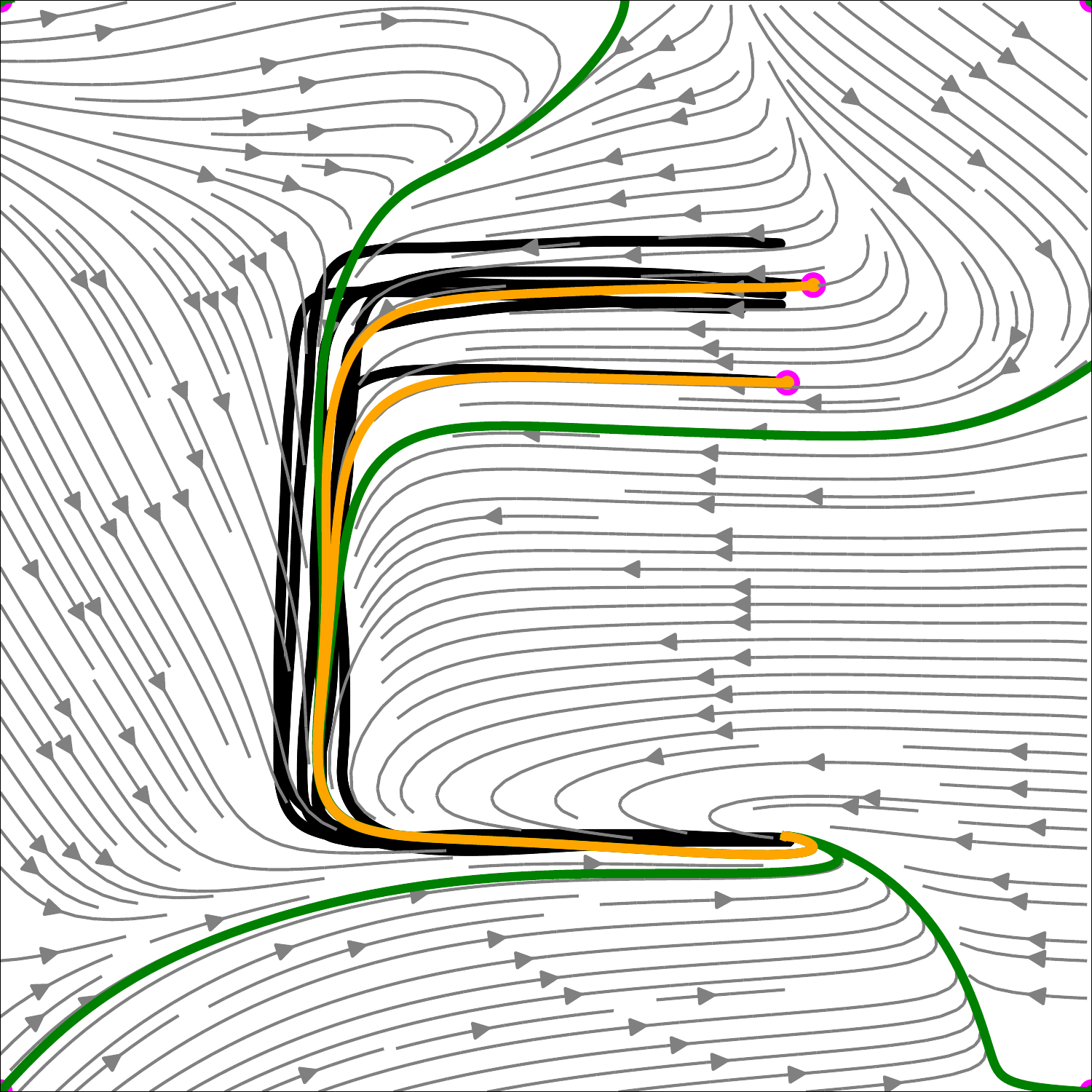}
  \end{subfigure}%
    \begin{subfigure}{.2\textwidth}
    \centering
    \includegraphics[width=1.0\textwidth]{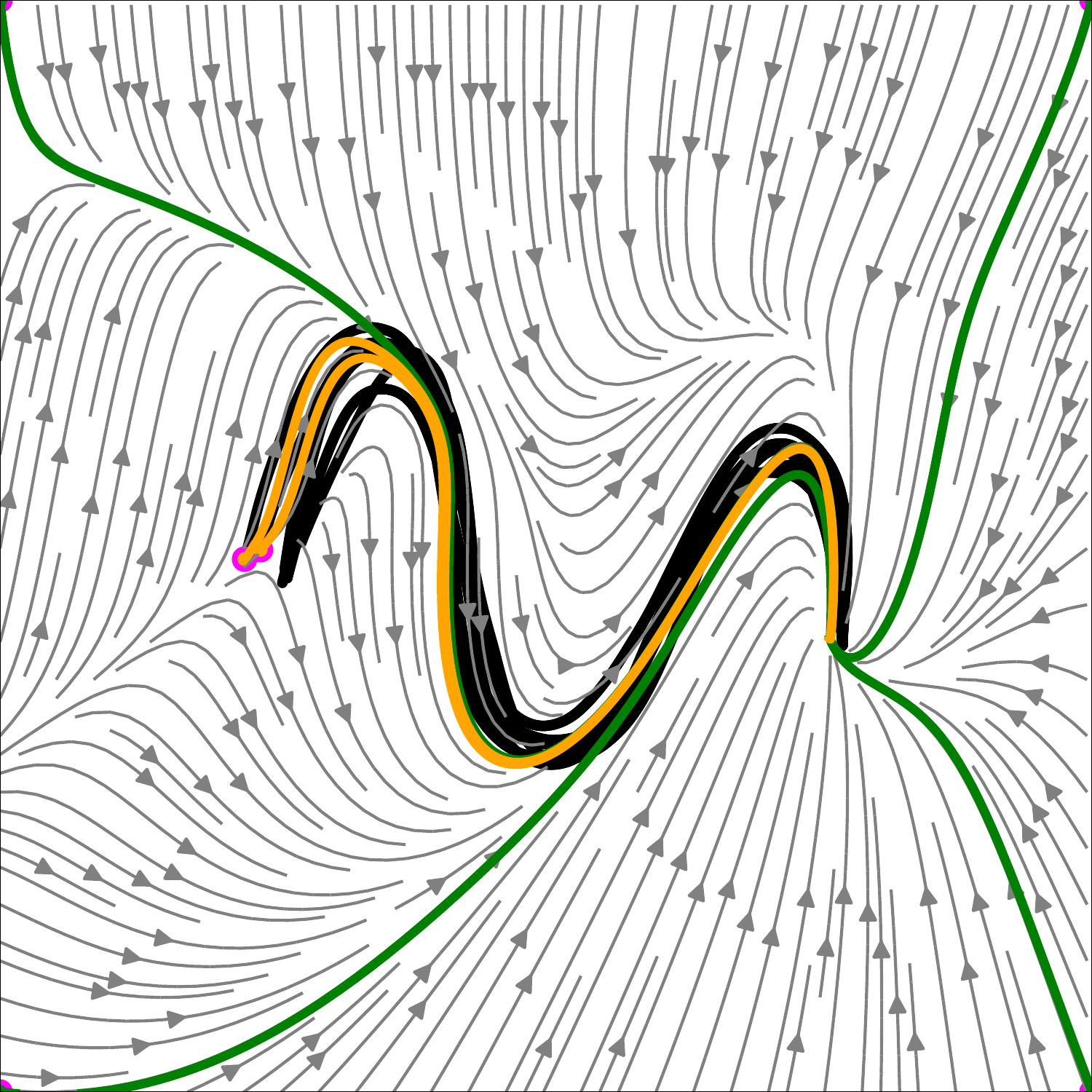}    
  \end{subfigure}%
  \caption{Visualization of the LASA-2D dataset: Gray contours represent the learned vector field, black trajectories depict demonstrations, and orange/green trajectories illustrate integral curves starting from the initial points of the demonstrations and plot corners. The magenta circles indicate the initial points of the integral curves.}
  \label{fig:LASA_Full_Dataset}
\end{figure*}
\begin{table*}[h!]
    \centering
    \small % Reduces font size of the entire table
    \renewcommand{\arraystretch}{1.3}
    \setlength{\tabcolsep}{16pt}
    \begin{tabular}{p{6cm} ccccc}
        \rowcolor{gray!15} % Light gray for the header row
        \textbf{Dataset} & \textsf{Angle} & \textsf{Multimodels} & \textsf{Line} & \textsf{SharpC} & \textsf{Sine} \\
        \midrule
        \rowcolor{gray!5} % Light gray for alternating row
        Constant Regularization & $516$ & $436$& $445$ & $490$ & $439$ \\ 
        State-independent Regularization Vector & $135$ & $105$ & $255$ & $209$ & $176$ \\ 
        \bottomrule
    \end{tabular}
    \caption{Average time steps spent outside the data region for $100$ integral curves, each consisting of $1000$ points, originating from an equidistant grid (lower is better). The datasets and NCDS models correspond to those displayed in Fig.~\ref{fig:LASA_Full_Dataset}.}
    \label{tab:contraction_comparison}
\end{table*}

To evaluate the efficiency of NCDS, we consider several synthetic and real tasks. Comparatively, we show that NCDS is the only method to scale gracefully to higher-dimensional problems, due to the latent structure. Through ablation studies in Sec.~\ref{sec:Ablation_studies}, we further analyze the effect of various activation functions, regularization techniques, and network architectures on the NCDS performance. We further demonstrate the ability to build dynamic systems on the Lie group of rotations and to avoid obstacles while ensuring stability. None of the baseline methods has such capabilities. Demonstration videos are available at: \href{https://sites.google.com/view/extendedncds/home}{https://sites.google.com/view/extendedncds/home}.
\paragraph{Datasets.}
There are currently no established benchmarks for contraction-stable robot motion learning, so we focus on two datasets. 
First, we test our approach on the LASA dataset \citep{Lemme2015:LasaDataset}, often used to benchmark asymptotic stability. 
This consists of $26$ different two-dimensional hand-written trajectories, which the system is tasked to follow. 
To evaluate our method, we have selected $5$ different letter shapes from this dataset. 
To ensure consistency and comparability, we preprocess all trajectories to stop at the same target state. 
Additionally, we omit the initial few points of each demonstration, to ensure that the only state exhibiting zero velocity is the target state.
To further complicate this easy task, we aim at learning NCDS on $2$, or $4$ stacked trajectories, resulting in $4$ (LASA-4D), or $8$-dimensional (LASA-8D) data, respectively. 
Secondly, we consider a dataset consisting of $5$ trajectories collected via kinesthetic teaching on a $7$-DoF Franka-Emika Panda robot. These trajectories form a \textsf{V}-shape path on the desk surface, with the orientation of the end-effector smoothly changing to follow the direction of motion.
To evaluate the model's performance in a more challenging setting, we use the KIT Whole-Body Human Motion Database~\citep{KrebsMeixner2021:HumanMotionDataset}.
We provide further details on how these datasets are used in the relevant sections.
\paragraph{Metrics.}
\label{sec:metrics}
We use dynamic time warping distance (DTWD) as the established quantitative measure of reproduction accuracy w.r.t.\@ a demonstrated trajectory, assuming equal initial conditions~\citep{Ravichandar2017LfDParitialContraction, Sindhwani2018:ImitationContractingVFs},\looseness=-1
%
%This is defined as,
\begin{align*}
    \operatorname{DTWD}({\tau_x}, \tau_{{x}^\prime}) =
     \sum_{j \in l({\tau}_{x^\prime})} \min_{i \in l(\tau_{x})} &\left( d(\tau_{x_i}, {\tau}_{x_j^\prime}) \right) +\\ \sum_{i \in l(\tau_{x})} \min_{j \in l({\tau}_{x^\prime})} &\left(d(\tau_{x_i}, {\tau}_{x_j^\prime}) \right),
\end{align*}
where ${\tau_x}$ and $\tau_{{x}^\prime}$ are two trajectories (e.g.\@ an integral curve and a demonstration trajectory), $d$ is a distance function (e.g.\@ Euclidean distance), and $l({\tau})$ is the length of trajectory ${\tau}$. 
To quantitatively assess the contraction properties of different vector fields, we introduce a metric based on integral curves. Specifically, we consider $N$ integral curves, each initiated from points on an equidistant grid in the state space. We rollout the trajectory over a period of $T$ time steps.
We define a \emph{demonstration region} by computing the convex hull $\mathcal{H}$ around a set of demonstration data. 
In practice, we utilize the computational geometry libraries such as \texttt{shapely}\footnote{Shapely: \href{https://shapely.readthedocs.io/en/stable/}{Python package}. Last checked: February 2025.}  to handle these operations. The convex hull $\mathcal{H}$ is computed as a polygon using \texttt{shapely.geometry.Polygon}. Point membership is then efficiently determined using standard point-in-polygon algorithms (e.g., ray-casting).
The hyperparameters determining this convex hull shape, such as the margin size extending beyond the data, are chosen experimentally to best capture the region of interest. For each integral curve $\mathbf{x}_i(t)$, where $i = 1, \dots, N$ and $t = 1, \dots, T$, we compute the number of time steps $n_i$ for which the trajectory resides within $\mathcal{H}$:
\begin{equation}
    n_i = \sum_{t=1}^{T} \mathfrak{b} \Big( \mathbf{x}_i(t) \in \mathcal{H} \Big),
\end{equation}
where $\mathfrak{b}(\cdot)$ is the indicator function, returning $1$ if its argument is true and $0$ otherwise. %Hence, $n_i \in \{0, 1, 2, \dots, T\}$.
Intuitively, the more time steps a trajectory spends within $\mathcal{H}$, the faster it converges toward it, indicating a higher contractive behavior towards the demonstrations region. Additionally, we introduce quantitative metrics to directly evaluate the system's contraction properties. Specifically, as mention before, we compute the contraction spread as the maximum absolute difference between the eigenvalues.
Note that the the contraction spread and rate reflect the overall contraction properties of the system, rather than its convergence toward the data support region.
\paragraph{Baseline methods.}
Our work is the first contractive neural network architecture, so we cannot compare NCDS directly to methods with identical goals. 
Indeed, we report in Table~\ref{table:literature_study} the main papers in the contraction literature, showing that 
a lack of consensus exists regarding the method or baseline for comparison, in the context of learning contractive dynamical systems. 
\begin{table*}[!ht]
\renewcommand{\arraystretch}{1.4} % Adds extra padding 
\small % Reduces font size of the entire table
\centering
\begin{tabular}{p{6cm} p{2.0cm} p{6.0cm} p{1.4cm}}
    \rowcolor{gray!15} % Light gray for the header row
    \textbf{Paper Title} & \textbf{Abbr.} & \textbf{Compared Against} & \textbf{Year} \\ 
    Safe Control with Learned Certificates: A Survey of Neural Lyapunov, Barrier, and Contraction Methods \citep{Dawson2022LearnedCertificates} & ~ & ~ & 2022 \\ 
    \rowcolor{gray!5} % Alternates row colors for readability
    Learning Contraction Policies from Offline Data \citep{Rezazadeh2022OfflineContraction} & ~ & MPC (ILQR), RL (CQL) & 2022 \\ 
    Neural Contraction Metrics for Robust Estimation and Control: A Convex Optimization Approach \citep{Tsukamoto2021NeuralContractionMetric} & NCM & CV-STEM, LQR & 2021 \\ 
    \rowcolor{gray!5}
    Learning Stabilizable Nonlinear Dynamics with Contraction-Based Regularization \citep{Singh2021ContractionRegularization} & CCM-R & ~ & 2021 \\ 
    Learning Certified Control Using Contraction Metric \citep{Sun2020LearningCertifiedControl} & C3M & SoS (Sum-of-Squares programming), MPC, RL (PPO) & 2020 \\ 
    \rowcolor{gray!5}
    Learning Position and Orientation Dynamics from Demonstrations via Contraction Analysis \citep{Ravichandar2019:PosOri_LfDcontraction} & CDSP & SEDS, CLF-DM, Tau-SEDS, NIVF & 2019 \\ 
    Learning Stable Dynamical Systems Using Contraction Theory \citep{Blocher2017LfDContraction} & C-GMR & SEDS & 2017 \\ 
    \rowcolor{gray!5}
    Learning Contracting Vector Fields for Stable Imitation Learning \citep{Sindhwani2018:ImitationContractingVFs} & CVF & DMP, SEDS, CLF-DM & 2017 \\ 
    Learning Partially Contracting Dynamical Systems from Demonstrations \citep{Ravichandar2017LfDParitialContraction} & CDSP (position only) & DMP, CLF-DM & 2017 \\ 
\end{tabular}
\caption{The present state-of-the-art literature on the learning of contractive dynamical systems.}
\label{table:literature_study}
\end{table*}
However, we compare our method to \emph{ELCD}, recently introduced by~\cite{jaffe2024:ELCD}, which leverages normalizing flows for learning contraction.
Additionally, we compare to existing methods that provide asymptotic stability guarantees. In particular, \emph{Euclideanizing flow} \citep{Rana2020:EuclideanizingFlows}, \emph{Imitation flow} \citep{Urain2020ImitationFlow} and \emph{SEDS} \citep{KhansariZadeh2011:StableEstimatorDS}. 
For Imitation flow, we use a network of $5$ layers, where each was constructed using \textsc{CouplingLayer}, \textsc{RandomPermutation}, and \textsc{LULinear} techniques, as recommended by~\citet{Urain2020ImitationFlow} for managing high-dimensional data. 
We train for $1000$ epochs, with a learning rate of $10^{-3}$. 
For Euclidenizing flow, we used a coupling network with random Fourier features, using $10$ coupling layers of $200$ hidden units, and a length scale of $0.45$ for the random Fourier features. This is trained for $1000$ epochs, with a learning rate of $10^{-4}$. 
Lastly, for SEDS, we use a Gaussian mixture model of $5$ components that are trained for $1000$ epochs with an MSE objective.
In all cases, hyperparameters were found experimentally and the best results have been selected to be compared against. 
For ELCD, we used the public codebase\footnote{ELCD codebase: \href{https://github.com/seanjaffe1/Extended-Linearized-Contracting-Dynamics}{GitHub Repository}. Last checked: February 2025.}. We followed their suggested parameter set, using $2$ flow steps and a hidden dimension of $16$. The model was trained for $100$ epochs with a learning rate of $10^{-3}$.
\subsection{LASA dataset}
\label{sec:res:lasa}
We first evaluate NCDS on two-dimensional trajectories for ease of visualization. 
Here data are sufficiently low-dimensional that we do not consider a latent structure. 
The Jacobian network was implemented as a neural network with two hidden layers, each containing $500$ nodes. The network's output was reshaped into a square matrix format. 
For integration, we used the efficient \textsf{odeint} function from the \textsf{torchdiffeq} Python package~\citep{Chen2018torchdiffeq}, which supports various numerical integration methods. 
In our experiments, we employed the widely used \textsf{Runge-Kutta} and \textsf{dopri5} methods for solving ordinary differential equations.
\begin{table*}
    \centering
    \small % Reduces font size of the entire table
    \renewcommand{\arraystretch}{1.3}
    \setlength{\tabcolsep}{23pt}
    \begin{tabular}{lcccc} % Use 'lcccc' instead of 'p{...}' for simpler alignment
        \rowcolor{gray!15} % Light gray for the header row
        \textbf{Dataset} & \textbf{LASA-2D} & \textbf{LASA-4D} & \textbf{LASA-8D} & \textbf{7 DoF Robot} \\ 
        Euclideanizing Flow &  $\textbf{0.72} \pm \textbf{0.12}$ & $3.23 \pm 0.34$ & $10.22 \pm 0.40$ & $5.11 \pm 0.30$ \\ 
        \rowcolor{gray!5} % Light gray for alternating rows
        Imitation Flow & $0.80 \pm 0.24$ & $\textbf{0.79} \pm \textbf{0.22}$ & $4.69 \pm 0.52$ & $2.63 \pm 0.37$ \\ 
        SEDS & $1.60 \pm 0.44$ & $3.08 \pm 0.20$ & $4.85 \pm 1.64$ & $2.69 \pm 0.18$ \\ 
        \rowcolor{gray!5} % Light gray for alternating rows
        NCDS & $1.37 \pm 0.40$ & $0.98 \pm 0.15$ & $\textbf{2.28} \pm \textbf{0.24}$ & $\textbf{1.18} \pm \textbf{0.16}$ \\ 
    \end{tabular}
    \caption{Average dynamic time warping distances (DTWD) between different approaches. NCDS shows competitive performance in low-dimensional spaces and outperforms others in higher dimensions.}
    \label{tab:comparison_results}
\end{table*}
\begin{table*}[h!]
    \centering
    \small % Reduces font size of the entire table
    \renewcommand{\arraystretch}{1.3}
    \setlength{\tabcolsep}{20pt}
    \begin{tabular}{p{6cm} cccc}
        \rowcolor{gray!15} % Light gray for the header row
        \textbf{Input Dim} & \textbf{2D} & \textbf{3D} & \textbf{8D} & \textbf{44D} \\
        \midrule
        \rowcolor{gray!5} % Light gray for alternating row
        Input Space (Without VAE) & $1.23$ ms & - & $40.9$ ms & - \\ 
        Latent Space (With VAE) & - & $10.6$ ms & $20.2$ ms & $121.0$ ms \\ 
        \bottomrule
    \end{tabular}
    \caption{Average execution time (in milliseconds) of a single NCDS integration step for learning the vector field in different spaces.}
    \label{tab:time_comp}
\end{table*}
Figure~\ref{fig:LASA_Full_Dataset} shows the learned vector fields (gray contours) for $5$ different trajectory shapes chosen according to their difficulty from the LASA dataset, covering a wide range of demonstration patterns (black curves) and dynamics. 
The top row shows the behavior of the learned dynamics for vanilla NCDS, while the bottom row depicts the dynamics learned with state-independent vector regularization (which showed the strongest contraction behavior as reported in Fig.~\ref{fig:regularization_comparison_violin}).
We observe that both NCDS approaches effectively capture and replicate the underlying dynamics.
This is observed in the orange integral curves, starting from the initial points of the demonstrations, showing that both approaches can reproduce the demonstrated trajectories accurately.\looseness=-1

\begin{figure}
    \includegraphics[width=1.0\linewidth]{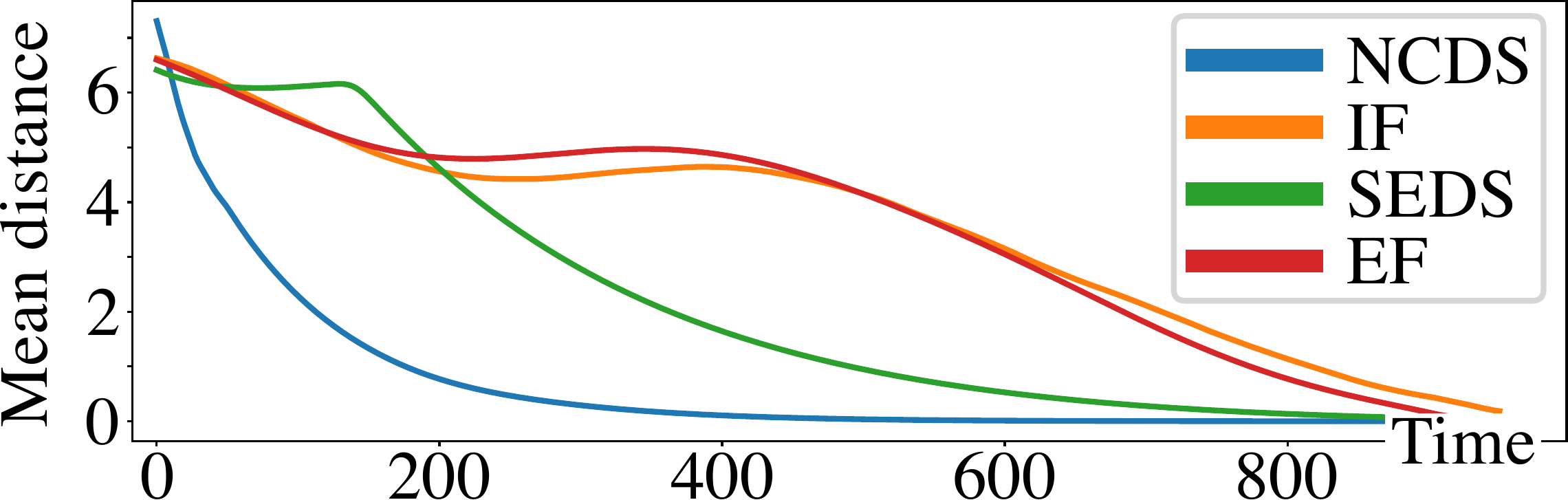}
    \caption{Average distance between random nearby trajectories over time for LASA-2D. Only NCDS monotonically decreases, i.e.\@ it is the only contractive method.}
    \label{fig:average_distance}
\end{figure}

\begin{figure*}
  \includegraphics[width=\linewidth]{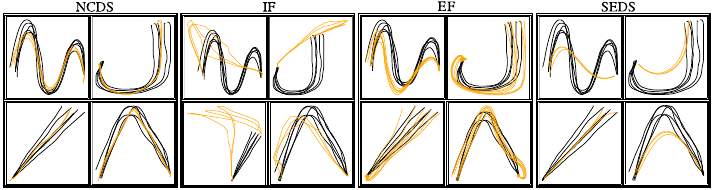}
  \caption{integral curves generated using different methods on LASA-8D. Black curves are training data, while yellow are the learned integral curves. To construct the 8D dataset, we have concatenated 4 different 2D datasets.}
  \label{fig:traj_comparison_8d}
\end{figure*}

We compute additional integral curves starting from outside the data support (green curves) to assess the generalization capability of both NCDS approaches beyond the observed demonstrations. 
Table~\ref{tab:contraction_comparison} lists the average number of time steps that $100$ integral curves, each consisting of $1000$ points and originating from an equidistant grid, spent outside the data region for the examples shown in Fig.~\ref{fig:LASA_Full_Dataset}. 
Low values indicate that the trajectories converge quicker toward the demonstration region, reflecting improved contractive behavior.
These results suggest that state-independent vector regularization enhance the contractive behavior (see also Fig.~\ref{fig:regularization_comparison_violin}).
This is evidence that the contractive construction is a viable approach to controlling the extrapolation properties of the neural network.
\begin{figure*}
  \centering
    \begin{subfigure}{.23\linewidth}
        \centering
        \includegraphics[width=1.0\linewidth]{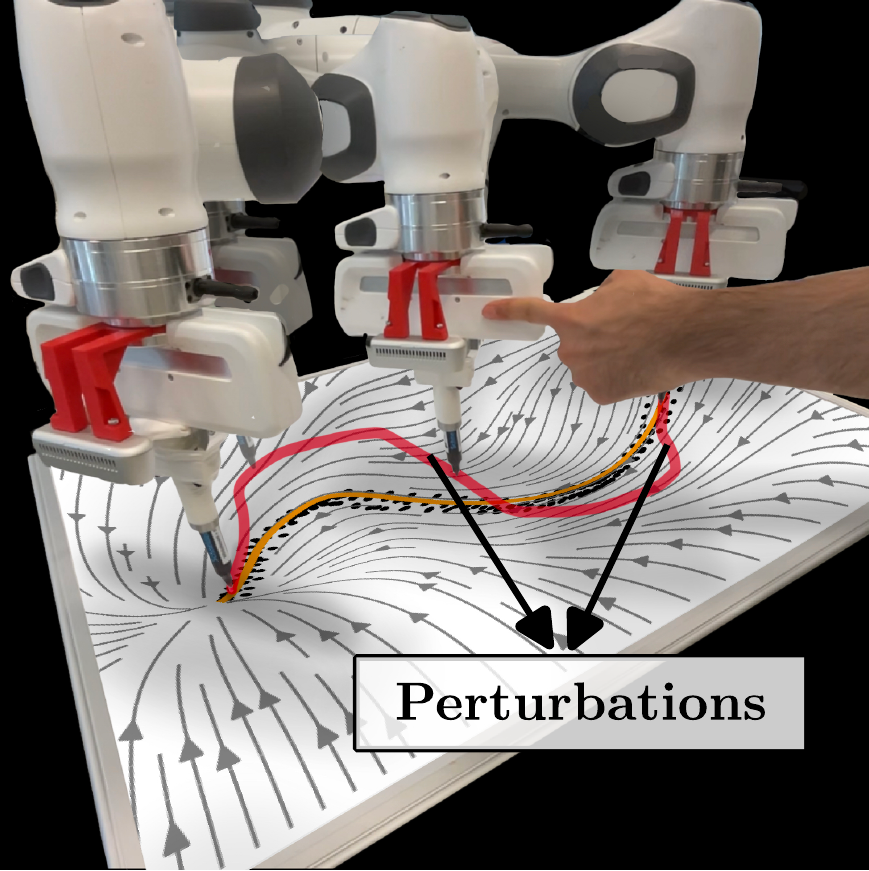}
  \end{subfigure}%
  \begin{subfigure}{.229\linewidth}
    \centering
    \includegraphics[width=1.0\textwidth]{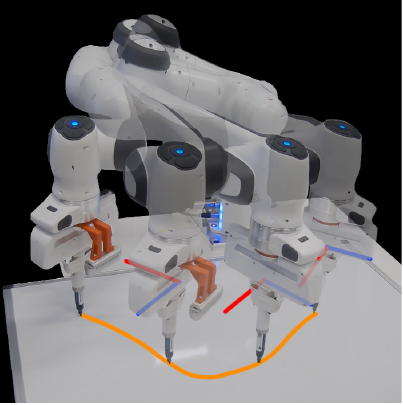}
  \end{subfigure}%
    \begin{subfigure}{.236\linewidth}
    \centering
    \includegraphics[width=1.0\textwidth]{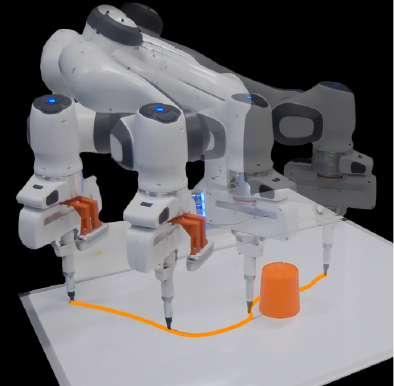}
  \end{subfigure}%
  \begin{subfigure}{.231\linewidth}
    \centering
    \includegraphics[width=1.0\textwidth]{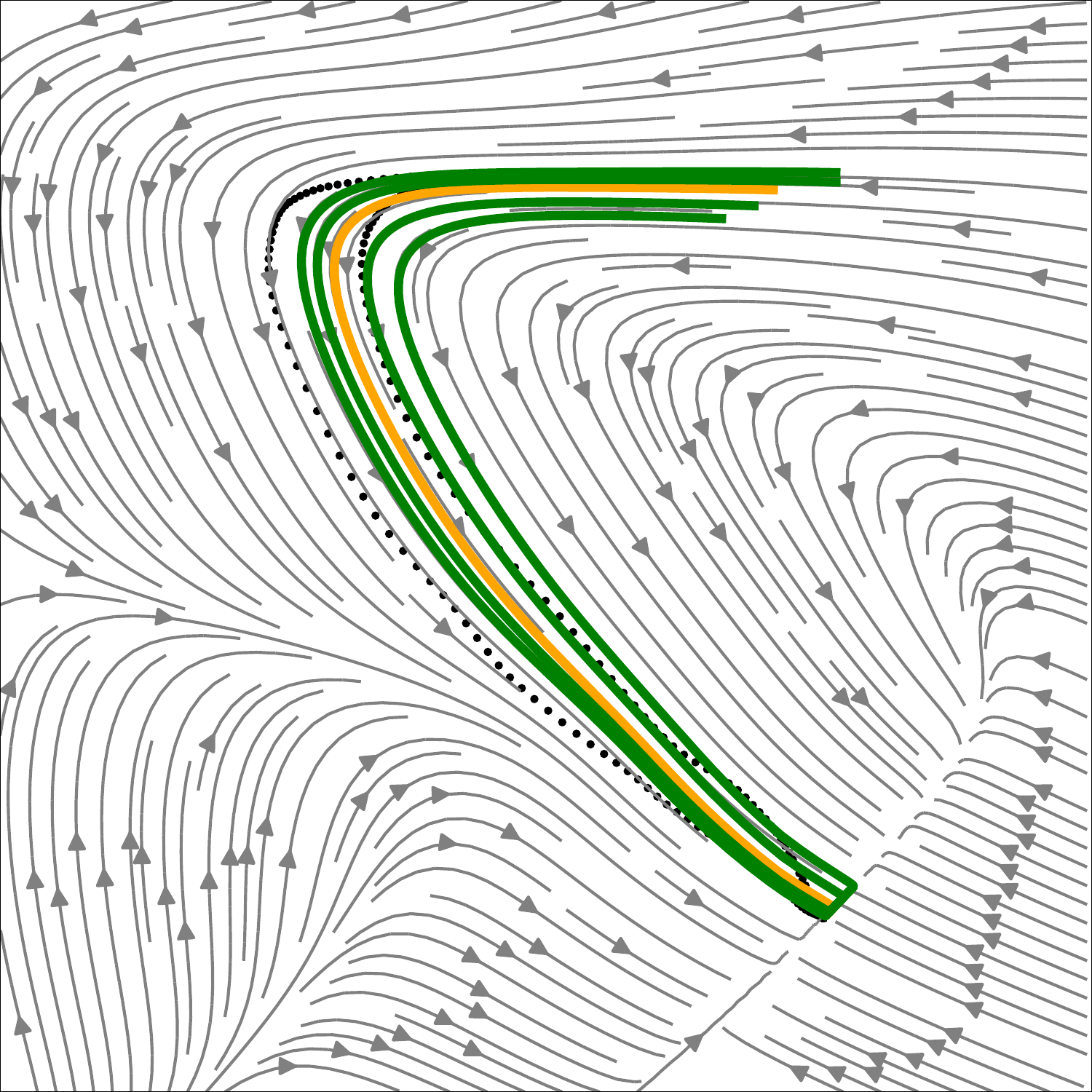}
  \end{subfigure}
  \caption{Robot experiments. The left two panels show robot experiments with orange and red paths illustrating unperturbed and perturbed motion. The first panel shows a learned vector field in $\R^3$ with a constant orientation. In the second panel, the vector field expands to $\R^3 \times \SO$, aligning the end-effector's orientation with the motion direction. Superimposed robot images depict frames of executed motion. 
  The third panel shows the robot successfully avoiding the obstacle (orange cylinder) using the modulation technique utilized in the vanilla NCDS. The right panel illustrates the contours of the latent vector field in the background, with the green and yellow curves representing the integral curves, while the black dots indicate the demonstrations in latent space.}
  \label{fig:real_robot_experiment_vannila}
\end{figure*}
\paragraph{Comparative Study on the LASA Dataset: } 
Table~\ref{tab:comparison_results} shows average DTWD distances among five generated integral curves and five demonstration trajectories for both NCDS and baseline methods. 
For the two-dimensional problem, both Euclideanizing flow and Imitation flow outperform NCDS and SEDS, though all methods perform quite well. 
To investigate stability, Fig.~\ref{fig:average_distance} displays the average distance over time among five integral curves starting from random nearby initial points for different methods. 
We observe that only for NCDS does this distance decrease monotonically, which indicates that it is the only contractive method.
To test how these approaches scale to higher-dimensional settings, we consider the LASA-4D and LASA-8D datasets, where we train NCDS with a two-dimensional latent representation. 
From Table~\ref{tab:comparison_results}, it is evident that the baseline methods quickly deteriorate as the data dimension increases, and only NCDS gracefully scales to higher dimensional data. 
This is also evident from Fig.~\ref{fig:traj_comparison_8d}, which shows the training data and reconstructed trajectories of different LASA-8D dimensions with different methods. 
These results clearly show the value of having a low-dimensional latent structure.

\begin{figure}[h!]
    \centering
    % First row: subfigures 0, 1, and 2
    \begin{subfigure}{0.33\linewidth}
        \centering
        \includegraphics[width=\linewidth]{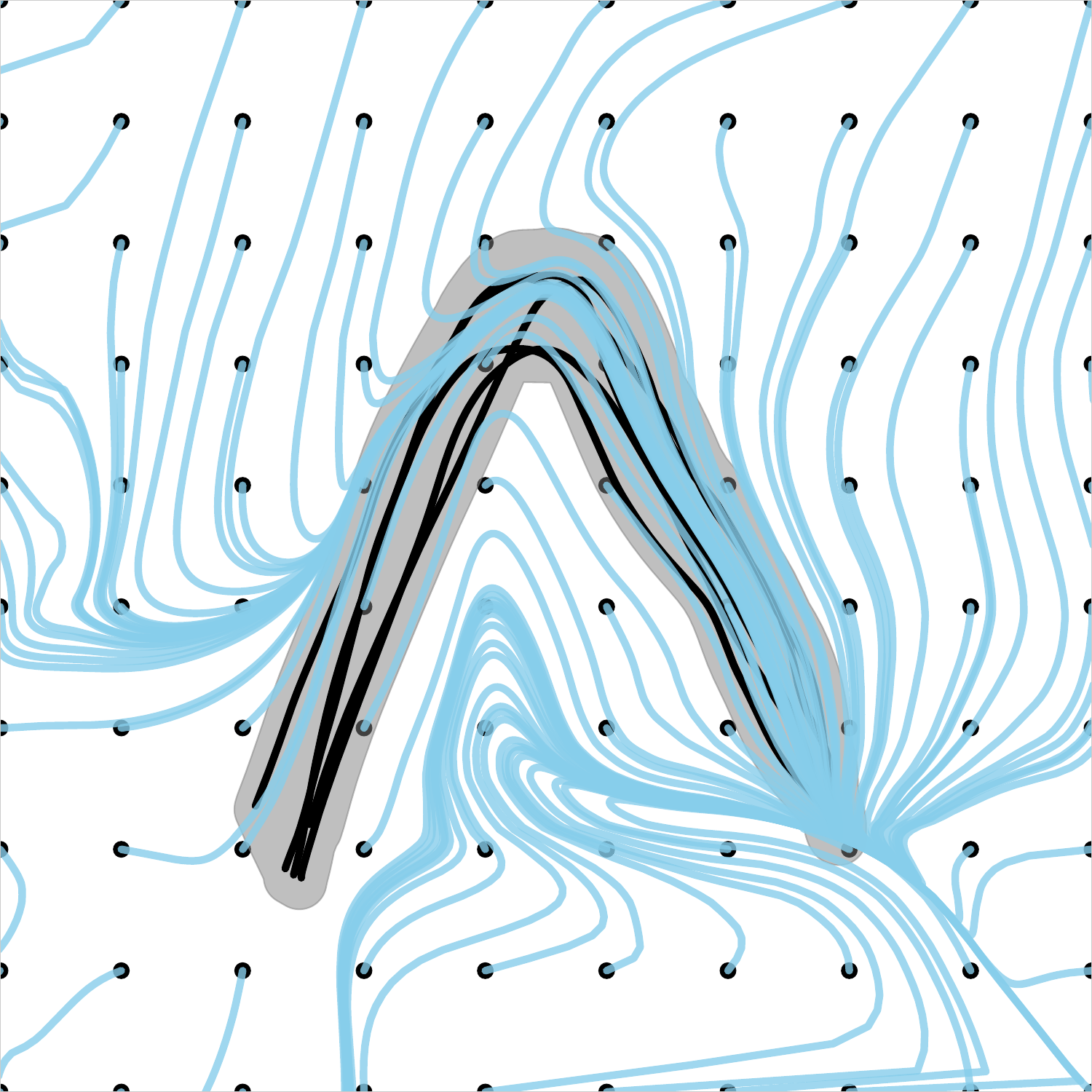}
        \label{fig:subfig2}
    \end{subfigure}\hspace{-5pt}
    \begin{subfigure}{0.33\linewidth}
        \centering
        \includegraphics[width=\linewidth]{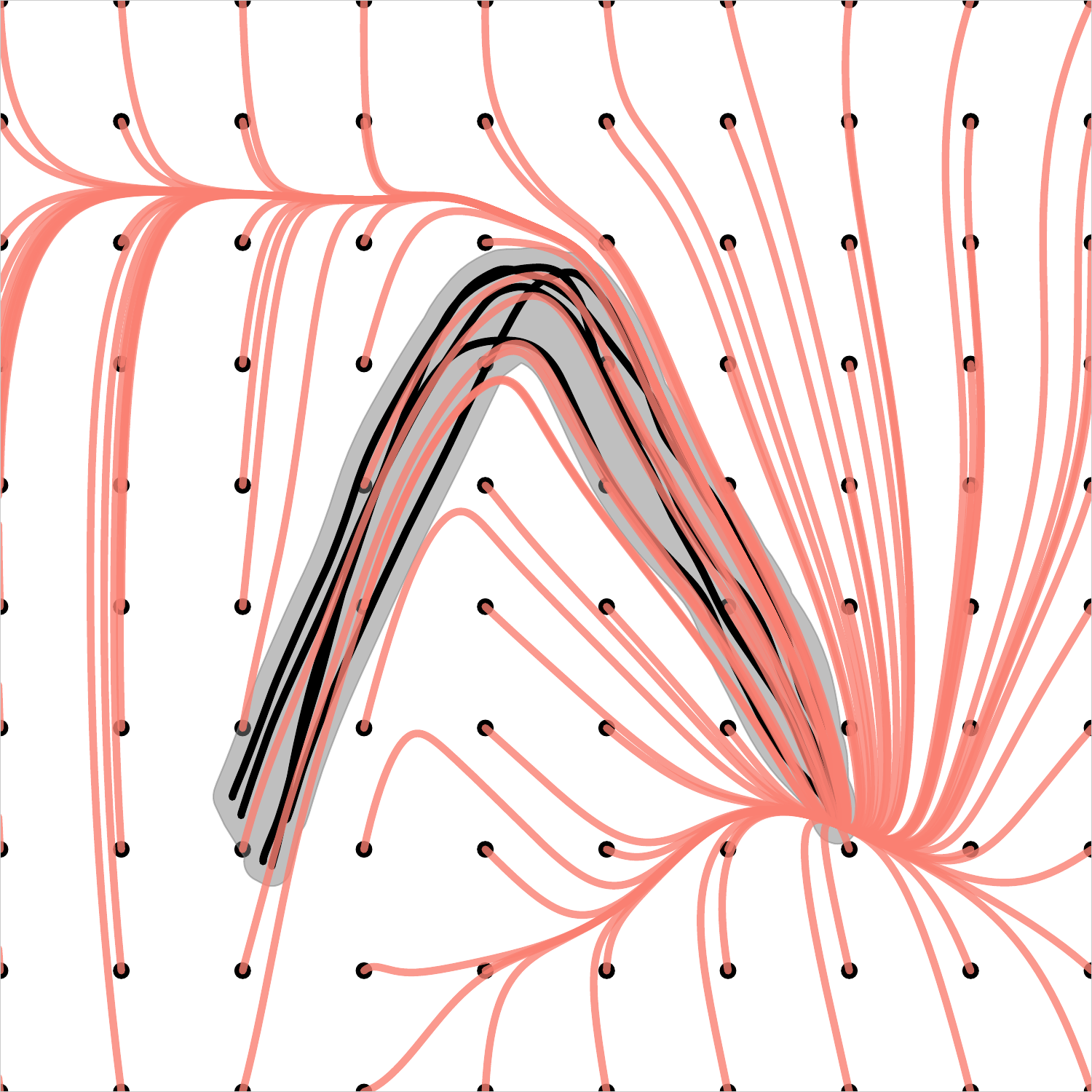}
        \label{fig:subfig0}
    \end{subfigure}\hspace{-5pt}
    \begin{subfigure}{0.33\linewidth}
        \centering
        \includegraphics[width=\linewidth]{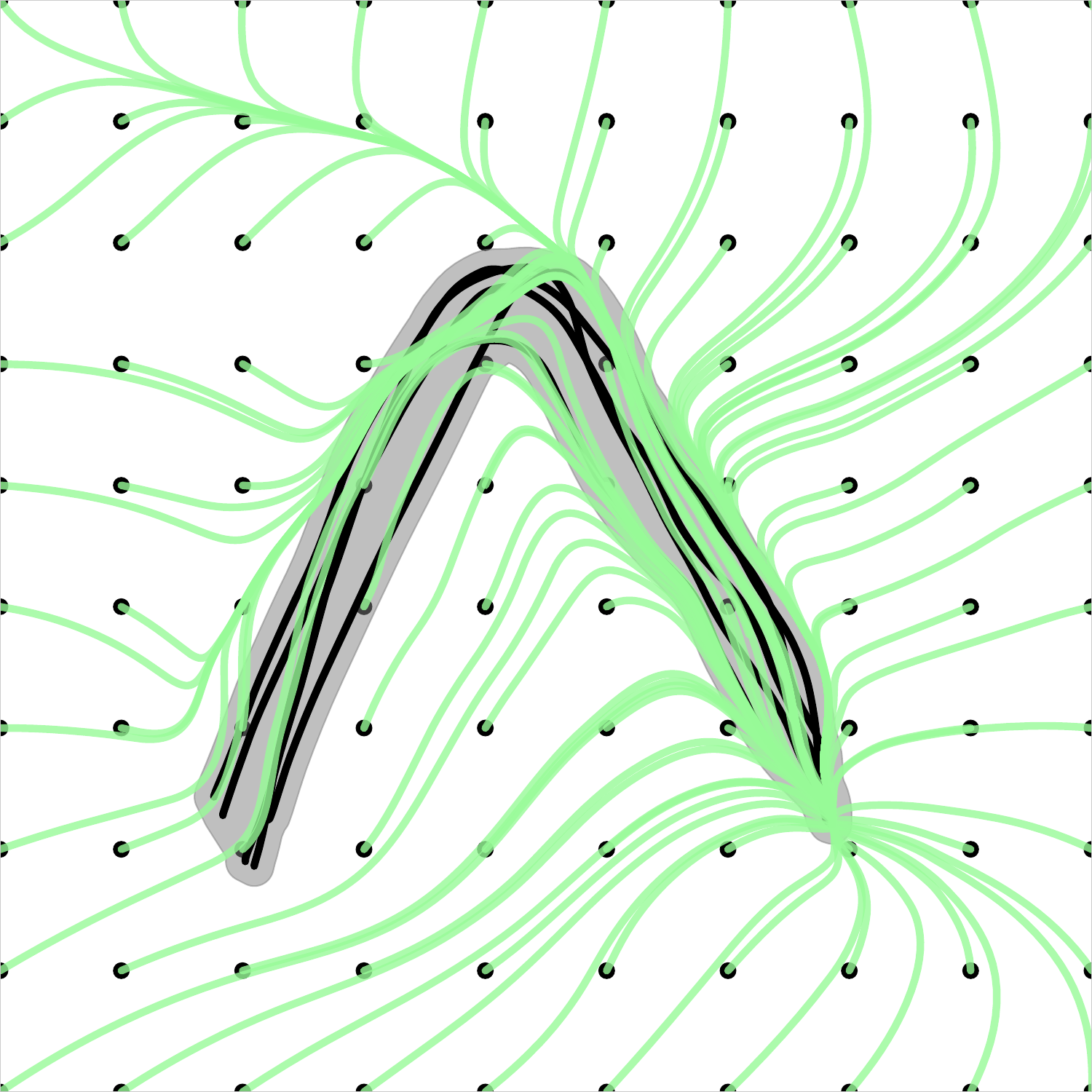}
        \label{fig:subfig1}
    \end{subfigure}\\
    \vspace{-13pt}
    \begin{subfigure}{0.33\linewidth}
        \centering
        \includegraphics[width=\linewidth]{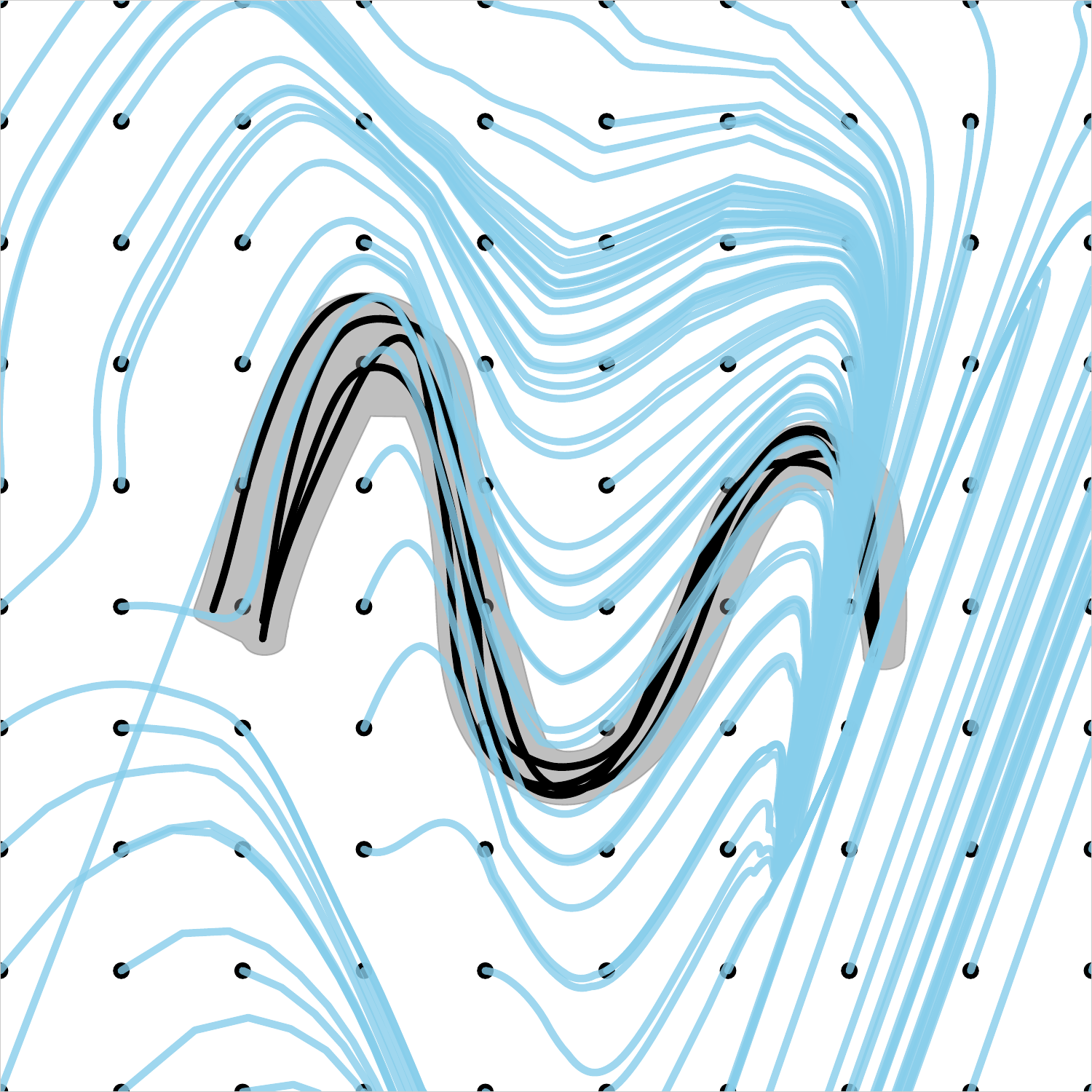}
        \label{fig:subfig2}
    \end{subfigure}\hspace{-5pt}
    \begin{subfigure}{0.33\linewidth}
        \centering
        \includegraphics[width=\linewidth]{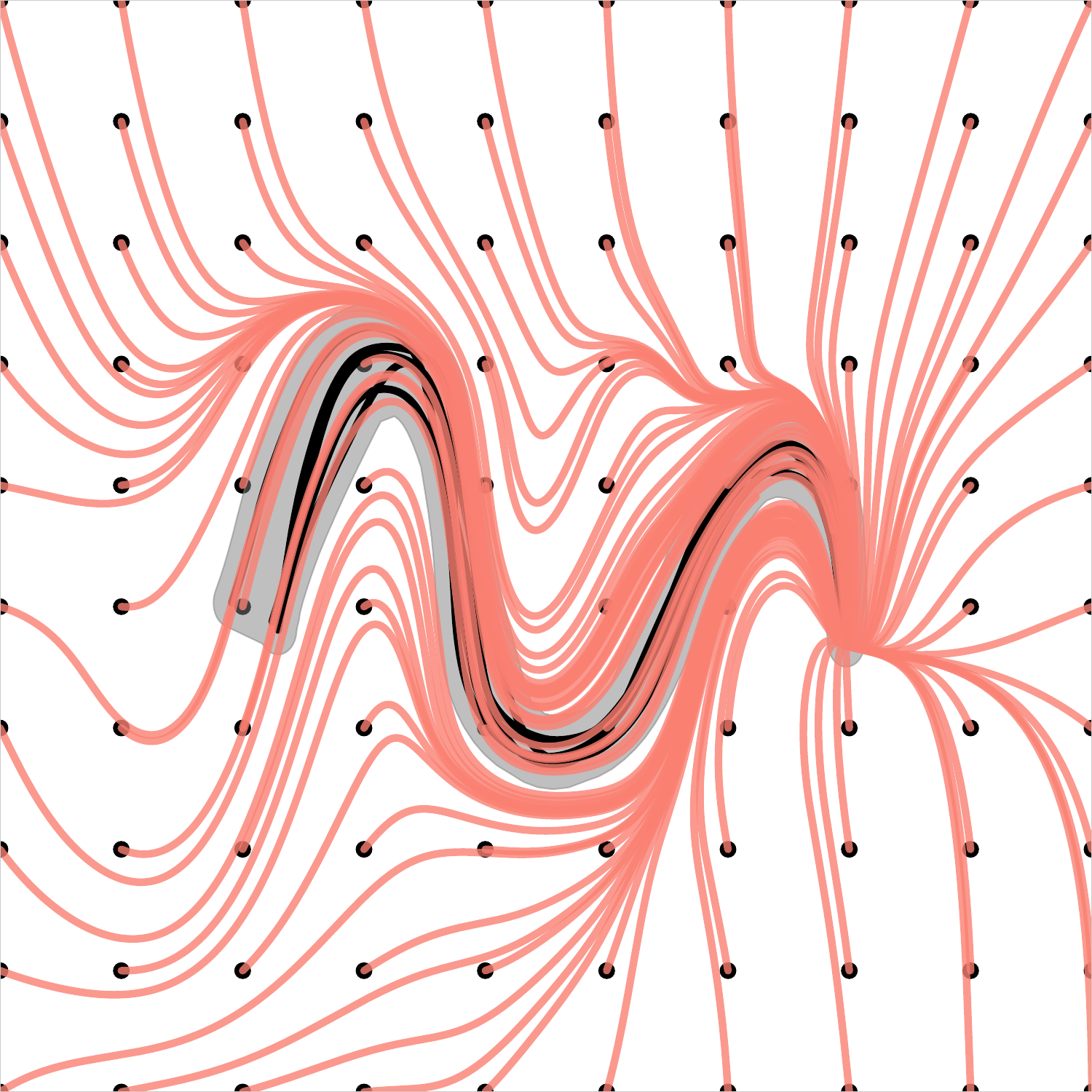}
        \label{fig:subfig0}
    \end{subfigure}\hspace{-5pt}
    \begin{subfigure}{0.33\linewidth}
        \centering
        \includegraphics[width=\linewidth]{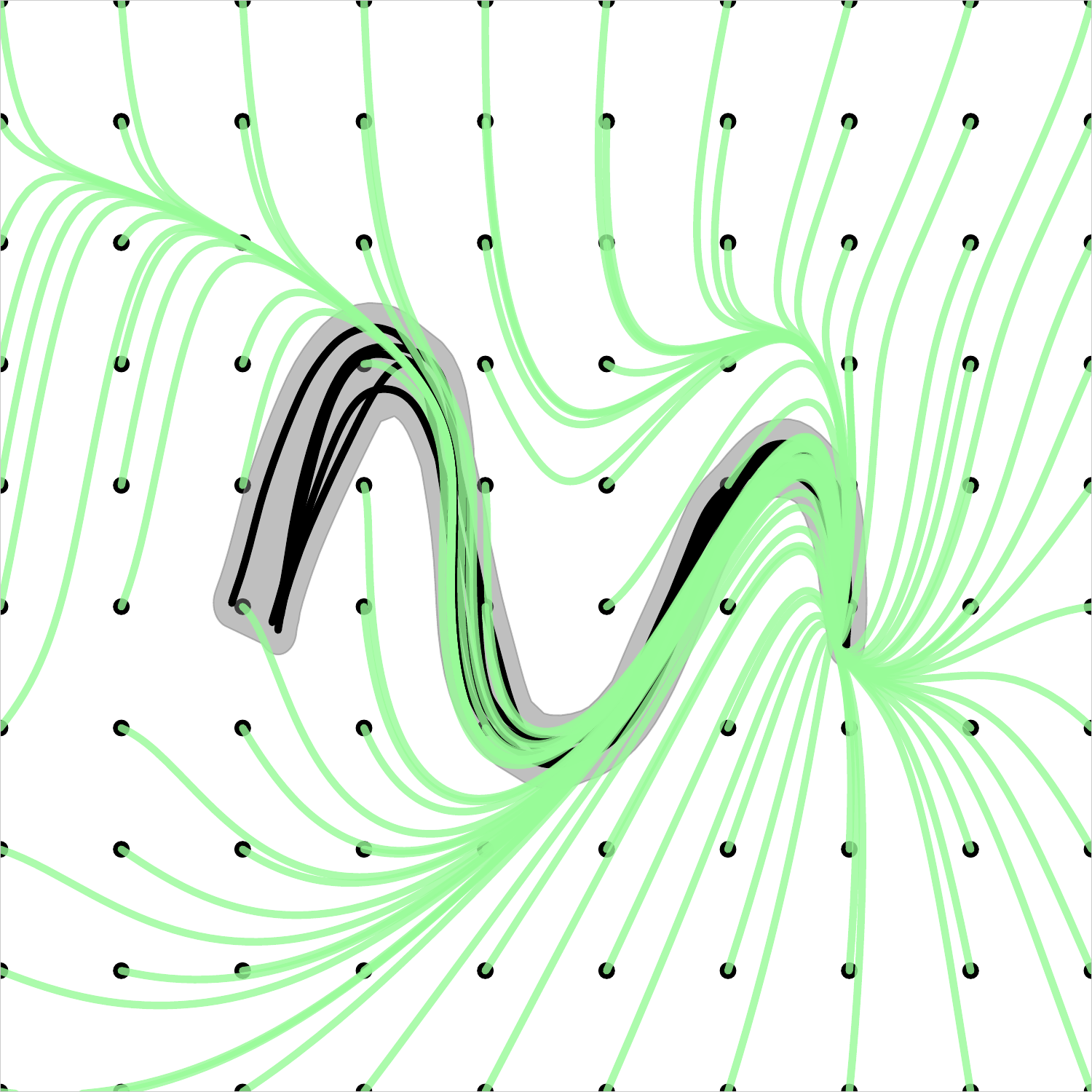}
        \label{fig:subfig1}
    \end{subfigure}
    \caption{Comparison of vector fields generated by NCDS and ELCD~\citep{jaffe2024:ELCD}.  
    The left, middle, and right columns depict the integral curves produced by ELCD, vanilla NCDS, and NCDS with state-independent regularization, respectively.
    A total of $100$ integral curves, each consisting of $1000$ points, are generated on an equidistant grid around the demonstration data and evaluated on two LASA datasets: \emph{Angle} and \emph{Sine}.}
    \label{fig:ELCD:Comparison}
\end{figure}

\begin{figure}[h!]
    \centering
    \begin{minipage}{\linewidth}
        \centering
        \includegraphics[width=\linewidth]{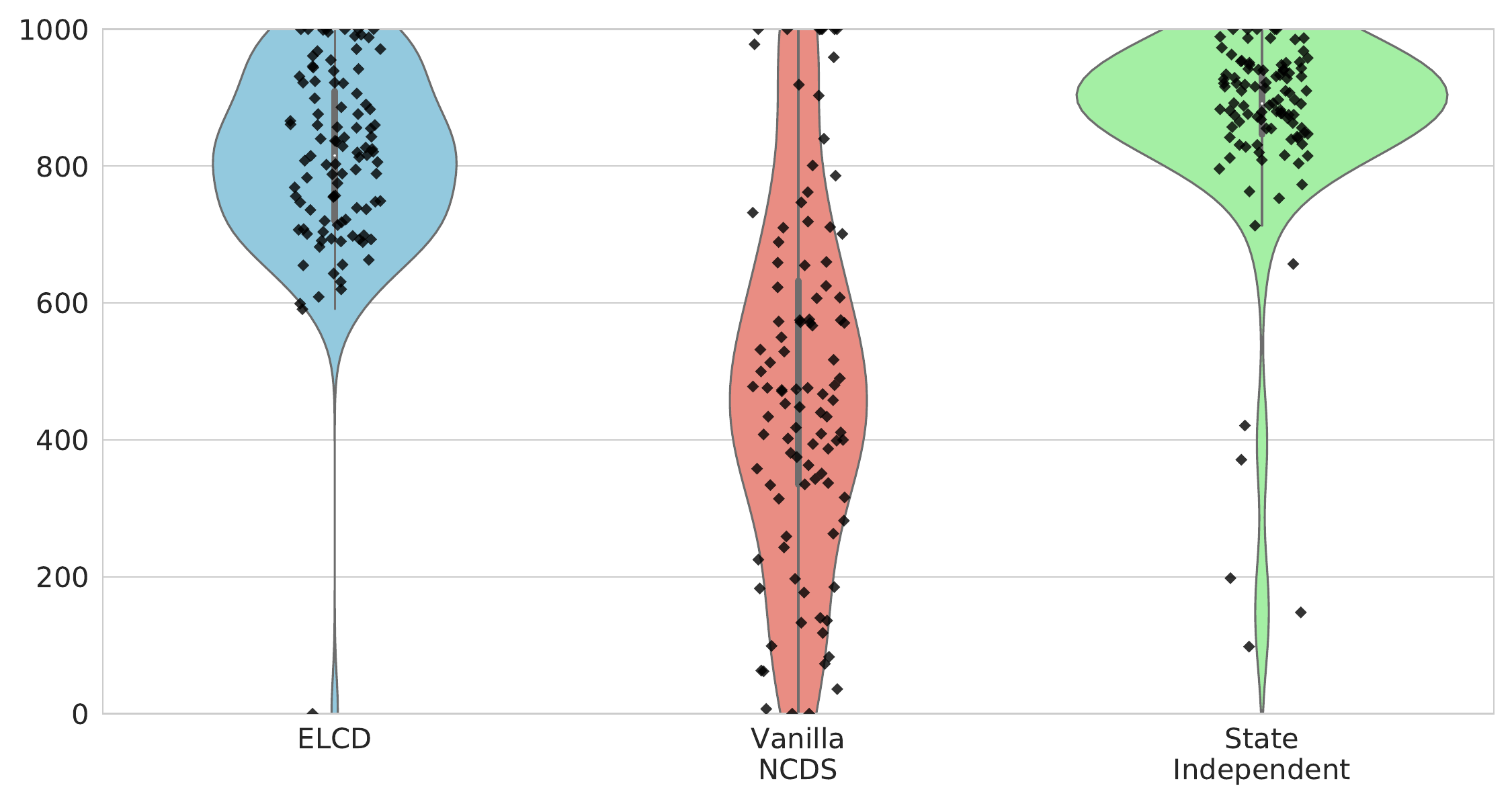}
    \end{minipage}\\
    \vspace{-14pt}
        \begin{minipage}{\linewidth}
        \centering
        \includegraphics[width=\linewidth]{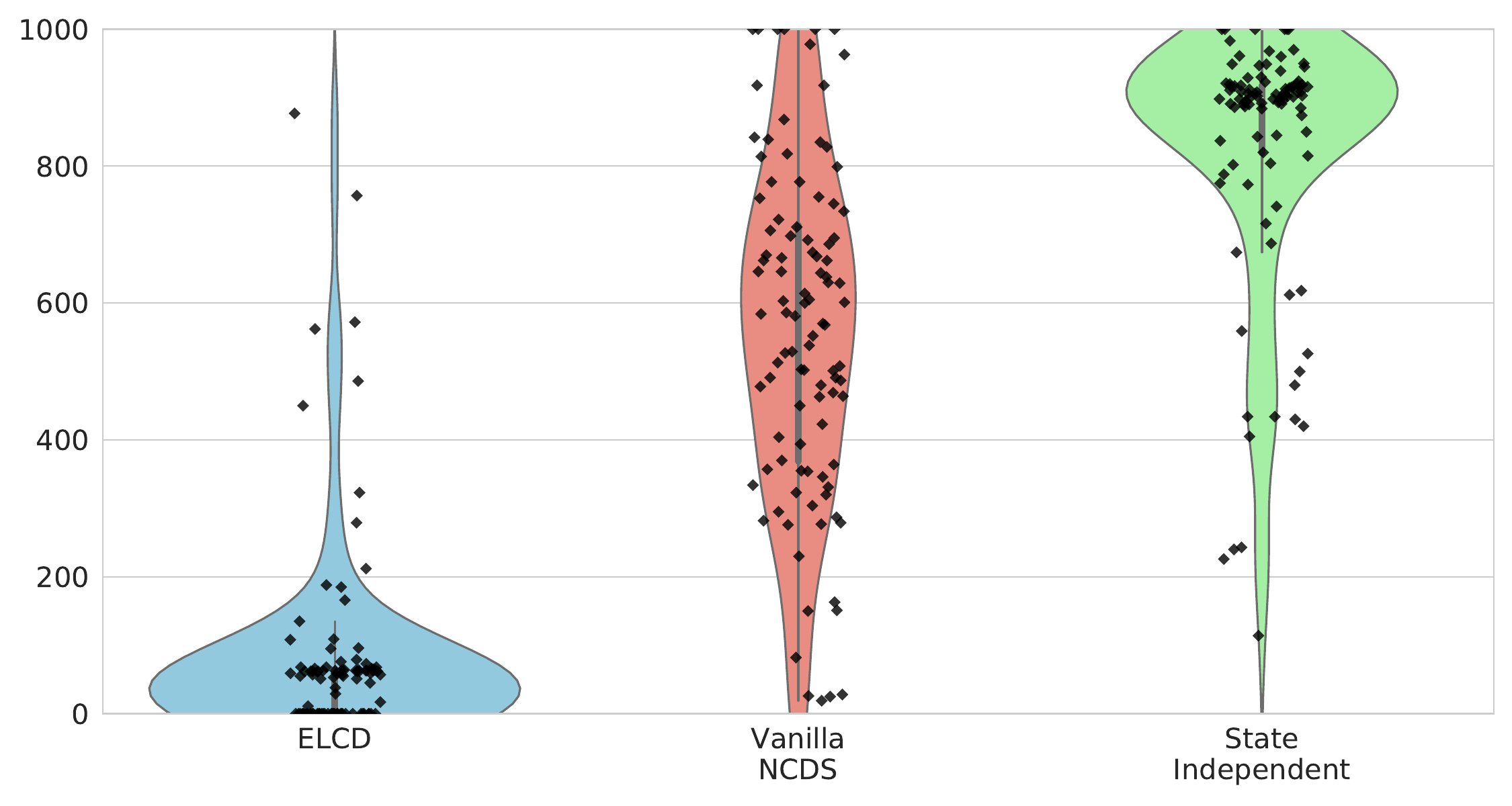}
    \end{minipage}\\
        \vspace{-14pt}
    \begin{minipage}{\linewidth}
        \centering
        \renewcommand{\arraystretch}{1.2} % Increase space between rows
        \setlength{\tabcolsep}{20pt} % Increase space between columns
        \begin{tabular}{l c c}
            \rowcolor{gray!15}
            \textbf{Method} & \textbf{Angle} & \textbf{Sine} \\
            \rowcolor{customblue!80}
            ELCD & 809 & 82   \\
            \rowcolor{customred!80}
            Vanilla NCDS & 484 & 561  \\
            \rowcolor{customgreen!80}
            Independent Reg.& \textbf{862} & \textbf{824}  \\
        \end{tabular}
    \end{minipage}
    \caption{Comparison of contractivity across methods. The top and middle plots show the number of trajectory points inside the demonstration region for models trained on \emph{Angle} and \emph{Sine} datasets, respectively, where higher values indicate stronger contraction. The bottom table presents the average number of points per integral curve within the demonstration region.}
    \label{fig:ELCD:convexhull_time}
\end{figure}

Next, we compare our method to ELCD~\citep{jaffe2024:ELCD} using two different LASA datasets. Figure~\ref{fig:ELCD:Comparison} illustrates $100$ integral curves, each generated on an equidistant grid with $1000$ time steps. The results indicate that all methods successfully recover the demonstration while maintaining contractivity. However, ELCD shows unnecessary deviations outside the data support. Instead of smoothly contracting toward the demonstrations region, some trajectories take detours, making extra turns or oscillations before eventually converging.
Figure~\ref{fig:ELCD:convexhull_time}-\emph{top} quantifies how many of the $1000$ points in each integral curve remain inside the demonstration region. A higher value means that the trajectory stays within the data support for more time steps, indicating stronger contraction. The table in Fig.~\ref{fig:ELCD:convexhull_time}-\emph{bottom} presents the average number of points spent within the demonstration region across all integral curves. The results show that NCDS with state-independent regularization keeps trajectories more focused on the demonstrations region. Particularly when using the \emph{Angle} dataset, ELCD exhibits stronger contraction than vanilla NCDS, as shown in Fig.~\ref{fig:ELCD:convexhull_time}-\emph{bottom}. However, as illustrated in Fig.~\ref{fig:ELCD:Comparison}, ELCD also introduces more unnecessary motion patterns outside the demonstration region.
Moreover, Fig.~\ref{fig:ELCD:convexhull_time}-\emph{middle} shows the results for the \emph{Sine} dataset, where ELCD, despite our efforts to enhance its performance, does not perform as well. This trend is further supported by Fig.~\ref{fig:ELCD:convexhull_time}-\emph{bottom}. 

\paragraph{Comparative study on the robot dataset:}
To further compare our method, we consider a robotic setting where we evaluate all the methods on a real dataset of joint-space trajectories collected on a 7-DoF robot. 
The last column of Table~\ref{tab:comparison_results} shows that only NCDS scales gracefully to high-dimensional joint-space data, and outperforms the baseline methods. 
The supplementary material includes a video showcasing the simulation environment in which the robot executes a drawing task, depicting \textsf{V}-shape trajectories on the table surface.

\paragraph{Dimensionality and execution time:} Table~\ref{tab:time_comp} compares $5$ different experimental setups. 
Both the Variational Autoencoder (VAE) and Jacobian network $\hat{\bm{J}}_{\bm{\theta}}$, were implemented using PyTorch \citep{Paszke2019Pytorch}. The VAE employs an injective generator based on $\mathcal{M}$-flows~\citep{Brehmer2020MFlow}, using rational-quadratic neural spline flows with three coupling layers.
The experiments were conducted on a system with an Intel\textsuperscript{\textregistered} Xeon\textsuperscript{\textregistered} Processor E3-1505M v6 (8M Cache, $3.00$ GHz), $32$ GB of RAM, and an NVIDIA Quadro M2200 GPU.

Table~\ref{tab:time_comp} reports the average execution time when dimensionality reduction is not used and the dynamical system is learned directly in the input space. 
As the results show, increasing the dimensionality of the input space from $2$ to $8$ entails a remarkable $40$-fold increase in execution time for a single integration step. This empirical relationship reaffirms the inherent computational intensity tied to Jacobian-based computations. 
This table also shows that our VAE distinctly mitigates the computational burden. 
In comparison, the considerable advantages of integrating the contractive dynamical system with the VAE into the full pipeline become evident as it leads to a significant halving of the execution time in the $8$D scenario. 
We further find that increasing the dimensionality from $8$ to $44$ results in a sixfold increase in running time. 
The results indicate that our current system may not achieve real-time operation. 
However, the supplementary video from our real-world robot experiments demonstrates that the system's rapid query response time is sufficiently fast to control the robot arm, devoid of any operational concerns.
\subsection{Generalization outside the decoder’s manifold}
\label{sec:Outside_Manifold}
In our framework, although the latent and ambient spaces share the same overall dimensionality due to the padding operation, the encoder projects data into a lower-dimensional representation by discarding certain dimensions (via unpadding). These discarded (or extra) dimensions, while not directly used in the final projection, retain information that is indicative of whether a point lies on the decoder’s manifold. We define the decoder’s manifold as the set of latent points for which the extra dimensions are exactly zero. Formally, let $\z=[\z_m,\z_e]$ be the latent vector,
where $\z_m$ represents the dimensions used for the final projection and $\z_e$ the extra dimensions, then a point is considered on the decoder's manifold if $\parallel \! \z_e \! \parallel = 0$.

Conversely, any nonzero values in $\z_e$ indicate that the corresponding state lies outside the decoder’s manifold.
Although deviations from the decoder’s manifold are generally minimal, in certain real-robot scenarios measurement noise or ambient perturbations can occasionally cause the latent representation to stray from the decoder's manifold. In these cases, our approach incorporates a transition phase to correct the deviation. Specifically, if a latent point is identified as being off the manifold, we compute a transition vector that incrementally steers the point back toward the manifold. This is achieved by integrating along a path in latent space until the condition $\parallel \! \z_e \! \parallel \simeq 0$ is effectively reached, thereby ensuring compatibility with the decoder’s assumptions.
This corrective mechanism is not invoked routinely but serves as an important safeguard to enhance system robustness in the rare instances when deviations occur.

\begin{figure*}[t!]
    \centering
    % First row: subfigures 0, 1, and 2
    \begin{subfigure}[b]{0.24\linewidth}
        \centering
        \includegraphics[width=\linewidth]{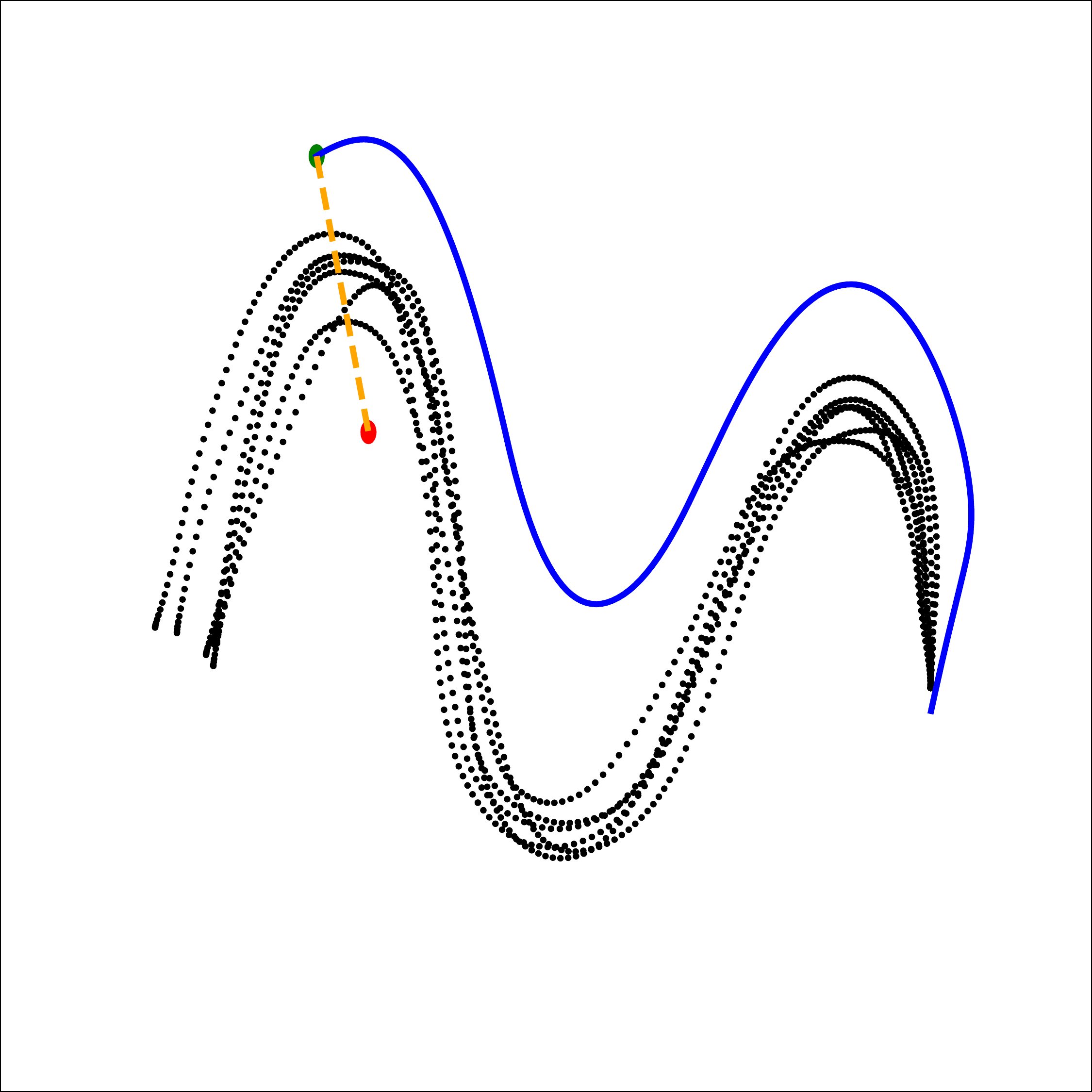}
        \label{fig:subfig0}
    \end{subfigure}
    \begin{subfigure}[b]{0.24\linewidth}
        \centering
        \includegraphics[width=\linewidth]{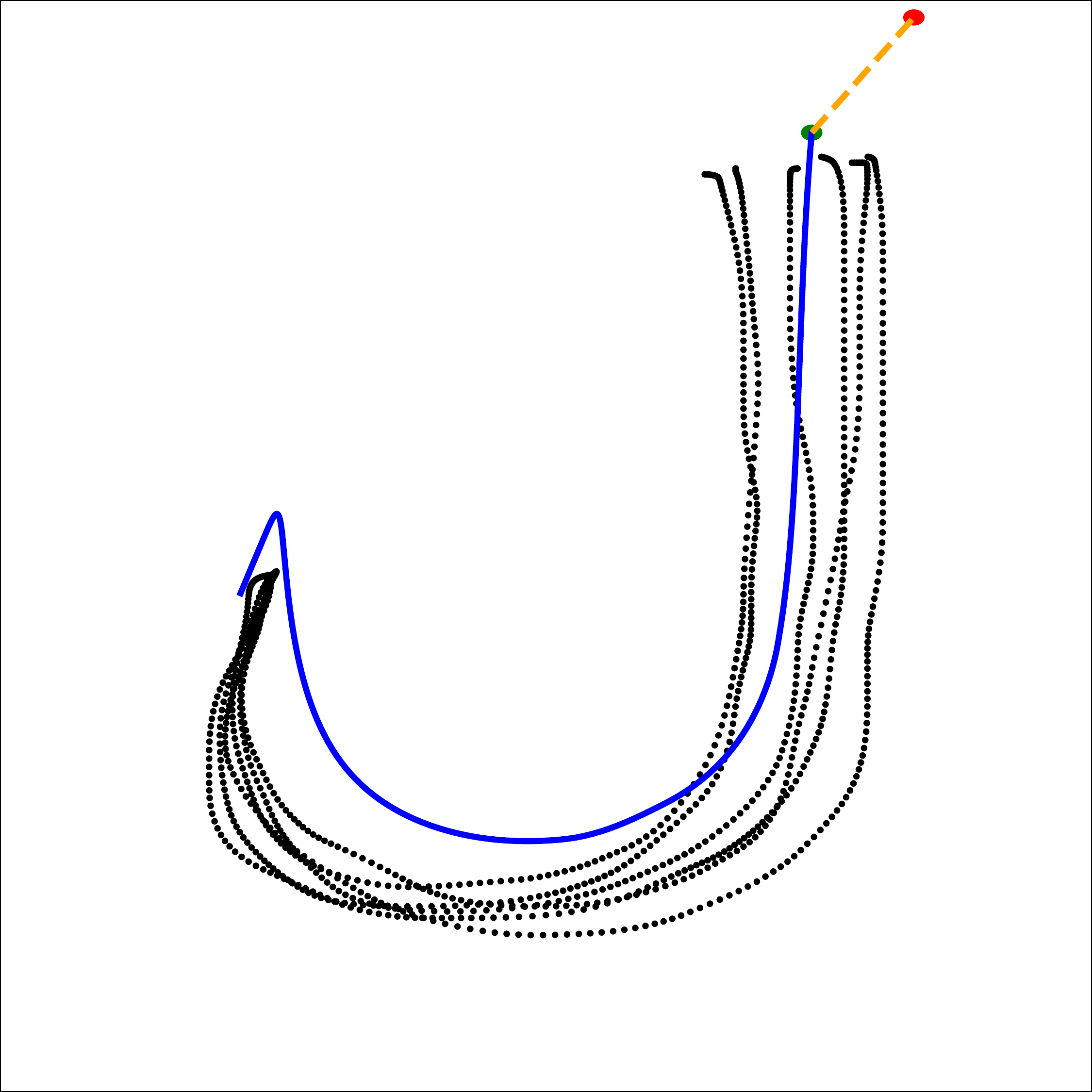}
        \label{fig:subfig1}
    \end{subfigure}
    \begin{subfigure}[b]{0.24\linewidth}
        \centering
        \includegraphics[width=\linewidth]{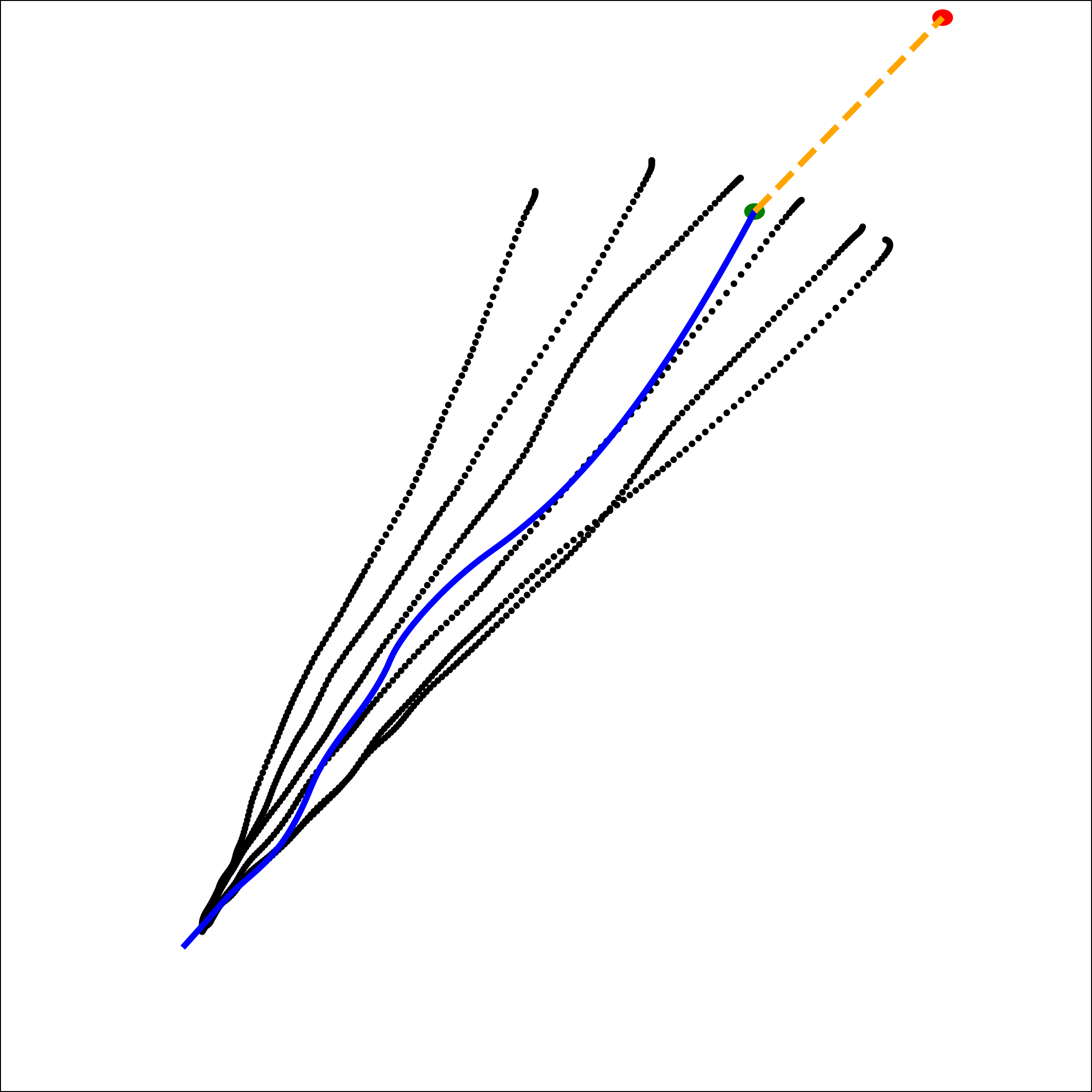}
        \label{fig:subfig2}
    \end{subfigure}
    \begin{subfigure}[b]{0.24\linewidth}
        \centering
        \includegraphics[width=\linewidth]{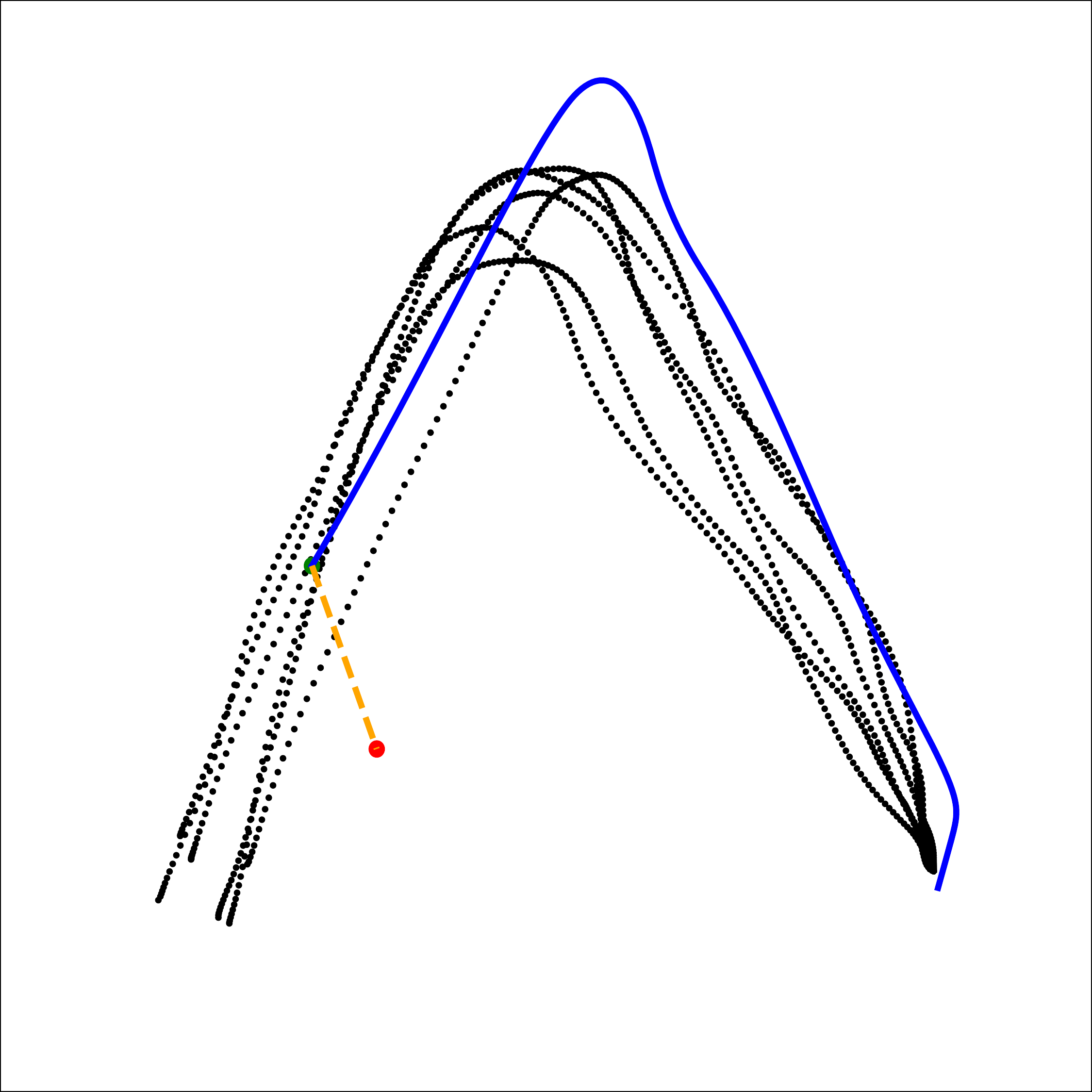}
        \label{fig:subfig3}
    \end{subfigure}

    \caption{Generalization under deviation from the decoder's manifold using the 8D LASA dataset. Red circles indicate initial positions, the orange dashed line shows the transition vector to the manifold (green circle), and the blue curve represents the integral curve.}
    \label{fig:outside_of_manifold_8D}
\end{figure*}

Figure~\ref{fig:outside_of_manifold_8D} illustrates this procedure using the 8D LASA dataset. In the figure, red markers denote the initial latent positions that deviate from the demonstration’s starting point. The orange dashed line represents the transition vector that guides the point toward the manifold, and the green marker indicates the latent point after successful projection onto the manifold. The blue curve depicts the integrated path taken during the transition phase.
By quantifying the deviation from the manifold e.g., via the Euclidean norm $\parallel \! \z_e \! \parallel$, we can effectively measure and correct the discrepancy between the actual latent representation and the ideal manifold state.

\subsection{Learning robot motion skills}
To demonstrate the application of NCDS to robotics, we conducted several experiments on a $7$-DoF Franka-Emika robotic manipulator, where an operator kinesthetically teaches the robot drawing tasks.
Both the Variational Autoencoder (VAE) and the Jacobian network $\hat{\bm{J}}_{\bm{\theta}}$, were implemented using the PyTorch framework \citep{Paszke2019Pytorch}.
For the VAE, we used an injective generator based on $\mathcal{M}$-flows~\citep{Brehmer2020MFlow}, specifically employing rational-quadratic neural spline flows with three coupling layers. 
In each coupling transform, half of the input values underwent element-wise transformation through a monotonic rational-quadratic spline.
These splines were parameterized by a residual network with two residual blocks, each containing a hidden layer of $30$ nodes.
We used \textsf{Tanh} activations throughout the network, without batch normalization or dropout. 
The rational-quadratic splines used ten bins, evenly spaced over the range $(-10, 10)$.
To learn these skills, we employ the same architecture described in Sec.~\ref{sec:res:lasa}.
The learned dynamics are executed using a Cartesian impedance controller. 
First, we demonstrate $7$ sinusoidal trajectories while keeping the end-effector orientation constant, such that the robot end-effector dynamics only evolve in $\mathbb{R}^3$.
As Fig.~\ref{fig:real_robot_experiment_vannila}-\emph{left} shows, the robot was able to successfully reproduce the demonstrated dynamics by following the learned vector field encoded by NCDS.
Importantly, we also tested the extrapolation capabilities of our approach by introducing physical perturbations to the robot end-effector, under which the robot satisfactorily adapted and completed the task (see the supplementary video).
The second experiment tests NCDS ability to learn orientation dynamics evolving in $\mathbb{R}^3 \times \SO$, i.e.\@ full-pose end-effector movements. 
We collect several \textsf{V}-shape trajectories on a table surface on which the robot end-effector always faces the direction of the movement, as shown in the second panel from the left in Fig.~\ref{fig:real_robot_experiment_vannila}. 
However, the orientation trajectories of the robot experiments tend to show relatively simple dynamics.
To evaluate the performance of this setup on a more complex dataset, we generate a synthetic LASA dataset in $\mathbb{R}^3 \times \SO$.
To introduce complexity within $\SO$, we project the LASA datasets onto a $3$-sphere, thus generating synthetic quaternion data.
Although we consider orientation in $\SO$ here, it is worth noting that NCDS can also handle quaternions directly.
Next, we transform these into rotation matrices on $\SO$, and apply the $\operatorname{Log}$ map to project onto the Lie algebra. 
To construct the position data in $\mathbb{R}^3$, we embed the 2D LASA dataset in $\mathbb{R}^3$ by concatenating with a zero. 
Together, this produces a state vector $\x$ in $\mathbb{R}^6$.
To increase the level of complexity, we employ two distinct LASA letter shapes for the position and orientation dimensions.\looseness=-1
Figure~\ref{fig:real_robot_experiment_vannila}-\emph{right}, depicts the 2D latent dynamical system, where the black dots represent the demonstrations. Moreover, the yellow integral curves start at the initial point of the demonstrations, while the green integral curves start at random points around the data support. 
This shows that NCDS can learn complex nonlinear contractive dynamical systems in $\mathbb{R}^3 \times \SO$.

\subsubsection{Obstacle avoidance in vanilla NCDS:}
\label{sec:obstacle_avoidance_experiment}

Unlike baselines methods, our vanilla NCDS framework also provides dynamic collision-free motions.
To demonstrate this in practice, we experiment with a real robot, where we block motion demonstrations with an object.
Here, the modulation matrix is computed based on the formulation in~\eqref{eq:modulation}. 
The third panel of Fig.~\ref{fig:real_robot_experiment_vannila} shows the obstacle avoidance in action as the robot's end-effector navigates around the orange cylinder.
This demonstrates how the dynamic modulation matrix approach leads to obstacle-free trajectories while guaranteeing contraction.
\subsection{Learning human motion}
\label{sec:human_motion}
To assess the model's performance in a more complex environment, we consider the KIT Whole-Body Human Motion Database~\citep{KrebsMeixner2021:HumanMotionDataset}.
This captures subject-specific motions, which are standardized based on the subject's height and weight using a reference kinematics and dynamics model known as the master motor map (MMM). 

\begin{figure*}
  \centering
    \begin{subfigure}{.30\textwidth}
        \centering
        \includegraphics[width=1.0\textwidth]{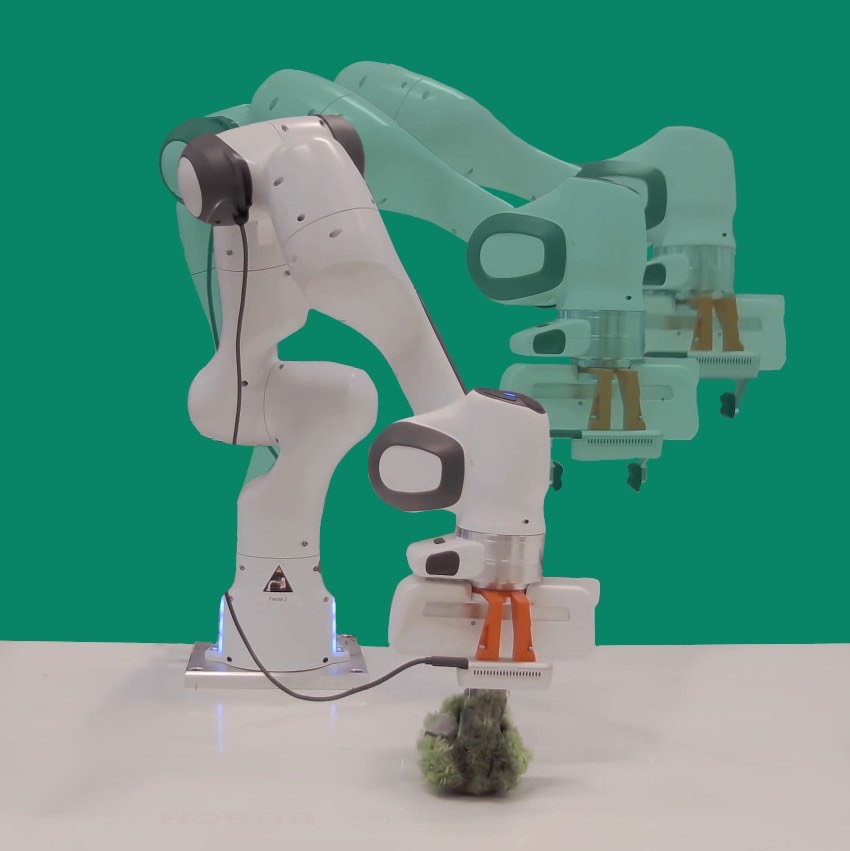}
  \end{subfigure}%  
  \space
  \begin{subfigure}{.30\textwidth}
    \centering
    \includegraphics[width=1.0\textwidth]{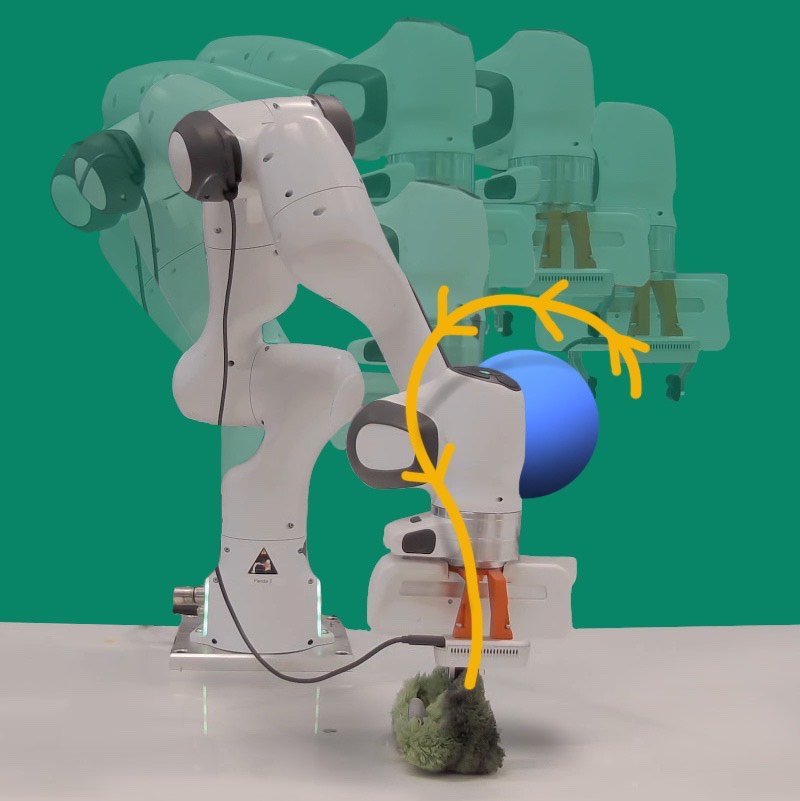}
  \end{subfigure}%
  \space      
  \begin{subfigure}{.30\textwidth}
        \centering
        \includegraphics[width=1.0\textwidth]{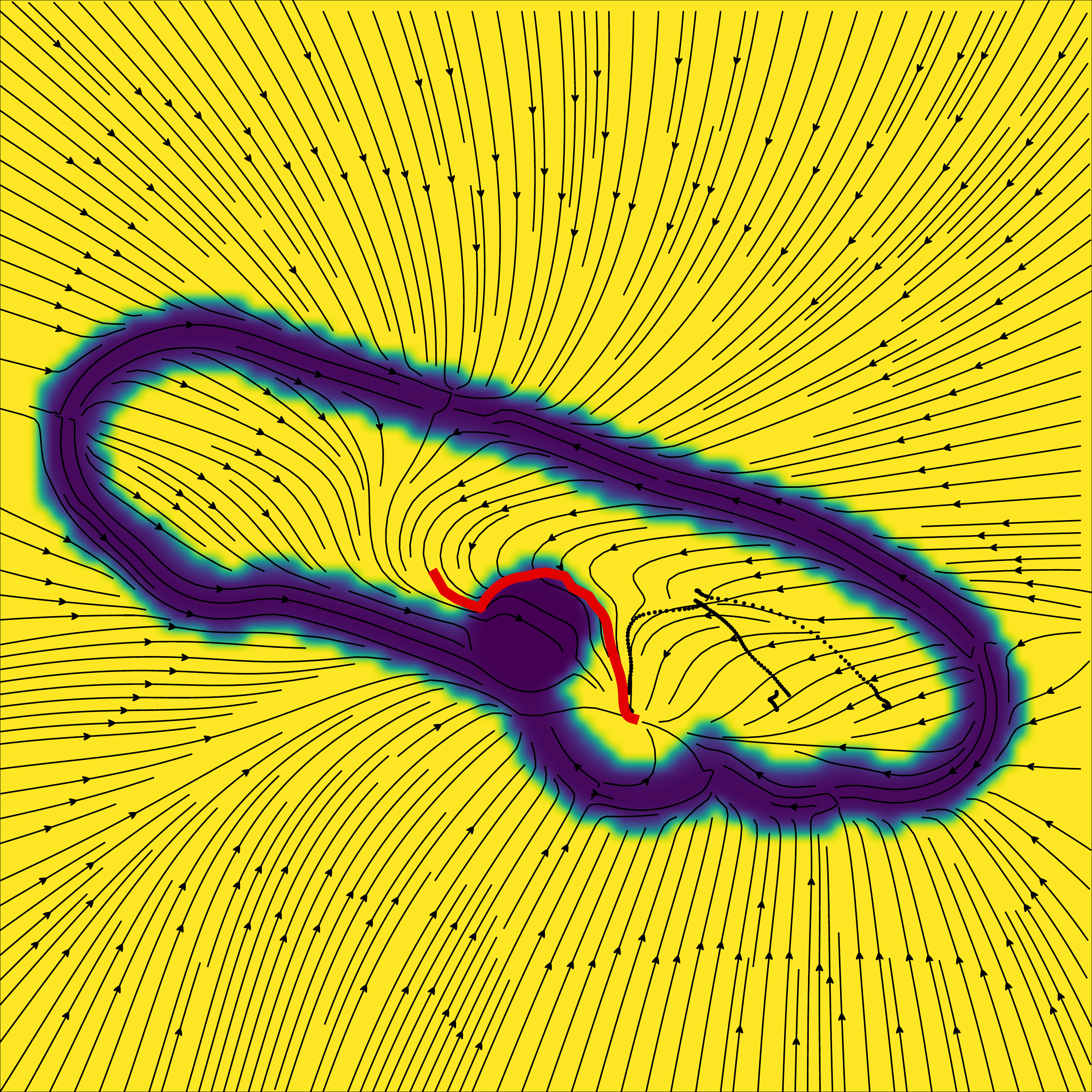}
  \end{subfigure}%  
    \caption{Obstacle avoidance via a Riemannian pullback metric. 
        \emph{Left}: The robot performs a grasping task without obstacle avoidance. 
        \emph{Middle}: The robot navigates around the obstacle by avoiding both the obstacle and the unsafe regions of the manifold. 
        \emph{Right}: The latent modulated vector field is locally reshaped around the obstacle and in areas outside the data support.
        }
  \label{fig:real_robot_riemannian_modulation}
\end{figure*}
\begin{figure}
    \centering
    \begin{subfigure}{0.48\linewidth}
        \centering
        \includegraphics[width=1.0\linewidth]{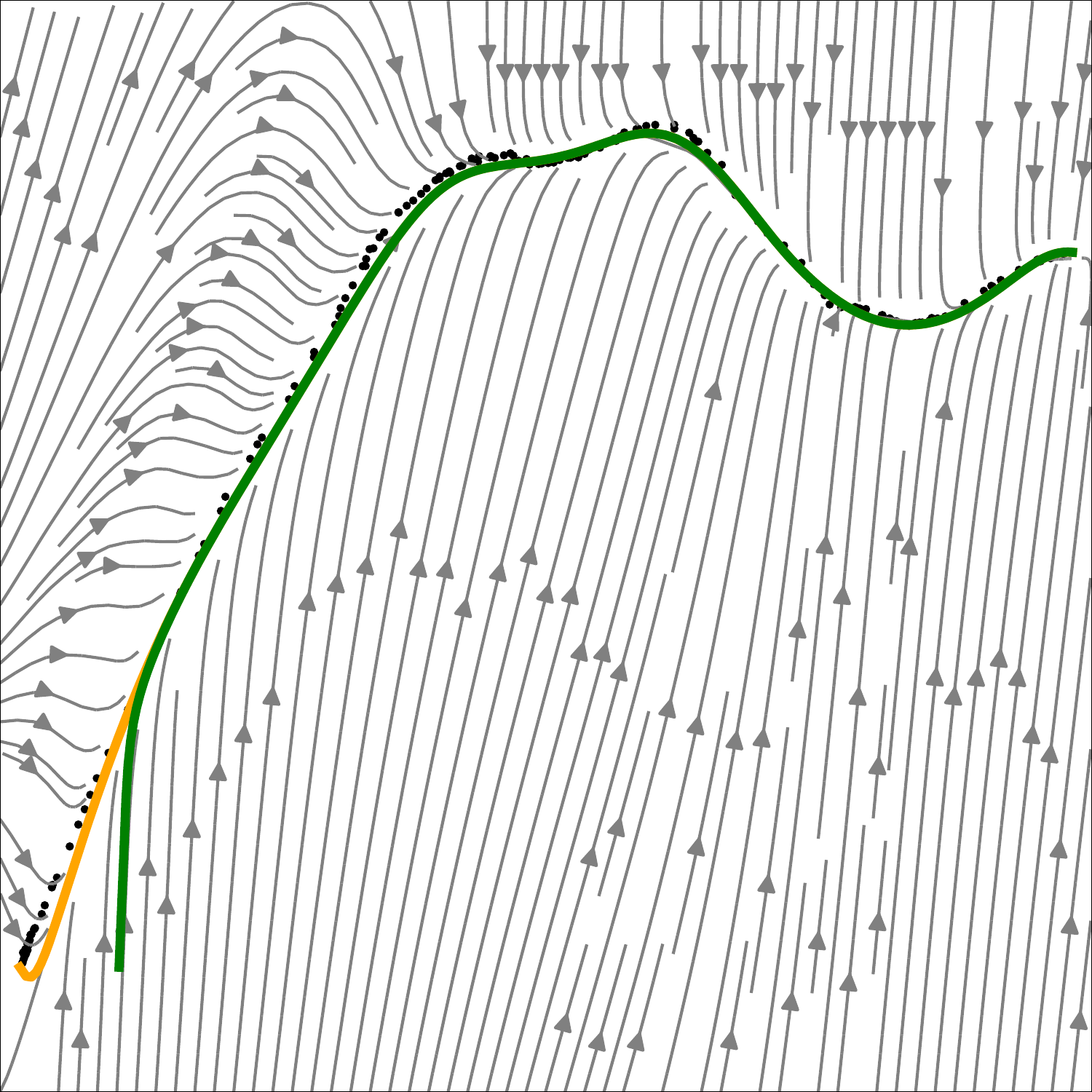}
    \end{subfigure}%
    \begin{subfigure}{0.48\linewidth}
        \centering
        \includegraphics[width=1.0\linewidth]{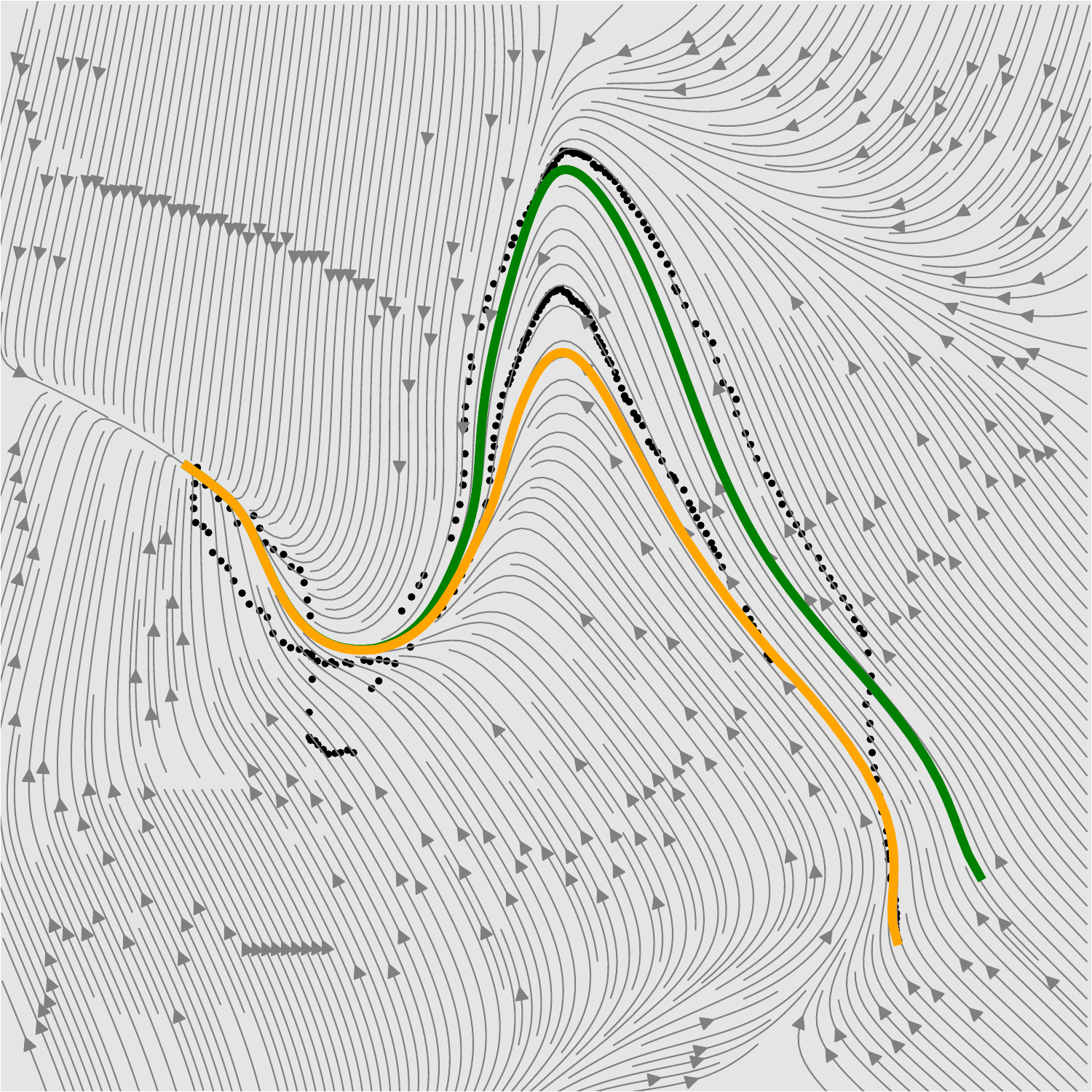}
  \end{subfigure}%
  \caption{\emph{Left:} Latent integral curves with NCDS trained in joint space $\mathbb{R}^{44}$. \emph{Right:} Latent integral curves with NCDS trained on full human motion space: $\mathbb{R}^{44} \times \mathbb{R}^3 \times \SO$. The background depicts the contours of the latent vector field, green and yellow curves depict the integral curves, and black dots represent demonstration in latent space. }
  \label{fig:Human_Latent_Dynamics}
\end{figure}
\begin{figure*}[t]
    \centering
        \begin{subfigure}{0.51\linewidth}
        \centering
        \includegraphics[width=1.0\textwidth]{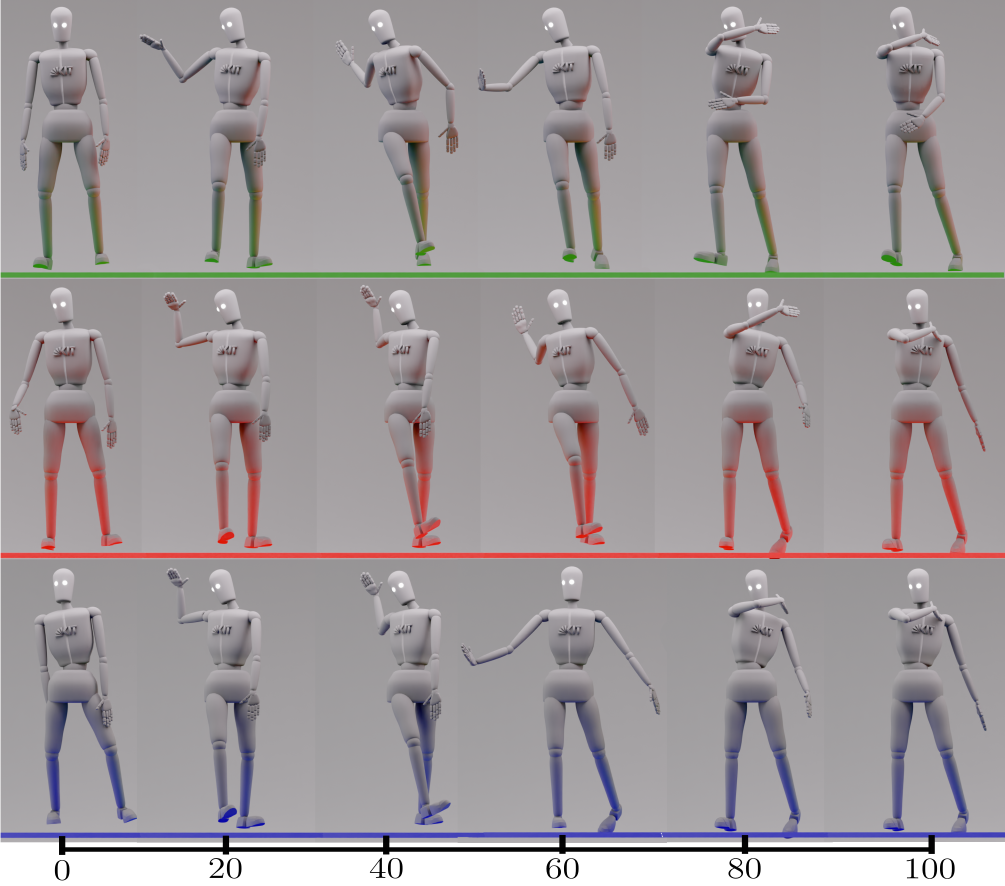}
    \end{subfigure}%
    \centering
        \begin{subfigure}{0.46\linewidth}
        \centering
        \includegraphics[width=1.0\textwidth]{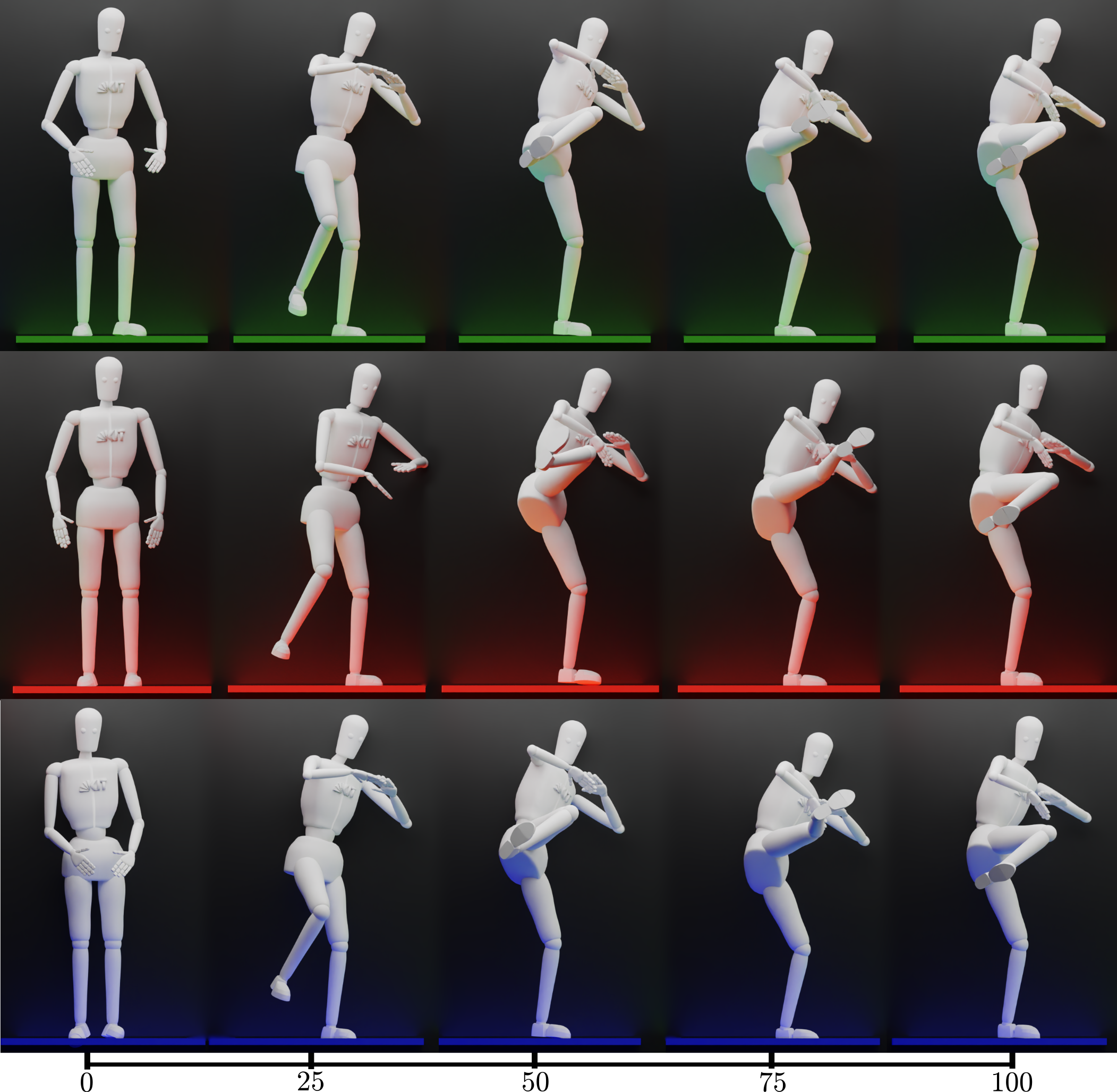}
    \end{subfigure}%
        \caption{\emph{Left}: Generated Human Motion with NCDS trained in joint space $\mathbb{R}^{44}$: The upper row depicts the progressive evolution of the demonstration motion from left to right. Meanwhile, the middle and bottom rows illustrate the motions generated by NCDS when the initial point is respectively distant from and coincident with the initial point of the demonstration. \emph{Right}: Generated Human Motion with NCDS trained on full human motion space $\mathbb{R}^{44} \times \mathbb{R}^3 \times \SO$: The upper row depicts the progressive evolution of the demonstration motion from left to right. Meanwhile, the middle and bottom rows illustrate the motions generated by NCDS when the initial point is respectively distant from and coincident with the initial point of the demonstration. }
        \label{fig:human_motion_joint}
\end{figure*}

First, we focus on learning the motion in the $44$-dimensional joint space of the human avatar. 
We chose the motion that corresponds to a human performing a tennis forehand skill. The architecture is the same as in Sec.~\ref{sec:res:lasa} except that the Jacobian network was implemented as a neural network with two hidden layers, each containing $50$ nodes. For this specific skill, we use a single demonstration to learn the skill, showing that NCDS is able to learn and generalize very complex skills with very few data. 
The results are shown in Fig.~\ref{fig:human_motion_joint}--\emph{left}, where the motion progresses from left to right over time.
The top row displays the demonstrated motion, the middle row shows the motion generated by an integral curve starting from one of the demonstration's initial points, and the bottom row displays the motion generated by an integral curve starting far from the demonstration's initial points.
These results demonstrate that NCDS is able to learn high-dimensional dynamics. 
Figure~\ref{fig:Human_Latent_Dynamics}--\emph{left} illustrates the integral curves generated by NCDS on a $2$D latent space.
The previous human motion experiment focused on reconstructing motions in the human joint space alone, thereby failing to model the global position and orientation of the human, which may be insufficient for some particular human motions. 
The KIT motion database also includes the base link pose of the human avatar, located at the hip. 
We have chosen a leg kick action to emphasize the importance of learning the base-link pose. 
Without incorporating the base-link pose and only focusing on joint motion, the movement appears unnatural, making the human avatar look as if it is floating or disconnected from the ground. 
Learning the base-link pose, which anchors the motion at the hip, helps maintain realistic contact with the ground, resulting in a more grounded and natural-looking movement. 
The architecture remained the same as in the joint-space experiment, except that the Jacobian network was implemented as a neural network with two hidden layers of $200$ nodes each. For this skill, we used two demonstrations to learn the motion.
Here each point of the motion trajectory is defined in a $44$-dimensional joint space $\R^{44}$, along with the position and orientation of the base link $\R^{3} \times \SO$, leading to a $50$-dimensional input space $\R^{44} \times \R^{3} \times \SO$.
Figure~\ref{fig:Human_Latent_Dynamics}--\emph{right} shows the latent integral curves generated by NCDS, while Fig.~\ref{fig:human_motion_joint}--\emph{right} shows the corresponding motion. 
The upper row of the latter plot depicts the progressive evolution of the demonstration motion from left to right. 
Meanwhile, the middle and bottom rows illustrate the motions generated by NCDS when the initial point is respectively distant from and coincident with the initial point of the demonstration.
Interestingly, we experimentally found that even this complex skill can be effectively learned using only a $2$-dimensional latent space. 
This may be task-dependent, as some tasks might require a higher-dimensional latent space to be accurately encoded~\citep{BeikMohammadi23:ReactiveMotion}. 
We have not observed this necessity in our experiments.

\begin{figure*}
  \centering
      \begin{subfigure}{.1625\linewidth}
            \centering
            \includegraphics[width=1.\linewidth]{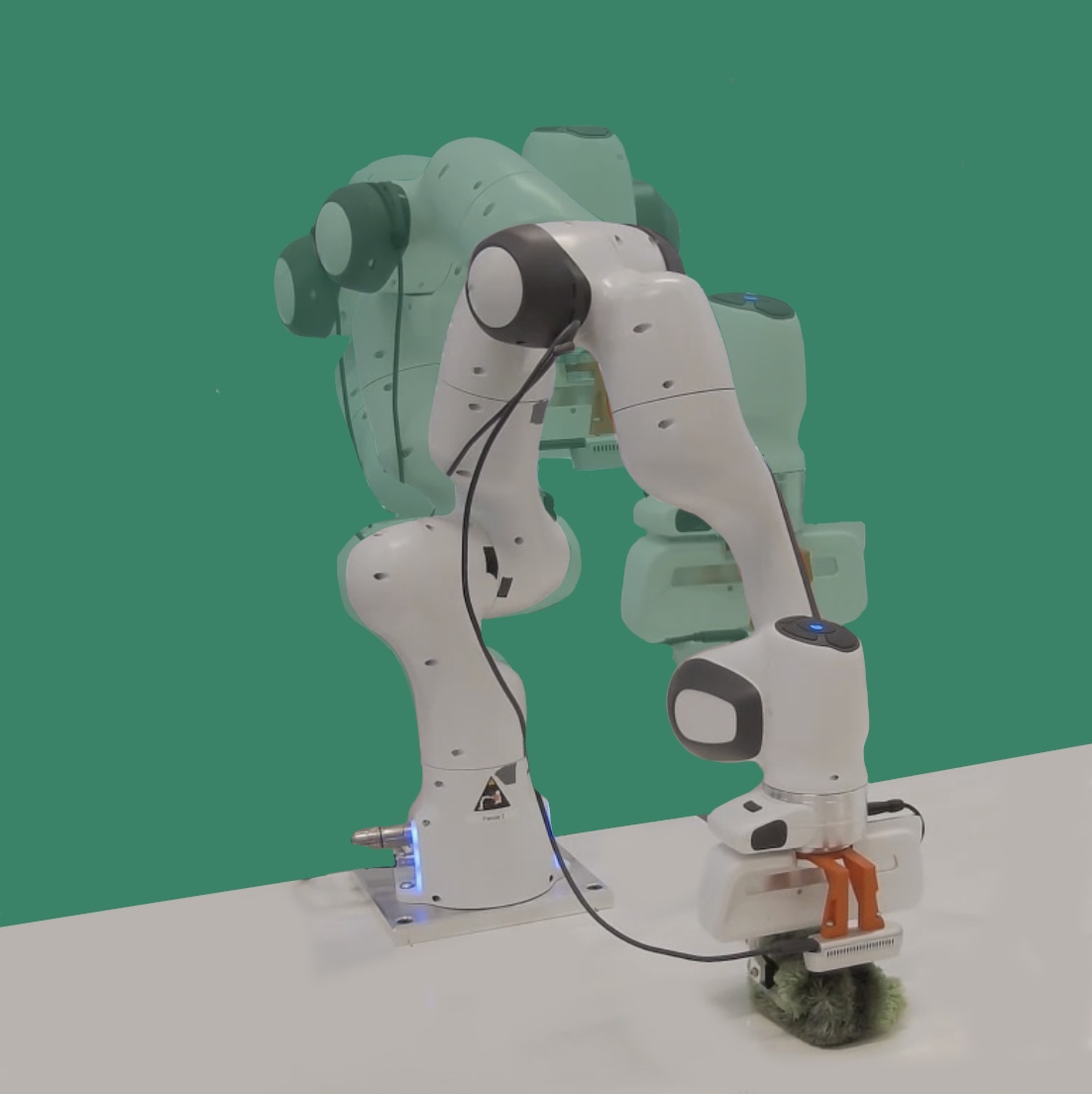}
            \caption*{\centering Grasping right object}
      \end{subfigure}
        \begin{subfigure}{.1625\linewidth}
            \centering
            \includegraphics[width=1.0\linewidth]{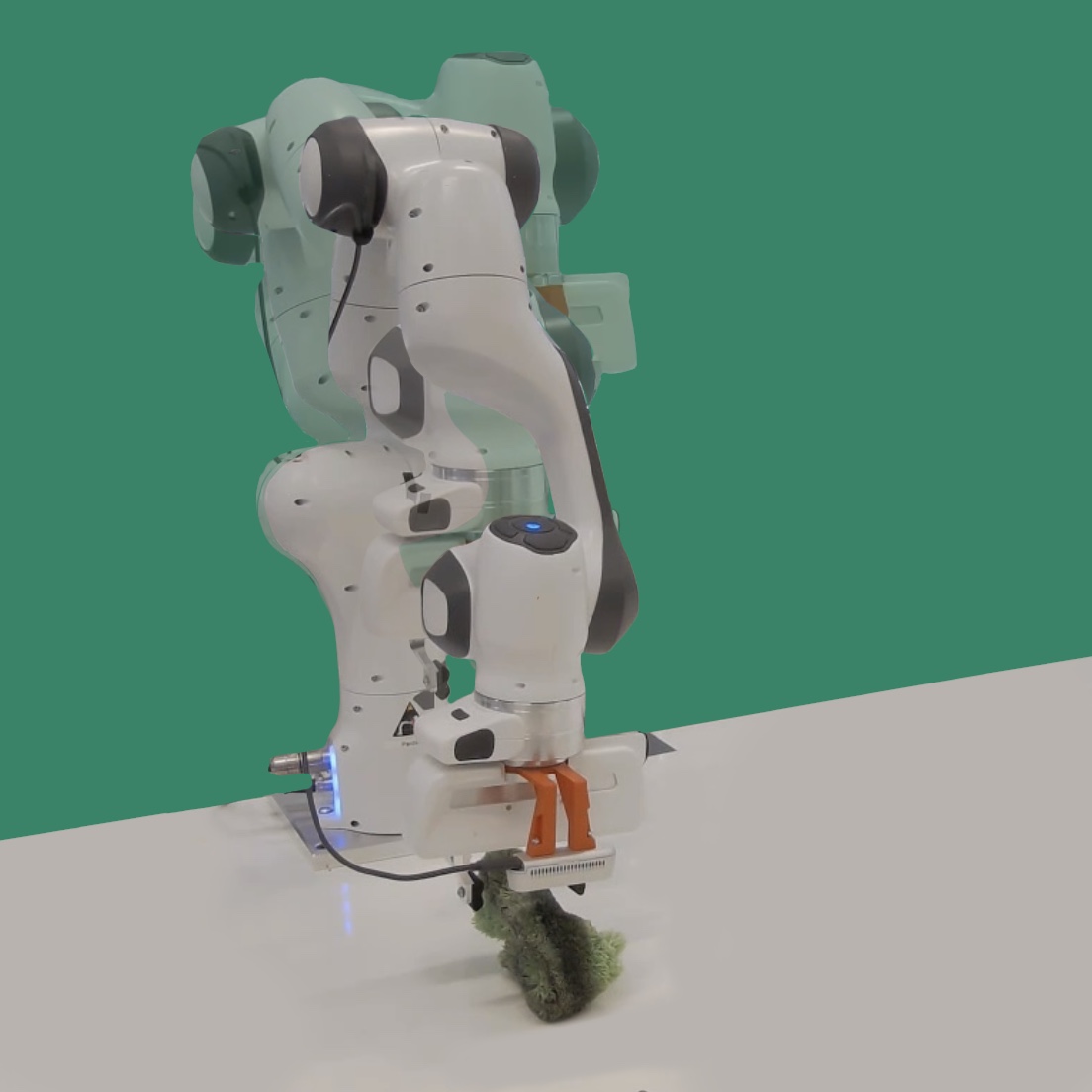}
            \caption*{\centering Grasping middle object}
      \end{subfigure}
          \begin{subfigure}{.1625\linewidth}
        \centering
        \includegraphics[width=1.0\linewidth]{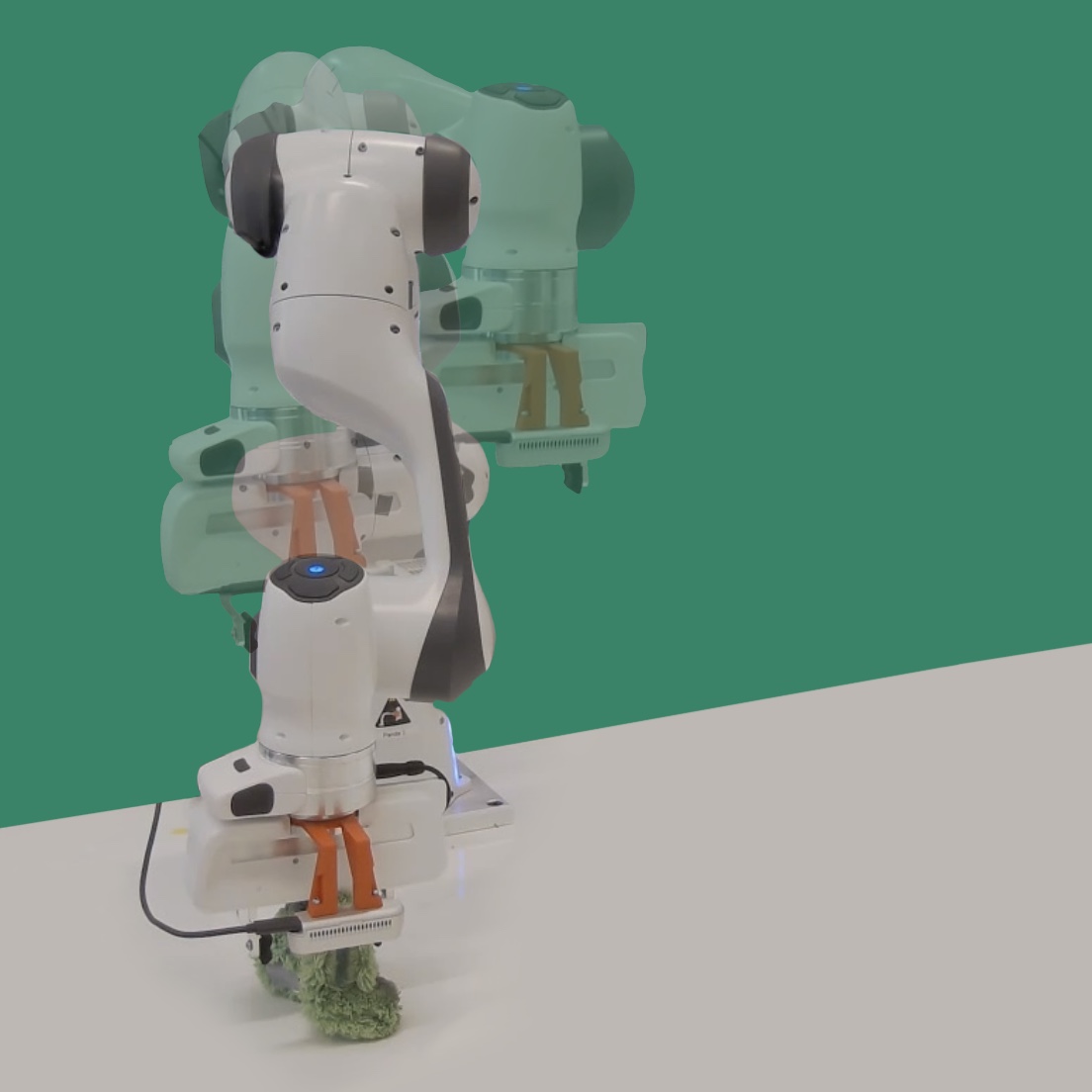}
        \caption*{\centering Grasping left object}
      \end{subfigure}
            \begin{subfigure}{.1625\linewidth}
            \centering
            \includegraphics[width=1.\linewidth]{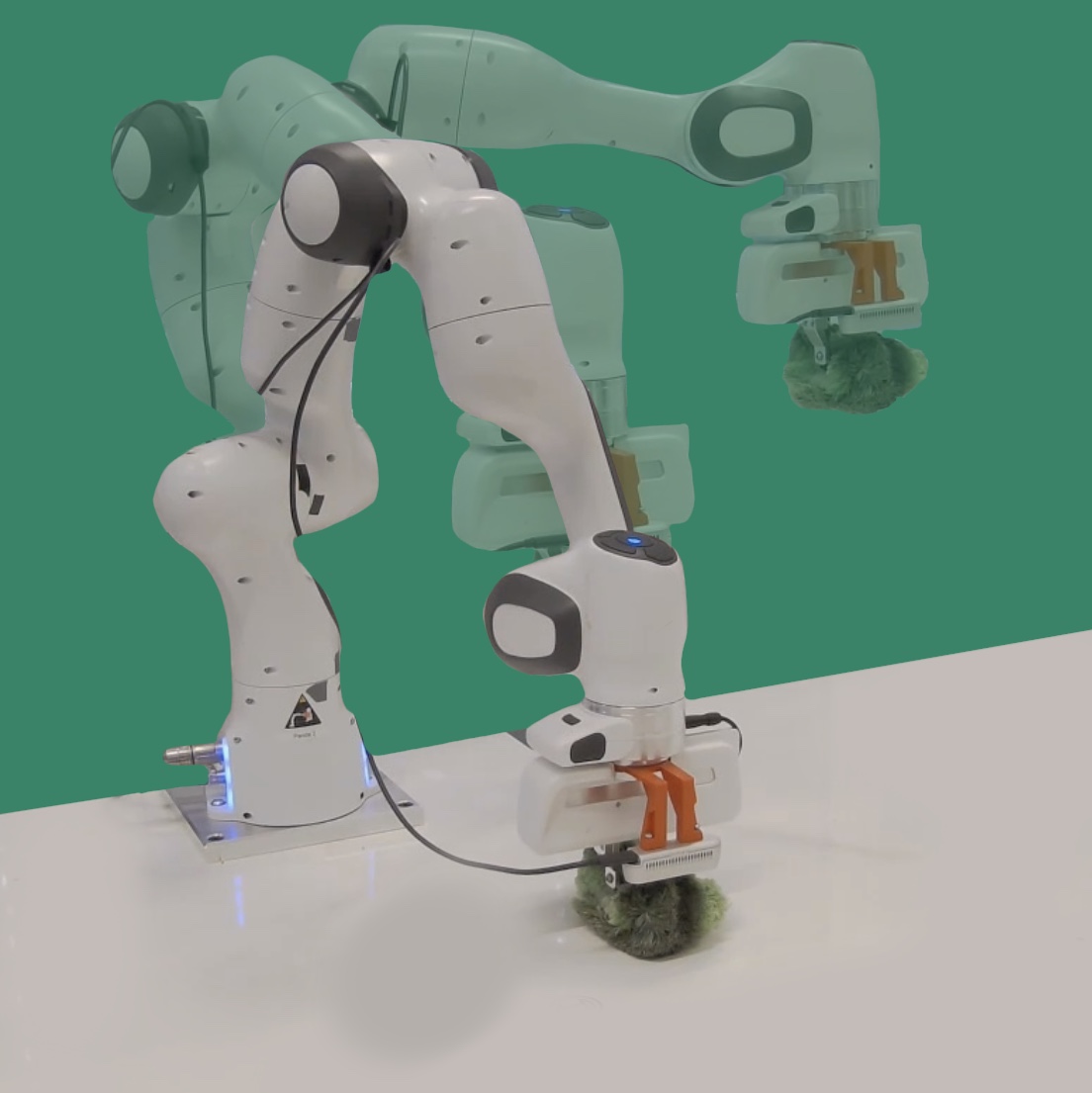}
            \caption*{\centering Dropping right object}
      \end{subfigure}
        \begin{subfigure}{.1625\linewidth}
            \centering
            \includegraphics[width=1.0\linewidth]{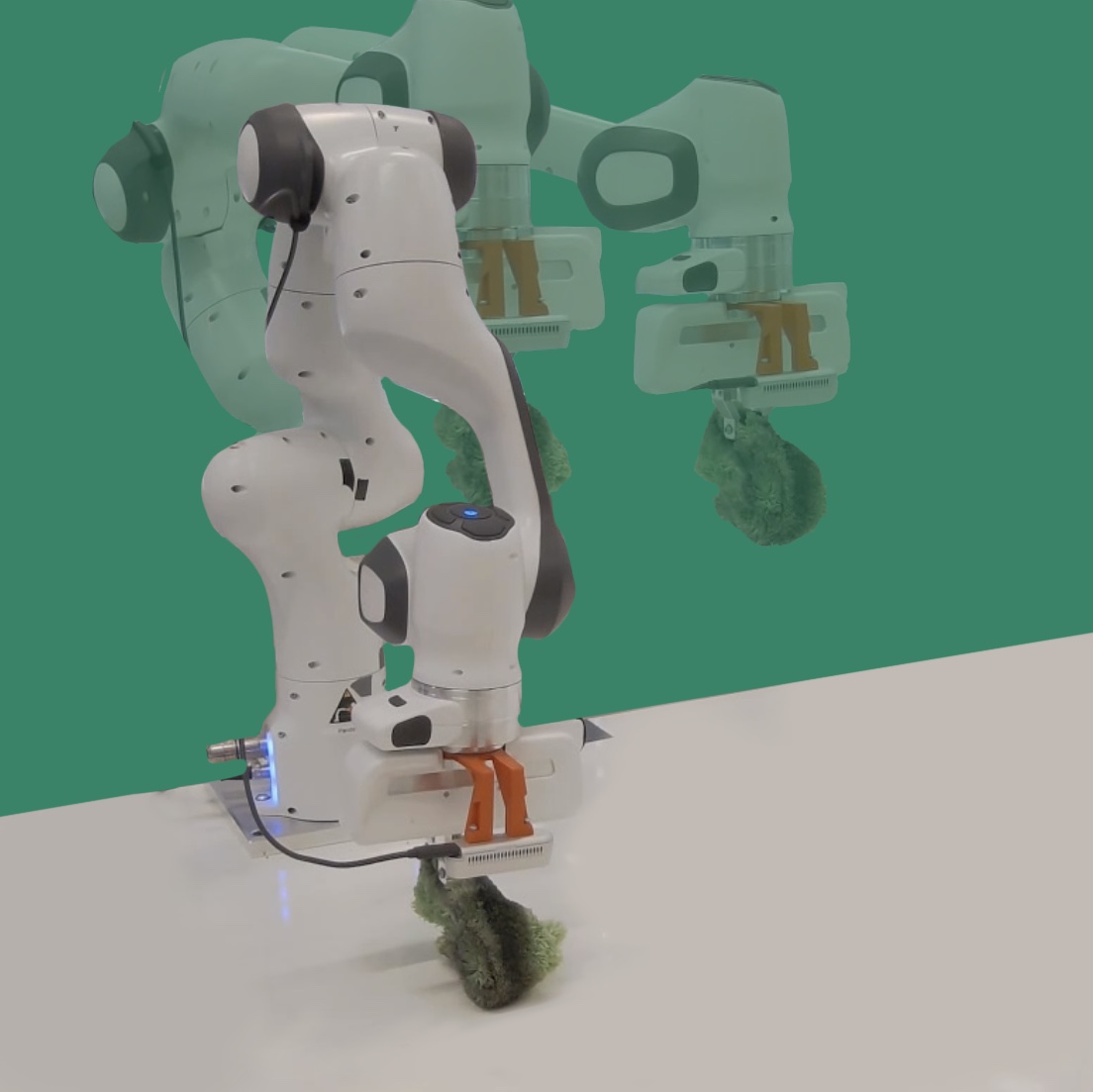}
            \caption*{\centering Dropping middle object}
      \end{subfigure}
          \begin{subfigure}{.1625\linewidth}
        \centering
        \includegraphics[width=1.0\linewidth]{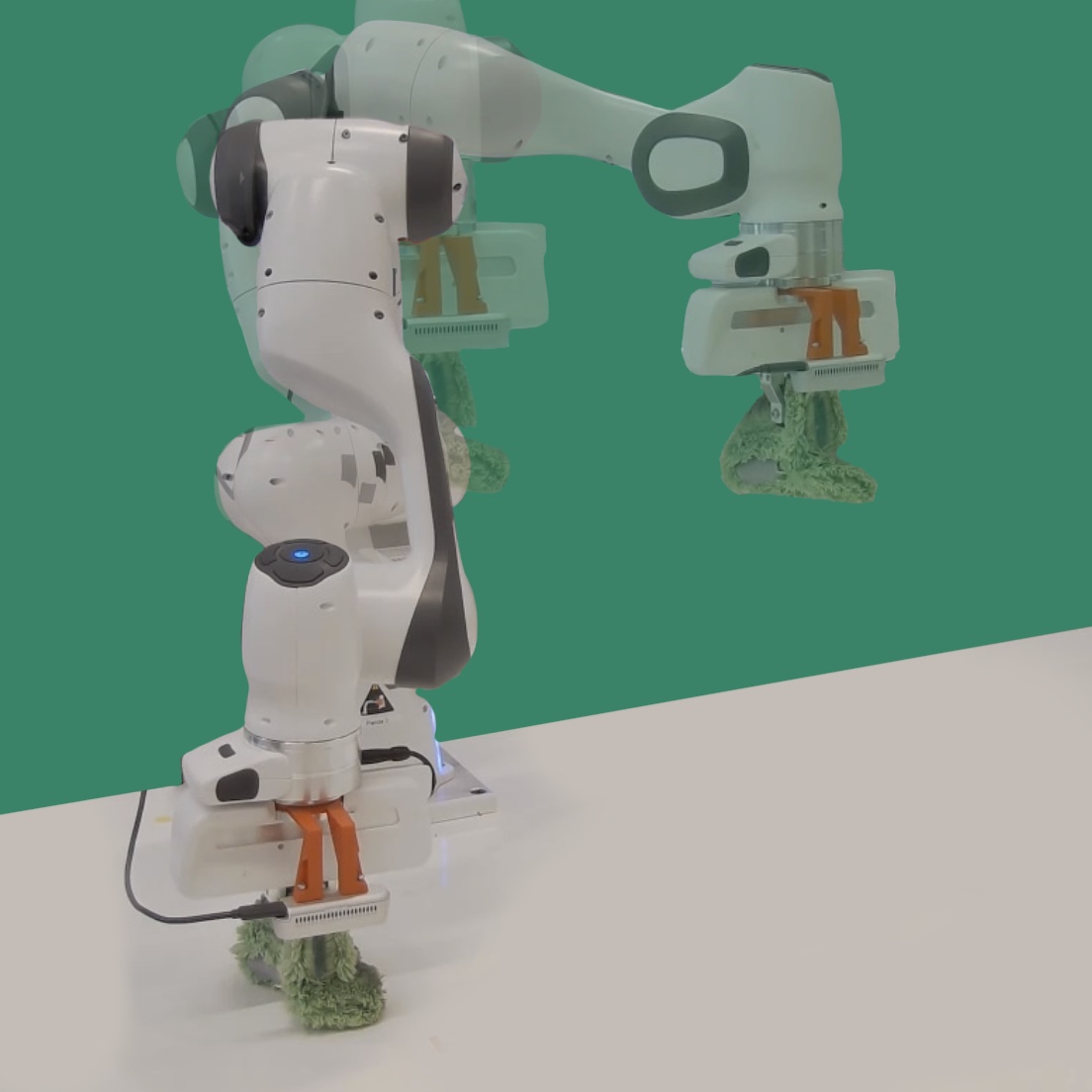}
        \caption*{\centering Dropping left object}
      \end{subfigure}
      \caption{Grasping and dropping actions learned by the CNCDS for a soft object in three positions, conditioned on image input. The CNCDS is trained using cross-entropy loss, velocity reconstruction, and regularization with \emph{ResNet18} image embeddings.}
        \label{fig:contional_robot}
\end{figure*}
\subsection{Vision-based grasp-and-drop with CNCDS}
\label{sec:exp:rob:conditional}
The Conditional Neural Dynamical System (CNCDS) can learn multiple skills using a single network, which we demonstrate using a vision perception system. 
We place a soft object in three distinct positions on a table, and a camera captures an image of the scene for each condition. 
We use a pre-trained \textsc{ResNet18} model to generate a $1000$-dimensional feature vector, which is then passed through a fully connected network to produce a $3$-dimensional embedding vector that serves as the conditioning input.
During training, only the \textsc{ResNet18} model is kept frozen, while the fully connected network and the NCDS are trained end-to-end. The training optimizes a combined loss function comprising cross-entropy loss $\mathcal{L}_{\text{CE}}$, velocity reconstruction loss $\mathcal{L}_{\text{vel}}$, and regularization loss $\mathcal{L}_\epsilon$, as defined in the NCDS framework.
The cross-entropy loss is given by $\mathcal{L}_{\text{CE}} = -\sum_{i} y_i \log(\hat{y}_i)$, where $y_i$ represents the true label, and $\hat{y}_i$ is the predicted probability for class $i$.
The fully connected layer consists of a single layer with $1000$ nodes.
The Jacobian network consists of a single hidden layer with $100$ nodes, using a \textsf{ReLU} activation function. The VAE encoder/decoder is designed with two layers, containing $200$ and $100$ nodes, respectively.
Two separate CNCDS models are trained: one for grasping the object and another for dropping it, each functioning as an independent CNCDS conditioned on the initial image of the object on the table.
For both the grasping and dropping skills, we collected $7$ demonstrations, each containing $500$ datapoints that capture the full trajectory of the motion. 
Additionally, for each object location relevant to the task, we gathered $50$ initial snapshot images that represent the initial state of the object on the table.
To build the complete dataset, each data point in the motion trajectories was paired with each of the initial snapshot images.
The first three panels in Fig.~\ref{fig:contional_robot} show the grasping actions from each initial position, while the second three panels show the corresponding dropping actions. 
By conditioning on visual information, the CNCDS framework effectively adapts to varying initial object positions, successfully learning and reproducing the desired skills.
\subsection{Riemannian safety regions}
\label{sec:obstacle_avoidance_riemannian}
In this section, we employ the pullback metric derived from the decoder's Jacobian to formulate the modulation of a contractive dynamical system. 
Specifically, the robot uses modulation to navigate around obstacles and avoid uncertain regions far from the demonstrations, under the assumption that out-of-support data regions are unsafe.
As explained in Sec.~\ref{Sec:RiemannianModualtion}, we use a Riemannian pullback metric computed via~\eqref{eq:RiemMetricWithObstacle}, to formulate the modulation matrix in~\eqref{eq:modulation_riemannian}. Specifically, we define the ambient metric $\Metric_\ambient$ for the end-effector position as in \eqref{eq:obstacle_metric}.
Note that we use a Gaussian-like function to represent the obstacle here; however, more detailed representations, such as meshes constructed from point clouds, can also be employed.
We demonstrate this approach by applying the dynamics learned from the grasping skill experiment reported in Sec.~\ref{sec:exp:rob:conditional}, where vector field modulation is incorporated based on the Riemannian formulation introduced in Sec.~\ref{Sec:RiemannianModualtion}.
In Fig.~\ref{fig:real_robot_riemannian_modulation}, the \emph{left} panel illustrates the robot performing a skill by grasping an object from the table without considering obstacle avoidance or designated safety regions. 
The \emph{middle} panel displays the same action, this time with obstacle avoidance and safety region constraints enabled. 
The \emph{right} panel represents the associated latent space, where darker areas represent unsafe zones. The dark oval-shaped region represents the Riemannian manifold derived from skill demonstrations, marking the boundary of safety regions within the latent space. The dark circular region inside represents the latent representation of the obstacle, obtained by pulling back the ambient metric into the latent space.
These results demonstrate that modulation based on the Riemannian metric allows the robot to effectively avoid obstacles while remaining within learned safety regions.
\begin{figure*}
    \begin{center}
        \begin{subfigure}{0.23\textwidth}
            \centering
            \includegraphics[width=\linewidth]{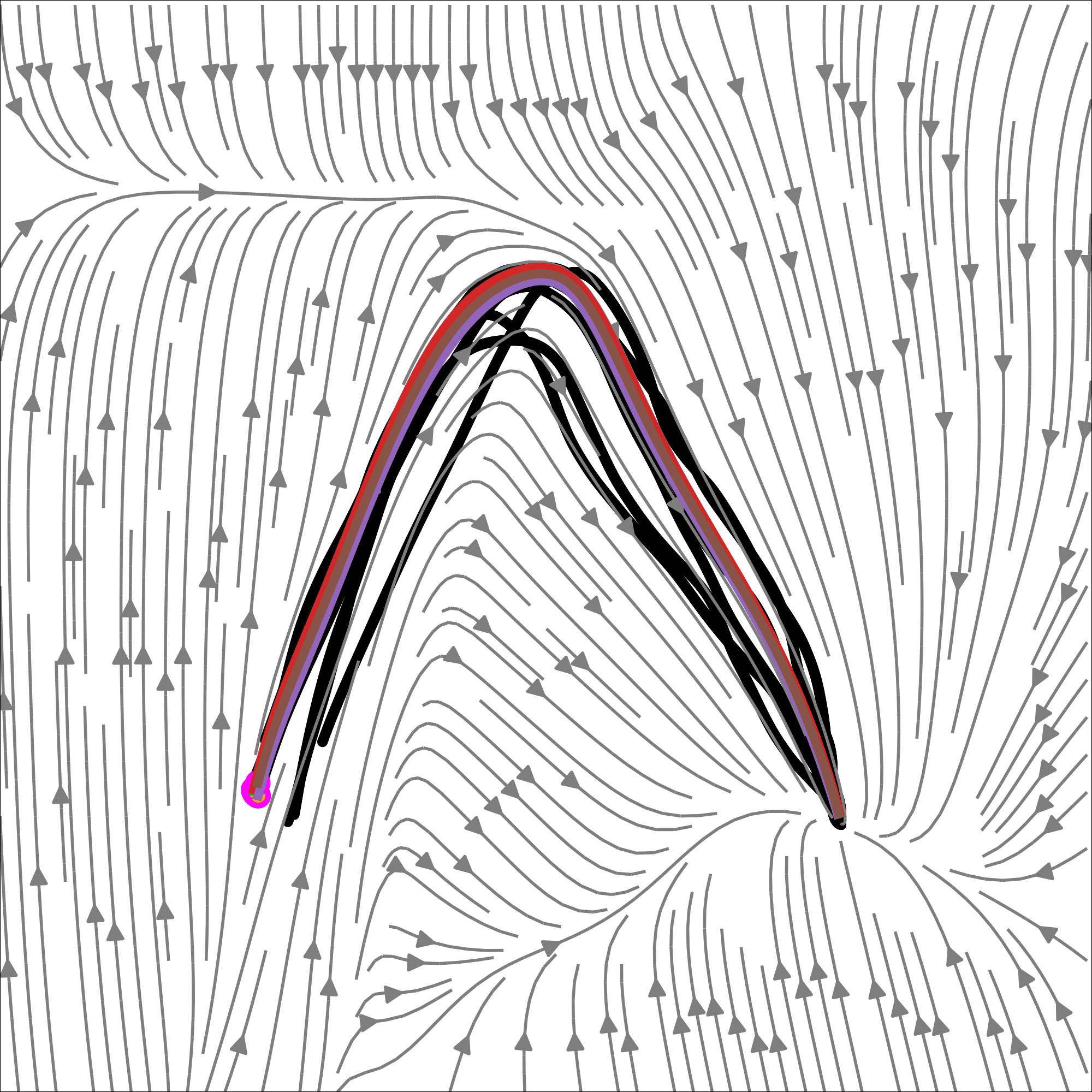}
            \caption{$\epsilon = 10^{-4}$}
        \end{subfigure}
        \begin{subfigure}{0.23\textwidth}
            \centering
            \includegraphics[width=\linewidth]{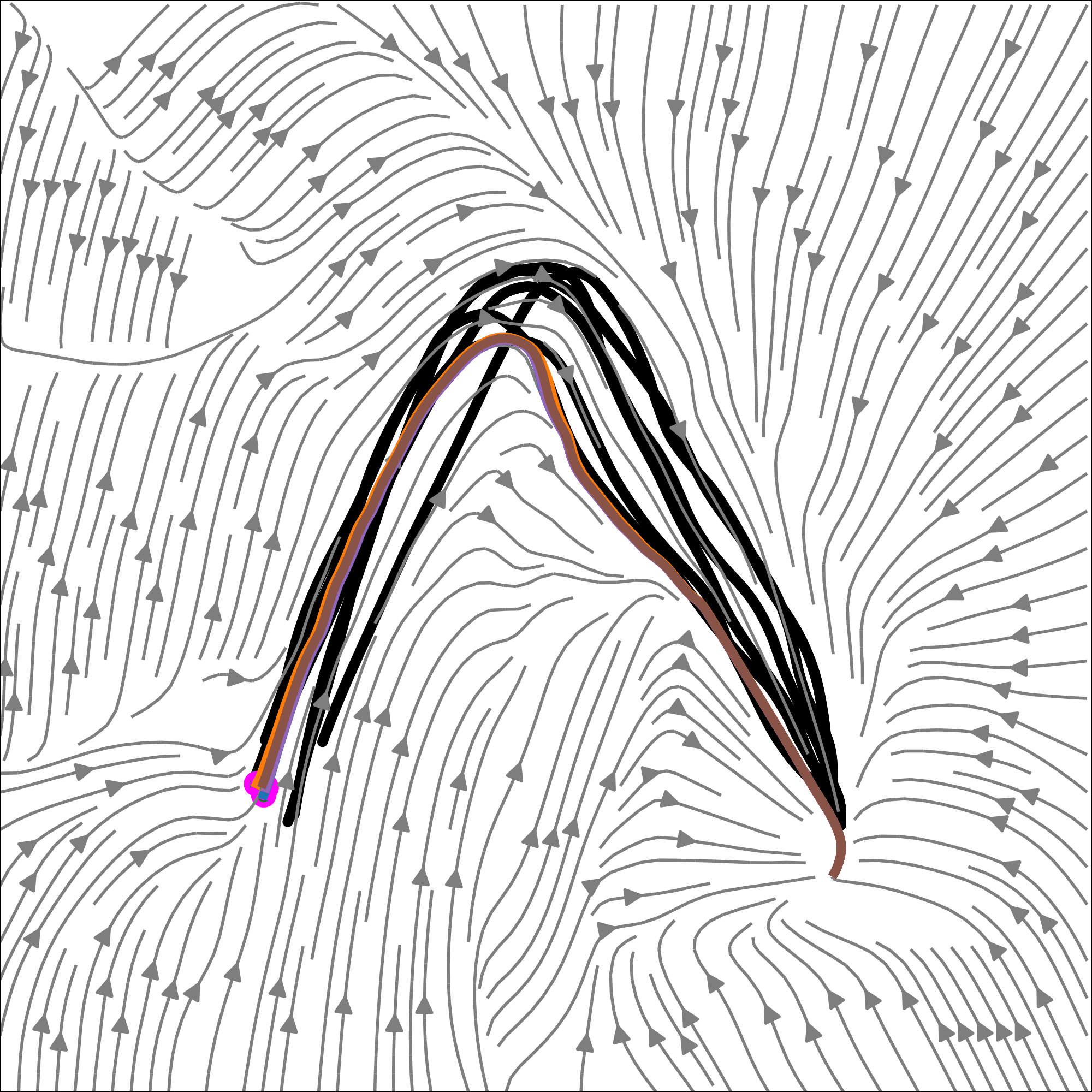}
            \caption{$\epsilon = 10^{-2}$}
        \end{subfigure}
        \begin{subfigure}{0.23\textwidth}
            \centering
            \includegraphics[width=\linewidth]{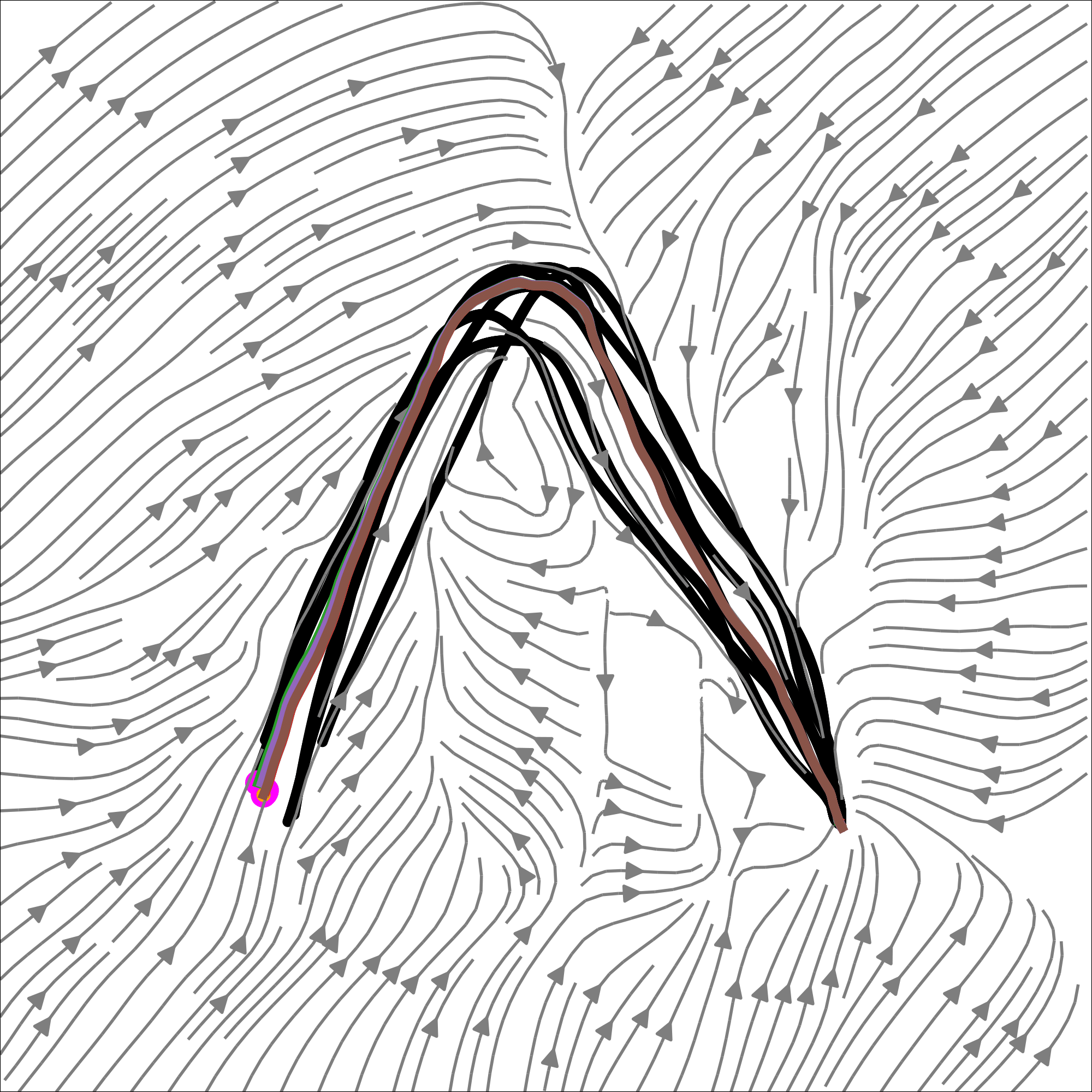}
            \caption{$\epsilon = 10^{-1}$}
        \end{subfigure}
        \begin{subfigure}{0.23\textwidth}
            \centering
            \includegraphics[width=\linewidth]{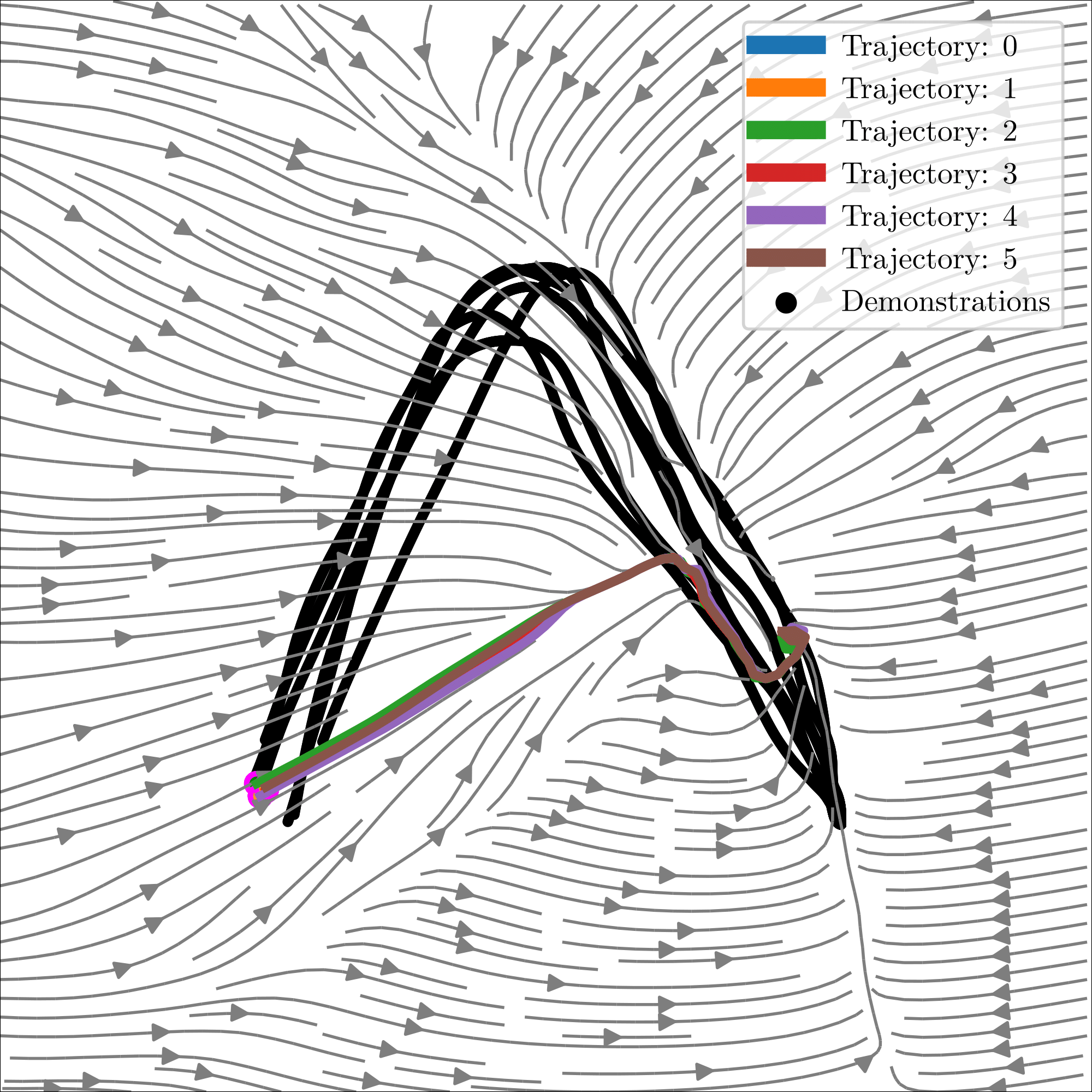}
            \caption{$\epsilon = 10$}
        \end{subfigure}
        \caption{Integral curves generated by NCDS trained using different regularization term $\epsilon$.}
        \label{fig:traj_epsilon}
    \end{center}
\end{figure*}
\begin{figure*}[h!]
    \begin{center}
        \begin{subfigure}{0.23\textwidth}
            \centering
            \includegraphics[width=\linewidth]{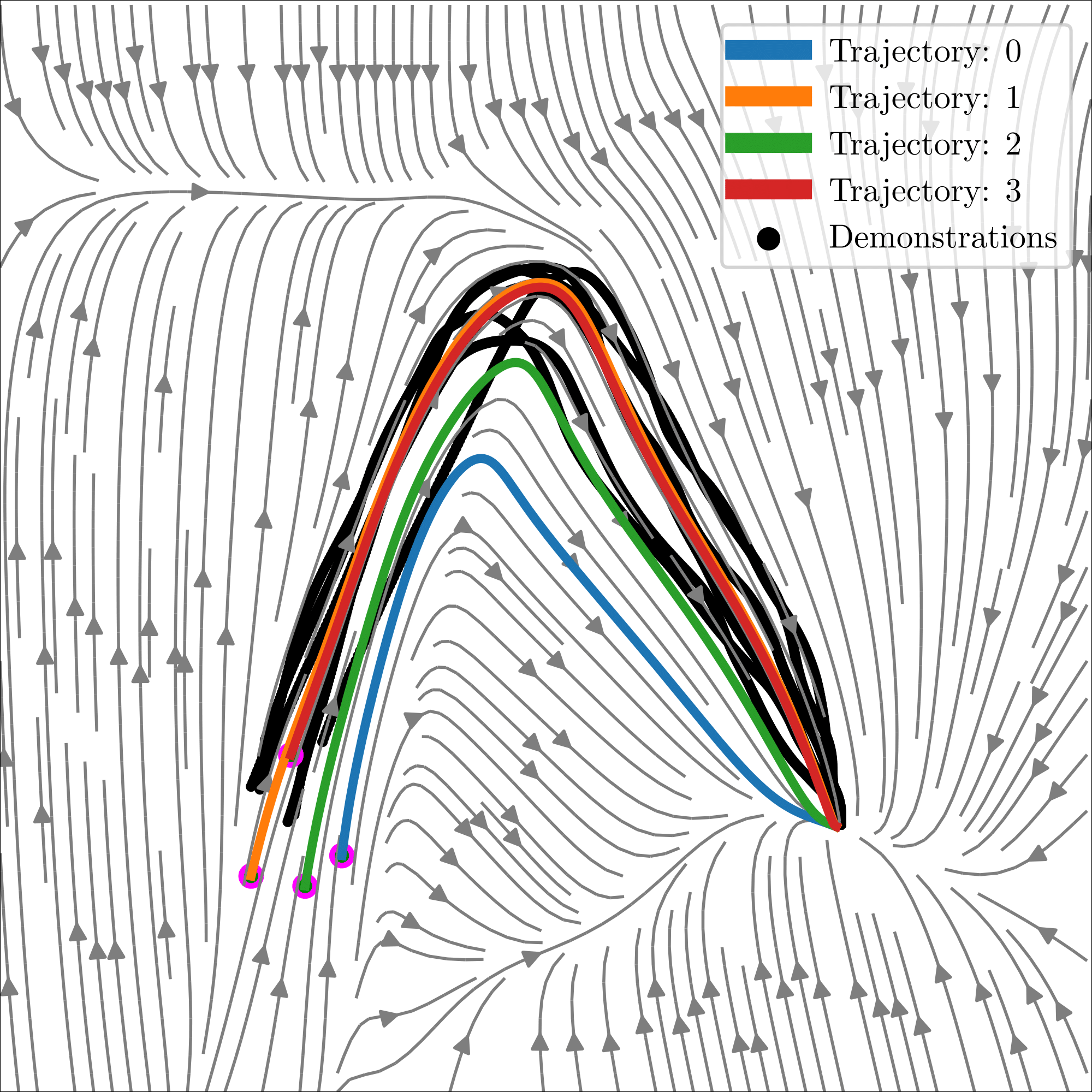}
            \caption{Contractive}
        \end{subfigure}
        \begin{subfigure}{0.23\textwidth}
            \centering
            \includegraphics[width=\linewidth]{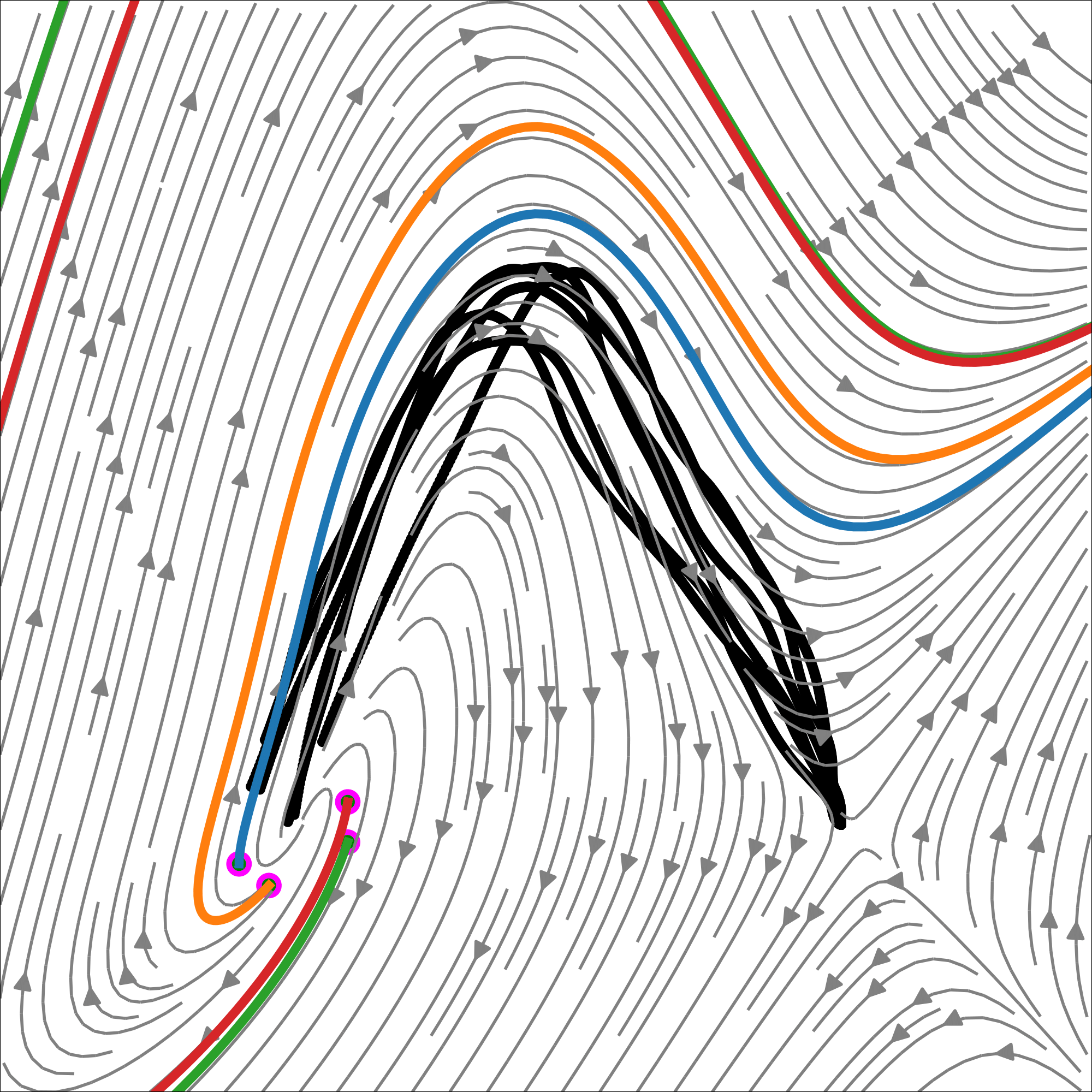}
            \caption{Unconstrained}
        \end{subfigure}
        \begin{subfigure}{0.23\textwidth}
            \centering
            \includegraphics[width=\linewidth]{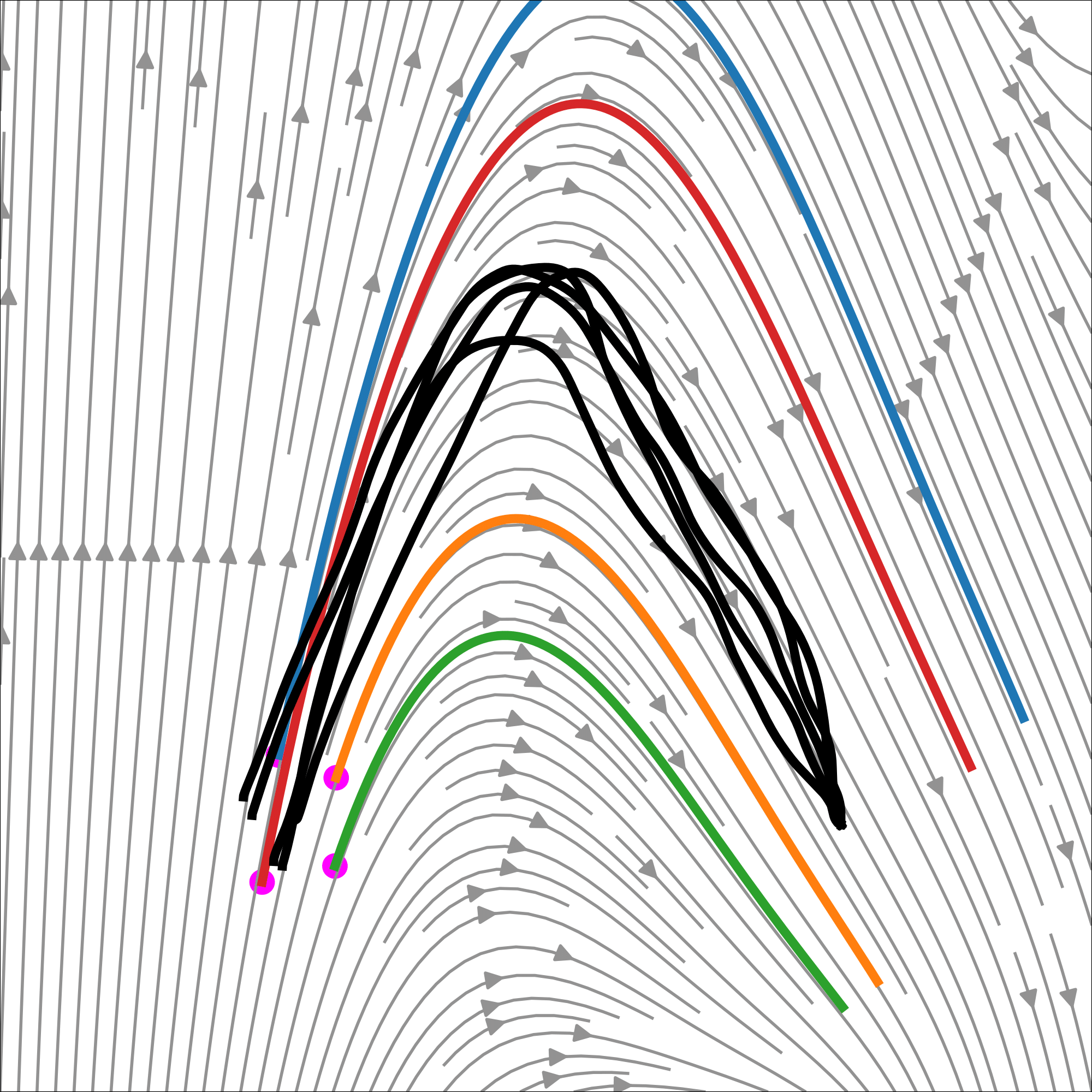}
            \caption{MLP}
        \end{subfigure}
        \begin{subfigure}{0.23\textwidth}
            \centering
            \includegraphics[width=\linewidth]{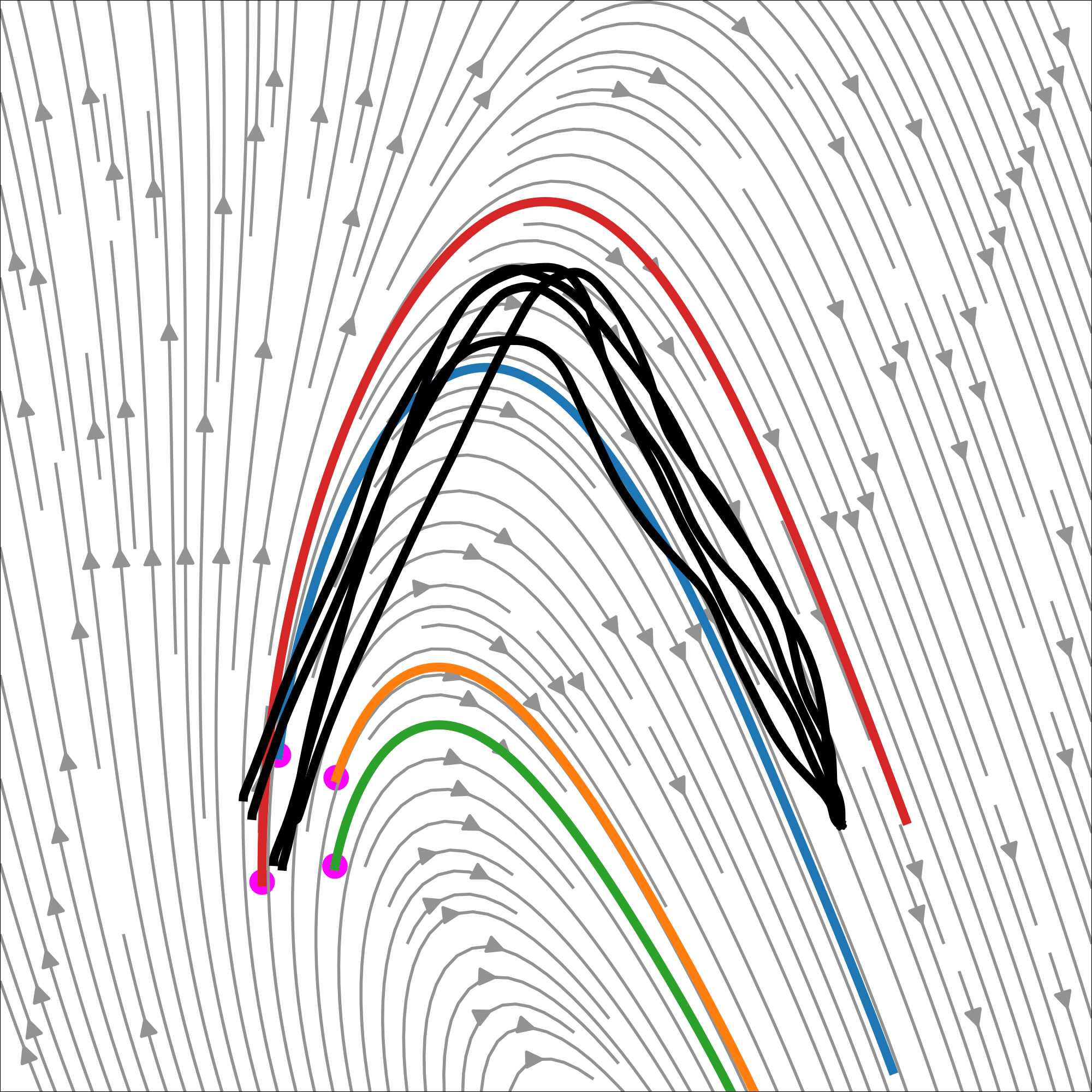}
            \caption{Neural ODE}
        \end{subfigure}
        \caption{Integral curves generated under the Neural Contractive Dynamical Systems (NCDS) setting, along with baseline comparisons using Multilayer Perceptron (MLP) and Neural Ordinary Differential Equation (NeuralODE) models.}
        \label{fig:path_integrals_unconstained}
    \end{center}
\end{figure*}
\begin{figure*}[h!]
    \begin{center}
    \begin{subfigure}{0.23\textwidth}
        \centering
        \includegraphics[width=\linewidth]{Sections/Plots/Quantitative/Angle_evolution_1e-4-1.png}
        \caption{\textsf{Tanh}}
    \end{subfigure}
    \begin{subfigure}{0.23\textwidth}
        \centering
        \includegraphics[width=\linewidth]{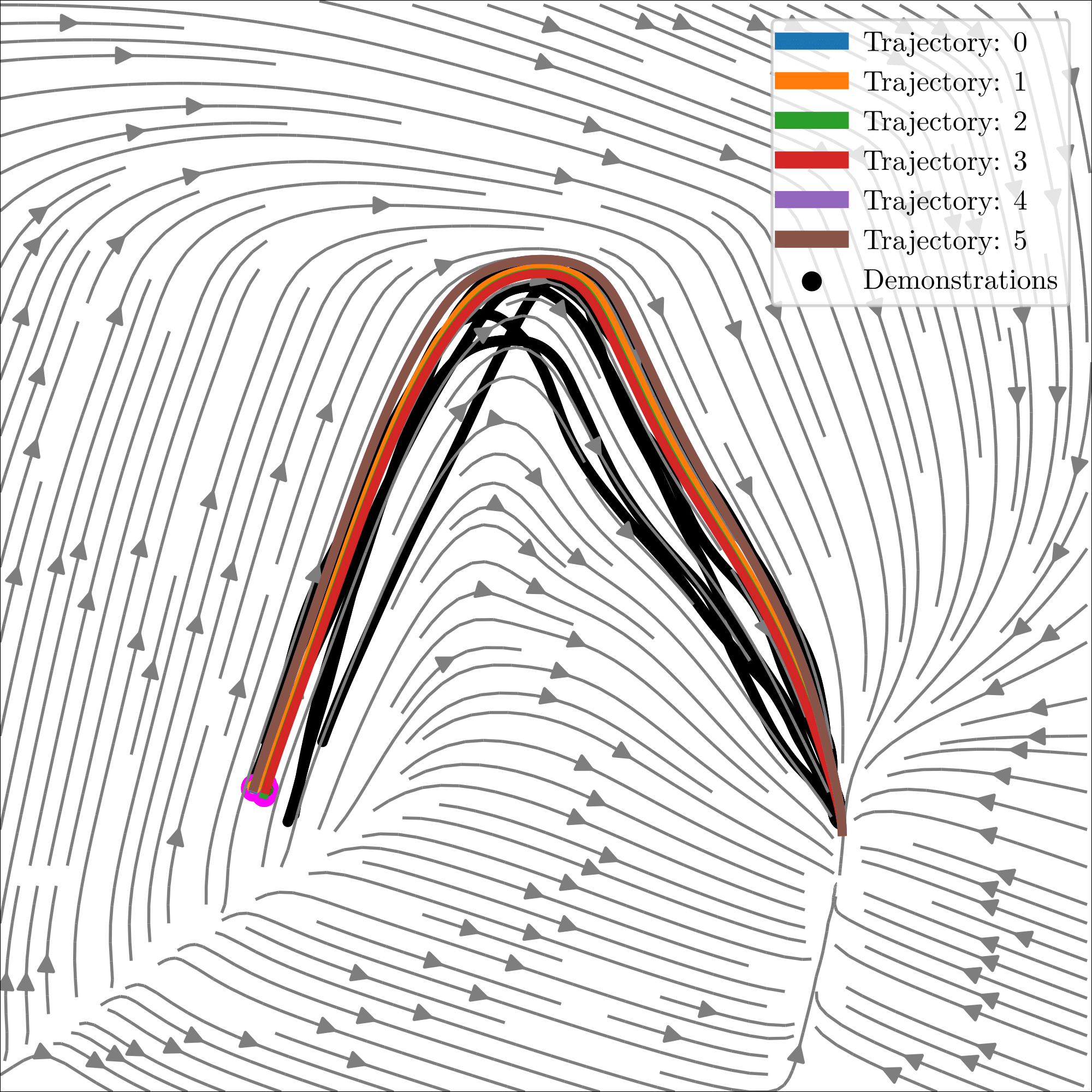}
        \caption{\textsf{Softplus}}
    \end{subfigure}
        \begin{subfigure}{0.23\textwidth}
        \centering
        \includegraphics[width=\linewidth]{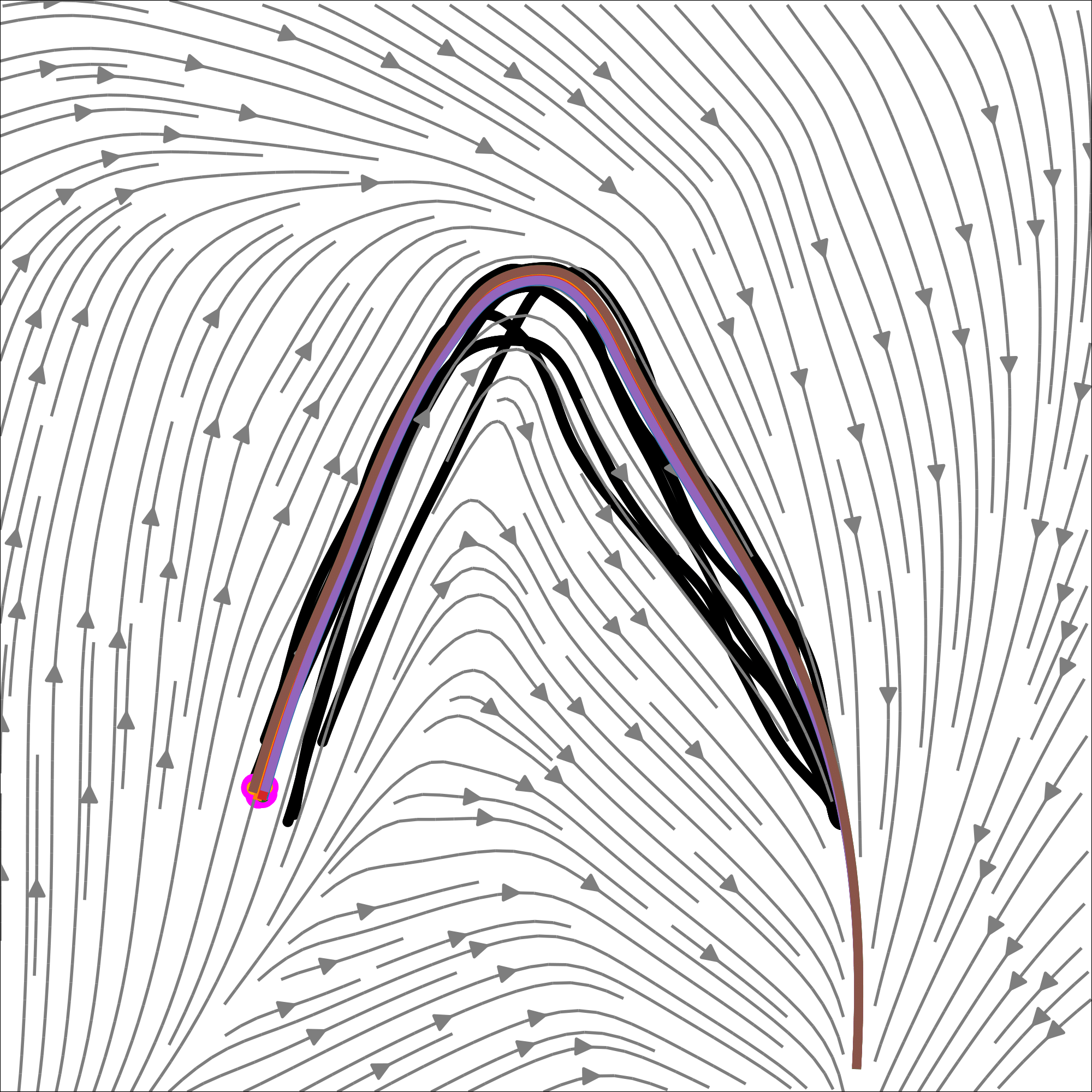}
        \caption{\textsf{Sigmoid}}
    \end{subfigure}
    \caption{Integral curves generated by NCDS trained using different activation functions.}
    \label{fig:activation_function}
    \end{center}
\end{figure*}
\subsection{Ablation studies}
\label{sec:Ablation_studies}
In this section, we perform ablation studies to analyze the impact of various components of our NCDS framework. By systematically altering or removing key elements, we aim to highlight the significance of the contraction constraint, the choice of network architecture, and the effect of different activation functions.
\paragraph{Unconstrained dynamical system:}
To illustrate the influence of having a negative definite Jacobian in NCDS, we contrast our NCDS against an identical system, except that we remove the negative-definiteness constraint on the Jacobian $\hat{\bm{J}}_{\bm{\theta}}$. 
Figure~\ref{fig:path_integrals_unconstained} shows the integral curves generated by both contractive (Fig.~\ref{fig:path_integrals_unconstained}-a) and unconstrained (Fig.~\ref{fig:path_integrals_unconstained}-b) dynamical systems. 
As anticipated, the dynamics generated by the unconstrained system lack stable behavior. 
However, the vector field aligns with the data trends in the data support regions. 
Furthermore,  we performed similar experiments with two different baseline approaches: an MLP and a Neural Ordinary Differential Equation (NeuralODE) network, to highlight the effect of the contraction constraint on the behavior of the dynamical system. 
The MLP baseline uses a neural network with $2$ hidden layers each with $100$ neurons with $\textsf{Tanh}$ activation function. 
The NeuralODE baseline is implemented based on the code provided by \citet{torchdyn:poli}, with $2$ hidden layers each with $100$ neurons. 
The model is then integrated into a NeuralODE framework configured with the \textsf{adjoint} sensitivity method and the \textsf{dopri5} solver for both the forward and adjoint passes. 
It is worth mentioning that these baselines directly reproduce the velocity according to $\dot{\x}= f(\x)$. 
Both networks are trained for $1000$ epochs with the ADAM~\citep{Kingma2014:AdamAM}. 
Figure~\ref{fig:path_integrals_unconstained}-c and \ref{fig:path_integrals_unconstained}-d show that both models effectively capture the observed dynamics (vector field); however, as anticipated, contraction stability is only achieved by NCDS.
\paragraph{Activation function:} The choice of activation function certainly affects the generalization in the dynamical system in areas outside the data support. 
To show this phenomena, we ablate a few common activation functions. 
For this experiment, we employ a feedforward neural network with two hidden layers, each of $100$ units. 
As shown in Fig.~\ref{fig:activation_function}, both the \textsf{Tanh} and \textsf{Softplus} activation functions give superior performance, coupled with satisfactory generalization capabilities. 
This indicates that the contour of the vector field aligns with the overall demonstration behavior outside of the data support. 
Conversely, the \textsf{Sigmoid} activation function yields commendable generalization beyond the confines of the data support, but it fails to reach and stop at its target. 

\begin{figure}
    \centering
    % First row: subfigures 0, 1, and 2
    \begin{subfigure}[b]{0.53\linewidth}
        \centering
        \includegraphics[width=\linewidth]{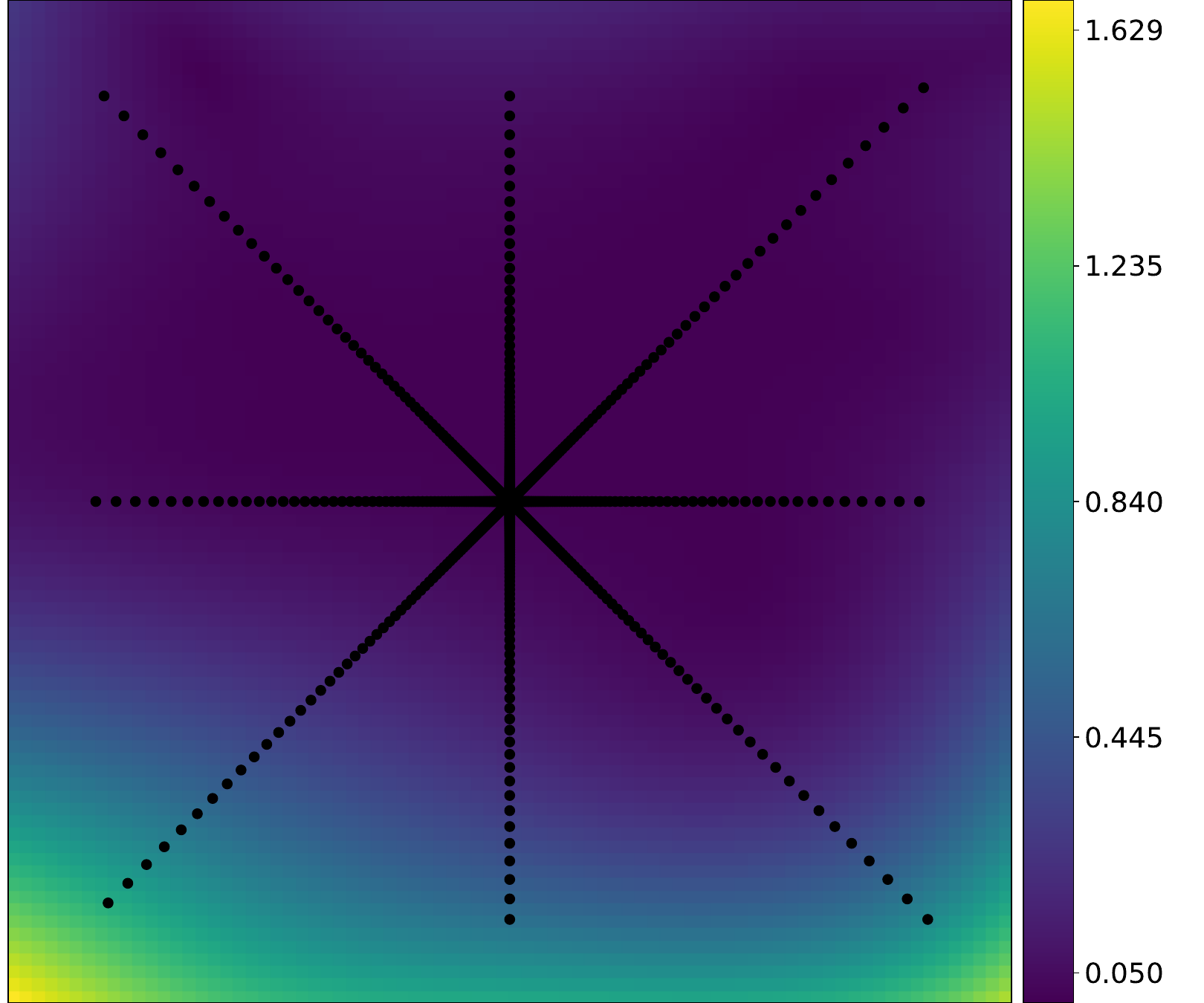}
        \label{fig:subfig0}
    \end{subfigure}
    \begin{subfigure}[b]{0.44\linewidth}
        \centering
        \includegraphics[width=\linewidth]{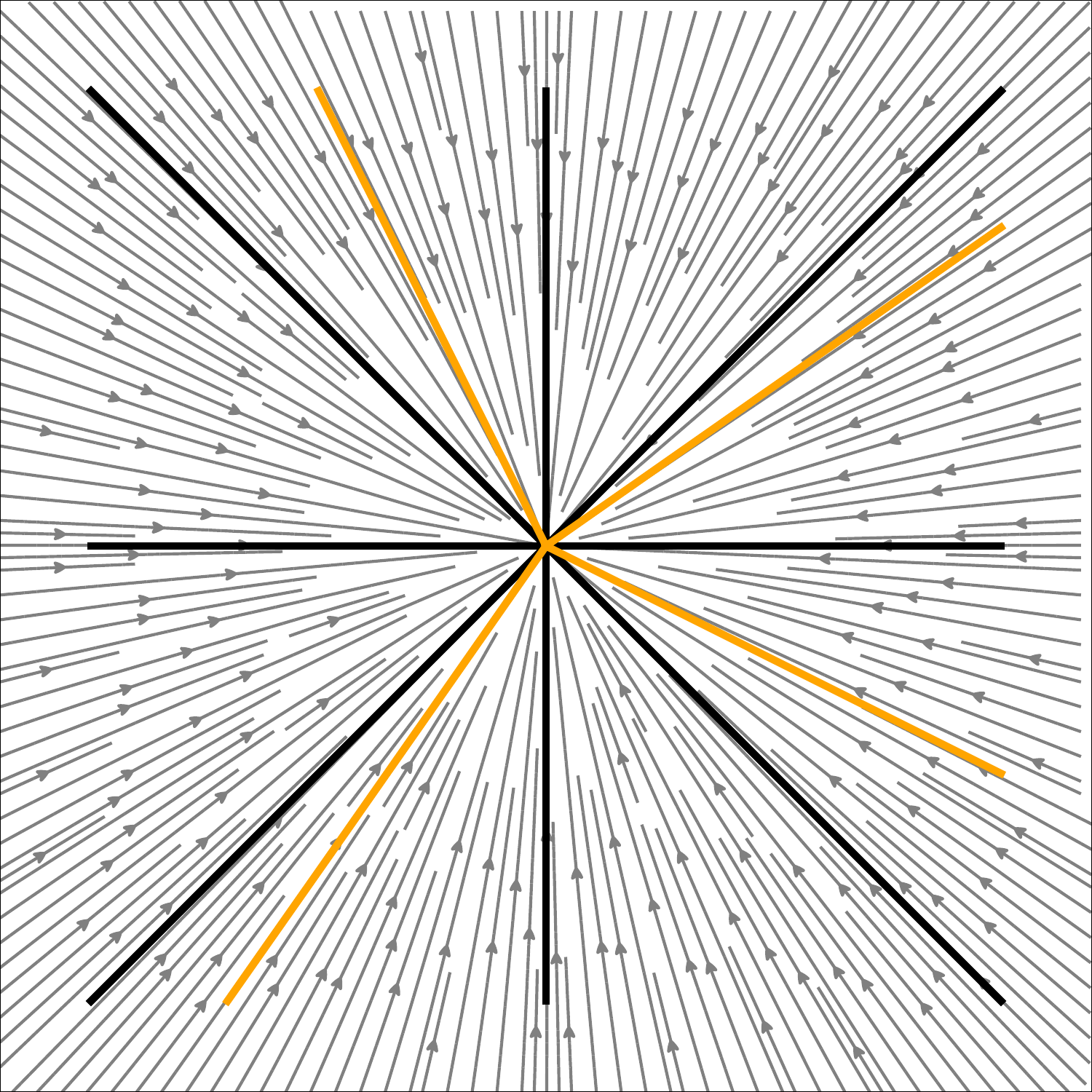}
        \label{fig:subfig1}
    \end{subfigure}
    \caption{Learned vector field with zero contraction spread. \emph{Left:} The contraction spread, scaled by $1000$, highlights minimal eigenvalue differences. \emph{Right:} Orange integral curves closely follow the training demonstrations (black), with gray contours outlining the learned vector field.}
    \label{fig:ELC_Comparison}
\end{figure}

\paragraph{Zero contraction spread vector fields:}
In our formulation, the regularization loss terms serve as soft constraints that guide the learning process without imposing a rigid structure on the solution. For example, when the ground-truth vector field (see Fig.~\ref{fig:ELC_Comparison}, left panel) exhibits straight-line trajectories converging to the origin, the network prioritizes an accurate reconstruction of such demonstrated behavior. This is achieved without enforcing a large disparity in the regularization vector components or the eigenvalues spread.
To illustrate this phenomenon, Fig.~\ref{fig:ELC_Comparison} shows a comparative analysis. In the right panel, the black trajectories represent the training demonstrations, while the orange trajectories are the integral curves generated by the model. The results indicate that the model effectively captures the linear behavior of the ground-truth vector field. Additionally, the scaled contraction spread (i.e., $\times 1000$) demonstrates that the differences between the eigenvalues are effectively negligible.

%% file: Sections/Discussion.tex
In this paper, we proposed an enhanced framework for designing neural networks that are contractive, building upon the original Neural Contractive Dynamical Systems (NCDS, \cite{NCDS:2023BeikMohammadi}). This allows us to create a flexible class of models capable of learning dynamical systems from demonstrations while retaining contractive stability guarantees, offering a more controlled generalization behavior. Key improvements include the introduction of Jacobian matrix regularization, where we explore both state-independent and state-dependent regularization strategies, as well as eigenvalue-based methods, to fine-tune the contraction properties and improve robustness and generalization.

Furthermore, we extended the NCDS framework by introducing asymmetry into the Jacobian matrix, incorporating both symmetric and skew-symmetric components. However, our evaluation showed that this modification did not provide any significant advantage over using only the symmetric part. This suggests that the NCDS with symmetric Jacobian is sufficient to locally approximate the behavior of dynamical systems with non-symmetric Jacobians, eliminating the need for the added complexity. 

We also introduced the Conditional NCDS (CNCDS), which allows us to handle multiple motion skills by conditioning on task-related variables, such as target states, allowing a single NCDS module to adapt to multiple task conditions. \looseness=-1

Finally, to enhance real-world applicability, we improve the model's obstacle avoidance capabilities by employing Riemannian safety regions on the NCDS latent space. Inspired by modulation matrix techniques, this approach reshapes the latent learned vector field near unsafe areas, enabling the system to avoid obstacles and out-of-data support regions, while maintaining contractive stability guarantees.

Overall, our extended NCDS framework significantly addresses the key limitations of the original model, introducing improvements in flexibility, robustness, and adaptability.

However, several limitations persist. The model remains sensitive to the choice of numerical integration schemes, particularly the reliance on adaptive step sizing to ensure reliable performance, which increases computation time and poses challenges for real-time applications. Moreover, the need to compute the Jacobian of the decoder at every step, in order to calculate the ambient velocity and Riemannian metric, further impacts performance, adding computational overhead. Furthermore, a limitation of implementing the conditional variable in CNCDS is its limited reactivity to changes in the conditioning variable. In other words, if the conditioning variable changes during skill execution, it is uncertain whether CNCDS will adapt seamlessly to these changes. We consider these issues minor compared to the many benefits that the NCDS framework provides.

%% file: Sections/Appendix.tex
\subsection{Modulation diagonal matrix D}
\label{App:matrixD}

Based on these requirements, matrix $\bm{D}$ can be designed according to the following sigmoid function,
\begin{align}
    \Xi(\rho, \nu, \lambda_{init}, \lambda_{end}, k) = \lambda_{init} + \\ \cfrac{\lambda_{end} - \lambda_{init}}{1 + \exp\left(-k \left[\mathcal{S}(\x) - \frac{(\rho+\nu)}{2}\right]\right)}, \text{with}
    \label{eq:sigma}
\end{align}
\begin{equation*}
    \lambda_n(\x) = \Xi(\rho=1, \nu=10, \lambda_{init}=0, \lambda_{end}=1, k=2),
\end{equation*}
\begin{equation*}
    \lambda_\tau(\x) = \Xi(\rho=1, \nu=10, \lambda_{init}=2, \lambda_{end}=1, k=2),
\end{equation*}

\begin{figure}[ht!]
    \centering
    \includegraphics[width=1.0\linewidth]{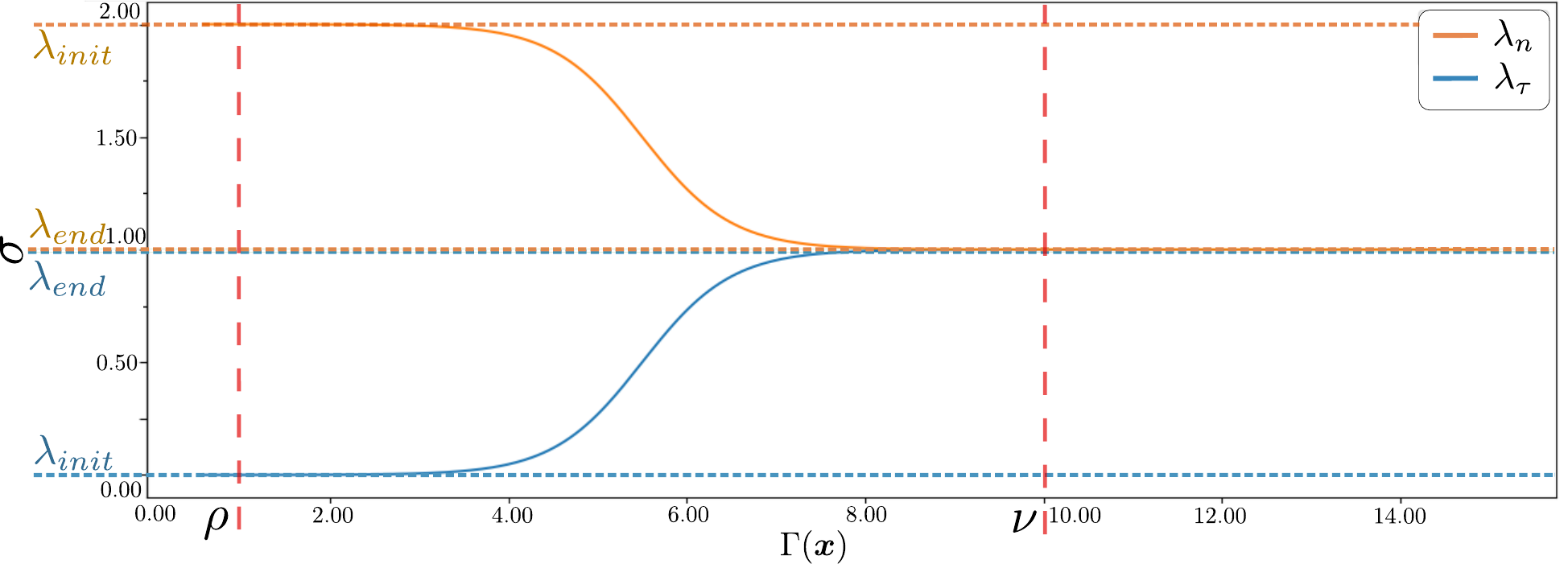}
    \caption{The behavior of the elements of the matrix $\bm{D}$ in response to distance from obstacle.}
    \label{fig:matrix_d}
\end{figure}

The parameters are defined as follows: $\rho$ specifies the distance at which the obstacle becomes impenetrable, while $\nu$ indicates the distance from which the modulation is inactive. 
The initial and final values of the sigmoid function $\Xi(...)$ are given by $\lambda_{init}$ and $\lambda_{end}$, respectively. 
For example, in Fig.~\ref{fig:matrix_d}, $\lambda_{init}$ and $\lambda_{end}$ (shown in blue) are the initial and target values for $\lambda_\tau$. 
This means that $\lambda_\tau$ takes the value of $\lambda_{init}$ when inside the obstacle and similarly it takes the values of $\lambda_{end}$ when outside of the obstacle. 
The same applies for $\lambda_n$, depicted in orange. 
Lastly, $k$ controls the smoothness of the transition between these values.
The specific values for these parameters are provided in~\cite{Koptev2023:ImplicitDistanceFunctions} to configure the matrix $\bm{D}$ for correct modulation activation. 
In this specific setup, the modulation activates $\nu$ units away from the obstacle surface and progressively intensifies until it reaches $\rho$ unit from the surface, at which point the surface becomes impenetrable. 
These values can be adjusted to suit different configurations or experimental conditions.
Figure~\ref{fig:matrix_d} illustrates how the elements of matrix $\bm{D}$ behave as a function of the distance from the obstacle.
Note that the values of $\lambda_n(\x)$ begin to decrease towards $0.0$, while $\lambda_t(\x)$ values start increasing towards $2.0$ as the distance drops below $\nu$ units. 

\begin{figure*}[htb!]
    \centering
        \begin{subfigure}{.24\linewidth}
            \includegraphics[width=1.0\linewidth]{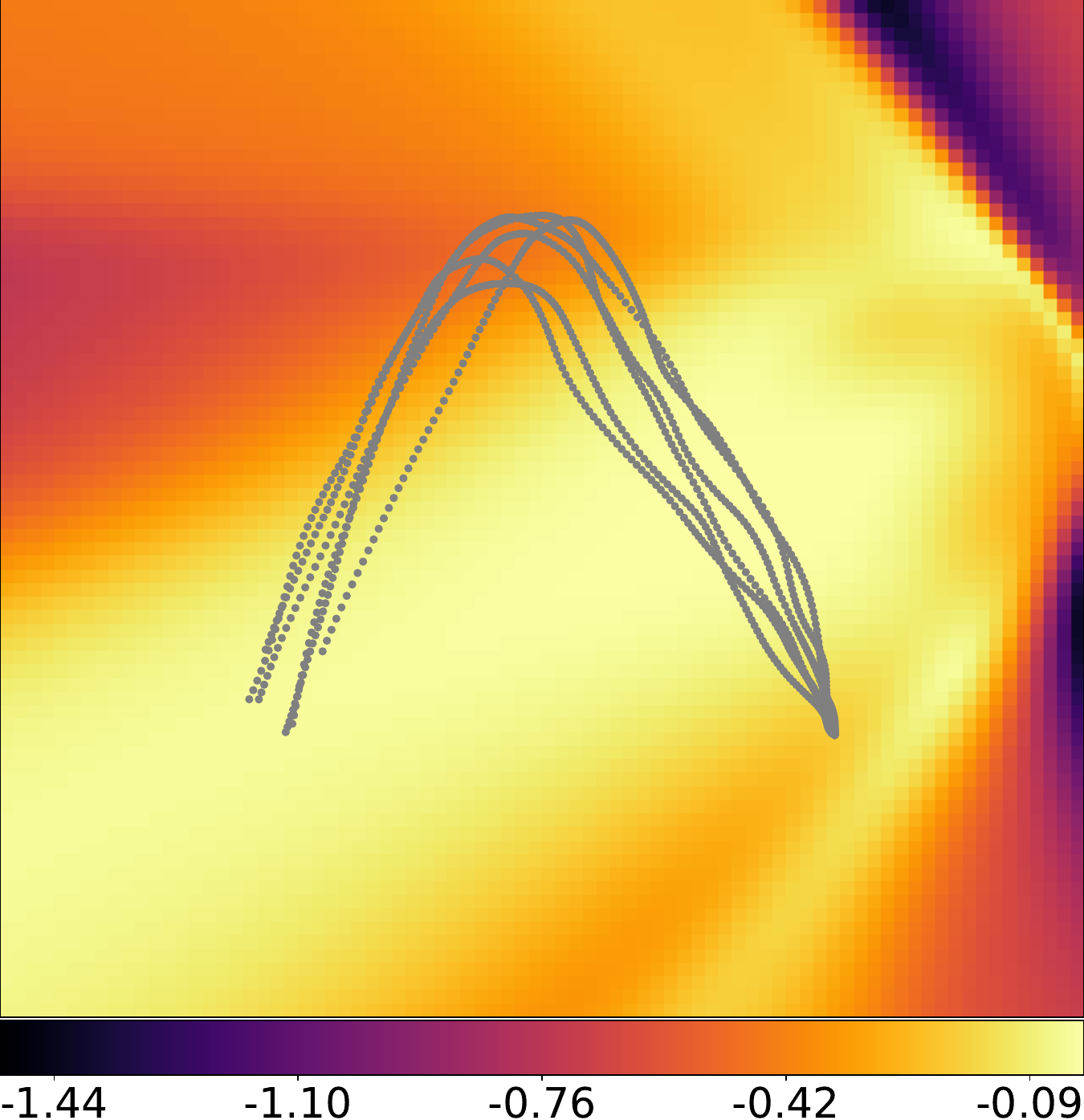}
        \end{subfigure}
        \begin{subfigure}{.24\linewidth}
            \includegraphics[width=1.0\linewidth]{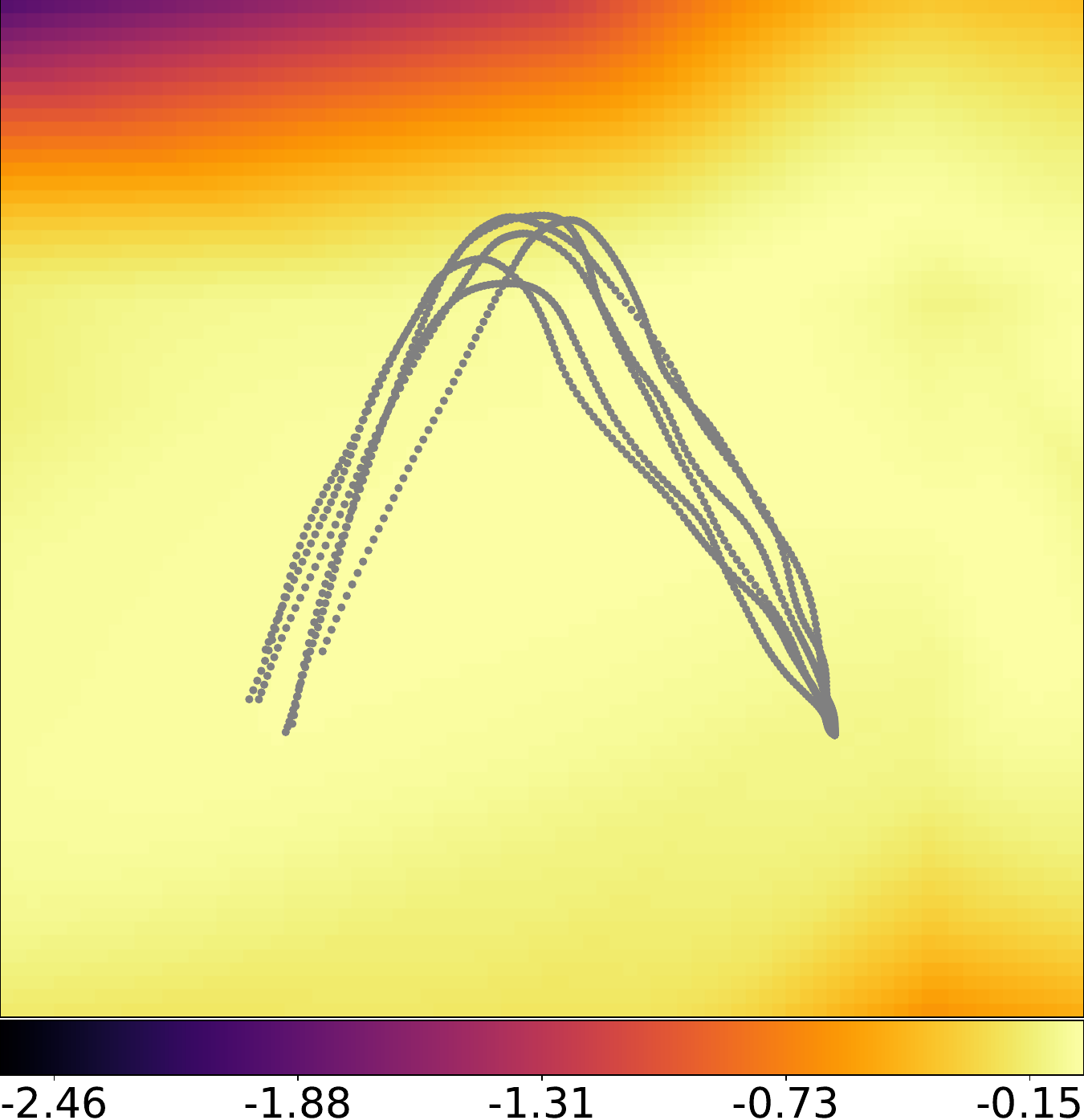}
        \end{subfigure}
        \begin{subfigure}{.24\linewidth}
            \includegraphics[width=1.0\linewidth]{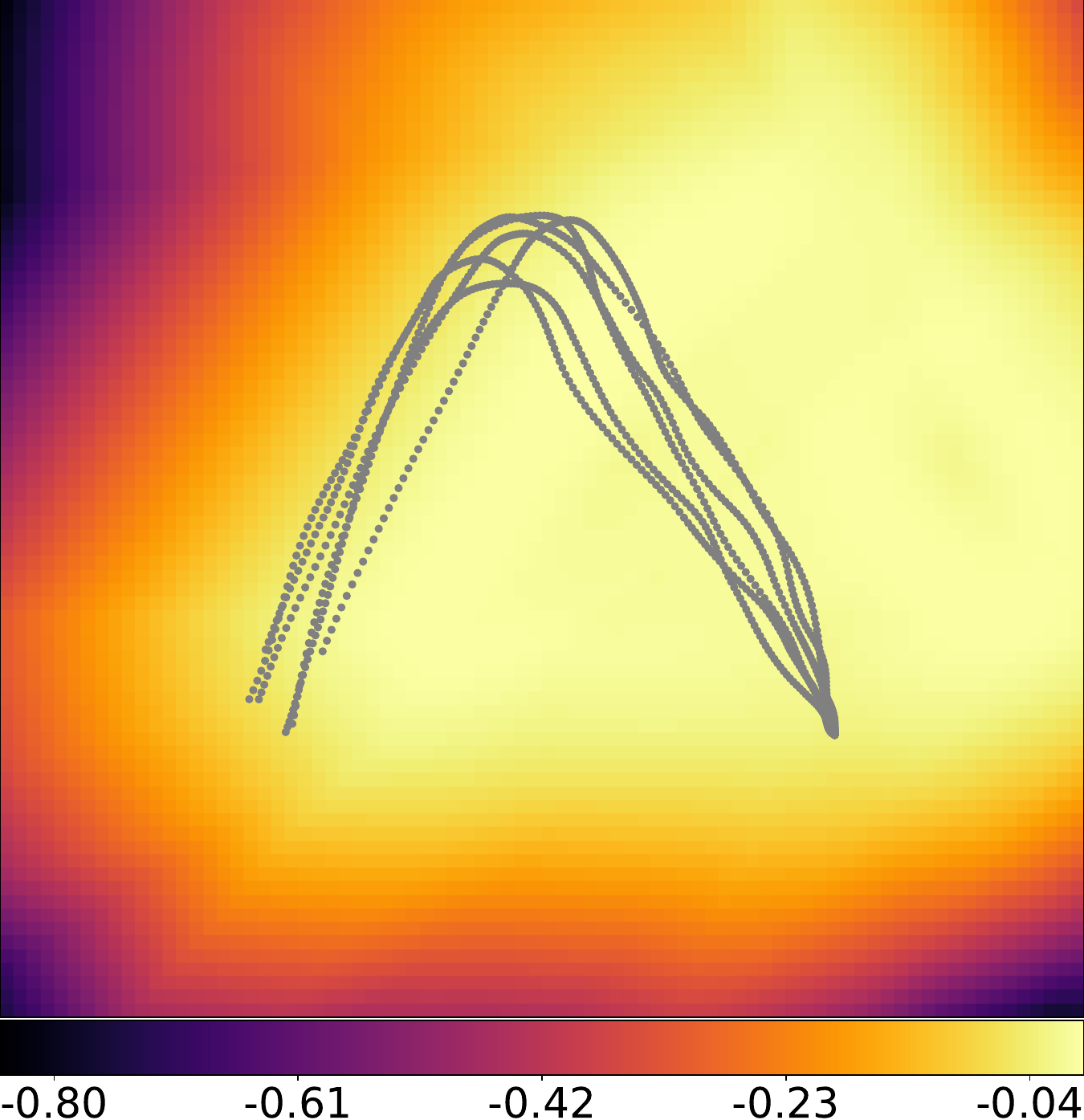}
        \end{subfigure}
        \begin{subfigure}{.24\linewidth}
            \includegraphics[width=1.0\linewidth]{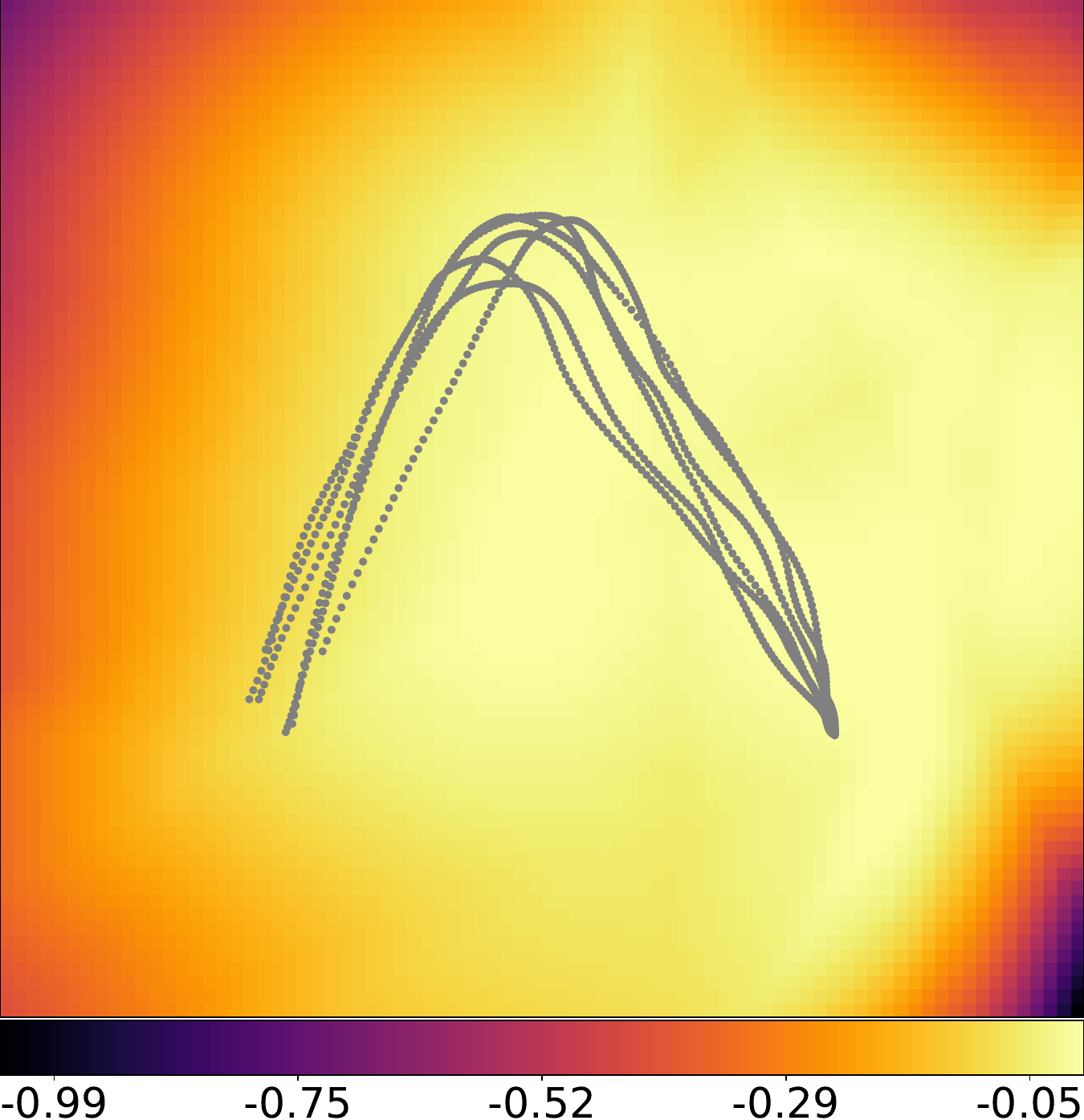}
        \end{subfigure}
        \begin{subfigure}{.24\linewidth}
            \includegraphics[width=1.0\linewidth]{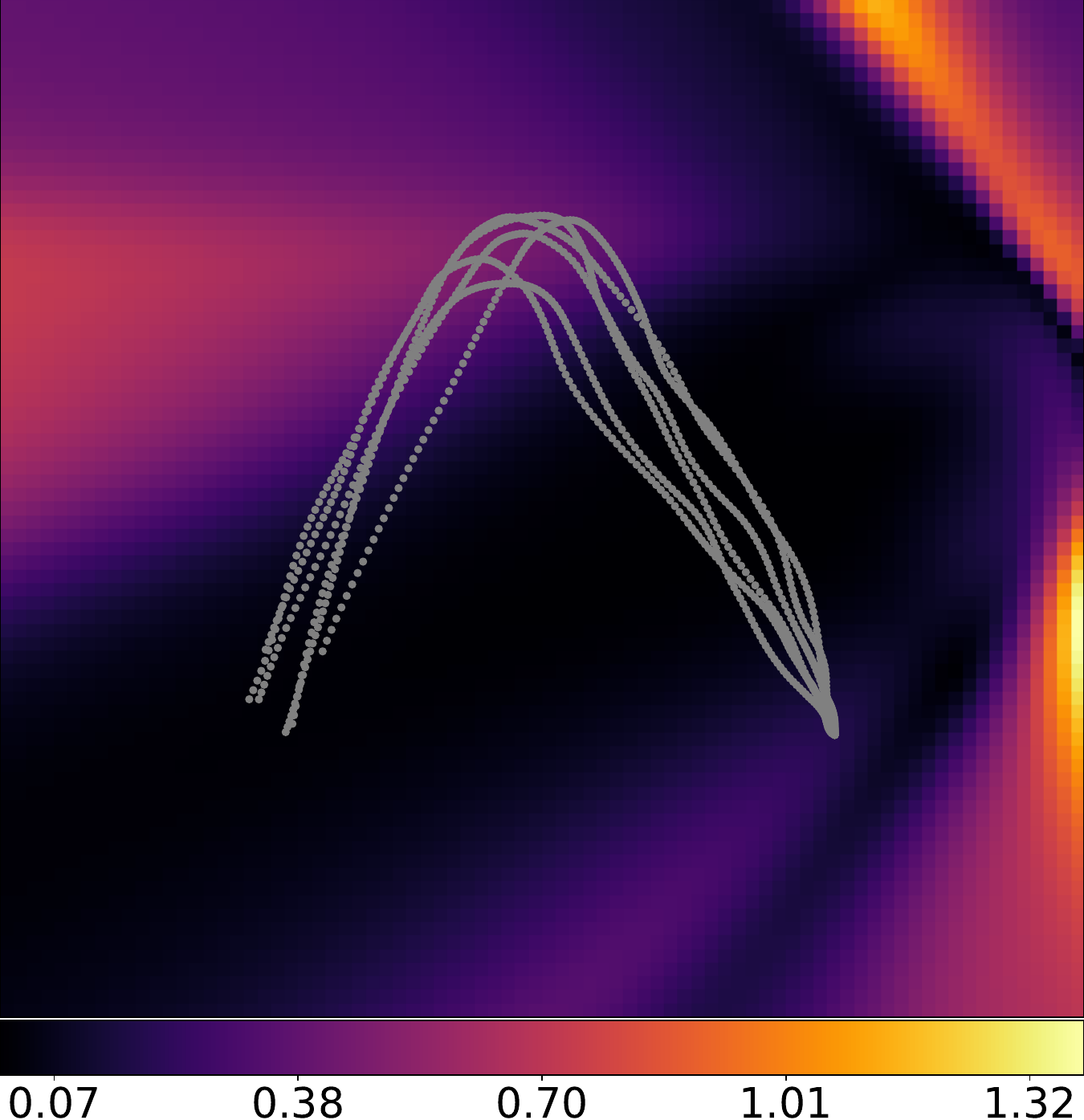}
            \caption{}
        \end{subfigure}
        \begin{subfigure}{.24\linewidth}
            \includegraphics[width=1.0\linewidth]{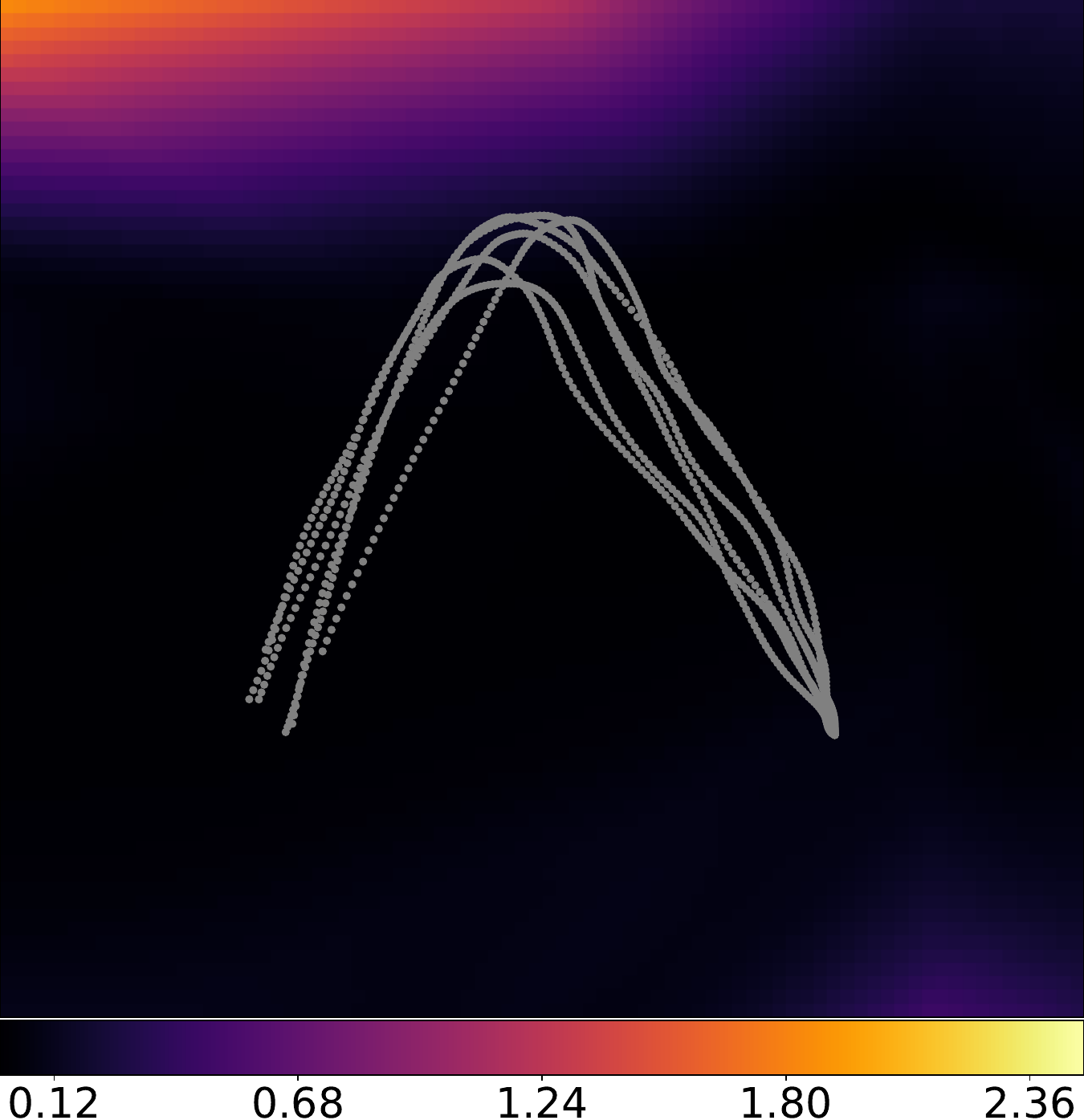}
            \caption{}
        \end{subfigure}
        \begin{subfigure}{.24\linewidth}
            \includegraphics[width=1.0\linewidth]{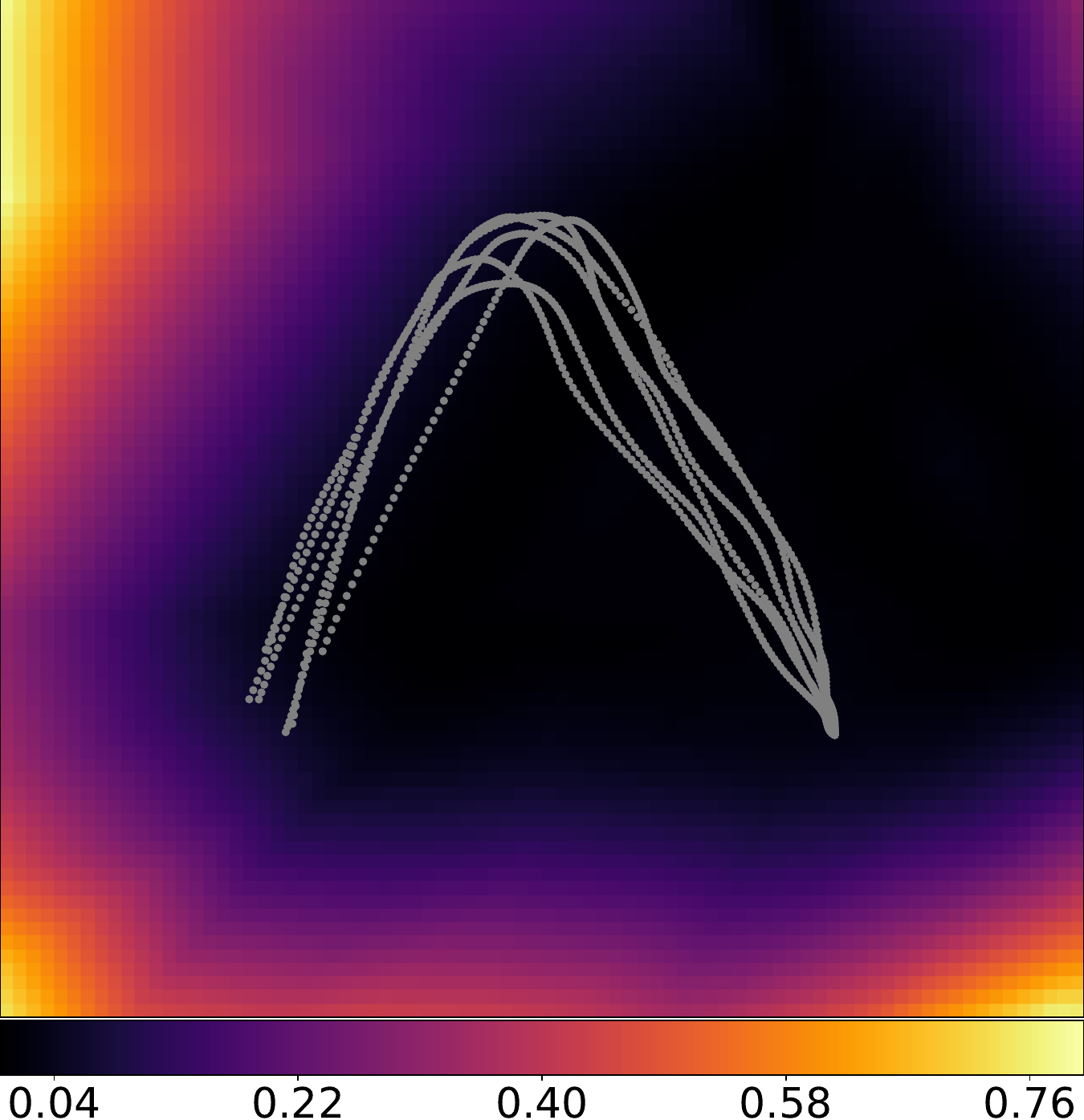}
            \caption{}
        \end{subfigure}
        \begin{subfigure}{.24\linewidth}
            \includegraphics[width=1.0\linewidth]{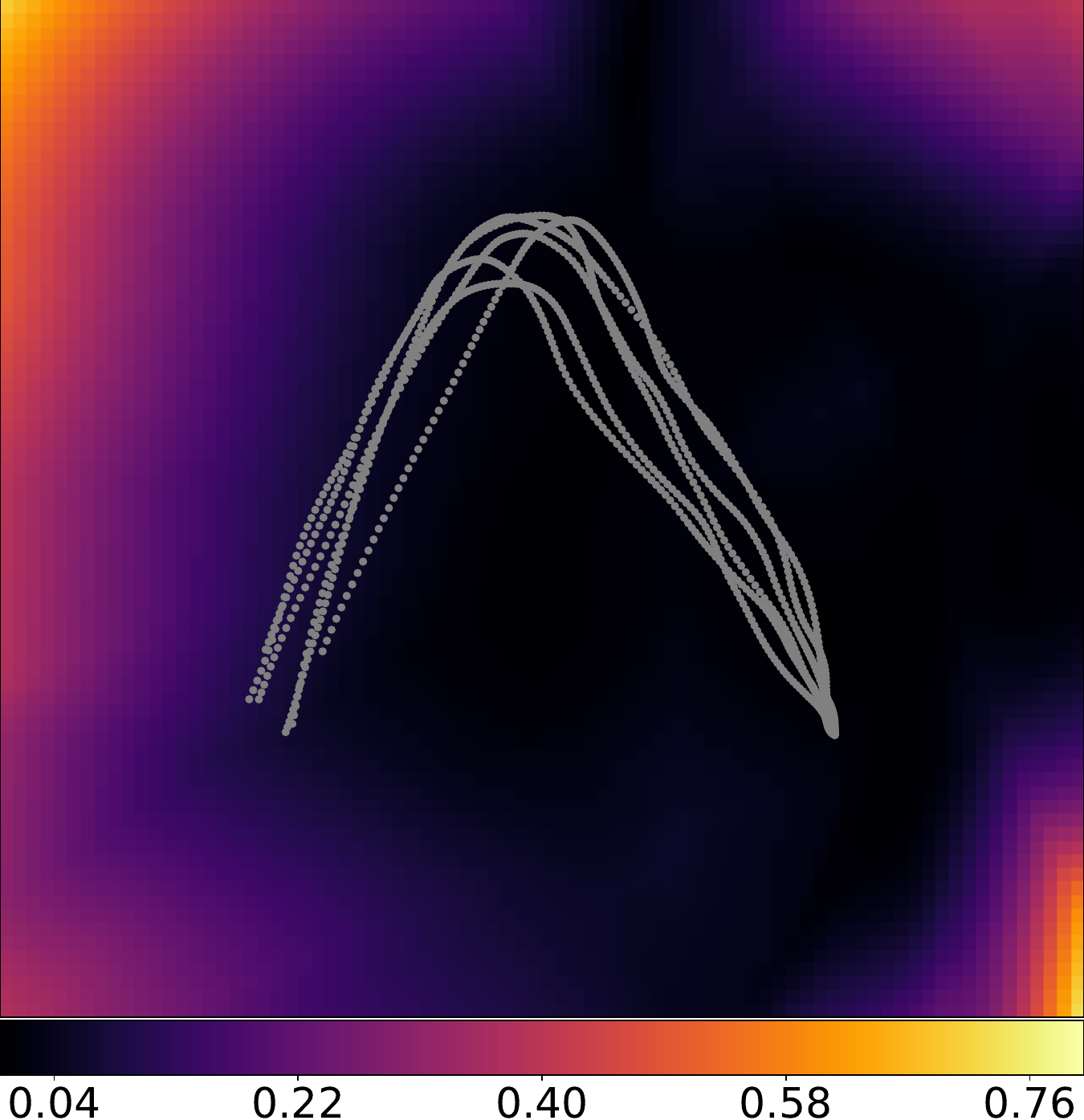}
            \caption{}
        \end{subfigure}
            \caption{Comparison of 2D vector fields learned by using the following regularization approaches: 
            \textbf{(a)} constant regularization,
            \textbf{(b)} state-independent regularization (using basic optimization),
            \textbf{(c)} state-dependent regularization (using a neural network), and
            \textbf{(d)} eigenvalue regularization.
            The \textit{top row} displays a map of the maximum eigenvalue, while the \textit{bottom row} shows the absolute difference between eigenvalues (i.e., the contraction spread).
            }
        \label{fig:regularization_heatmaps}
\end{figure*}
\subsection{Modulation scaling factor}
\label{App:alpha}
To compute the weight $\alpha$, we discretize the data manifold with an equidistant mesh grid in the latent space $\latent$, followed by the computation of the corresponding metric volumes $\mathcal{V}(\z)$ for each point on the mesh grid. 
Later, the maximum and the minimum volumes, $\mathcal{V}_{\max}$ and $\mathcal{V}_{\min}$, are used to normalize all volumes $\mathcal{V}(\z)$ as follows,
\begin{equation*}
\mathcal{V}(\z) = \cfrac{\mathcal{V}(\z) - \mathcal{V}_{\min}}{\mathcal{V}_{\max} - \mathcal{V}_{\min}}.
\end{equation*}
Afterward, $\alpha$ is experimentally chosen to align with the specified distance range defined by $\rho$ and $\nu$ in~\eqref{eq:sigma}. 
This ensures that the scalar field is scaled appropriately, such that:
\begin{align*}
    \mathfrak{S}(\z)(\mathbf{z}) <& \rho, \quad \mathbf{z} \in \text{obstacle region}, \\ \mathfrak{S}(\z)(\mathbf{z}) >& \nu, \quad \mathbf{z} \notin \text{obstacle region}.
\end{align*}

\subsection{Eigenvalue metric maps of regularization methods}
\label{app:reg_maps}

To better assess the impact of each regularization method, Fig.~\ref{fig:regularization_heatmaps} presents heatmaps depicting the distribution of various eigenvalue metrics within a selected square region around the demonstration data. The results for state-independent regularization show that this method exhibits larger negative values across the spectrum, indicating overall stronger contraction. Specifically, within the selected region, the contraction rate reaches $-0.15$, the highest among all methods. Additionally, as shown in the figure, this method achieves the largest contraction spread, suggesting a more directionally dependent contractive behavior.

\subsection{Eigenvalue metric maps of symmetric vs. asymmetric Jacobian}
\label{app:sym_maps}

\begin{figure*}[htb!]
    \centering
    \begin{subfigure}[b]{0.24\linewidth}
        \centering
        \includegraphics[width=\linewidth]{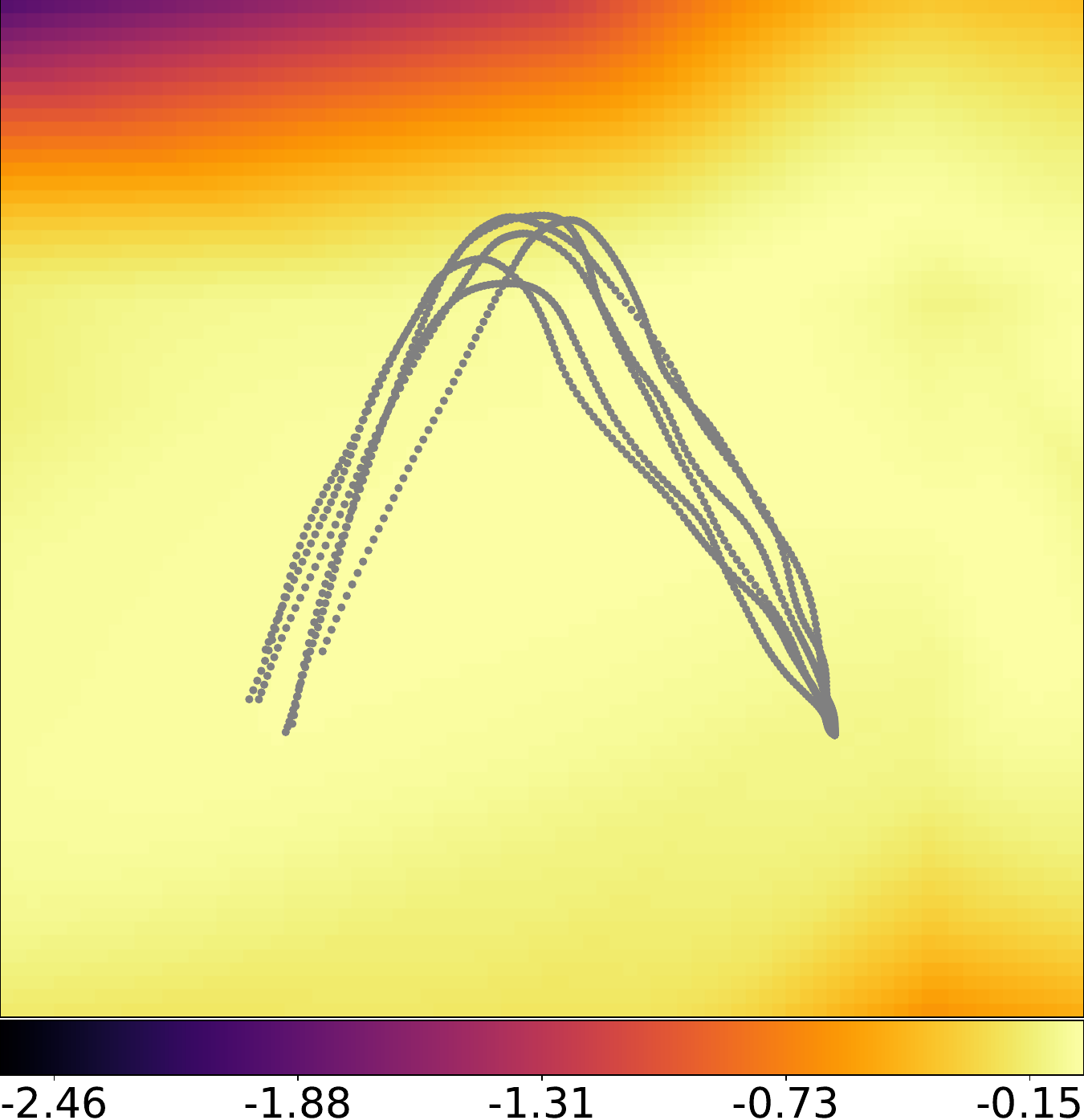}
        \label{fig:subfig0}
    \end{subfigure}
    \begin{subfigure}[b]{0.24\linewidth}
        \centering
        \includegraphics[width=\linewidth]{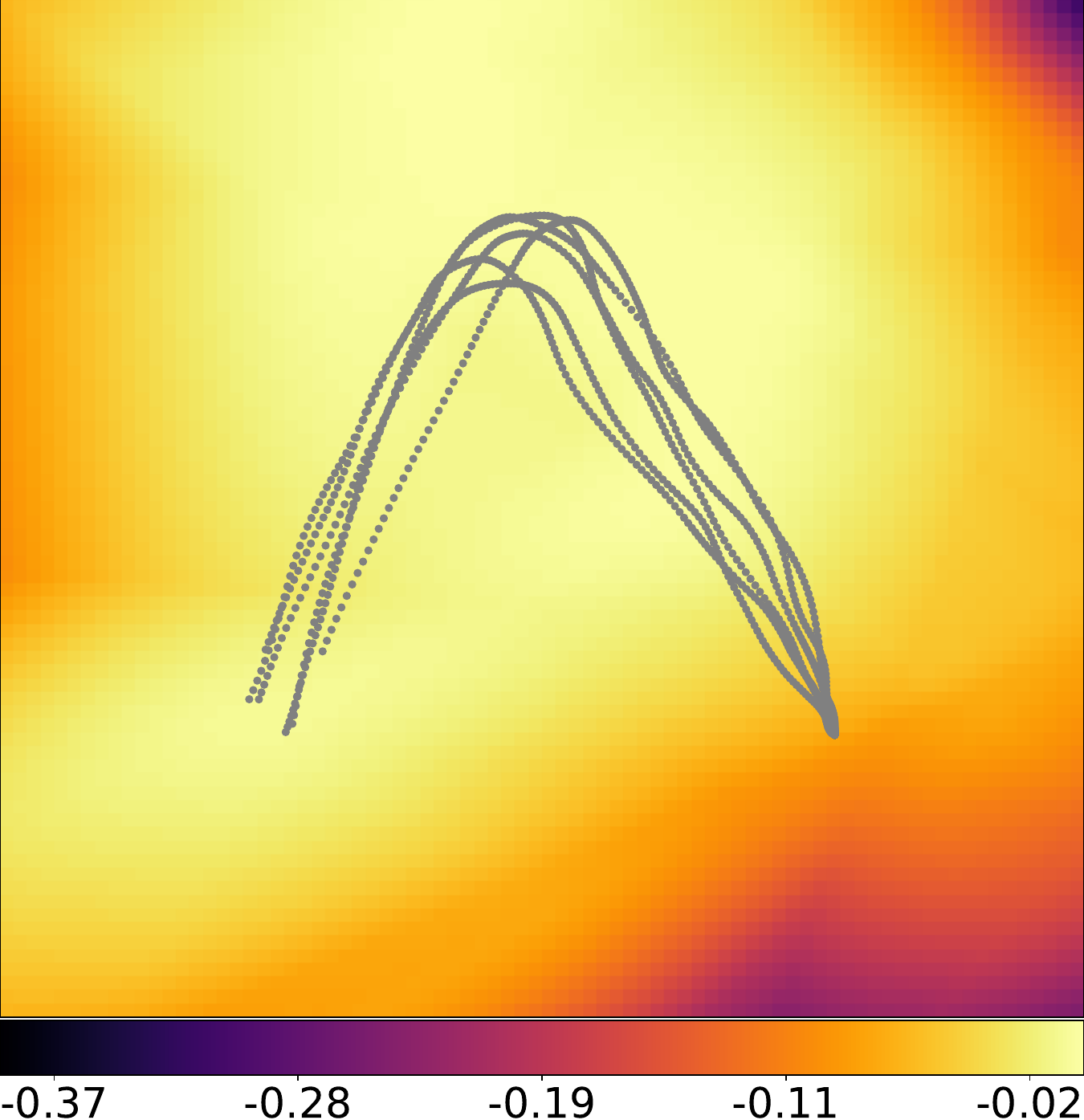}
        \label{fig:subfig1}
    \end{subfigure}
    \begin{subfigure}[b]{0.24\linewidth}
        \centering
        \includegraphics[width=\linewidth]{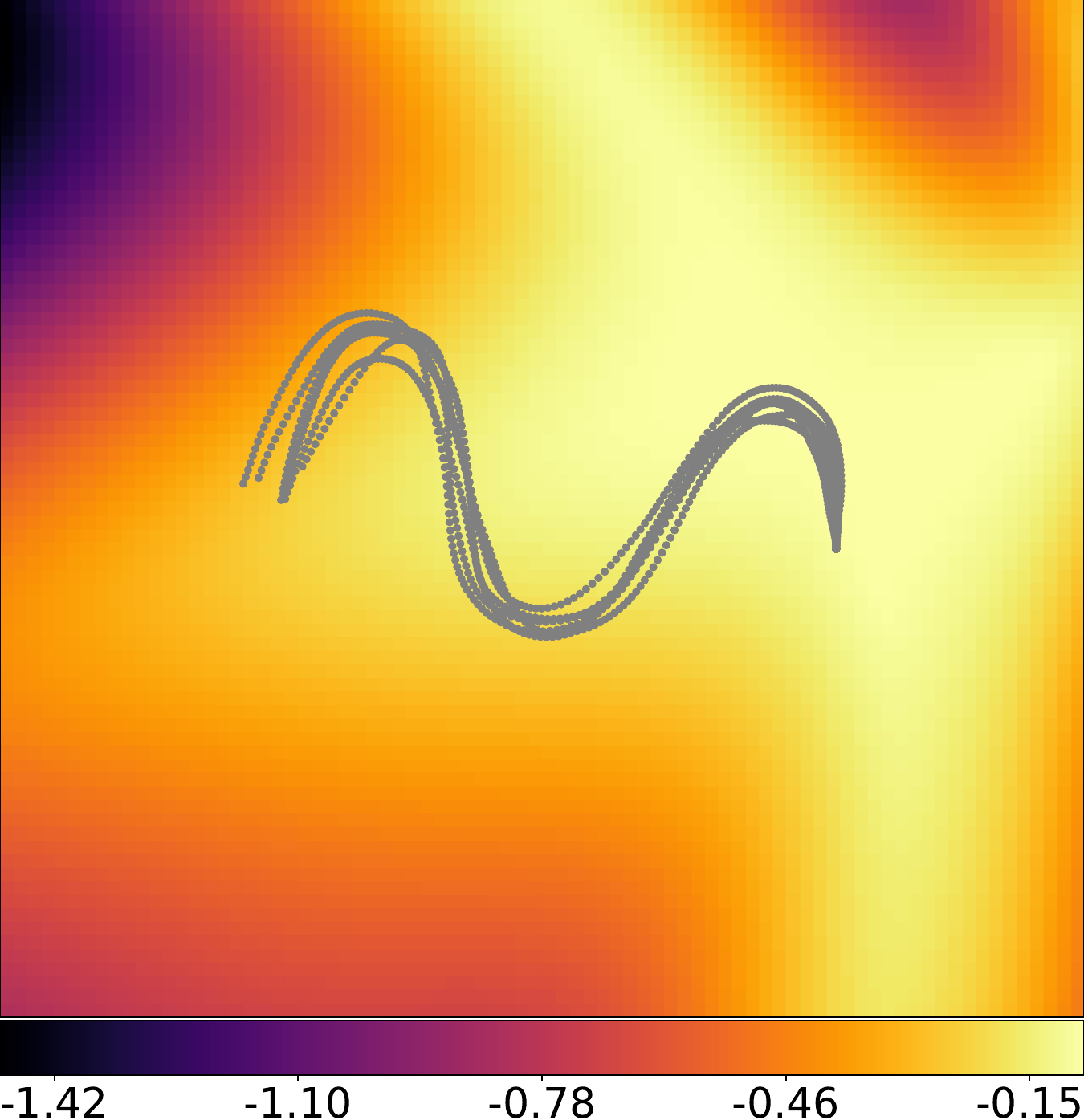}
        \label{fig:subfig2}
    \end{subfigure}
    \begin{subfigure}[b]{0.24\linewidth}
        \centering
        \includegraphics[width=\linewidth]{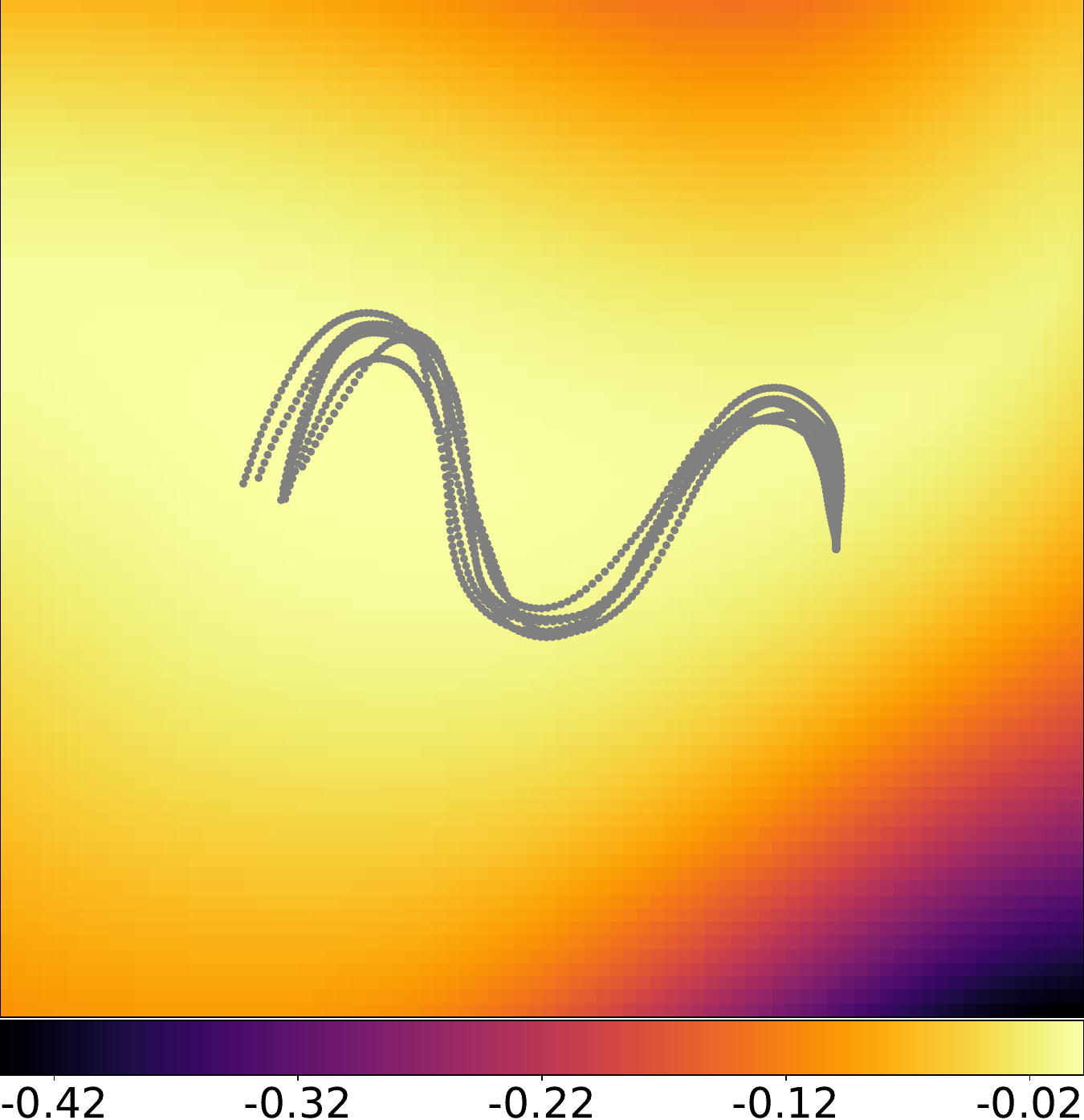}
        \label{fig:subfig3}
    \end{subfigure}\\ \vspace{-10pt}
        \begin{subfigure}[b]{0.24\linewidth}
        \centering
        \includegraphics[width=\linewidth]{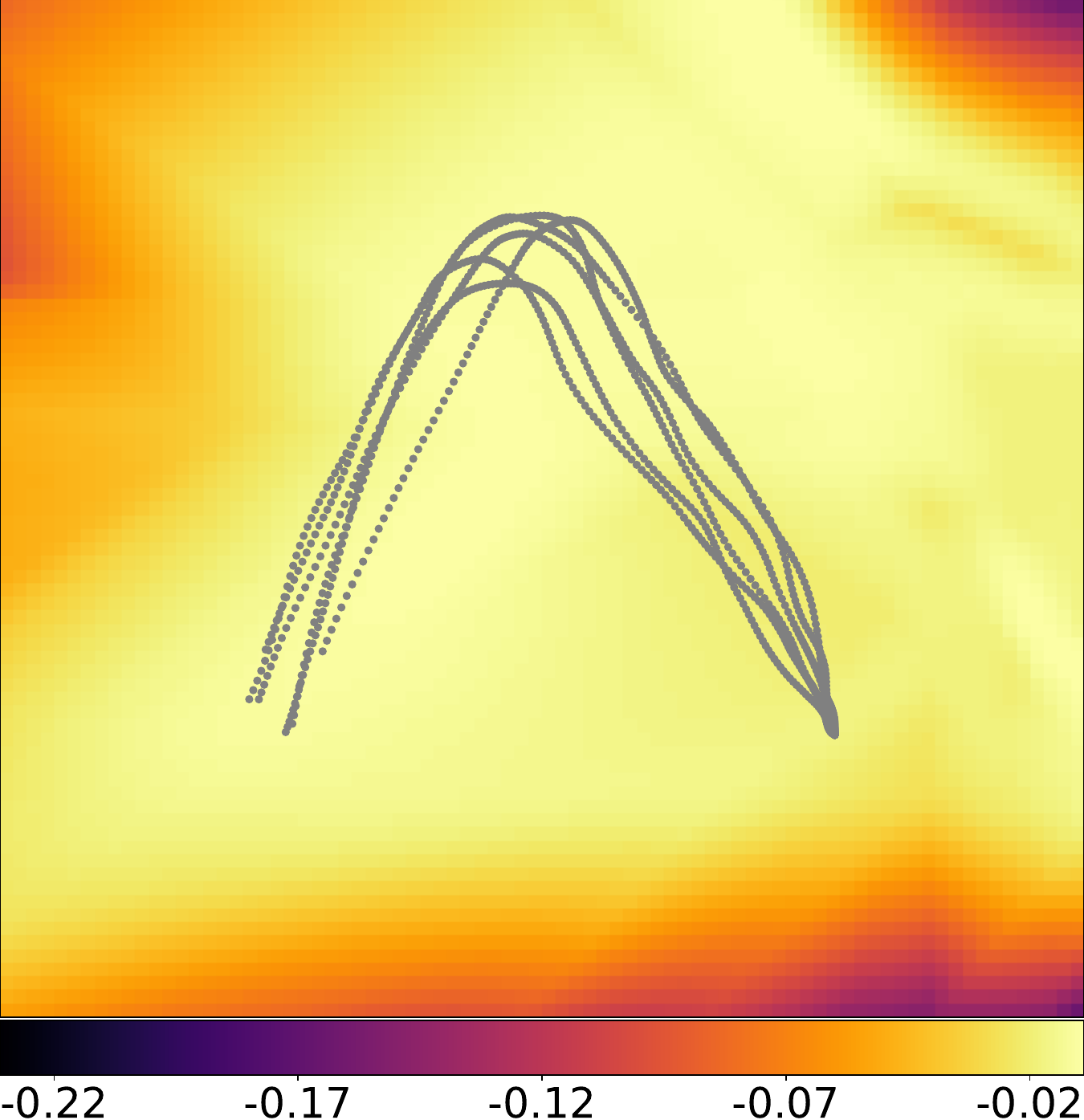}
        \label{fig:subfig0}
    \end{subfigure}
    \begin{subfigure}[b]{0.24\linewidth}
        \centering
        \includegraphics[width=\linewidth]{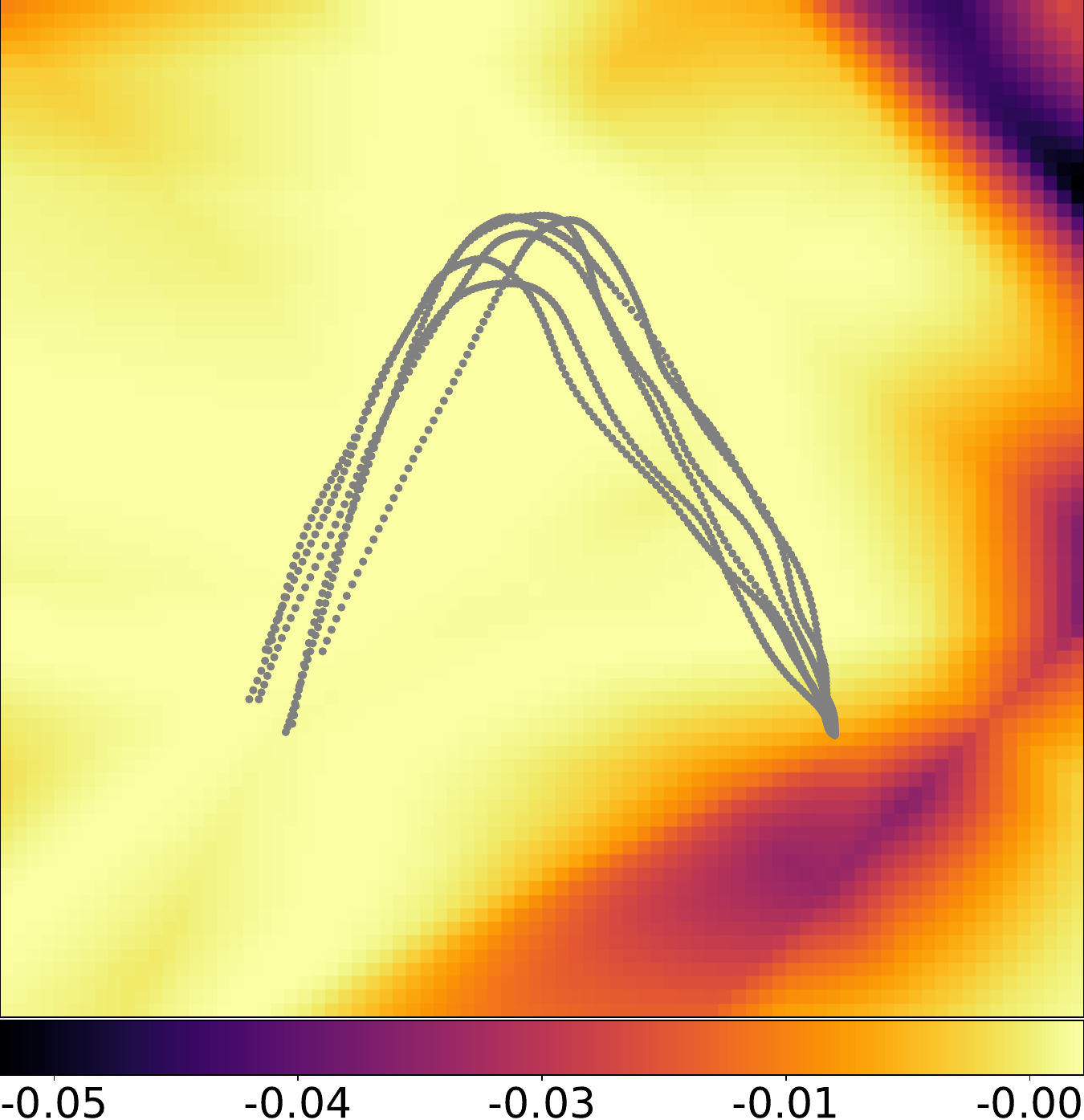}
        \label{fig:subfig1}
    \end{subfigure}
    \begin{subfigure}[b]{0.24\linewidth}
        \centering
        \includegraphics[width=\linewidth]{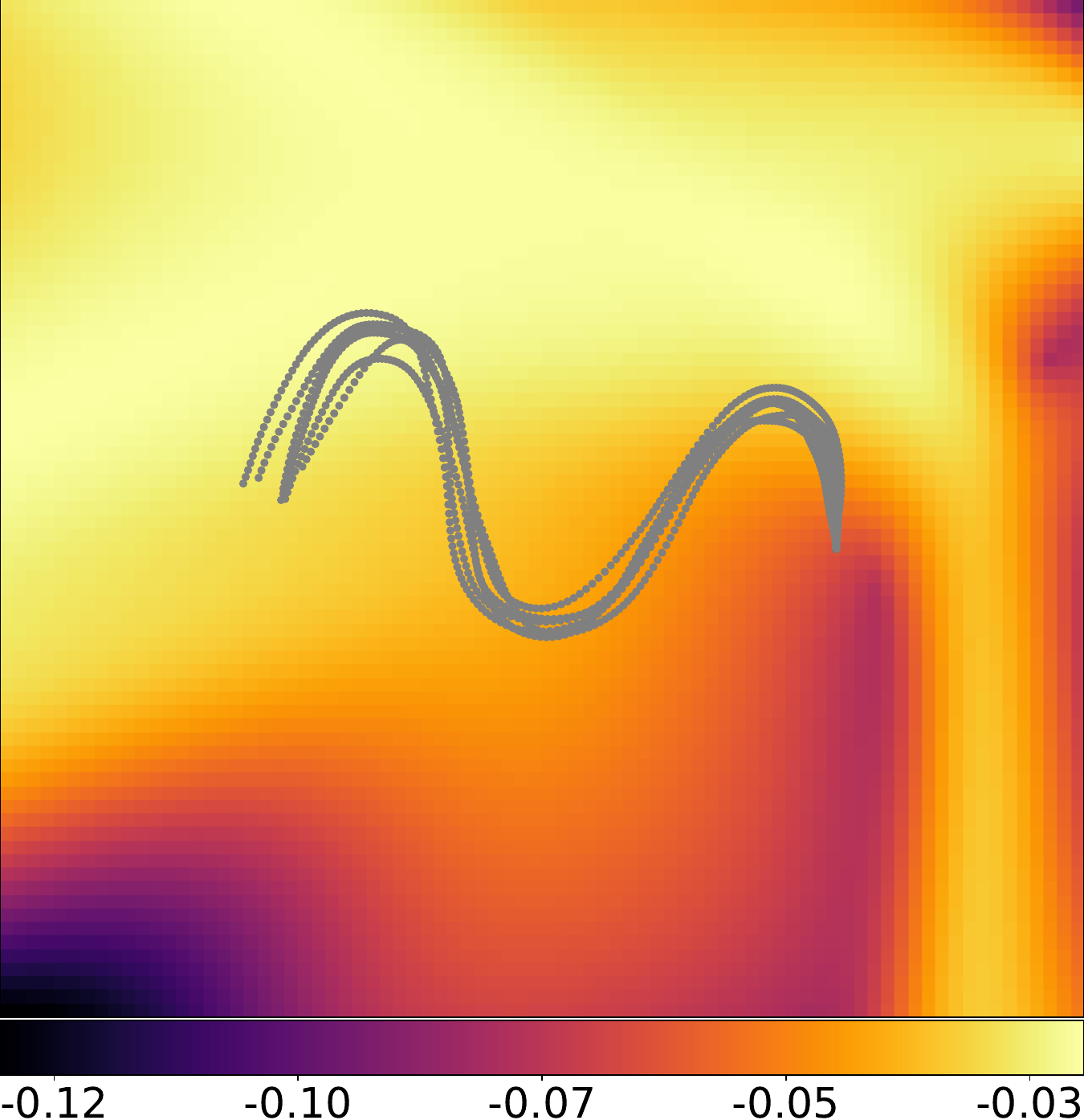}
        \label{fig:subfig2}
    \end{subfigure}
    \begin{subfigure}[b]{0.24\linewidth}
        \centering
        \includegraphics[width=\linewidth]{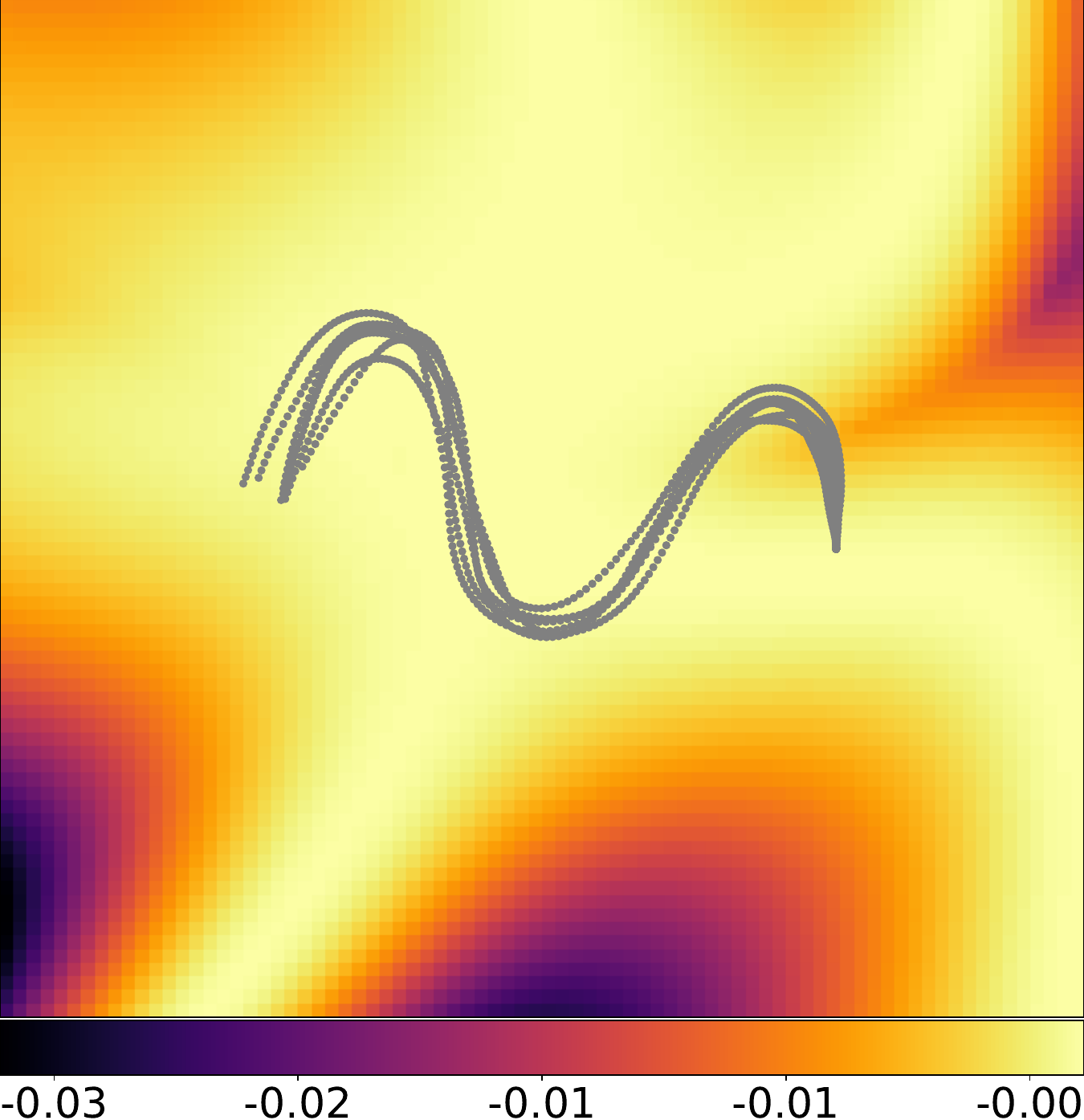}
        \label{fig:subfig3}
    \end{subfigure}\\ \vspace{-10pt}
        \begin{subfigure}[b]{0.24\linewidth}
        \centering
        \includegraphics[width=\linewidth]{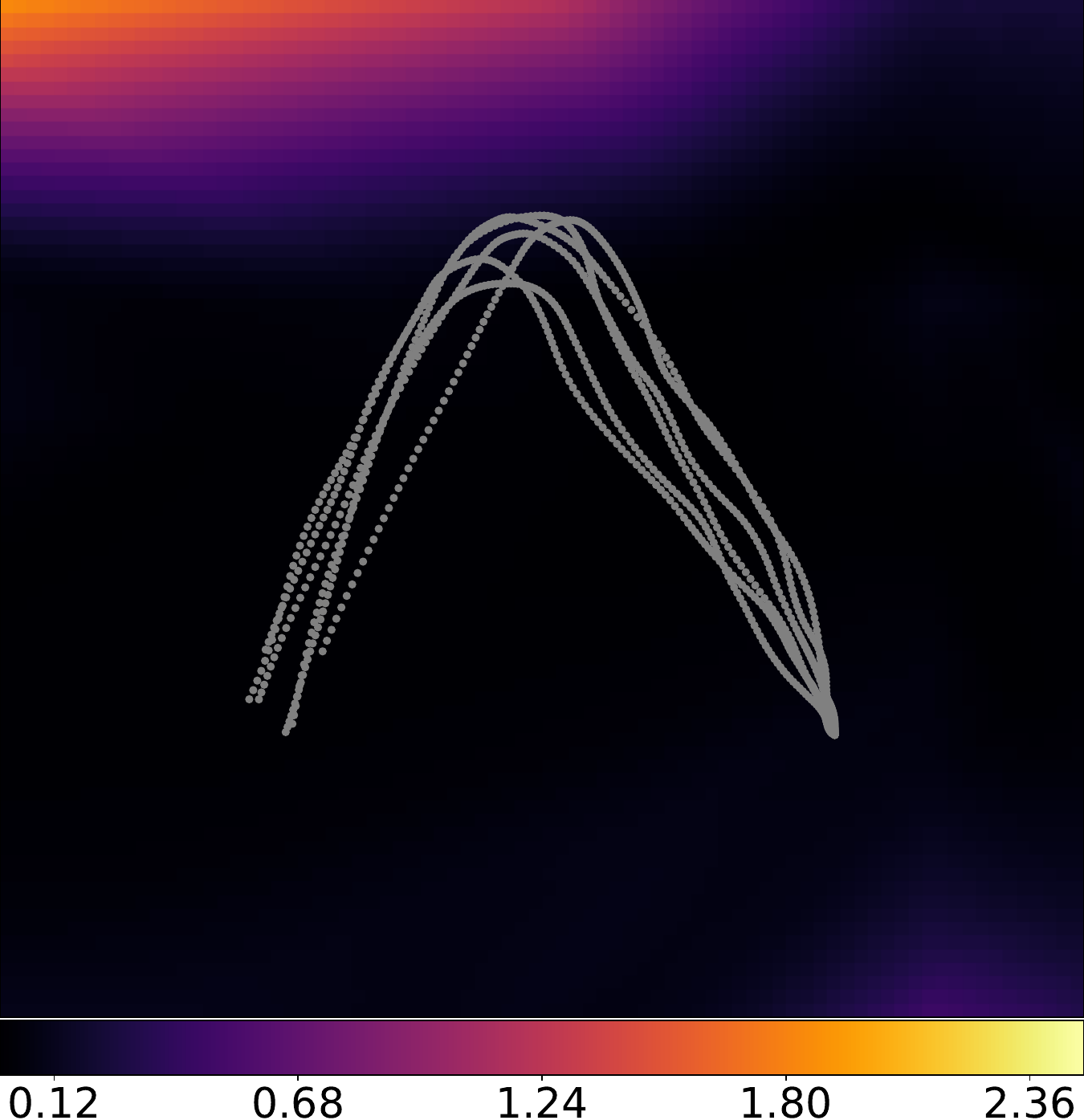}
        \label{fig:subfig0}
    \end{subfigure}
    \begin{subfigure}[b]{0.24\linewidth}
        \centering
        \includegraphics[width=\linewidth]{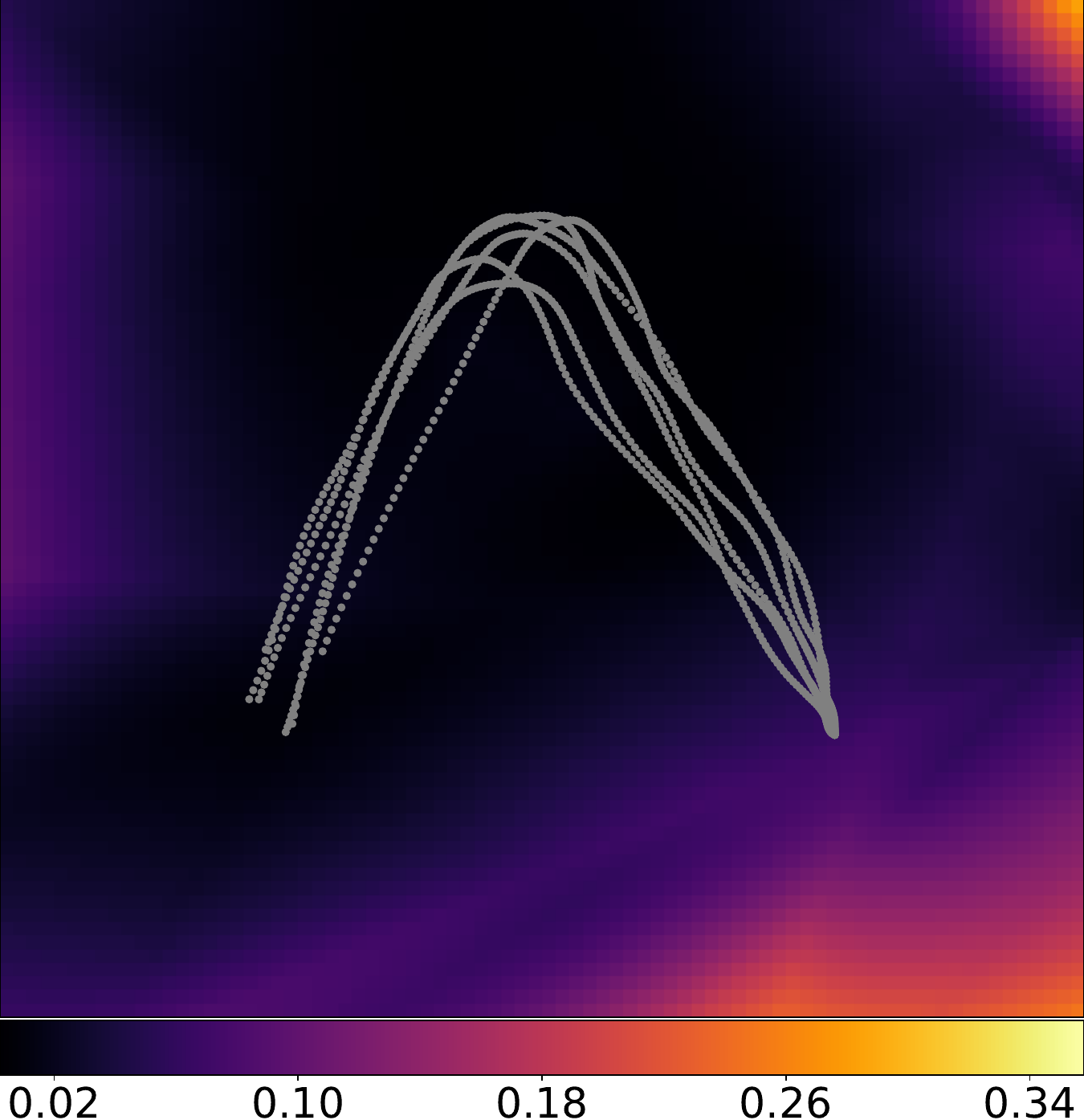}
        \label{fig:subfig1}
    \end{subfigure}
    \begin{subfigure}[b]{0.24\linewidth}
        \centering
        \includegraphics[width=\linewidth]{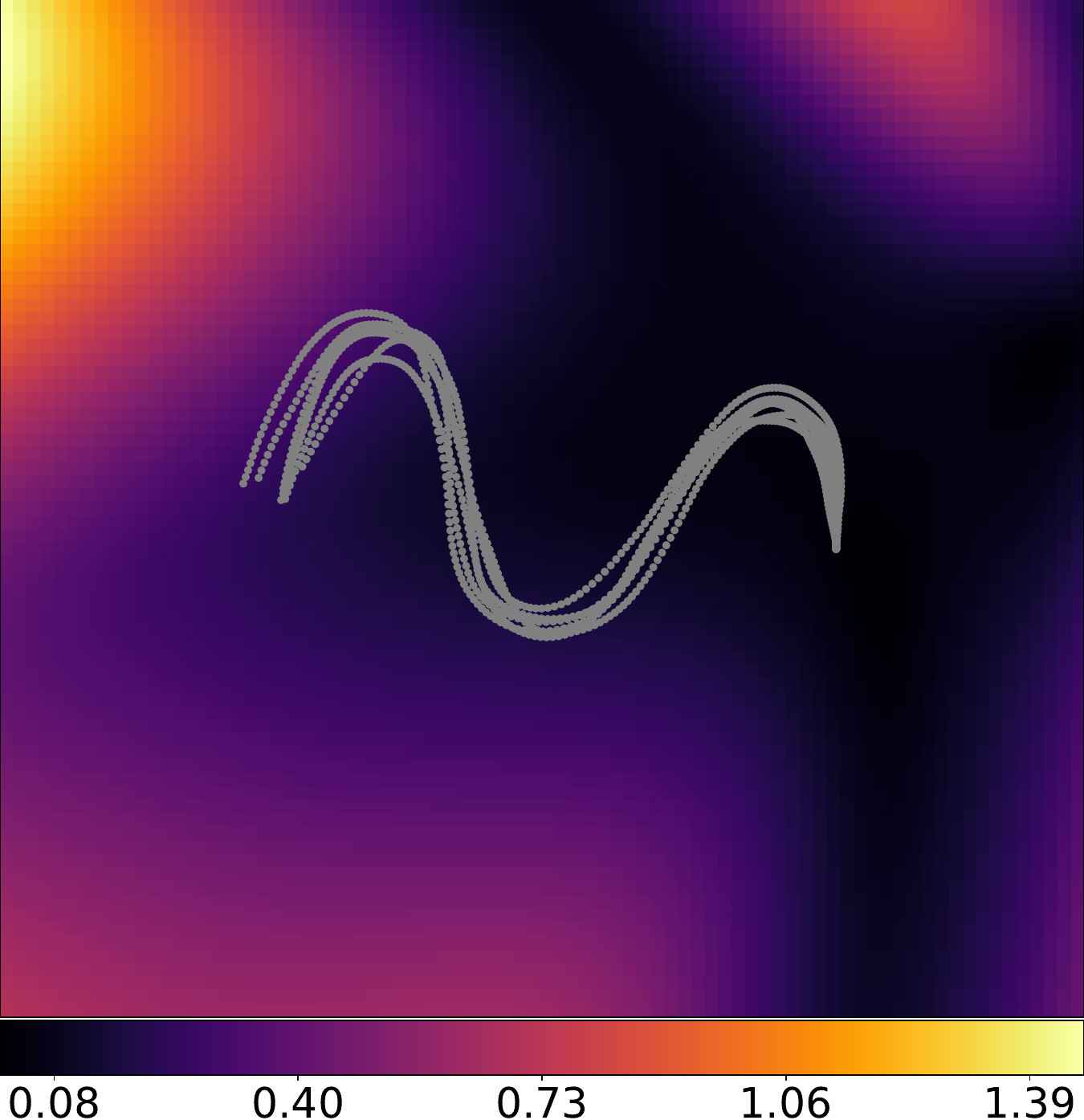}
        \label{fig:subfig2}
    \end{subfigure}
    \begin{subfigure}[b]{0.24\linewidth}
        \centering
        \includegraphics[width=\linewidth]{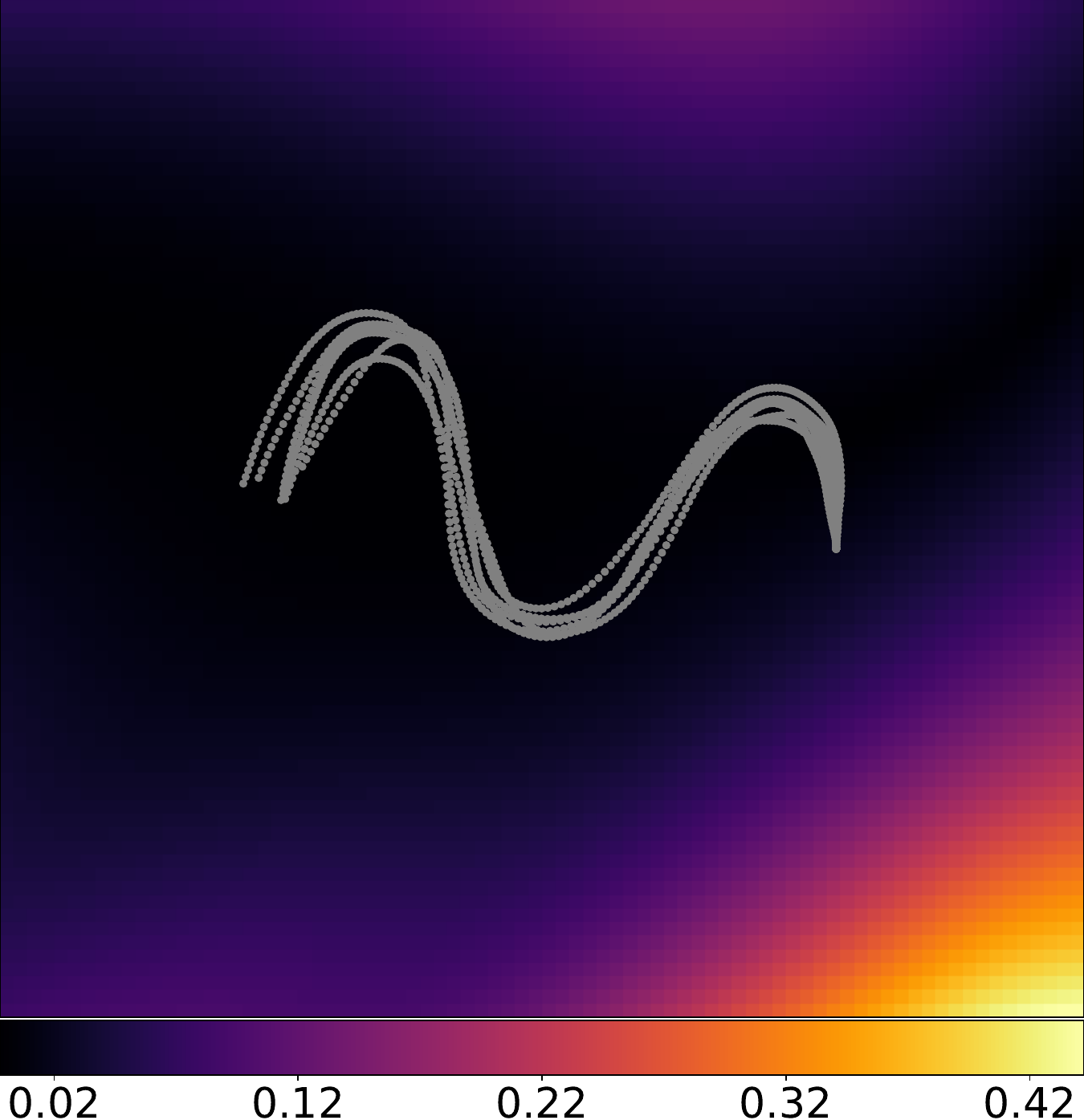}
        \label{fig:subfig3}
    \end{subfigure} 
    \caption{Eigenvalue metric heatmaps of the learned contractive dynamics shown in Fig.~\ref{fig:symmetry_assymetry}. The \textit{top} and \textit{middle} rows display the minimum and maximum eigenvalues computed at each point on the equidistant grid, respectively. The \textit{bottom} row shows the contraction spread, defined as the absolute difference between these eigenvalues. The first and third columns correspond to the results obtained with a symmetric Jacobian for the \emph{Angle} and \emph{Sine} datasets, respectively, while second and forth columns show the corresponding results for an asymmetric Jacobian.}
    \label{fig:assymetric_contraction_metric_heatmaps}
\end{figure*}

Figure~\ref{fig:assymetric_contraction_metric_heatmaps} shows different eigenvalue metric heatmaps of the learned contractive dynamics. The top row shows that symmetric Jacobian approach results in increased contraction along the ``dominant'' eigenvector. In the middle row, we observe that even the ``non-dominant'' eigenvector has stronger contraction under this method. Finally, the bottom row shows a higher overall contraction spread, indicating a more pronounced pull toward a specific eigenvector.

More precisely, in the \emph{Angle} dataset, the average minimum eigenvalue is $-0.266$ compared to $-0.046$ for the asymmetric case, while the average maximum eigenvalue remains more negative ($-0.034$ vs. $-0.006$). The maximum contraction spread, representing the difference between these eigenvalues, is significantly higher for the symmetric model ($2.489$ vs. $0.362$), confirming its more robust contractive behavior. Extreme values further support this trend, with the minimum eigenvalue reaching $-2.589$ for the symmetric case, compared to only $-0.388$ for the asymmetric model.

Similarly, in the \emph{Sine} dataset, the symmetric model maintains a average minimum eigenvalue of $-0.381$ versus $-0.063$ in the asymmetric case, with a more negative average maximum eigenvalue ($-0.042$ vs. $-0.003$). The maximum contraction spread is also notably higher ($1.457$ vs. $0.439$), reinforcing the role of symmetry in promoting directional contraction. The extreme minimum eigenvalue reaches $-1.489$ in the symmetric formulation, whereas it remains at $-0.439$ for the asymmetric model.

